\documentclass[11pt, a4paper, twoside, openright]{memoir}

\usepackage{graphicx}
\graphicspath{{images/}}

\usepackage{seb}\usepackage{layouts}
\captionnamefont{\small}
\captiontitlefont{\small}
\tcbuselibrary{breakable}
\usepackage{subfiles} 

\usepackage{amsmath,amsfonts,bm}

\def\eqref#1{equation~\ref{#1}}

\def\1{\bm{1}}

\newcommand{\data}{\mathcal{D}}

\newcommand{\pred}{\hat{y}}
\newcommand{\vpred}{\hat{\vy}}
\newcommand{\gaussian}{\mathcal{N}}

\def\rr{{\textnormal{r}}}

\def\rx{{\textnormal{x}}}
\def\ry{{\textnormal{y}}}

\def\rvtheta{{\mathbf{\theta}}}

\def\rvw{{\mathbf{w}}}
\def\rvx{{\mathbf{x}}}

\def\rmX{{\mathbf{X}}}

\def\vmu{{\bm{\mu}}}
\def\vtheta{{\bm{\theta}}}
\def\vphi{{\bm{\phi}}}
\def\vsigma{{\bm{\sigma}}}

\def\vw{{\bm{w}}}
\def\vx{{\bm{x}}}
\def\vy{{\bm{y}}}

\def\mX{{\bm{X}}}

\DeclareMathAlphabet{\mathsfit}{\encodingdefault}{\sfdefault}{m}{sl}
\SetMathAlphabet{\mathsfit}{bold}{\encodingdefault}{\sfdefault}{bx}{n}

\def\gX{{\mathcal{X}}}
\def\gY{{\mathcal{Y}}}

\newcommand{\pdata}{p_{\textup{data}}}

\newcommand{\bigO}{\mathcal{O}}

\newcommandx{\E}[2]{\operatorname{\mathbb{E}}_{#1}\left[#2\right]}
\newcommand{\Ls}{\mathcal{L}}

\newcommand{\KL}{D_{\mathrm{KL}}}
\DeclarePairedDelimiterX{\KLdivx}[2]{\big(}{\big)}{%
  #1\;\delimsize\|\;#2%
}
\newcommand{\KLdiv}{\KL \KLdivx}

\newcommand{\Var}{\mathrm{Var}}

\DeclareMathOperator*{\argmax}{arg\,max}
\DeclareMathOperator*{\argmin}{arg\,min}

\input{tikzit.sty}

\title{Understanding Approximation for Bayesian Inference in Neural Networks}
\author{Sebastian Farquhar}
\date{\today}

\chapterstyle{ger}
\begin{document}
\pagestyle{plain}
\thispagestyle{empty}
    \begin{center}
        \vspace{0.2cm}
        {\huge {\bfseries {Understanding Approximation for Bayesian Inference in Neural Networks}} \par}
        \vspace*{1ex}
        {\large \emph{DPhil Thesis} \par}
        {\large \vspace*{20mm} {{\includegraphics{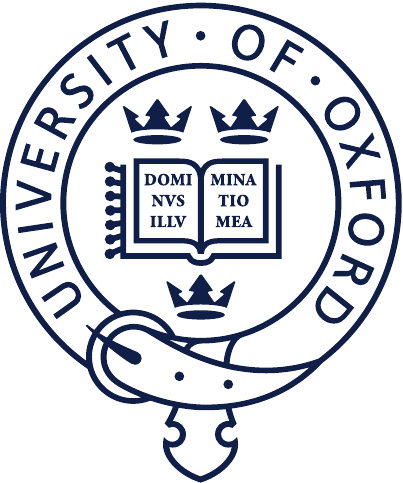}} \par} \vspace*{25mm}}
        {\large \vspace*{1ex}
            {{\Large Sebastian Farquhar} \par}
            {{Balliol College} \\ University of Oxford \par}
            \vspace*{1ex}
            {\texttt{sebastian.farquhar@cs.ox.ac.uk} \par}
            \vspace{1.5cm}
            {Supervised by Yarin Gal\par}
            \vspace*{1cm}
            {2022}
        }
    \end{center}
\newpage
\thispagestyle{empty}
\hfill
\clearpage

\section*{Acknowledgements}
I would like to thank Yarin for helping me become a scientist.
One of our first conversations was about how the opportunities created by empirical phenomena in machine learning in this decade were every bit as exciting as phenomena in electricity and magnetism of the nineteenth century.
That spirit of exploration has stuck with me, and become an essential part of how I see myself and my work, though we may never get any Maxwell's equations.

In addition, I am grateful to many others for the intellectual stimulation, mentorship, or friendship they have offered during my studies, especially (in alphabetical order): Milad Alizadeh, Joost van Amersfoort, David Burt, Adam Cobb, Tom Everitt, Angelos Filos, Andrew Foong, Aidan Gomez, Andreas Kirsch, Jannik Kossen, Yingzhen Li, Clare Lyle, Romy Minko, Mike Osborne, James Pavur, Tom Rainforth, Tim Rudner, Lewis Smith, and Mark van der Wilk.
In some cases they did things for me that they may have considered small, but meant a great deal to me.

I would also like to thank Professor Stephen Roberts and Professor Andrew Gordon Wilson for serving as examiners of this thesis.
Their expertise and diligence have greatly improved this thesis and more importantly have significantly changed how I think about these issues.

My intellectual progress has also depended essentially on the generosity of many anonymous reviewers.
I wish I could express my gratitude to them personally.
If you happen to have reviewed one of my papers and are reading this, I want you to understand that my appreciation is sincerely meant.

My brother, Gregory Farquhar, has also been a constant source of encouragement and an invaluable sounding-board for my research ideas.
I think that my favorite paper so far only exists because he encouraged me to just spend a fortnight trying to get something solid enough to persuade an initially-skeptical-Yarin it was worth pursuing further.

Besides the research of the last few years, I want to thank both of my parents for teaching me the habit of trying to understand what I see and say.
My father, Adam Farquhar, especially deserves credit for helping me to learn pragmatic idealism.
My mother, Angela Dappert, taught me so much that I value, but among them was the perseverance she showed in her own journey to getting a PhD, despite innumerable obstacles.
It saddens me unspeakably that she will never see my completed thesis, and I miss her every day.

I would especially like to thank my wife, Leah Broad, for her kindness, love, and guidance.
Her support makes it possible for me to read what Reviewer 2 has to say for the truths it contains without fear or anger.
Beyond that, her diligence and determination are an inspiration to me.
It does not surprise me one jot that she beat me in our race as to whether she would hand in her book before I handed in my thesis.

\chapter*{Abstract}
Bayesian inference has theoretical attractions as a principled framework for reasoning about beliefs.
However, the motivations of Bayesian inference which claim it to be the only `rational' kind of reasoning do not apply in practice.
They create a binary split in which all approximate inference is equally `irrational'.
Instead, we should ask ourselves how to define a spectrum of more- and less-rational reasoning that explains why we might prefer one Bayesian approximation to another.
I explore approximate inference in Bayesian neural networks and consider the unintended interactions between the probabilistic model, approximating distribution, optimization algorithm, and dataset.
The complexity of these interactions highlights the difficulty of any strategy for evaluating Bayesian approximations which focuses entirely on the method, outside the context of specific datasets and decision-problems.
For given applications, the expected utility of the approximate posterior can measure inference quality.
To assess a model's ability to incorporate different parts of the Bayesian framework we can identify desirable characteristic behaviours of Bayesian reasoning and pick decision-problems that make heavy use of those behaviours.
Here, we use continual learning (testing the ability to update sequentially) and active learning (testing the ability to represent credence).
But existing continual and active learning set-ups pose challenges that have nothing to do with posterior quality which can distort their ability to evaluate Bayesian approximations.
These unrelated challenges can be removed or reduced, allowing better evaluation of approximate inference methods.
\setcounter{tocdepth}{2
}
\tableofcontents*
\clearpage
\listoffigures*
\clearpage
\listoftables*
\clearpage


\chapter{Summary}
\label{chp:summary}
\section{Evaluating Bayesian Approximations}
Bayesian inference has many attractions.
When we know the structure of the distribution generating our data, and it allows exact inference, probabilistic updates to prior beliefs let us elegantly compress what we know and assess how ignorant we are.
In that happy case, Bayesian inference arguably follows directly from axioms that represent basic rationality principles \citep{coxAlgebra1961,jaynesProbability2003} and doing anything else can be exploited to your disadvantage \citep{ramseyTruth1926,definettiForesight1937}.
There are reasonable criticisms even in this case, but the conceptual foundations of Bayesian inference are at their strongest.

Unfortunately, in practice we need to make large approximations.
We need to restrict the class of models that we examine and to make approximations to exact inference in those models.
Insofar as Bayesian inference is entailed by principles of rationality, following arguments like those of \citep{jaynesProbability2003}, the principles only offer a binary split of the world into `rational' and `not rational'.
Approximate Bayesian inference, the only kind anybody does, is solidly in the `not rational' category.
But this really just demonstrates how unhelpful that narrow notion of `rationality' is.
The sorts of axiomatic or Dutch-book motivations of Bayesian inference do not really give us a way of deciding which kinds of automated reasoning we should prefer when choosing between practical options.

Methods inspired by Bayesian inference which make use of the approximate probabilistic inference can be very successful.
Moreover, they offer a range of capabilities that `non-Bayesian' methods do not necessarily: such as the ability to `introspect' and know what the model does not know.

So, what makes one Bayesian approximation better than another one?
I argue that it is not a matter of adherence to Bayesian doctrine.
The complicated interactions of the architecture of the model, the choice of approximate inference technique, the dataset, and the optimization strategies make it almost impossible to tell up-front which of two methods actually \textit{is} more `principled'.
An approximation that appears to be severe in one setting can be very mild in another, and vice-versa.
This thesis explores two settings where this sort of behaviour arises, and the significance of different kinds of approximation can vary greatly depending on the context.

In contrast, I try to answer the question of how to evaluate Bayesian approximations in a pragmatic, and largely largely frequentist, way---good expected utility of the approximate posterior under cross-validation.
In order to do this, we want to: identify desirable characteristic behaviours of Bayesian systems, identify decision-problem families that rely primarily on that characteristic behaviour, and develop evaluations for those decision-problems that test the ability of approximate Bayesian methods to provide the desired characteristic behaviour and which generalize to other examples of that decision-problem family.
We can then say that the quality of the Bayesian approximation lies in its ability to deliver the desired characteristic behaviours in a wide range of settings.
This will never be a guarantee---the behaviour might change in settings that appear to be very similar---but by operationalizing the things we want from approximate Bayesian methods we can get firmer purchase on the idea of what makes an approximation `good' than we can manage otherwise.

This thesis proposes two examples of a characteristic behaviour we want from Bayesian approximations.
First, the ability to sequentially update approximate posteriors in light of new information.
Second, the ability of an approximate posterior to represent uncertainty at a parameter-level in a way that lets us know what information would reduce our uncertainty the most.
This points towards decision-problem families of continual learning and active learning respectively (although other decision-problem families might be just as suitable for evaluating these characteristic behaviours).

\section{Approximating Distributions Suit Different Architectures}
The first example of the complicated interactions between architecture and approximation method which we explore (\cref{chp:parameterization}) focuses on methods that pick an approximating distribution over a parametric model.
Rather than allowing our inference to give us a fully general distribution, one often insists on some simplifying constraints: perhaps our parameters must all follow Gaussian distributions \citep{mackayPractical1992} or be independent of each other \citep{hintonKeeping1993}.
Many authors discuss the restrictiveness of these assumptions in absolute terms (including \citet{mackayPractical1992}).
But this misses the interaction between these approximating distributions and the ways that they are \textit{used}.

Imagine, for example, how limited a uniform distribution is as a way to express a posterior distribution.
But by being slightly clever about how we \textit{use} these uniform distributions, we can easily express a Gaussian distribution \citep{boxNote1958}.
In an analogous way, though much more complicated, a neural network which is parameterized with very simple approximating distributions can express very complex distributions over \textit{predictions}.
Only three arbitrarily wide layers are required to allow a multi-layer perceptron to express arbitrary predictive distributions, under otherwise mild assumptions.
In practice, methods including variational inference \citep{jordanIntroduction1999} and mean-field-ammortized stochastic-gradient Markov Chain Monte Carlo \citep{maddoxSimple2019} seem to be remarkably unaffected by the mean-field assumption, and this may be because they find good posterior predictive distributions from a suitable mode of the parameter posterior.

As a result, existing efforts to enrich the approximating distributions over parameters may be misguided for certain tasks (in cases where the actual distributions over parameters \textit{do} matter, such as transfer learning \citep{shwartz-zivPreTrain2022} or continual learning \citep{ritterScalable2018}, this is not true).
In practice a simple approximating distribution in a complicated network is often easier to work with, and computationally cheaper, than a complicated distribution in a simple parametric model.

\section{Optimizers Suit Different Approximating Distributions}
The second example considered in this thesis (\cref{chp:optimization}) of unexpected difficulties when understanding how `good' a Bayesian approximation comes from the interaction between the optimizer and approximating distribution in stochastic variational inference in neural networks.
Where the architectures used in modern neural networks suit independent Gaussian distributions over parameters very well, the most common choices of optimizer may not.
Stochastic gradient-based optimizers can be used effectively for variational inference in Bayesian neural networks with independent Gaussian approximating distributions \citep{blundellWeight2015}.
But researchers have found difficulties in using this approach in large neural networks without applying a number of tweaks which undermine the Bayesian principles underlying the methods.

Part of the reason for this lies in the strange sampling properties of independent Gaussian random variables in high-dimensional spaces.
In high-dimensions, with very high probability samples come from a hypersphere `soap-bubble'.
This dispersion increases the variance of gradient estimators which are based on samples from the approximating distribution.
This can be corrected by using a slightly modified approximating distribution, a Radial-Gaussian distribution, which does not demonstrate the `soap-bubble' pathology.

\section{How Good Is My Bayesian Approximation?}

Until now, we have loosely helped ourselves to notions of one approximate Bayesian posterior being `better' than another, but this idea is hard to pin down.
In practice, researchers often use metrics like accuracy or log-likelihood on a held-out test set as a way to compare the performance of approximate Bayesian models.
This choice has some principled motivations from modelling a Bayesian decision problem \citep{keyBayesian1999}.
But even if we wanted to use the marginal likelihood instead, as suggested by \citep{mackayBayesian1992}, this would be intractable.
An extension of the approach of \citet{keyBayesian1999} is to select a number of interesting applications that test parts of Bayesian approximations that we value, and examine the expected utility of the approximate posterior.

This approach shares motivations with loss-calibrated (approximate) inference \citep{lacoste-julienApproximate2011a, cobbLossCalibrated2018}.
Loss-calibrated inference aims to produce posterior distributions that are tailored to a specific utility function.
We are proposing using the degree to which an approximate posterior has high utility on a task that exemplifies a characteric behaviour as an \textit{evaluation} of good approximate inference.
That is, we are not just trying to find approximate posteriors that maximize expected utility, we are trying to identify specific applications where the goal of maximizing expected utility will require achieving the desired characteristic behaviours.

Two such applications are continual learning and active learning.
Continual learning, in a Bayesian fashion \citep{nguyenVariational2018}, involves using a Bayesian model to compress data observed up to a time, and then to sequentially update this model as new data arrives.
In true Bayesian inference, the prior can be updated sequentially with each data point in any order and reach the same posterior distribution.
Performing well at sequential learning therefore tests an approximation's ability to capture this aspect of Bayesian inference, which is at least partly a measure of the approximation's success compressing the available data.
Active learning, in contrast, involves picking which unlabeled data points should be labelled in order to learn as effectively as possible.
In a Bayesian setting, this problem can be interpreted as estimating expected information gain \citep{mackayInformationBased1992}.
Good active learning performance, therefore, can be interpreted as showing that an approximate Bayesian model is able to identify places where the uncertainty is most capable of being reduced by new information.

However, in both of these cases, the application raises considerable difficulty interpreting performance measures.
In the case of continual learning, deficiencies in standard benchmarks have hidden the failures of Bayesian methods.
Desiderata can be identified for continual learning evaluations that are both more suited to applications and better measures of Bayesian approximation.
For active learning, performance measures are complicated by the fact that active learning in practice relies on implicit biases.
The most useful approximate posteriors during active learning are the ones that encode these implicit biases well, rather than the ones that come closest to the true posterior distribution of the given model.
By removing the bias, we can flip the ordering that active learning evaluations assign to different approximate posteriors.
Instead, we may be able to use active testing \citep{kossenActive2021} as an evaluation, which is not dependent on implicit biases in the same way (though active testing may face its own challenges).

\section{Overview}

In \cref{chp:challenges}, I consider the challenges for approximate Bayesian inference.
I review the motivations for Bayesian inference in exact settings as well as the literature on various approximations.
Immediately, many of the ways in which `good' exact Bayesian inference was historically motivated become inapplicable.
Measuring quality in terms of the expected utility on representative decision-problems for desirable characteristic behaviours of an approximate posterior remains a sensible option.

In \cref{chp:parameterization}, I consider how the architecture of a neural network affects the family of predictive distributions which a parametric approximating distribution can represent.
Using deeper neural networks can allow even simple mean-field Gaussian approximating distributions over parameters to approximate arbitrarily complicated predictive posterior distributions to arbitrary precision as they become arbitrarily wide.
I investigate this phenomenon from several analytical angles, and suggest that in practice approximation methods seem to uncover these successful modes.

In \cref{chp:optimization}, I consider how parametric approximating distributions can affect optimization procedures used in some approaches to approximate Bayesian inference.
Because the approximating distribution affects the sampling properties of the optimization objective, we can prefer approximating distributions which lead to low-variance Monte Carlo estimators of the loss, even if these were a worse fit to the true posterior than the global optimum for another approximating distribution.
In particular, I show how avoiding pathologies of mean-field Gaussian distributions in high-dimensional spaces can improve variational approximations by reducing the variance of Monte Carlo loss estimators.

In \cref{chp:evaluation} we return to the use of expected posterior utility for model evaluation.
Active learning and continual learning each test different and important aspects of Bayesian approximations.
However, when used naively, each can give a very misleading picture of the effectiveness of the Bayesian approximations which are used.
For each, I examine some of these failures and suggest ways to use these applications to get a better understanding of the quality of the approximate posterior.

\ifSubfilesClassLoaded{
\bibliographystyle{plainnat}
\bibliography{thesis_references}}{}
\end{document}
\clearpage

\section*{Notation}
\label{notation}
\subsection*{General notation}
\begin{tabular}{C{2cm}l}
    A &proposition\\
    $\rx$ &\text{scalar random variable}\\
    $\rvx$ &\text{vector random variable}\\
    $\rmX$ &\text{matrix random variable}\\
    $x$ &\text{scalar}\\
    $\vx$ &\text{vector}\\
    $\mX$ &\text{matrix}
\end{tabular}

\subsection*{Machine learning notation}
\begin{tabular}{C{2cm}l}
    $\data$ &data\\
    $\vx_i$ &$i$'th input data point (generally vector)\\
    $\vy_i$, $y_i$ &$i$'th supervision data point (vector, scalar)\\
    $\vpred_i$, $\pred_i$ &$i$'th prediction of a model (vector, scalar)\\
    $\Ls(\vy_i, \vpred_i)$ &Loss for the $i$'th prediction\\
\end{tabular}

\subsection*{Neural network notation}
\begin{tabular}{C{2cm}l}
    $\vtheta$ &\text{neural network parameters}\\
    $f_\vtheta$ &neural network parameterized by $\vtheta$\\
\end{tabular}

\subsection*{Other notation}
\begin{tabular}{C{2cm}l}
    $\gaussian$ & Gaussian distribution\\
    $\KLdiv{\cdot}{\cdot}$ & Kullback-Leibler divergence\\
    $\E{p}{\rx}$ & Expectation of $\rx$ under distribution $p$\\
    $\bigO(\cdot)$ & Big-$O$\\
\end{tabular}

\subsection*{Acronyms}
\begin{tabular}{C{2cm}l}
    NN & neural network\\
    BNN & Bayesian neural network\\
    VI & variational inference\\
    MFVI & mean-field variational inference\\
    FCVI & full-covariance variational inference\\
    ELBO & evidence lower bound\\
    GP & Gaussian process\\
    MC & Monte Carlo\\
    MCMC & Markov chain Monte Carlo\\
    i.i.d.& independently and identically distributed\\
    p.d.f. & probability density function\\
    c.d.f. & cumulative distribution function
\end{tabular}

\chapter{Challenges for Evaluating Approximate Bayesian Inference in Neural Networks}
\label{chp:challenges}

How can we know if we are doing Bayesian inference well?
In some sense, the answer is trivial.
All we need to do is to select a model and prior that express our subjective beliefs and update it following the laws of probability.

In practice, this approach struggles.
We select models and priors that we know how to express and choose them partly for computational simplicity.
For complicated models like neural networks, we do not know what the stated priors actually mean---let alone whether they match our beliefs.
In addition, following the laws of probability is computationally intractable for most models.

This creates the challenge which this thesis addresses.
Can we construct evaluations which give us insight into how well our machine learning systems are pursuing Bayesian goals?
Falling short of idealised Bayesian inference is guaranteed---our task is to define a spectrum of degrees of Bayesian success.
Without making this precise, it is hard to know whether research in approximate Bayesian methods is making progress.

The first two sections of this chapter should be interpreted as an opinionated review of prior work.
The first considers the foundational motivations for an unrealisable Bayesian ideal about finding the `right' posterior distribution.
The second discusses obstacles to using this idealised approach in practice.
In the third section, I explain why aiming for an approximate posterior that is `close' to the true posterior is the wrong way to evaluate the quality of approximate inference.

\section{Ideal Bayesian Inference}
\label{s:challenges:Ideal_bayesian_inference}
Bayesian inference is a formal system for reasoning about beliefs.
This makes it a promising candidate for understanding machine learning.
With Bayesian inference, we represent the strength of our belief in various propositions as probabilities, and then adjust beliefs in light of new evidence by following the laws of probability.
Why would we want idealised Bayesian inference?

\subsection{Axioms for Reasoning}
\label{ss:challenges:axioms_for_reasoning}

\citet{jaynesProbability2003} develops an argument by \citet{coxAlgebra1961} that Bayesian inference is not just \emph{a} way to reason about information, it is the \emph{only} rational way to do so.
They prove that every system of reasoning that obeys some assumptions follows the laws of probability, which in turn entail the `updating' procedure commonly associated with Bayesian inference.
I sketch the assumptions and character of their argument, deliberately leaving the details aside.

Assume that reasoning is a procedure that lets us adapt the strength of our belief about the plausibility of claims as we accumulate evidence.
Assume further that we can represent degrees of belief using numbers, where more plausible claims get bigger numbers and infinitesimally more plausible things get infinitesimally bigger numbers.
These assumptions are not completely trivial, but they come easily to machine learning practitioners who are in the habit of assigning numbers to messy things.

Add some modest rules about how these degrees of belief behave.
Suppose I think I saw you in the supermarket yesterday.
But then you tell me that you were in a different city giving a talk.
This makes it less likely that I saw you in the supermarket, but has no bearing at all on whether the sun will rise tomorrow.
Our modest rules entail two things.
First, on learning you were in a different city, I ought to think less likely the conjunction that I saw you in the supermarket and that the sun will rise.
Second, on learning you were in a different city, I should think it more likely that I did \emph{not} see you in the supermarket (since we know \textit{a priori} that you cannot be in two cities at the same time).
This can be easily formalized and is plausible.

Last, we add some consistency conditions.
We assume that if any conclusion can be reached in multiple ways, all procedures of reasoning reach the same conclusion.
We assume that we always use all our evidence.
We assume that equivalent states of knowledge get assigned the same number.\footnote{This final assumption seems problematic to me. I have not seen a grounding of equivalent states of knowledge which seems non-circular and I believe this mostly reprises the same problems Laplace had in using symmetry to ground classical probabilities.}

In practice, none of these conditions describe how people actually reason.
Moreover, it seems likely that no physically implementable system could satisfy all these conditions.
Any definition of `rationality' which is so demanding as to be impossible to implement does not correspond to natural use of the term.
But there is a sense in which they might represent an ideal for how one ought to reason.
This raises the question of how we ought to evaluate \textit{degrees} of rationality along a spectrum where a system that follows these axioms is at the unattainable optimum.

\subsection{Dutch Books}
\label{ss:challenges:dutch_books}

There is another way to give teeth to ideal Bayesian inference.
A family of results \citep{ramseyTruth1926,definettiForesight1937}, called Dutch Book arguments, operationalize the mistake of not following these probability rules.
These arguments demonstrate that any system of reasoning which assigns numbers to degrees of belief about a proposition, which does not follow the laws of probability regarding the relationship between those numbers, can be taken advantage of.
They show that if you follow such a `non-Bayesian' system of reasoning, it is possible to construct bets such that if you set the price, and I am allowed to choose whether to buy or sell the bet, you are guaranteed to lose money no matter the outcome \citep{jacksonModified1976}.
Moreover, the converse holds and following the laws of probability protects you from having a Dutch book formed against you \citep{lehmanConfirmation1955, kemenyFair1955}

Once again, we can quibble with the strength of these arguments.
It's a narrow sense of `rationality', maybe.
Perhaps it is just as rational to opt out, wherever possible, of letting strangers pick which side of a bet they are on after you declare a price!

But equally, the argument should give us pause.
A Dutch book feels like a violation of the principles of our reasoning system in a way that can be separated from the pragmatics of forming bets with predatory strangers \citep{skyrmsCoherence1987}.

More than that, if we, as machine learning researchers, want to develop mathematical systems of reasoning that are \emph{not} robust to this sort of betting game, but we want to deploy our systems in practice, then we should make very sure that the decision-problems we work on are not isomorphic to this sort of betting game.
That is, when I roll out a face-recognition algorithm I ought to be sure that none of its users are doing something that is similar enough at a conceptual level that the same results apply.
This is especially true when machine learning systems might encounter interested adversaries who might construct situations that create a Dutch Book-style dynamic.
To save ourselves the difficulty of constructing our decision problems to avoid this, and proving that we have done so (if this is possible, which I am unsure of), we might find our lives greatly simplified by following the rules of probability.

\subsection{Bayes and Probability}
\label{ss:challenges:bayes_and_probability}

Suppose that degrees of belief are the sorts of things that you can apply the laws of probability to, and that we do so.
How does this relate to Bayesian inference?

Bayes' theorem follows from the product rule of probability and commutativity of conjunction.
Consider some claims about the world, $A$ and $B$.
For example, $A$ might be true if and only if I saw you at the supermarket yesterday.
We assign a credence to this claim, representing its degree of plausibility, $P(A)$.
We further define $P(A \mid B)$ to be the plausibility that $A$ is true given the evidence provided by $B$.
The product rule states that $P(AB) = P(A\mid B)P(B)$, where $AB$ is used to stand for the probability that both $A$ and $B$ are true at once.
The commutativity of probabilities states that $P(AB) = P(BA)$ (after all, $A$ and $B$ are true precisely when $B$ and $A$ are).
Then
\begin{align*}
    P(B\mid A)P(A) = P(BA) &= P(AB) = P(A\mid B)P(B) \\
    P(B\mid A)P(A) &= P(A\mid B)P(B) \\
    \tag{Bayes' Theorem}
    P(B\mid A) &= \frac{P(A\mid B)P(B)}{P(A)}. \label{eq:challenges:bayes_theorem}
\end{align*}
There are multiple sets of laws of probability, such as those by \citet{kolmogorovFoundations1950} and \citet{coxAlgebra1961}.
While they disagree about some subtleties, especially involving infinitesimals, they agree about these rules, and so they agree about Bayes' theorem.
Although abstract at this point, it offers a powerful rule for changing your mind about $B$ when you learn about $A$.

A few notational remarks.
Propositions are named with capital letters $A$, $B$, etc.
In work on probability it is common to name random variables with capital letters.
However, because it can be important in machine learning to distinguish scalar, vector, and matrix quantities, we adopt the practice of denoting scalar, vector, and matrix random variables using the letters like $\rx$, $\rvx$, and $\rmX$.
Their realizations are denoted $x$, $\vx$, and $\mX$ and are drawn from spaces which are defined where required but are typically calligraphic capitals (e.g., $\gX$).
Some special variables depart from these conventions for sake of consistency with other work, which will be noted as it arises.

\subsection{Bayes for Machine Learning}
\label{ss:challenges:bayes_for_machine_learning}

Earlier, we chose to view machine learning as a formal system for reasoning about beliefs.
Moreover, we wanted a way to move from observations to generalizations about what one might observe.

One way to describe those generalizations is to write down a parametric model.
We will be mostly concerned with neural networks, which are a particularly versatile family of non-linear parametric models.
\begin{align*}
    \vx \in \gX &: \text{ input}\\
    \vy \in \gY &: \text{ output}\\
    f_\vtheta : \gX \rightarrow \gY &: \text{ parametric model (sometimes just written $f$)}\\
    \vtheta &: \text{ parameters of the parametric model}
\end{align*}
In order to generalize beyond our observations $\data \coloneqq \{(\vx_i, \vy_i) \mid 0 \leq i < N\}$ we would like to infer a degree of plausibility for each possible value of the parameters in the context of that parametric model's structure.
In the notation of probability this may be written $p(\vtheta \mid \data; f)$ and by \ref{eq:challenges:bayes_theorem}
\begin{equation}
    p(\vtheta \mid \data; f) = \frac{\overbrace{p(\data \mid \vtheta; f)}^{\text{likelihood}}\overbrace{p(\vtheta; f)}^{\text{prior}}}{\underbrace{p(\data)}_{\text{marginal likelihood}}},\label{eq:challenges:bayesian_parameter_inference}
\end{equation}
where we have used the fact that the probability of the observations is unaffected by the parametric model's structure.
In what follows, I will mostly suppress the context of the function, as this is often taken for granted.
It is, however, very important when we consider model-selection and approximation, which enters into \cref{s:challenges:Common_approximations_to_bayesian_inference}.

Within the context of that parametric model, if we could state the three terms on the right-hand-side we could state the plausibility of any parameter value.
The plausibility of any output, $\vy$, given some input, $\vx$, can then be found by integrating over all possible parameter values
\begin{equation}
    p(\vy \mid \vx, \data) = \int p(\vy \mid \vx, \vtheta)p(\vtheta \mid \data) \,d\vtheta.\label{eq:challenges:bayesian_predictive_distribution}
\end{equation}
For many useful parametric models, the likelihood is provided directly by the model.
In principle, the prior can be written down \textit{a priori}---relying only on our subjective beliefs when we frame the problem.
The marginal likelihood is challenging because it requires integrating over $\vtheta$.

In this way, Bayes' rule offers a principled way to capture what we know about the parameters of a model given some observations.
Moreover, when the true observation-generating process can be expressed by $f_\vtheta$, in the limit of infinite observations our beliefs about $\vtheta$ will converge to reality \cite{doobApplication1949}. 
Suppose the data are generated by a process which is exactly a parametric likelihood function $f_{\vtheta^*}$.
Then, in the limit of infinite data, $N$, when we approximate $\vtheta^*$ by $\vtheta$, the Bernstein-von Mises theorem can be used to show that the posterior $p(\vtheta \mid \data)$ tends towards the normal distribution $\gaussian(\vtheta^*, \bigO(\frac{1}{N}))$.
This offers reassurance that Bayesian inference ought to approach the truth as we gather more data.

\subsection{Bayesian Model Comparison}
\label{ss:challenges:bayesian_model_comparison}

Another strength of Bayesian inference is that it guides which function one ought to select.
Suppose, for example, that we are choosing between $f$ and $g$.
We might denote the probability that $f$ is the true data-generating process by $p(f)$.

Conditioning on our observations then gives us the degree of plausibility of the model in light of our observations, $p(f \mid \data)$.
By \ref{eq:challenges:bayes_theorem} this can be written
\begin{equation}
    p(f \mid \data) = \frac{p(\data \mid f)p(f)}{p(\data)}.
\end{equation}
Estimating $p(f \mid \data)$ is difficult because both the model prior and the marginal likelihood are not obvious.
However, when choosing between two models about which we are equally ignorant, we can simplify the calculation \citep{mackayBayesian1992} by estimating the ratio of the probabilities of the two models
\begin{align}
    \frac{p(f \mid \data)}{p(g \mid \data)} &= \frac{p(\data \mid f)\cancel{p(f)}}{\cancel{p(\data)}}\frac{\cancel{p(\data)}}{p(\data \mid g)\cancel{p(g)}}\nonumber\\
    &= \frac{p(\data \mid f)}{p(\data \mid g)},
\end{align}
where we have assumed that the two models under consideration are equi-probable \textit{a priori}.

This is a useful trick in principle, because it reduces model selection to the problem of computing the marginal likelihood of the data given the model.
However, in practice even this can be difficult to compute for models of interest.
The marginal likelihood ratio is also not valid for model comparison in cases where we do not, in fact, want to assume that the two models are equally plausible \textit{a priori} and does not necessarily predict generalization error \citep{mackayBayesian1992, lotfiBayesian2022}.

\section{Common Approximations to Bayesian Inference}
\label{s:challenges:Common_approximations_to_bayesian_inference}

So far, we have assumed a best-case scenario setting aside pragmatic concerns.
Unfortunately almost all practical settings require substantial approximations.
These approximations can be divided into approximations in defining the problem setting (choice of likelihood function and prior) and computational approximations (computing the posterior).
However, these choices are linked.
Chapters \ref{chp:parameterization} and \ref{chp:optimization} focus on specific challenges created by the interactions of these two categories of approximations.
Chapter \ref{chp:evaluation} considers strategies for evaluation that are helpful when approximations have been made of either kind.

\subsection{Approximating the Model}
\label{ss:challenges:model_misspecification}

We never know the structure of the true data-generating process with certainty.
This means that, especially when we restrict our models so that we are able to write them down and compute them, we cannot guarantee that our model is capable of expressing the true process.
Instead, we would like to pick an approximation which is not too restrictive.

Fortunately, an approximate extension of the Bernstein-von Mises theorem applies even when the model cannot express the true data-generating process \citep{kleijnBernsteinVonMises2012}.
When the parameters which we are optimizing belong to the wrong parametric model, $g_{\vphi}$, nevertheless the posterior $p(\vphi \mid \data)$ converges to the Gaussian $\gaussian(\vphi^*, \bigO(\frac{1}{N}))$ such that
\begin{equation}
    g_{\vphi^*} = \argmin_{\vphi} \KLdiv{f_{\vtheta^*}}{g_{\vphi}}.
\end{equation}
When our parametric model is over-parameterized and expressive, this suggests that the error introduced due to the model failing to express the true process is unlikely to be large.
Partly for this reason, this thesis focuses mostly on computational approximations and the interactions between computational approximations and modeling approximations.

There are also some efforts to address model-misspecification more directly.
For example, \citet{grunwaldSafe2011} makes use of a `safe' maximum \textit{a posteriori} (MAP) estimator which is designed to handle misspecified settings well.
While there may be some promise in these sorts of approaches, they depart sufficiently from the basic Bayesian procedure to require separate treatment.

Meanwhile, \citet{keyBayesian1999} propose what is effectively computing the expected utility of a Bayesian posterior as a solution to model-misspecification, which they motivate as a solution to a properly framed Bayesian decision problem.
They additionally argue for the value of `generic' utility functions like the log-likelihood as a proxy for application-specific utilities.
Although this argument has received little vindication from theoreticians, it has received the ultimate compliment of being more-or-less the only way that modern Bayesian deep learning papers evaluate their approximate posteriors.
While they argue for this in the context of model-misspecification, I will return to this solution in more detail when we consider evaluating \emph{approximate} Bayesian inference.

\subsection{Parametric Approximate Bayesian Inference}
\label{ss:challenges:parametric_approximate_bayesian_inference}

Neural networks \citep{rosenblattPerceptron1958} are a family of parametric models which apply both linear and non-linear `layers' to an input.
Their architectures are diverse, and will be discussed in detail within later chapters where important.
In deterministic neural networks, the values of the parameters can be `learned', for example through backpropagation \citep{linnainmaarepresentation1970}.

From a Bayesian perspective, we can acknowledge subjective uncertainty over the parameters of a neural network.
Given a prior, it is then possible to infer the parameter distribution over neural network weights.
Bayesian neural networks (BNNs) are networks whose weights follow a posterior distribution inferred in this way \citep{tishbyConsistent1989,buntineBayesian1991}, or whose weights are approximately distributed as a posterior distribution inferred in this way.

There is no discussion in the literature of what sort of approximation suffices for a parametric model to `count' as approximately Bayesian.
This very question might reasonably be regarded as so vague as to be meaningless.
We will return to it later but only answer it partially---by identifying some characteristically Bayesian reasoning properties and proposing tests for their presence.

If a neural network, $f$, has parameters, $\vtheta$, then we can infer a posterior distribution over parameters given some observed data, $p(\vtheta \mid \data)$ following \cref{eq:challenges:bayesian_parameter_inference}.
In order to do this, we must be able to fix a prior, $p(\vtheta)$, and define the likelihood function $p(\data \mid \vtheta)$, both of which are generally feasible.

However, the marginal likelihood, $p(\data)$, is intractable.
It is constant w.r.t.\ $\vtheta$, so we can find $p(\vtheta \mid \data) \propto p(\data \mid \vtheta)p(\vtheta)$ as an unnormalized distribution.
But when we make predictions following \cref{eq:challenges:bayesian_predictive_distribution}, integrating over this unnormalized distribution is also intractable.

\subsubsection{Variational Inference}
\label{ss:challenges:variational_inference}
One alternative is variational inference \citep{jordanIntroduction1999}.
Instead of computing the posterior distribution over parameters, we instead define an approximating distribution, $q(\vtheta)$, and minimize a measurement of difference between this and the true posterior.
`Variational inference' is what we call this procedure when the distance is the Kullback-Leibler (KL) divergence between the approximating distribution and the true posterior, $\KLdiv{q(\vtheta)}{p(\vtheta \mid \data)}$.

In fact, this KL-divergence is not directly tractable either because we do not have access to the posterior distribution.
Instead, we optimize a bound.
\citet{barberEnsemble1998} show in the context of neural networks that:
\begin{equation}
    \label{eq:challenges:kl_bound}
    \KLdiv{q(\vtheta)}{p(\vtheta \mid \data)} = \log p(\data) - \textrm{ELBO}(q, p),
\end{equation}
where the evidence lower bound (ELBO) is defined
\begin{equation}
    \label{eq:challenges:ELBO_definition}
    \textrm{ELBO}(q,p) = \E{q(\vtheta)}{\log p(\data \mid \vtheta)} - \KLdiv{q(\vtheta)}{p(\vtheta)}.
\end{equation}
Because the marginal likelihood is constant w.r.t.\ $\vtheta$, minimizing the negative ELBO is equivalent to minimizing the KL-divergence between the approximating distribution and the true posterior.
This recovers the objective from earlier work based on minimum description lengths \citep{hintonKeeping1993}.

In neural networks, we can maximize the ELBO using gradient-based optimization of a Monte Carlo approximation of the integrals \citep{gravesPractical2011,blundellWeight2015}.
The integration approximation can be refined, in certain settings, with further approximations \citep{kingmaVariational2015,wenFlipout2018,khanFast2018,wuDeterministic2019}.
At the same time, while \citet{blundellWeight2015} use a fully factored Gaussian approximating distribution, other approximating distributions can be chosen.
For example, partially or hierarchically factorised distributions are preferable for some applications (e.g., \citep{kesslerHierarchical2021}).
\citet{galDropout2015} propose an even simpler approximating distribution with Bernoulli noise multiplied into hidden units.
On the other hand, one could use more complicated distributions which also capture correlations between weights \citep{louizosStructured2016} or apply normalizing flows \citep{rezendeVariational2015}.

 \subsubsection{Laplace Approximation}
\label{ss:challenges:laplace_approximation}
Laplace's method offers another approximation for the posterior of a neural network \citep{denkerTransforming1991,mackayPractical1992}.
As before, we choose an approximating distribution, often assumed Gaussian with independent parameters.
We then fit a Gaussian distribution centred on the maximum-a-posteriori parameter values.

First, we maximize directly the unnormalized posterior probability density over the parameters, $p(\vtheta \mid \data) \propto p(\data \mid \vtheta)p(\vtheta)$.
In the case of an independent Gaussian prior, this is just optimization w.r.t.\ the negative log-likelihood loss with an $L_2$-norm regularization \citep{goodfellowDeep2016}.
Second, we find the Hessian of the loss with respect to parameter-spaces
\begin{equation}
    H_{ij} = \frac{\partial^2 \Ls}{\partial \theta_i \theta_j}.
\end{equation}
Assuming that the distribution around the maximum is approximately an independent Gaussian, the variance in each dimension is given by the diagonal term of the Hessian.
The independence assumption may be fully relaxed, or partially relaxed---for example, using matrix-variate Gaussian structured covariance approximations \citep{ritterScalable2018}.

\subsubsection{Markov Chain Monte Carlo}
\label{ss:challenges:markov_chain_monte_carlo}
Markov chain Monte Carlo (MCMC) methods can be used to draw samples from the approximate posterior distribution of neural networks \citep{nealBayesian1993}.
\citet[Chapter 32]{mackayInformation2003} provides an excellent overview of MCMC methods as a whole.
MCMC methods have the advantage that, in the limit of infinite samples, they draw samples from the true posterior distribution.
In particular, this is likely to be an advantage when true posterior distributions are very unlike a Gaussian (say).
For example, a multi-modal posterior might in principle be much better approximated by MCMC.
However, they suffer from several disadvantages.
\begin{itemize}
    \item Although they give provably good samples given sufficient steps, there is no way to know for sure that sufficient steps have been taken.
    \item For multi-modal distributions, it can take a very long time to `hop' from one mode to another.
    \item Especially in high-dimensional spaces, exploration can be slow.
    \item The outputs are individual posterior samples. In deployment, these must be separately stored and loaded in and out of memory to compute \cref{eq:challenges:bayesian_predictive_distribution}. This is very slow compared to drawing multiple samples from the same multivariate Gaussian distribution.
    \item Because the outputs are individual posterior samples, they are hard to use as a prior in sequential or online learning.
    \item Samples are locally correlated which can create problems in some applications (e.g., bounding estimator variance).
\end{itemize}
For neural networks specifically, a number of adaptations improve exploration as well as performance in the high-dimensional parameter spaces.
Stochastic gradient Markov chain Monte Carlo (SG-MCMC) takes advantage of noise in stochastic gradient-based optimizers to perform the Markov chain stepping efficiently \citep{wellingBayesian2011}.
Hamiltonian dynamics can be included \citep{chenStochastic2014} and adaptive step-size schemes can improve efficient exploration \citep{hoffmanNoUTurn2014}.
The drawback of memory and storage can also be addressed by amortizing the samples into a Gaussian distribution \citep{maddoxSimple2019}.

As a result, in recent years some work has attempted to approximate large neural network posterior distributions using Hamiltonian Monte Carlo \citep{izmailovWhat2021}.
Although experimental HMC samples cannot be equated with the true posterior distribution they represent the best strategy currently available for high-fidelity inspection of the true posterior distribution.

\subsubsection{Other Methods}
\label{ss:challenges:other_methods}
A number of other methods to approximate Bayesian inference in neural networks exist.
These include expectation propagation \citep{minkaFamily2001,jylankiExpectation2014}, probabilistic back-propagation \citep{hernandez-lobatoProbabilistic2015}, alpha-divergence minimization \citep{hernandez-lobatoBlackbox2016}, and sequential Monte Carlo \citep{doucetSequential2001}.
However, because these are not currently used to a significant degree, I do not cover them in detail here.
They are covered in more detail in other reviews of the literature for approximate Bayesian inference in neural networks \citep{galUncertainty2016,alquierApproximate2020}.

\section{Conceptual Challenges for Evaluating Approximate Bayes}
\label{s:challenges:conceptual_challenges_for_approximate_bayes}
\Cref{s:challenges:Ideal_bayesian_inference} discussed motivations for Bayesian inference through rationality considerations and Dutch books.
Unfortunately, these binary motivations do not give a way to express being more or less rational.

This is a problem for the approximate Bayesian.
Even if we assume that the true data-generating distribution is with the support our probabilistic model, our approximations mean that we do not expect to actually produce the true posterior distribution.
This means that we have not, in fact, followed the `rational' rules of probability.
Nor does it defend us against Dutch books.
This has been termed the `approximate inference conundrum' \citep{ghahramaniShould2008}.

It is natural to try to provide a partial answer by considering the distance between an approximate posterior and the true Bayesian posterior, in much the same way as variational inference is motivated.
However, unfortunately, there are a number of plausible candidates.
To name a few:
\begin{itemize}
    \item the forwards KL-divergence (expectation propagation);
    \item the backwards KL-divergence (variational inference);
    \item the Wasserstein distance between the approximate and true posterior;
    \item the maximum density difference between the approximate and true posterior.
\end{itemize}
And for each of these we could consider the distance in parameter-space, function-space, or predictive-space.
It is not clear to me, \textit{a priori}, that any of these is the `correct' sense of distance to the true posterior distribution.
Moreover, it is clearly true that the optimal parameter distribution on one of these distances will not be optimal on another.
For example, if the true posterior is multi-modal, the optimal approximations on the forwards and backwards KL-measures are radically dissimilar (one fits a single mode and the other covers them all).
This means that our inability to select a `correct' sense of distance also robs us of an ordering over approximate posterior quality.

It is my (unproven) belief that we will not be able to solve this problem without adding a few extra tools: a utility function and a data-distribution.
These allow us to frame our problems as Bayesian decision problems.
On this picture \citep{keyBayesian1999}, we care about the expected utility of the approximate posterior distribution for the problem of interest
\begin{equation}
    \E{}{U} = -\E{\vy, \vx \sim \pdata, \vtheta \sim q(\vtheta; \data)}{\int p(\vpred \mid \vx, \vtheta) \Ls(\vy, \vpred) \,d\vpred}.
\end{equation}
Here, we have specialized the utility, $U$, to be the negative model loss, $\Ls(\vy, \vpred)$, for an actual $\vy$ and a predicted $\vpred$.
This is appropriate for decision-rules based on predictive functions, which we will focus on.
The expectation is over both the data-distribution, $\pdata$, and the parameter distribution, $q(\vtheta ; \data)$ and represents the expected utility of our model for the problem we care about.
\citet{keyBayesian1999} proposed this expected utility as a solution to the model-misspecification problem.
In \cref{chp:evaluation} we will consider how this approach can be extended to a more general family of approximations including computational approximations.

A key assumption is that the actual decision problem we will face is in the context of data generated from the same distribution as that which $\data$ was drawn from.
This `distribution matching' assumption is more often associated with frequentism than Bayesian methods.
However, for a Bayesian who cares about anything remotely practical this distinction is a mistake.
In practice, it is never \textit{important} to infer a posterior distribution for parameters except in the hope of generalizing to some other data.
Even when we just want to explore data out of curiosity, it is our expectation that the data is related in some generalizable way to something else in the universe that makes the exploration meaningful.
So a Bayesian cannot escape some sort of assumption connecting the data which have been observed to some problem of significance.

A further assumption is that we can write down a loss or utility function that is what we actually care about.
A common choice, echoing \citet{keyBayesian1999} is the cross-entropy or negative log-likelihood loss which can supposedly stand in as a \textit{universal} utility function.
Note that almost all current work in approximate Bayesian inference in practice uses the cross-validation expected utility of the approximate posterior as a main assessment.

However, we also care about \textit{using} Bayesian approximations for various downstream tasks.
In my view, the philosophical commitment of a Bayesian approach is that probability distributions represent strengths of beliefs in states of affairs.
This means that a Bayesian approach should offer introspection and compression.
`Introspection' in the sense that our posterior distribution ought to offer guidance about which parts of our knowledge are most uncertain and, at least in principle, most amenable to correction in light of new observation.
`Compression' in the sense that Bayesian inference ought to support sequential inference by representing the total state of belief after incorporating some evidence.
The posterior from one set of observations can become a prior when approaching another dataset.
Both of these aspects appear within the axioms of constructive formulations of both objective and subjective Bayesianism.

As a result, in \cref{chp:evaluation} we will look at two applications that most ideally characterize these two aspects of Bayesian inference: active learning and continual learning.
It is difficult to structure these tasks in a way that genuinely tests the quality of the approximate posterior in some generalizable sense, rather than too narrowly evaluating unimportant details of the application.
However, some progress can be made.
In this way, a utility-centred framework for evaluating approximate posterior distributions can step past a reliance on purely general utility functions and start to evaluate different aspects of the Bayesian promise separately.


\ifSubfilesClassLoaded{
\bibliographystyle{plainnat}
\bibliography{thesis_references}}{}
\end{document}
\end{document}

\chapter{Network Parameterization Determines True Posterior and Approximation}
\label{chp:parameterization}

\newtheorem{lemma}{Lemma}
\newtheorem{theorem}{Theorem}
\newtheorem{proposition}{Proposition}
\newtheorem{remark}{Remark}
\newenvironment{proofsketch}{%
  \renewcommand{\proofname}{Proof Sketch}\proof}{\endproof}
\newenvironment{elaboration}{%
  \renewcommand{\proofname}{Elaboration}\proof}{\endproof}
\newcommand{\tp}{p(\vtheta \mid \data)}
\newcommand{\qf}{\hat{q}_{\textup{full}}(\vtheta)}
\newcommand{\qd}{\hat{q}_{\textup{diag}}(\vtheta)}


\textit{\textbf{Notational remark:} Because the proof of \cref{thm:uat} depends heavily on distinctions between random variables and realizations, but very little on distinctions between scalars/vectors/matrices, we adopt the standard notation from statistics that lower case letters are the realizations of upper case random variables, departing from the notation of other sections of the paper, such as the proofs of \cref{thm:matnorm} where the matrix structure is key.}
\vspace{6mm}

While performing approximate Bayesian inference in Bayesian neural networks (BNNs) researchers often make what they regard as a severe approximation.
This is variously called the `mean-field' or `diagonal' approximation.
It assumes that the approximate posterior distribution over each weight is independent of all the others and such that the distribution fully factorizes in the following way:
\begin{equation}
  q(\vtheta ; \data) = \prod q(\theta_i ; \data).
\end{equation}
This is called a diagonal approximation because the covariance matrix for the vector random variable $\vtheta$ is diagonal.
It is called mean-field by analogy to the mean-field solution to Ising models, in which local correlations between neighboring spins are ignored.
In particular, a further approximation is often made that $q(\theta_i; \data)$ is the Gaussian $\gaussian(\mu_i, \sigma_i^2)$.
This approximation has been widely applied in neural networks for example for variational inference \citep{hintonKeeping1993,gravesPractical2011,blundellWeight2015} or amortized SG-MCMC \citep{maddoxSimple2019}, although it is sometimes useful to make different factorisation assumptions based on hierarchical structures, for example \citet{kesslerHierarchical2021}.

Although they have often used the mean-field approximation, researchers have assumed that it is a severe limitation.
After all, while we might choose to restrict the \textit{approximate posterior} in this way, we rarely actually believe it to be true of the \textit{true posterior}.
In an influential paper on approximate Bayesian inference for neural networks \citep{mackayPractical1992}, David MacKay framed the problem for the field, writing
\begin{quotation}
   The diagonal approximation is no good because of the strong posterior correlations in the parameters.
\end{quotation}

This has motivated extensive exploration of approximate posteriors that explicitly model correlations between weights.
For example, for variational inference \citet{barberEnsemble1998} use a full-covariance Gaussian while later work has developed more efficient structured approximations \citep{louizosStructured2016,zhangNoisy2018,mishkinSLANG2019,ohRadial2019}.
Still further work has tried to significantly increase the expressive power of the approximate posterior distribution \citep{jaakkolaImproving1998,mnihNeural2014,rezendeVariational2015,louizosMultiplicative2017,sunFunctional2019}.
Similarly, for the Laplace approximation full- \citep{mackayPractical1992} and structure-covariance \citep{ritterScalable2018} have been employed and structured-covariance for amortized SG-MCMC \citep{maddoxSimple2019}.

In addition to these research efforts, \citet{foongExpressiveness2020} have identified pathologies in mean-field approximate posteriors for neural networks with a single hidden-layer used for regression, and have conjectured that these might exist in deeper models as well and might apply in other contexts.

This chapter interrogates the prevailing assumption that mean-field methods are overly restrictive.
The substance of MacKay's objection to using diagonal approximate posteriors seems to have been an argument something like this: empirically, there are strong correlations between parameters in the true posterior distribution.
If we want a good approximation, our approximate posterior distribution needs to be able to capture all the important properties of the true posterior, one of which is the presence of off-diagonal correlations.
So a diagonal distribution will make a low-quality approximate posterior.
Our investigation will mirror the steps of this argument:
\begin{itemize}
  \item Does the \textit{true} posterior distribution have strong correlations between the parameters?
  \item Do there exist approximate posterior distributions that are `good' even if they are mean-field?
  \item Do actual methods for approximate Bayesian inference uncover these `good' approximations?
\end{itemize}
In doing so, we will explore the ways in which neural network \textit{architectures} interact with the choice of approximating distribution and the ways in which an approximate posterior can be good or bad.

For the first question, we will provide empirical evidence that some modes of the true posterior for BNNs are approximately mean-field, especially in deeper fully-connected networks (while finding some evidence that more off-diagonal correlations are present in parts of the true posterior for deep residual convolutional networks).
This suggests that a mean-field approximate posterior might be a good approximation for some neural network architectures even directly in \textit{parameter-space}.
For the second, we will provide a proof that for fully-connected networks with at least two hidden-layers it is possible to approximate an arbitrary \textit{predictive} posterior distribution arbitrarily closely as the network becomes wider.
For many applications, we care far more about the predictive distribution than we do about the distributions over parameters, which makes this perspective more useful for considering the restrictiveness of approximation assumptions.
For the third, we will provide a theoretical construction demonstrating how fully-connected neural networks might in practice develop patterns that allow `good' mean-field approximate posteriors and provide empirical evidence suggesting that this might emerge in practice.
However, we do not prove or demonstrate that methods like variational inference in fact find these `nearly optimal' predictive distributions.
Despite this, there are things that a mean-field approximate posterior cannot capture, which makes most of our results contingent not only on the architecture under consideration but also the \textit{dataset} and \textit{problem}.

\begin{table}
  \centering
  \begin{tabular}{lll}
    \toprule
    & \multicolumn{2}{c}{Complexity} \\ \cmidrule(lr){2-3}
    & Time            & Parameter    \\
    \midrule
    Mean-field VI \citep{hintonKeeping1993}& $K^2$           & $K^2$        \\
    Full-covariance VI \citep{barberEnsemble1998}               & $K^{12}$        & $K^4$        \\
    Matrix-variate Gauss. \citep{louizosStructured2016} & $K^3$           & $K^2$        \\
    MVG-Inducing Point [ibid.] & $K^2 + P^3$     & $K^2$        \\
    Noisy KFAC \citep{zhangNoisy2018} & $K^3$           & $K^2$       \\
    SWAG \citep{maddoxSimple2019} & $K^2N$ & $K^2N$ \\
    \bottomrule
  \end{tabular}
  \caption[Time complexity of mean-field, full-covariance, and structured-covariance approximations]{Complexity of a forward pass in $K$---the number of hidden units for a square weight layer. Mean-field VI has better time complexity and avoids a numerically unstable matrix inversion. Inducing point approximations can help, but inducing dimension $P$ then becomes a bottleneck.
  The structured-covariance approximation in \citet{maddoxSimple2019} is in terms of $N$, the number of iterates used for computation, which can be \textit{much} smaller than $K$ (e.g., 20) but is not suitable for VI-based methods.}
  \label{tbl:parameterization:complexity}
\end{table}

This matters for three reasons.
First, the rejection of mean-field methods comes at a price.
Relatively faithful structured covariance methods have worse time complexity (see \cref{tbl:parameterization:complexity}).
Even efficient implementations take over twice as long to train an epoch as comparable mean-field approaches \citep{osawaPractical2019}.
In contrast parsimonious structured approximations such as the one used by \citet{maddoxSimple2019} can have a minimal impact on time and memory complexity (although it is not immediately clear how variational inference methods could incorporate this specific approximation).
Researchers may benefit from exploring the most computationally feasible end of the spectrum of covariance-expressiveness rather than seeking ever-richer approximations.
Second, getting to grips with the question of how suitable the mean-field approximation provides a valuable case study for understanding how to evaluate the quality of approximate posteriors generally.
Third, the theoretical case against mean-field approximations has to handle some puzzles.
Ordinary non-Bayesian neural networks are arguably an extreme case of the mean-field approximation and they often work very well in many applications.
Moreover, recent work has succeeded in building mean-field BNNs which perform quite well (e.g., \citet{khanFast2018, maddoxSimple2019, osawaPractical2019,wuDeterministic2019,farquharRadial2020}) or which perform well despite making even \textit{more} stringent assumptions (e.g., \citet{swiatkowskiktied2020}).

The work in this chapter suggests that if a large amount of computational effort is required to approximate complicated posterior distributions then for many applications those resources might be better spent on using larger models with simpler Bayesian approximations instead.

\section{True Posterior Correlations}
\label{s:parameterization:true_posterior_correlations}
Let's consider the most basic question: does the true posterior distribution have strong correlations between the parameters?
MacKay was working with very small neural networks with few parameters, and in this setting he did find evidence of correlations between weights.

These correlations might be less present in larger, deeper networks.
Indeed, in later sections we will give some reasons to suspect that after three weight-layers---or two layers of hidden units---posterior correlations will be less necessary.
We can explore posterior correlations in larger neural networks using Hamiltonian Monte Carlo (HMC), which is generally regarded as providing a higher quality of posterior sample than methods based on a parametric approximating distribution.
This `microscope' suggests that there are modes of the true posterior that are approximately mean-field.
We examine a truly full-covariance posterior, not even assuming that layers are independent of each other, unlike \citet{barberEnsemble1998} and most structured covariance approximations.

\subsection{Methodology}
\label{ss:parameterization:true_posterior_correlations:methodology}
We use No-U-turn HMC sampling \citep{hoffmanNoUTurn2014} to approximate the true posterior distribution, $\tp$.
\footnote{We use No-U-turn HMC in order to adaptively set the step-size and trajectory length in a way that is `fair' to the different architectures which are considered here (as opposed to picking settings on one depth or having to individually tune each).
There is a growing literature on more efficient sampling for neural networks, including methods based on semi-separable HMC \citep{zhangSemiSeparable2014,cobbSemiseparable2019} and symmetric splitting \citep{cobbScaling2021} which I did not use because public implementations were not available at the time of execution, but might have allowed better coverage of the posterior.}
We then aim to determine how much worse a mean-field approximation to these samples is than a full-covariance one.
To do this, we fit a full-covariance Gaussian distribution to the samples from the true posterior---$\qf$---and a Gaussian constrained to be fully-factorized---$\qd$.

  \begin{figure*}
        \centering
        \includegraphics[width=0.5\columnwidth]{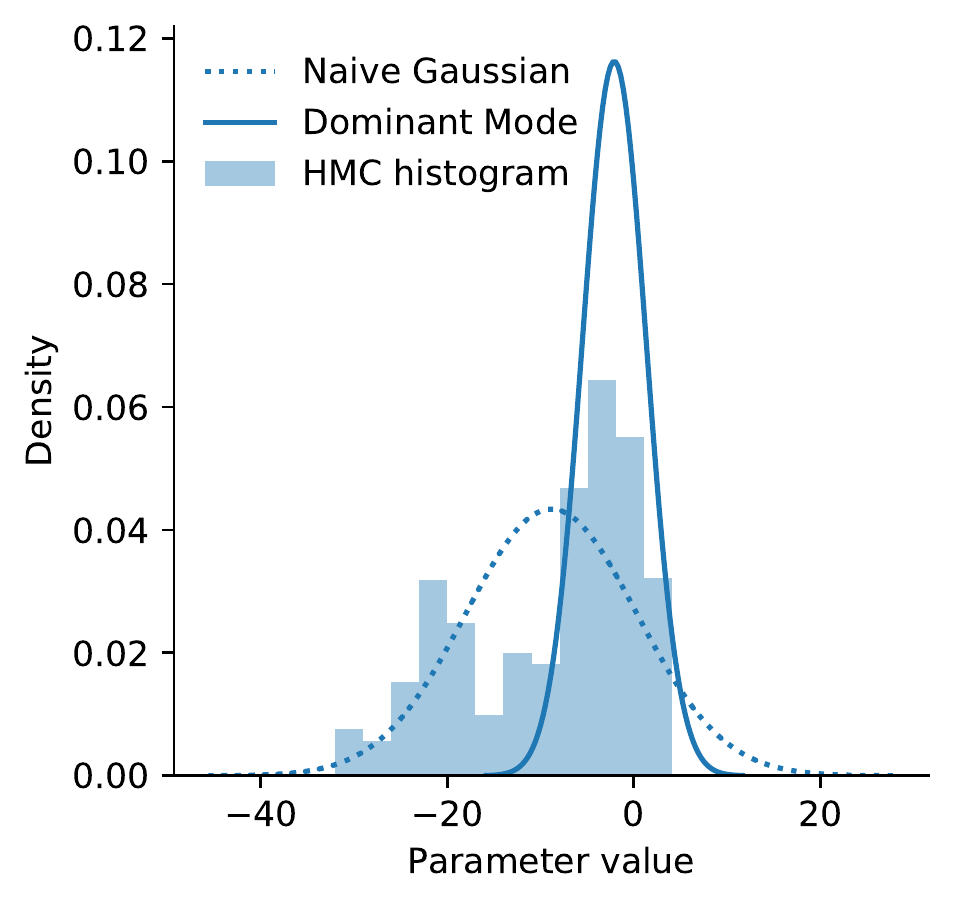}
    \caption[Parameter density histogram from HMC]{Example density for randomly chosen parameter from a ReLU network with three hidden layers. The HMC histogram is multimodal. If we picked the naive Gaussian fit, we would lie between the modes. By using a mixture model, we select the dominant mode, for which the Gaussian is a better fit.}
    \label{fig:mode_selection}
    \hfill
    
\end{figure*}
We deliberately do not aim to sample from multiple modes of the true posterior---we are assessing the quality of approximation to a single mode by a Gaussian distribution---and for this reason it is less problematic than normal that HMC can struggle to explore multiple separated modes in high-dimensional spaces.
A Gaussian is clearly unsuitable for approximating multiple modes at the same time, but a mixture of Gaussians can be used to approximate a multimodal posterior \citep{wilsonBayesian2020}.
Here, we fit a mixture of Gaussians to the HMC samples, using the Bayesian Information Criterion to select the number of mixture components, and then focus on the most highly weighted component for our analysis.
\footnote{The Bayesian Information Criterion was chosen for ease of implementation, but a more principled approach might have been to sample with a Dirichlet prior allowing shrinkage.}
See \cref{a:parameterization:hmc} for details.

We consider two measures of distance:
\begin{enumerate}
  \item \textbf{Wasserstein Distance: } We estimate the $L_2$-Wasserstein distance between samples from the true posterior and each Gaussian approximation. Define the Wasserstein Error:
  \begin{equation}
    E_W = W(\tp, \qd) - W(\tp, \qf).
  \end{equation}  
  If the true posterior is fully factorized, then $ E_W= 0$.
  The more harmful a fully-factorized assumption is to the approximate posterior, the larger $E_W$ will be.
  \item \textbf{KL-divergence: } We estimate the KL-divergence between the two Gaussian approximations. Define the KL Error:
  \begin{equation}
    E_{KL} = \KLdiv{\qf}{\qd}.
  \end{equation}
  This represents a worst-case information loss from using the diagonal Gaussian approximation rather than a full-covariance Gaussian, measured in nats (strictly, the infimum information loss under any possible discretization \citep{grayEntropy2011}).
  $E_{KL} = 0$ when the mode is naturally diagonal, and is larger the worse the approximation is.\footnote{$\tp$ is not directly involved in this error measure because KL-divergence does not follow the triangle inequality.}
\end{enumerate}
Either or both of these distances could increase even if either the ELBO or problem-specific loss function improved (c.f., discussion of the challenge for evaluating Bayesian approximations using the `distance' to the true posterior distribution in \cref{s:challenges:conceptual_challenges_for_approximate_bayes}).

Each point on the graph represents an average over 20 restarts (over 2.5 million model evaluations per point on the plot).
For all depths, we adjust the width so that the model has roughly 1,000 parameters and train on the binary classification `two-moons' task.
We report the sample test accuracies and acceptance rates in \cref{a:parameterization:hmc} and provide a full description of the method.

Note that we are only trying to establish how costly the \emph{mean-field} approximation is relative to full covariance, not how costly the \emph{Gaussian} approximation is.
Our later results, however, will show that any target predictive distribution can be approximated arbitrarily closely by a sufficiently large neural network with a Gaussian approximate posterior.

\subsection{Results}
\label{ss:parameterization:true_posterior_correlations:results}
\begin{figure}
  \begin{subfigure}[b]{0.48\textwidth}
      \centering
      \includegraphics[width=\textwidth]{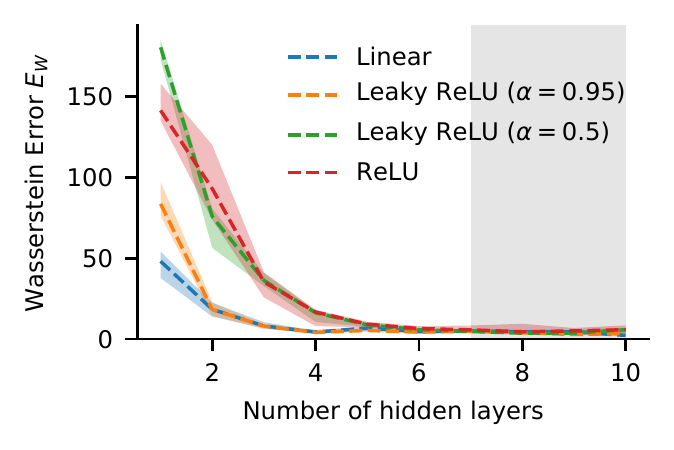}
      \caption{Wasserstein Error.}
      \label{fig:hmc_wasserstein}
  \end{subfigure}
  \hfill
  \begin{subfigure}[b]{0.48\textwidth}
    \centering
      \includegraphics[width=\textwidth]{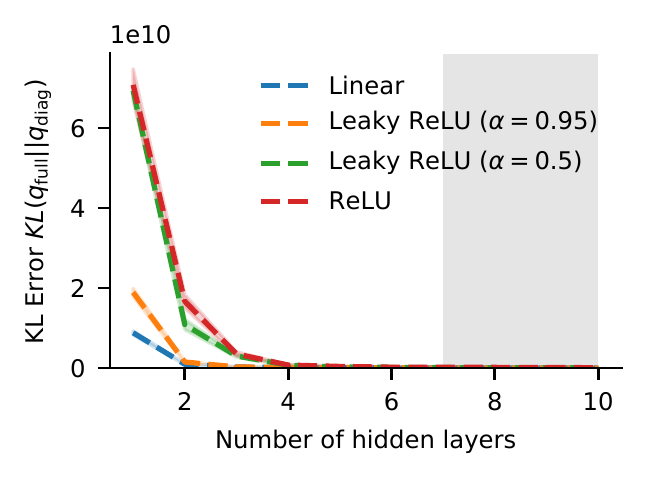}
      \caption{KL-divergence Error.}
      \label{fig:hmc_kl}
  \end{subfigure}
  \caption[`Cost' of the mean-field approximation relative to full-covariance for approximate Gaussian]{For all activations and both error measures, large error in shallow networks almost disappears with depth. All models have $\sim$1,000 parameters. Shaded depths: HMC samples in these depths show substantially lower test accuracy for the ReLU model (<95\% compared with ~99\%) so I would not generally depend on them (accuracies for the other models remain good though). See \cref{a:parameterization:hmc}.}
\end{figure}

In \cref{fig:hmc_wasserstein} we find that the Wasserstein Error introduced by the mean-field approximation is large in shallow fully-connected neural networks but falls rapidly as the models become deeper.\footnote{We do not examine HMC samples from more sophisticated network architectures with convolutions, skip connections, or attention. This is a limitation of the analysis in this section as well as the linear and piece-wise linear analysis in the next section. However, the empirical results in the final section examine networks with convolutions and skip connections like ResNets.} In \cref{fig:hmc_kl} we similarly show that the KL-divergence Error is large for shallow networks but rapidly decreases.
Although these models are small, this is very direct evidence that there are mean-field modes of the true posterior of a deeper Bayesian neural network.
That is, MacKay's assertion that there are ``strong posterior correlations in the parameters'' may not be true at all for larger neural networks than the ones he was looking at.

In all cases, this is true regardless of the activation we consider, or whether we use any activation at all.
Indeed, we find that a non-linear model with a very shallow non-linearity (LeakyReLU with $\alpha=0.95$) behaves very much like a deep linear model, while one with a sharper but still shallow non-linearity ($\alpha=0.5$) behaves much like a ReLU.

\subsection{Observational Investigation of Correlations in Large Neural Networks}

Performing HMC in the small setting considered above gives us the computational ability to observe the change in the information lost by the diagonal assumption as we vary the depth of the network while keeping the number of parameters approximately constant.
However, it may well be that the insights from this small setting would not extend to larger neural networks, or networks with non-fully-connected layers.

In order to examine this, I use the publicly available HMC samples provided by \citet{izmailovWhat2021} for ResNet-20 with filter response normalization \citep{singhFilter2020} trained on CIFAR10 and CIFAR100 and for a CNN-LSTM on the IMDB dataset.
These samples have the advantage of exploring much larger neural networks with more complicated structures involving convolution and recurrence.
The advantage of being able to examine these more complicated settings is partly offset by the fact that the computational scale means we are no longer able to vary the depth or complexity of the networks, so we have to rely purely on observations of the existing samples.

Similarly to the section above we fit a Gaussian mixture model to the combined HMC samples and examine the most highly weighted mode.
We only consider one chain at a time in order to avoid introducing correlations caused simply by the chains exploring different parts of the parameter space.
This is because we are interested here in whether a mean-field approximation is suitable for a unimodal approximation method.
If correlations are introduced by the presence of multiple modes then the appropriate solution is to represent those multiple modes separately \citep{wilsonBayesian2020} rather than to force a unimodal approximation to model those correlations.
In the figures in the main body of this document we show the correlation plots corresponding to the first chain of HMC samples, while the others can be found in \cref{a:cifar10_hmc}.
In addition, we only examine layers with less than or equal to 1000 parameters because of the computational difficulty of computing the covariance matrix for much larger layers and the variance of the estimators of the covariance terms with relatively few samples (240 per chain for CIFAR10, 150 per chain for CIFAR100, 400 per chain for IMDB).

\begin{figure}
  \centering
  \includegraphics[width=.8\textwidth]{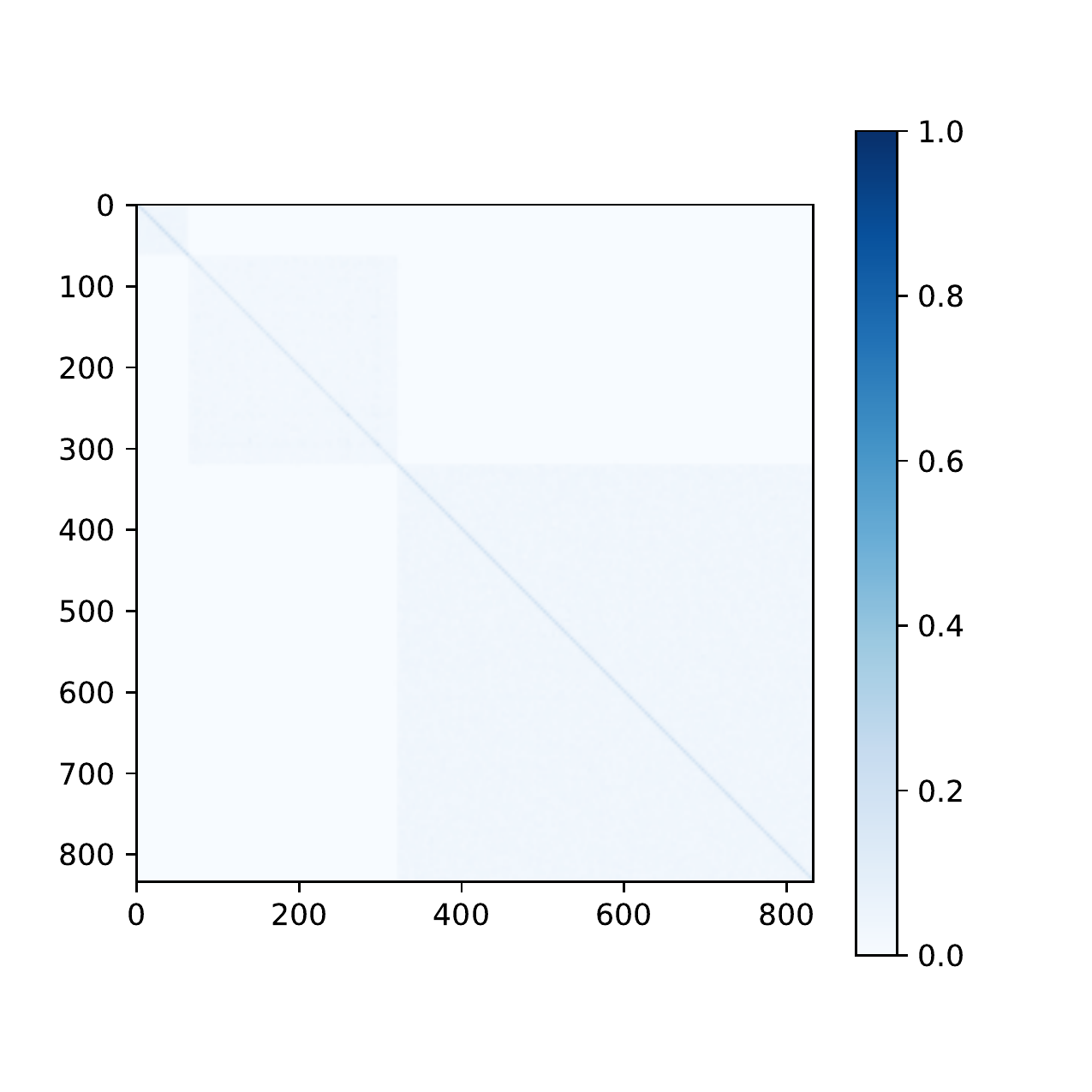}
  \caption[IMDB LSTM covariance plots]{IMDB, LSTM: We show the posterior sample covariance for the layers of the LSTM with fewer than 1000 weights. The HMC posterior samples show a mostly diagonal structure.}
  \label{fig:imdb_hmc}
\end{figure}

\begin{figure}
  \vspace{-15mm}
  \centering
  \begin{subfigure}[b]{\textwidth}
    \centering
    \includegraphics[width=0.8\textwidth]{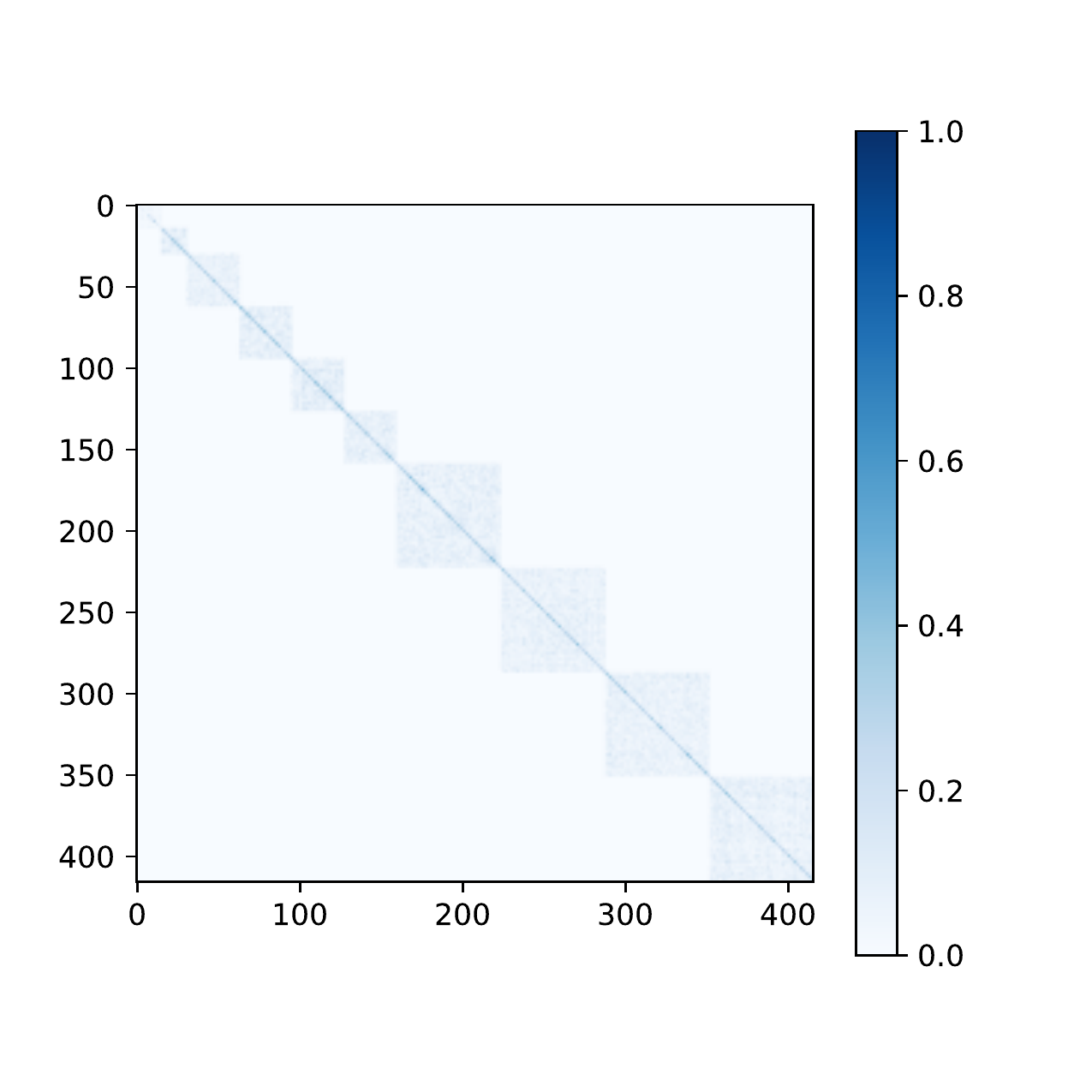}
    \vspace{-5mm}
    \caption{}
  \end{subfigure} \hfill

    \vspace{-5mm}
  \begin{subfigure}[b]{\textwidth}
    \centering
    \includegraphics[width=0.8\textwidth]{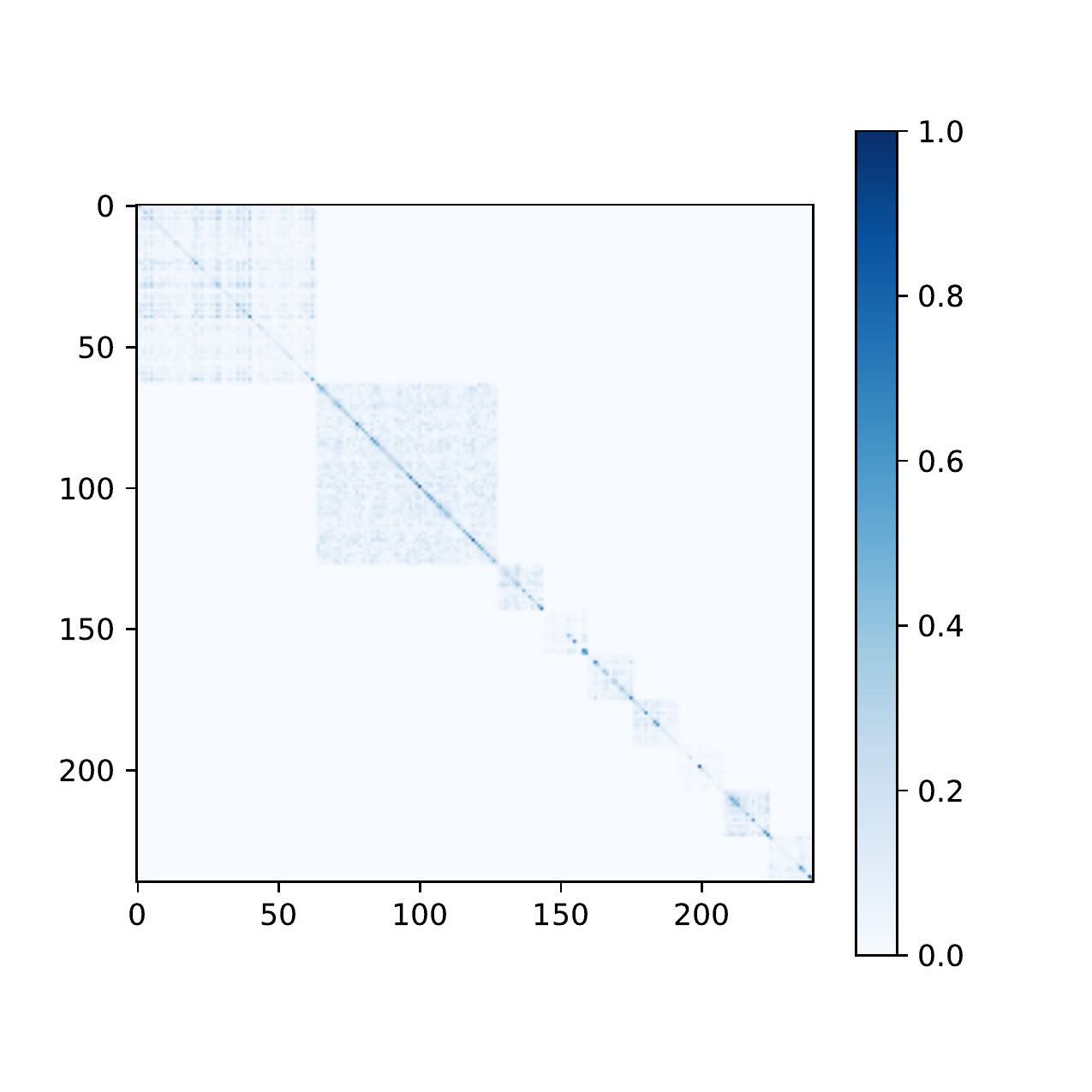}
    \vspace{-5mm}
    \caption{}
  \end{subfigure}\hfill
  \caption[CIFAR100, ResNet-20-FRN, covariance plot]{CIFAR100, ResNet-20-FRN: We show the HMC posterior sample covariance for small layers with fewer than 100 parameters. (a) Early in the network (first 10 small layers), convolutional layers have mostly diagonal covariances. (b) In the middle of the network (61st-70th small layers), there is more off-diagonal covariance.}
  \label{fig:cifar100_hmc}
\end{figure}

\begin{figure}
  \centering
    \includegraphics[width=0.8\textwidth]{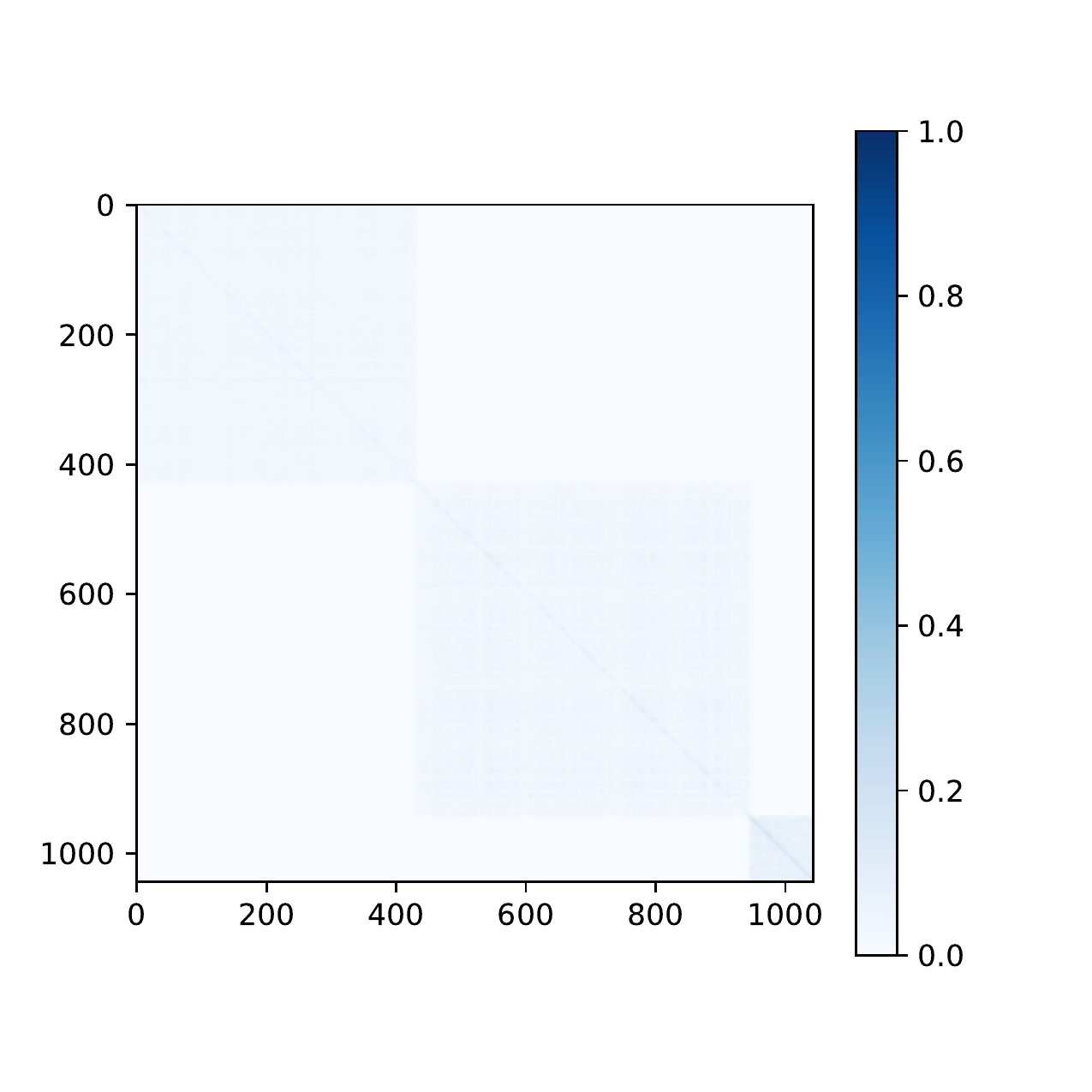}
  \caption[CIFAR100, ResNet-20-FRN, covariance plot, large]{CIFAR100, ResNet-20-FRN: We show the HMC posterior sample covariance for larger layers (between 100 and 1000 parameters). These seem to show more off-diagonal covariance. However, estimation error is greater for these layers since there are only 150 samples to fit the Gaussian mixture model with many more degrees of freedom than there are samples.}
  \label{fig:cifar100_hmc_large}
\end{figure}

\paragraph{Results} Indeed, we find that the covariance structure of the HMC samples is broadly diagonal.
For the LSTM, we find that most of the layers are fairly close to being diagonal \cref{fig:imdb_hmc}.
For the ResNet, many of the layers are almost entirely diagonal while some show off-diagonal covariances.
In \cref{fig:cifar100_hmc} and \cref{fig:cifar100_hmc_large}, we show the results from examining the samples of the ResNet-20-FRN on CIFAR100.
It seems that the middle and later layers of the ResNets show more off-diagonal correlations than are found in fully-connected networks.
These may be caused by the residual structure, by the convolutions, or by the very great depth of the network.
Our observational data is not enough to establish the cause.
However, it is also possible that these are estimation artifacts.
Note for example, that later results in \cref{s:parameterization:findable} examining the role of off-diagonal correlations in SWAG suggest that for predictive distributions the off-diagonal components may not be very important.

For computational reasons, we discard layers with more than 1000 parameters.
Computing the covariance matrix for these layers is unreliable because of both computational difficulties with inversion of large matrixes and because there are significantly more degrees of freedom in the Gaussian mixture model than there are samples.
However we do separately examine layers with between 100 and 1000 parameters, and find that these also show significant off-diagonal correlations.

Results for CIFAR10 are similar and are shown in \cref{a:cifar10_hmc} alongside the plots for the second and third HMC chains.


\section{Are There `Good' Approximate Posteriors?}
\label{s:parameterization:good_approximate_posteriors}

If the true posterior has nearly mean-field modes, it would suggest that a mean-field approximate posterior could be a good approximation to the \textit{parameter} distribution.
However, we can make an even more general argument that even when the true posterior is not approximately mean-field, the \textit{predictive} distribution can still be arbitrarily well approximated by a unimodal mean-field approximate posterior, even if the predictive distribution has multiple modes.

We show this using the universal approximation theorem (UAT) due to \citet{leshnoMultilayer1993} in a stochastic adaptation by \citet{foongExpressiveness2020}.
This shows that a BNN with a mean-field approximate posterior with at least two layers of hidden units can induce a function-space distribution that matches any true posterior distribution over function values arbitrarily closely, given arbitrary width.
Our proof formalizes and extends a remark by \citet[p23]{galUncertainty2016} concerning multi-modal posterior predictive distributions.

\begin{restatable}{proposition}{theoremuat}\label{thm:uat}
    Let $p(\mathbf{y}=\mathbf{Y}|\mathbf{x}, \mathcal{D})$ be the probability density function for the posterior predictive distribution of any given multivariate regression function, with $\mathbf{x} \in \mathbb{R}^D$, $\mathbf{y} \in \mathbb{R}^K$, and $\mathbf{Y}$ the posterior predictive random variable.
    Let $f(\cdot)$ be a Bayesian neural network with two hidden layers.
    Let $\hat{\mathbf{Y}}$ be the random vector defined by $f(\mathbf{x})$.
    Then, for any $\epsilon, \delta > 0$, there exists a set of parameters defining the neural network $f$ such that the absolute value of the difference in probability densities for any point is bounded:
    \begin{equation}
      \forall \mathbf{y}, \mathbf{x}, i: \quad \textup{Pr}\left(\abs{p(y_i=\hat{Y}_i) - p(y_i=Y_i|\mathbf{x}, \mathcal{D})} > \epsilon \right) < \delta,
    \end{equation}
    so long as: the activations of $f$ are non-polynomial, non-periodic, and have only zero-measure non-monotonic regions, the first hidden layer has at least $D+1$ units, the second hidden layer has an arbitrarily large number of units, the cumulative density function of the posterior predictive is continuous in output-space, and the probability density function is continuous and finite non-zero everywhere.
    Here, the probability bound is with respect to the distribution over a subset of the weights described in the proof, $\thet_{\textup{Pr}}$, while one weight distribution $\thet_{Z}$ remains to induce the random variable $\hat{\mathbf{Y}}$.
 \end{restatable}

The full proof is provided in \cref{a:parameterization:uat}.
Intuitively, we define a $q(\vtheta)$ to induce an arbitrary distribution over hidden units in the first layer and using the remaining weights and hidden layer we approximate the inverse cumulative density function of the true posterior predictive by the UAT.
The proof is made much more complicated by the fact that we aim for an architecture which is entirely a stochastic MLP.
In fact, all that is required is a single noise variable, $Z$, and a deterministic single-layer MLP which takes both $Z$ and $\mathbf{x}$ as inputs.
Our construction essentially uses the first layer to set up this simpler setting with $Z$ and $\mathbf{x}$, and then handles the stochasticity of the final layer with a probabilistic generalization of \citet{leshnoMultilayer1993}.

It follows from \cref{thm:uat} that there exists a mean-field approximate posterior which induces the true posterior distribution over predictive functions, whether or not the true posterior has an approximately mean-field mode.
Our proof strengthens a result by \citet{foongExpressiveness2020} which considers only the first two moments of the posterior predictive.

There are important limitations to this argument to bear in mind.
First, the UAT requires arbitrarily wide models in order to reach arbitrarily precise distribution matching.
Second, to achieve arbitrarily small error $\delta$ it is necessary to reduce the weight variance.
Both of these might result in very low weight-space evidence lower-bounds (ELBOs).
Third it may be difficult in practice to choose a prior in weight-space that induces the desired prior in function space.
Fourth, although the distribution in weight space that maximizes the marginal likelihood will also maximize the marginal likelihood in function-space within that model class, the same is not true of the weight-space ELBO and functional ELBO.
As a result, it is not entailed that any particular approximation technique, especially variational inference, is likely to find the approximate distributions that would result in a good predictive distribution approximation.

\clearpage
\begin{tcolorbox}[parbox=false,breakable,title=Discussion of \citet{foongExpressiveness2020}, colback=white]
  \label{a:foong}
  \citep{foongExpressiveness2020} also discuss expressivity of BNNs, reaching different conclusions although with no conflict on any proofs or direct results.
  Here, we consider possible reasons for our differing conclusions.

  Empirically, they focus on posterior distributions over function outputs in small models.
  They find that learned function distributions with MFVI for regression are overconfident.
  However, their largest experiment uses data with only 16-dimensional inputs, with only 55 training points and very small models with 4 layers of 50 hidden units.
  In contrast, our work analyses much larger models and datasets. 
  This is where MFVI would be more typical for deep learning but it also makes it harder to compare to a reference posterior in function-space.
  It is now possible to compare to samples from a reference posterior computed using HMC, for example \citet{wilsonEvaluating2021}.
  
  Theoretically, they show that single-layer mean-field networks cannot have appropriate `in-between' uncertainty, and conjecture that this extends to deeper networks and classification tasks.
  Their Theorem 1 states that the variance of the function expressed by a single-layer mean-field network between two points cannot be greater than the sum of the variances of the function at those points, subject to a number of very important caveats.
  The theoretical result applies only to regression models, not to classification.
  It is also strongest in a 1-dimensional input space.
  In higher dimensions, they show as a corollary that the `in-between' variance is bounded by the sum of the variances of the hypercube of points including that space, which grows exponentially with dimensionality: even only 10 dimensions, the in-between variance could be 1024 times greater than the average edge variance.
  The bound is therefore incredibly loose in, for example, computer vision.
  Last, the result only applies for certain line segments and is sensitive to translation and rotation in input-space.

  They also prove their Theorem 3, which establishes that deeper mean-field networks do not have the pathologies that apply in the limited single-layer regression settings identified in Theorem 1.
  We consider a similar result (\cref{thm:uat}) which is more general because it considers more than the first two moments of the distribution.
  Unlike \citet{foongExpressiveness2020} we see this result as potentially promising that deep mean-field networks can be very expressive.

  We note also that we find that deeper networks are able to show in-between uncertainty in regression, if not necessarily capture it as fully as something like HMC. In \cref{fig:updated-in-between} we show how increasing depth increases the ability of MFVI neural networks to capture in-between uncertainty even on low-dimensional data.

  {
    \centering
    \includegraphics{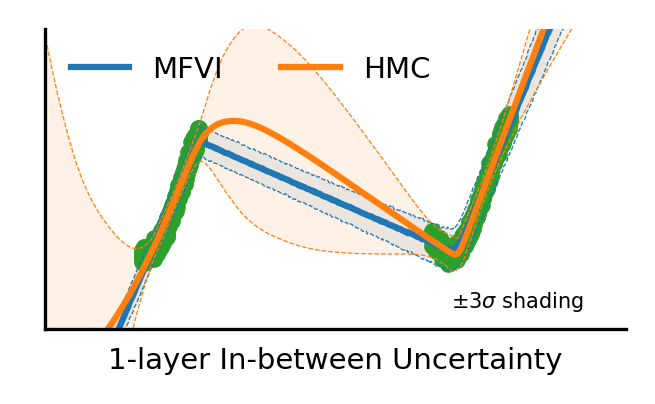}
    \captionof{figure}{1-layer BNN shows little `in-between' uncertainty.}
    \includegraphics{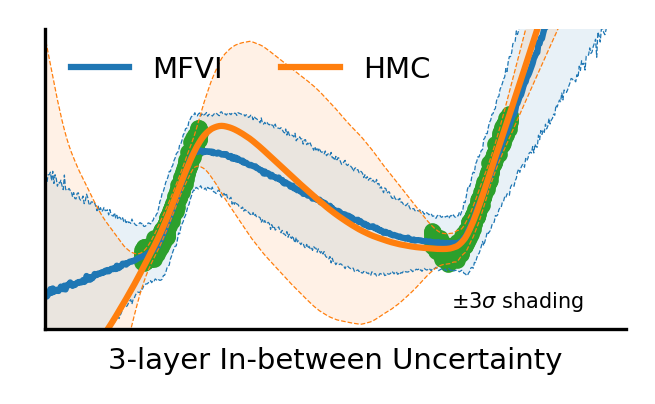}
    \captionof{figure}{3-layer BNN shows significant `in-between' uncertainty (but not as much as HMC).}
  \label{fig:updated-in-between}
  }
  \vspace{6mm}
  Here, each layer has 100 hidden units trained using mean-field variational inference on a synthetic dataset in order to demonstrate the possibility of `in-between' uncertainty.
  Full experimental settings are provided in \cref{tbl:hypers-toy}.
  The toy function used is $y = \sin(4(x-4.3)) + \epsilon$ where $\epsilon \sim \mathcal{N}(0, 0.05^2)$.
  We sample 750 points in the interval $-2 \leq x \leq -1.4$ and another 750 points in the interval $ 1.0 \leq x \leq 1.8$.
  We considered a range of temperatures between 0.1 and 100 in order to select the right balance between prior and data.
  Note of course that while our figure demonstrates the existence of deep networks that perform well, of course a single case of a one-layer network performing badly does not show that all one-layer networks perform badly.
\end{tcolorbox}

\clearpage
\section{Do Approximate Inference Methods Find Good Mean-field Approximations?}
\label{s:parameterization:findable}

Although we have shown empirically that there are nearly-mean-field modes of the true posterior for deep neural networks in parameter-space, and theoretically that mean-field approximate posteriors can be arbitrarily close to the true predictive posterior, we have not yet shown that approximate inference methods succeed in taking advantage of these properties.

Moreover, it is reasonable to hold reservations about the analysis in the previous sections.
Although HMC samples are our best lens into the behaviour of the true posterior, it is conceivable that HMC has specific failures in its ability to capture complex correlations that might interfere with our investigatiosn.
Similarly, although the universal approximation theorem is powerful, one might wonder if results in wide limits have practical implications for real networks.

In the following sections, we develop a motivating construction showing how depth can allow mean-field networks to induce the same predictive distributions as full-covariance approximate posteriors of shallower networks.
We provide a strong correspondence in deep linear models, showing that every matrix-variate Gaussian (MVG) structured covariance approximating distribution can be represented by a three-layer mean-field approximating distribution.
We then bridge this correspondence towards neural networks with piecewise-linear activation functions by introducing local product matrices as an analytical tool.
These arguments are intended to motivate an understanding of how a deep mean-field model could `naturally' converge on a good solution.
This extends on the hypothesis by \citet{hintonKeeping1993} that even with a mean-field approximating distribution, during optimization the parameters will find a version of the network where this restriction is least costly.
Note, however, that the true posterior over the weights of the shallow network will be different from the true posterior over the weights of the deeper network.
Our hypothesis is that, for sufficiently deep networks, both models are sufficiently good (see \cref{ss:challenges:model_misspecification}).
In this case, finding an approximate posterior for the larger model which is similar in predictive distribution to the smaller model would be sufficient, and is possibly better.

Finally, we examine the question empirically, observing that mean-field approximate posteriors are not obviously worse than structured-covariance approximations in very large models.
Moreover, for small models for which full-covariance approximate inference is tractable, the benefits are small and decrease with depth.

\subsection{Emergent Covariance in Deep Linear Mean-Field Networks}
\label{ss:parameterization:linear}
Although we are most interested in neural networks that have non-linear activations, linear neural networks can be analytically useful \citep{saxeExact2014}.
Setting the activation function of a neural network, $\phi(\cdot)$, to be the identity turns a neural network into a deep linear model.
Without non-linearities the weights of the model just act by matrix multiplication.
$L$ weight matrices for a deep linear model can therefore be `flattened' through matrix multiplication into a single weight matrix which we call the \emph{product matrix}---$M^{(L)}$.
For a BNN, the weight distributions induce a distribution over the elements of this product matrix.
Because the model is linear, there is a one-to-one mapping between distributions induced over elements of this product matrix and the distribution over linear functions $y = M^{(L)}\mathbf{x}$.
This offers us a way to examine exactly which sorts of distributions can be induced by a deep linear model on the elements of a product matrix, and therefore on the resulting function-space.

\subsubsection{Covariance of the Product Matrix}
We derive the analytic form of the covariance of the product matrix in \cref{a:derivation}, explicitly finding the covariance of $M^{(2)}$ and $M^{(3)}$ as well as the update rule for increasing $L$.
These results hold for \emph{any} factorized weight distribution with finite first- and second-order moments, not just Gaussian weights.
Using these expressions, we show:

\begin{restatable}{proposition}{lemgreaterthanzero}\label{lemma:covariance}
  For $L\geq3$, the product matrix $M^{(L)}$ of factorized weight matrices can have non-zero covariance between any and all pairs of elements. That is, there exists a set of mean-field weight matrices $\{W^{(l)} | 1 \leq l < L\}$ such that $M^{(L)} = \prod W^{(l)}$ and the covariance between any possible pair of elements of the product matrix:
  \begin{equation}
    \textup{Cov}(m^{(L)}_{ab}, m^{(L)}_{cd}) \neq 0,
  \end{equation}
  where $m^{(L)}_{ij}$ are elements of the product matrix in the $i$\textsuperscript{th} row and $j$\textsuperscript{th} column, and for any possible indexes $a$, $b$, $c$, and $d$.
\end{restatable}

This shows that a deep mean-field linear model is able to induce function-space distributions which would require covariance between weights in a shallower model.
We do not show that all possible fully parameterized covariance matrices between elements of the product matrix can be induced in this way.\footnote{E.g., a full-covariance layer has more degrees of freedom than a three-layer mean-field product matrix (one of the weaknesses of full-covariance in practice).
An $L$-layer product matrix of $K\times K$ Gaussian weight matrices has $2LK^2$ parameters, but one full-covariance weight matrix has $K^2$ mean parameters and $K^2 (K^2 + 1) / 2$ covariance parameters.
Note also that the distributions over the elements of a product matrix composed of Gaussian layers are not in general Gaussian (see \cref{box:product_distributions} for more discussion of this point).}
However, we emphasise that the expressible covariances become very complex.
Below, we show that a lower bound on their expressiveness exceeds a commonly used structured-covariance approximate distribution.

\subsubsection{Numerical Simulation}
\begin{figure}[t]
  \centering
  \begin{subfigure}{0.3\textwidth}
      \includegraphics[width=\linewidth]{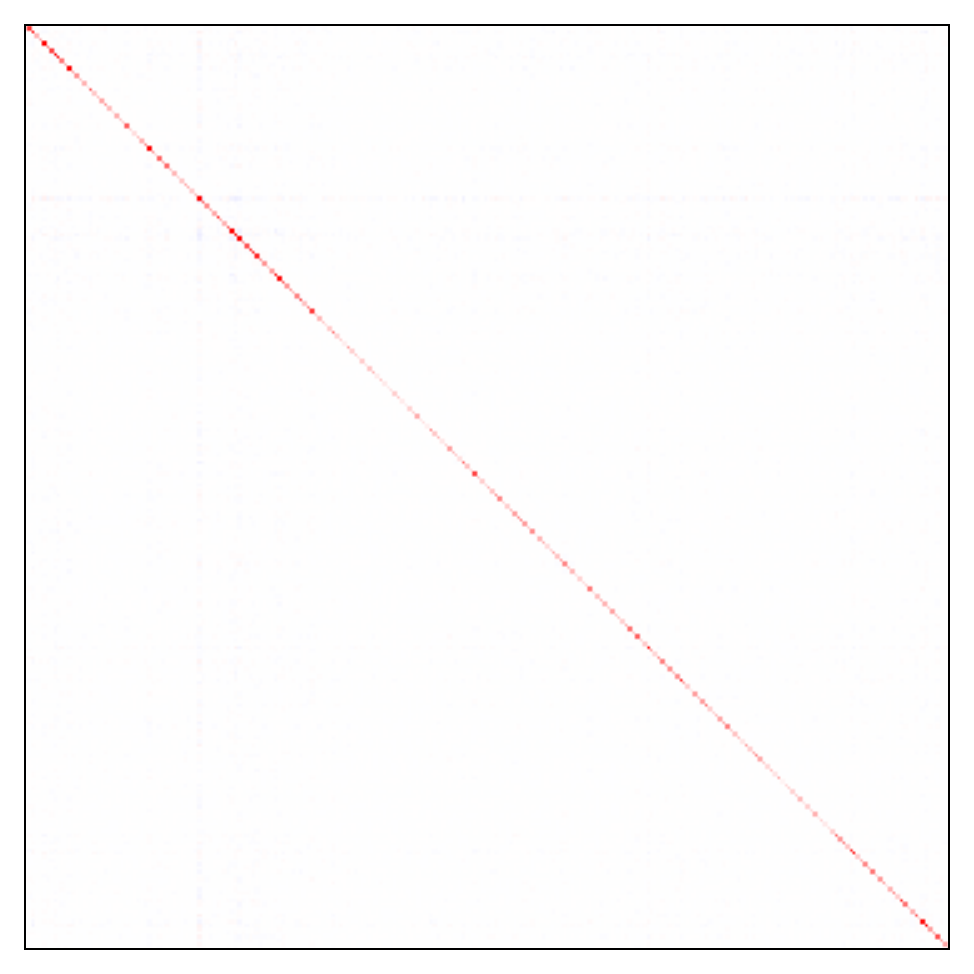}
      \caption{One weight matrix.\\\quad}
      \label{fig:cov_heatmap-a}
  \end{subfigure}\
  \begin{subfigure}{0.3\textwidth}
      \includegraphics[width=\linewidth]{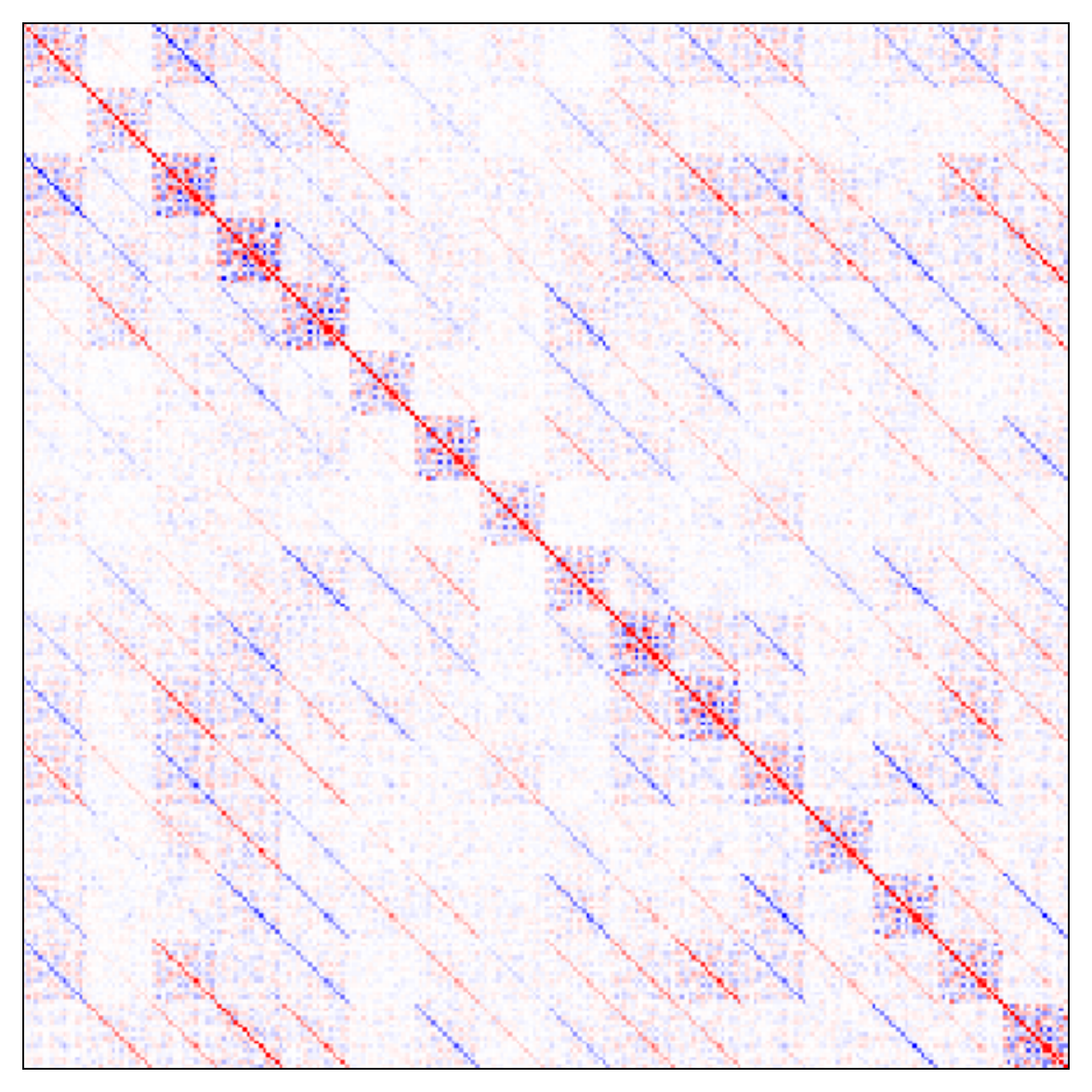}
      \caption{5-layer product matrix. (Linear)}
  \end{subfigure}
  \
  \begin{subfigure}{0.3\textwidth}
      \includegraphics[width=\linewidth]{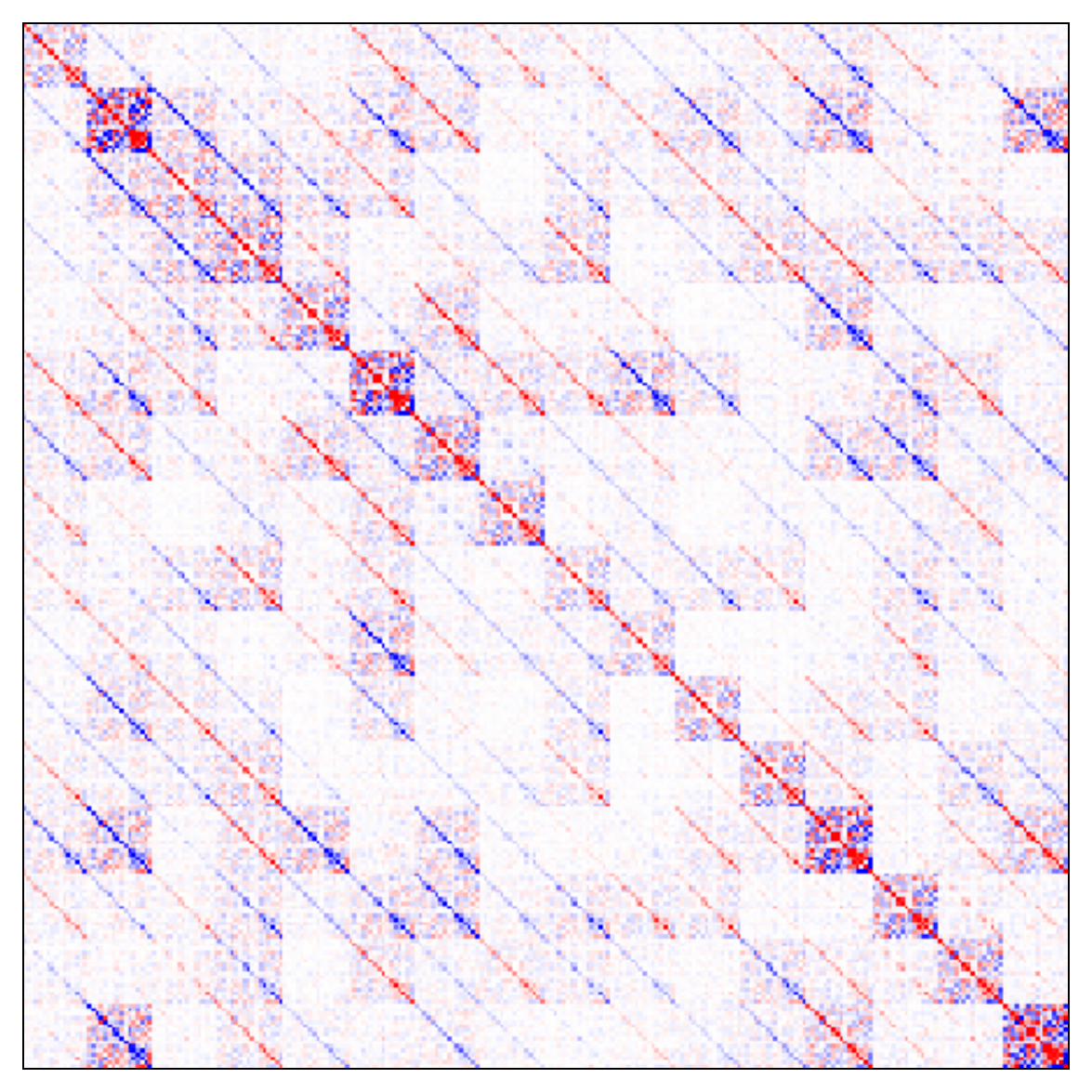}
      \caption{10-layer product matrix. (Linear)}
      \label{fig:cov_heatmap-c}
  \end{subfigure} \\
  \begin{subfigure}{0.3\textwidth}
    \includegraphics[width=\textwidth]{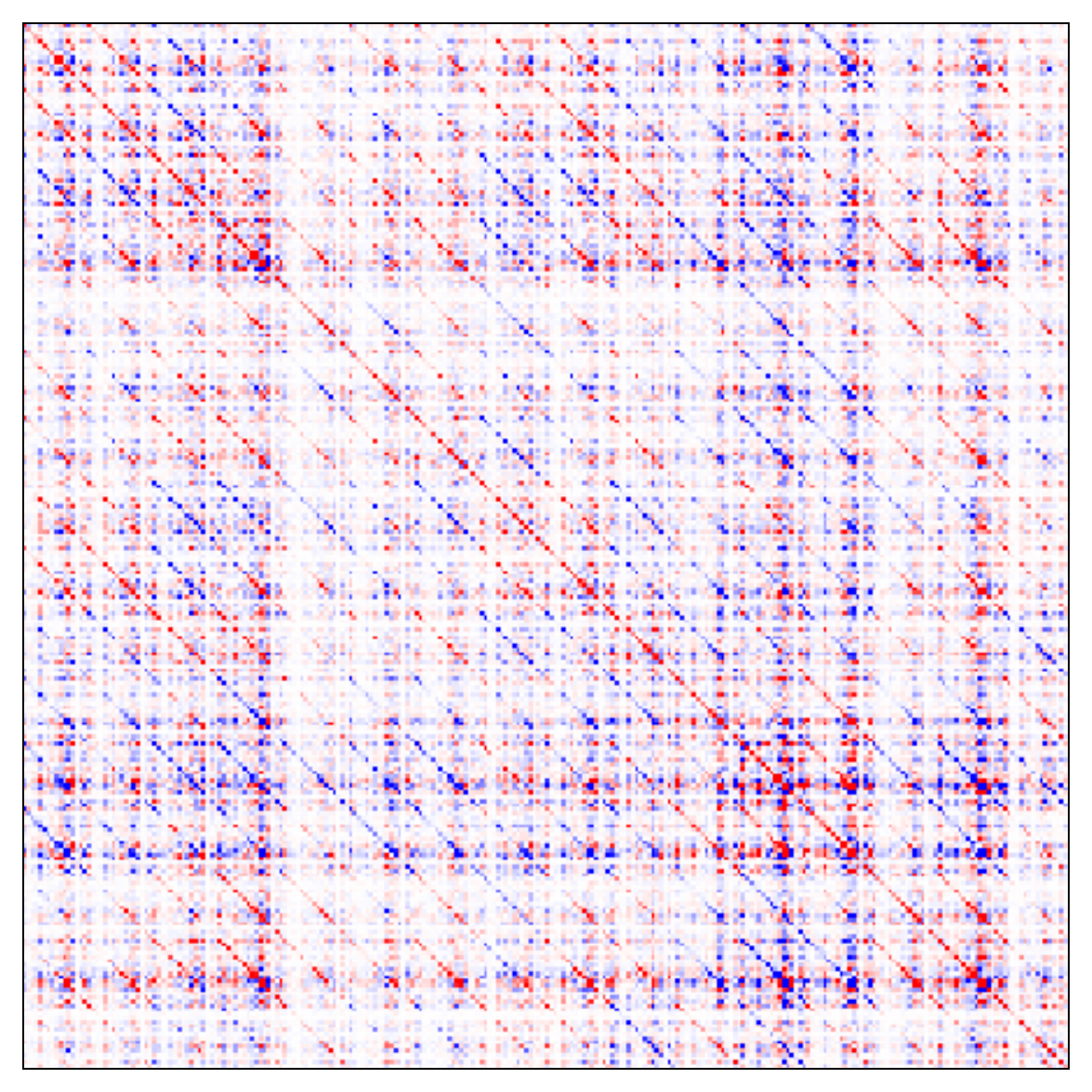}
    \caption{5-layer local product matrix. (Leaky ReLU)}
    \label{fig:cov_heatmap-d}
\end{subfigure}
\
\begin{subfigure}{0.3\textwidth}
    \includegraphics[width=\textwidth]{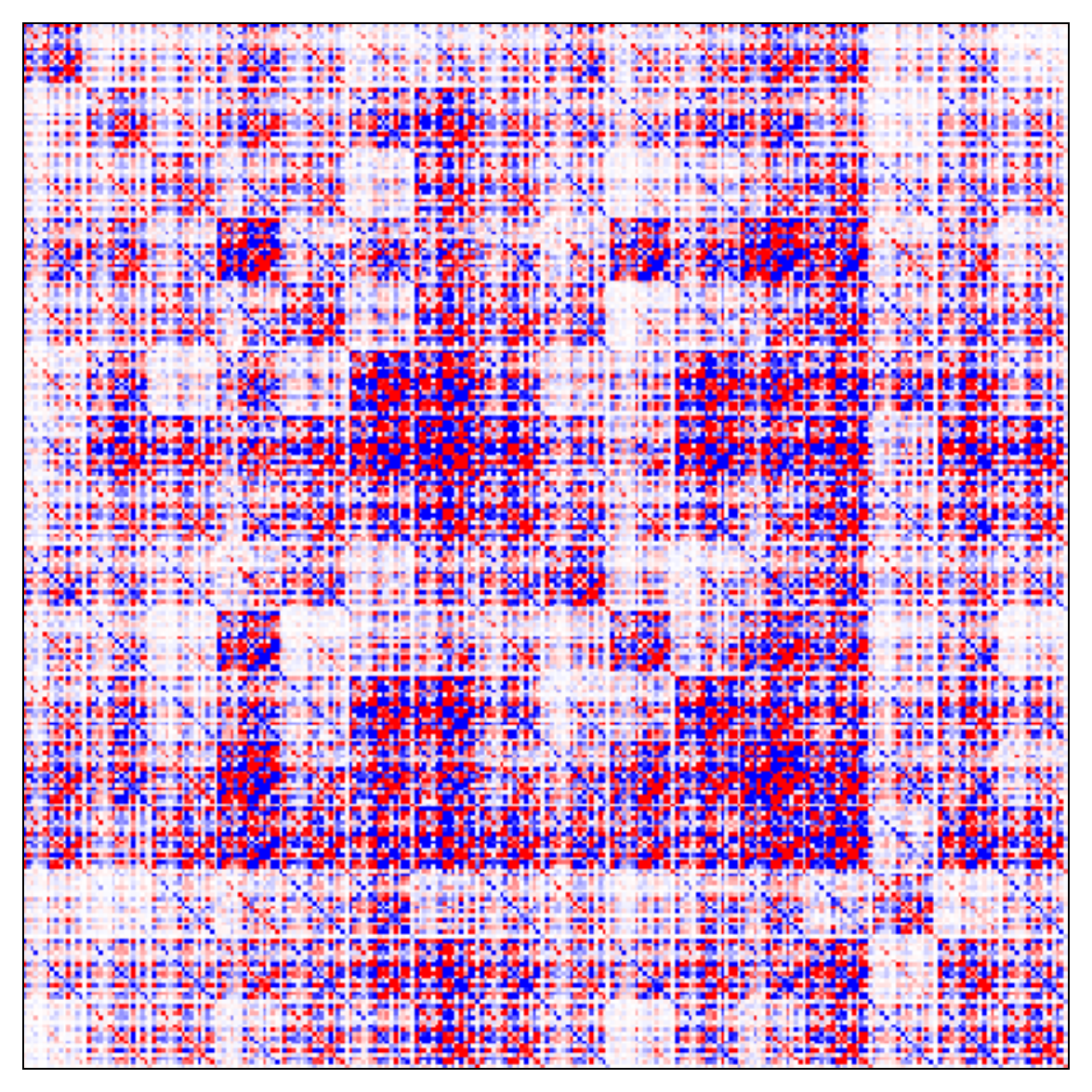}
    \caption{10-layer local product matrix. (Leaky ReLU)}
    \label{fig:cov_heatmap-e}
\end{subfigure}
\
\begin{subfigure}{0.3\textwidth}
  \includegraphics[width=\textwidth]{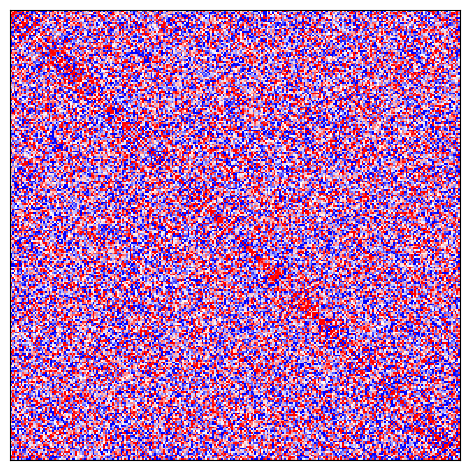}
  \caption{5 randomly sampled linear layers.}
  \label{fig:no_block}
\end{subfigure}
  \caption[Mean-field layers induce rich covariance product matrices]{Covariance heatmap for mean-field approximate posteriors trained on FashionMNIST. (a) A single layer has diagonal covariance. (b-c) In a deep linear model the \emph{product matrix} composed of $L$ mean-field weight matrices has off-diagonal covariance induced by the mean-field layers. Redder is more positive, bluer more negative. (d-e) For piecewise non-linear activations we introduce `local product matrices' (defined in \cref{ss:parameterization:piecewise}) with similar covariance. Shared activations introduce extra correlations. This lets us extend results from linear to piecewise-linear neural networks. (f) Untrained (randomly sampled) product matrices visually have a noisier and less structured covariance although the mathematical properties of this random matrix are not well understood.}
  \label{fig:cov_heatmap}
\end{figure}

To build intuition, in \cref{fig:cov_heatmap-a}--\subref{fig:cov_heatmap-c} we visualize the covariance between entries of the product matrix from a deep mean-field VI linear model trained on FashionMNIST.
Even though each weight matrix makes the mean-field assumption, the product develops off-diagonal correlations.
The experiment is described in more detail in \cref{a:heatmap}.

\subsubsection{How Expressive is the Product Matrix?}
We show that the Matrix Variate Gaussian (MVG) distribution is a special case of the mean-field product matrix distribution.
The MVG distribution is used as a structured-covariance approximation by e.g., \citet{louizosStructured2016}, \citet{zhangNoisy2018} to approximate the covariance of weight matrices while performing variational inference.\footnote{In some settings, MVG distributions can be indicated by the Kronecker-factored or K-FAC approximation. In MVGs, the covariance between elements of an $n_0 \times n_1$ weight matrix can be described as $\Sigma = V \otimes U$ where $U$ and $V$ are positive definite real scale matrices of shape $n_0 \times n_0$ and $n_1 \times n_1$.}
We prove in \cref{a:mvg}:

\begin{restatable}{proposition}{thmmatrixnormal}\label{thm:matnorm}
    The Matrix Variate Gaussian (Kronecker-factored) distribution is a special case of the distribution over elements of the product matrix.
    In particular, for $M^{(3)} = ABC$, $M^{(3)}$ is distributed as an MVG random variable when $A$ and $C$ are deterministic and $B$ has its elements distributed as fully factorized Gaussians with unit variance.
\end{restatable}

\Cref{thm:matnorm} directly entails in the linear case that for any deep linear model with an MVG weight distribution, there exists a deeper linear model with a mean-field weight distribution that induces the same posterior predictive distribution in function-space.
This is a \textbf{lower bound} on the expressiveness of the product matrix.
We have made \emph{very} strong restrictions on the parameterization of the weights for the sake of an interpretable result.
The unconstrained expressiveness of the product matrix covariance given in \cref{a:derivation} is much greater.
Also note, we do not propose using this, it is purely an analysis tool.

\begin{tcolorbox}[breakable,lower separated=false, title=Distribution of the Product Matrix, parbox=false, colback=white]
  \label{box:product_distributions}
  What is the distribution over elements of the product matrix made of Gaussian layers?
  It is not Gaussian.
  For scalars, the product of two independent Gaussian distributions is a generalized $\chi^2$ distribution.
  The product of $N$ Gaussians with arbitrary mean and variance is unknown outside of special cases (e.g., \citet{springerDistribution1970}).
  An example of a distribution family that \emph{is} closed under multiplication is the log-normal distribution.
  
 Matrix multiplication is important for neural network weights.
 To preserve distribution through matrix multiplication we would like a distribution which is closed under both addition and multiplication (such as the Generalized Gamma convolution \citep{bondessonClass2015}) but these are not practical.
  
  However, perhaps even a simple distribution like the Gaussian can maintain roughly similar distributions over product matrix elements as the network becomes deeper.
  Sometimes, people appeal to the central limit theorem, arguing that the sum of random variables is Gaussian so the distribution is preserved (e.g., \citep{kingmaVariational2015,wuDeterministic2019}).
  For only one layer of hidden units, provided $K$ is sufficiently large, this is true.
  For two or more layers, however, the central limit theorem fails because the elements of the product matrix are no longer independent.
  
  For almost all Gaussians, numerical simulation shows that the resulting product matrix is not remotely Gaussian.
  However, In fact, although the resulting product matrix is not a Gaussian, we show through numerical simulation that products of matrices with individual weights distributed as $\mathcal{N}(0, 0.23^2)$ have roughly the same distribution over their weights.
  This, combined with the fact that our choice of Gaussian distributions over weights was somewhat arbitrary in the first place, might reassure us that the increase in depth does not change the model prior in an important way.
  In \cref{fig:product_of_gaussians} we plot the probability density function of an arbitrarily chosen entry in the product matrix with varying depths of diagonal Gaussian prior weights.
  The p.d.f. for 7 layers is approximately the same as the single-layer Gaussian distribution with variance $0.23^2$.
  \tcblower
  \centering
  \includegraphics[width=0.4\linewidth]{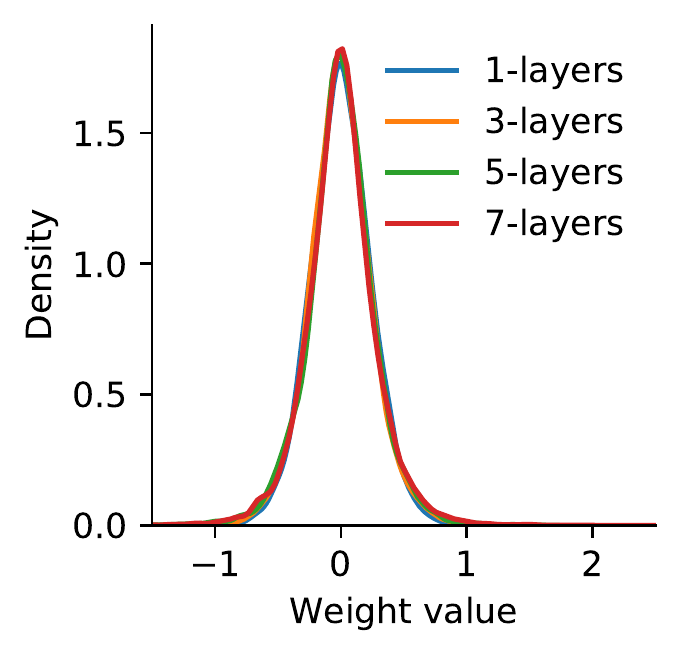}
  \captionof{figure}[Product matrix elements can approximate Gaussians]{Density over arbitrary element of product matrix for $L$ diagonal prior Gaussian weight matrices whose elements are i.i.d. $\mathcal{N}(0, 0.23^2)$. Product matrix elements are not strictly Gaussian, but very close.}
  \label{fig:product_of_gaussians}
\end{tcolorbox}

\subsection{Product Matrices in Piecewise-Linear Mean-field BNNs}
\label{ss:parameterization:piecewise}
Neural networks use non-linear activations to increase the flexibility of function approximation.
On the face of it, these non-linearities make it impossible to consider product matrices.
In this section we show how to define the \emph{local product matrix}, which is an extension of the product matrix to widely used neural networks with piecewise-linear activation functions like ReLUs or Leaky ReLUs.
For this we draw inspiration from a proof technique by \citet{shamirSimple2019} which we extend to stochastic matrices.
This analytical tool can be used for any stochastic neural network with piecewise linear activations.
Here, we use it to extend \cref{lemma:covariance} to neural networks with piecewise-linear activations.

\begin{tcolorbox}[parbox=false,lower separated=false, title=Multimodal local product matrices,colback=white]
 Researchers often critique a Gaussian approximate posterior because it is unimodal \textit{in parameter-space}.
 We confirm empirically that multiple mean-field layers induce a multi-modal product matrix distribution.
\Cref{fig:multi_modal} show a density over an element of the local product matrix from three layers of weights in a Leaky ReLU BNN with $\alpha = 0.1$ from a 3-layer network trained on FashionMNIST.

  {\centering
        \includegraphics[width=0.48\columnwidth]{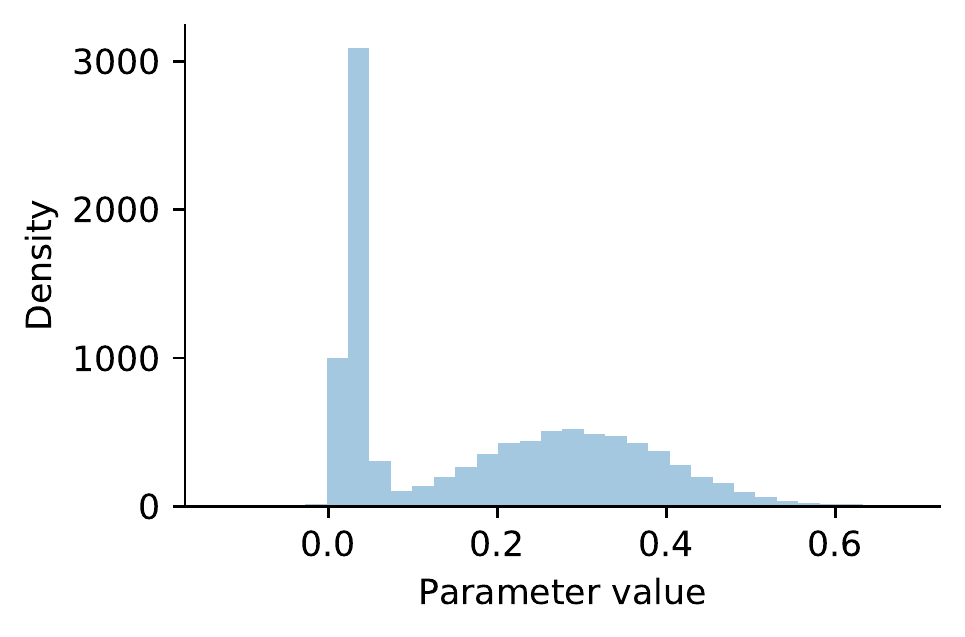}
        \captionof{figure}{Induced product matrices can have multimodal densities}
        \label{fig:multi_modal}
        \vspace{6mm}
  }

  Visually inspecting a sample of 20 elements of this product matrix showed that 12 were multi-modal.
  We found that without the non-linear activation, none of the product matrix entry distributions examined were multimodal, suggesting that the non-linearities in fact play an important role in inducing rich predictive distributions by creating modes corresponding to activated sign patterns.
\end{tcolorbox}

\subsubsection{Defining a Local Product Matrix}
Neural nets with piecewise-linear activations induce piecewise-linear functions.
These piecewise-linear neural network functions define hyperplanes which partition the input domain into regions within which the function is linear.
Each region can be identified by a sign vector that indicates which activations are `switched on'.
We show in \cref{a:local_linearity}:

\begin{restatable}{lemma}{linear}\label{lemma:linearity}
  Consider an input point $\mathbf{x}^* \in \mathcal{D}$. Consider a realization of the model weights $\vtheta$. Then, for any $\mathbf{x}^*$, the neural network function $f_{\vtheta}$ is linear over some compact set $\mathcal{A}_{\vtheta} \subset \mathcal{D}$ containing $\mathbf{x}^*$. Moreover, $\mathcal{A}_{\vtheta}$ has non-zero measure for almost all $\mathbf{x}^*$ w.r.t. the Lebesgue measure.
  \end{restatable}

Using a set of $N$ realizations of the weight parameters $\Theta = \{\thet_i \text{ for } 1\leq i \leq N\}$ we construct a product matrix within $\mathcal{A} = \bigcap_i \mathcal{A}_{\thet_i}$.
Since each $f_{\thet_i}$ is linear over $\mathcal{A}$, the activation function can be replaced by a diagonal matrix which multiplies each row of its `input' by a constant that depends on which activations are `switched on' (e.g., 0 or 1 for a ReLU).
This allows us to compute through matrix multiplication a product matrix of $L$ weight layers $M_{\mathbf{x}^*, \thet_i}^{(L)}$ corresponding to each function realization within $\mathcal{A}$.
We construct a local product matrix random variate $P_{\mathbf{x}^*}$, for a given $\mathbf{x}^*$, within $\mathcal{A}$, by sampling these $M_{\mathbf{x}^*, \thet_i}^{(L)}$.
The random variate $P_{\mathbf{x}^*}$ is therefore such that $y$ given $\mathbf{x}^*$ has the same distribution as $P_{\mathbf{x}^*}\mathbf{x}^*$ within $\mathcal{A}$.
This distribution can be found empirically at a given input point, and resembles the product matrices from linear settings (see \cref{fig:cov_heatmap-d}--\subref{fig:cov_heatmap-e}).

\subsubsection{Covariance of the Local Product Matrix}
We can examine this local product matrix in order to investigate the covariance between its elements.
We prove in \cref{a:proof_linearized_product_matrix} that:

\begin{restatable}{proposition}{maintheorem}
  \label{thm:main}
  Given a mean-field distribution over the weights of neural network $f$ with piecewise linear activations, $f$ can be written in terms of the local product matrix $P_{\mathbf{x}^*}$ within $\mathcal{A}$. 
  
  For $L \geq 3$, for activation functions which are non-zero everywhere, there exists a set of weight matrices $\{W^{(l)}|1 \leq l < L\}$ such that all elements of the local product matrix have non-zero off-diagonal covariance:
  \begin{equation}
    \textup{Cov}(p^{\mathbf{x}^*}_{ab}, p^{\mathbf{x}^*}_{cd}) \neq 0,
  \end{equation}
  where $p^{\mathbf{x}^*}_{ij}$ is the element at the $i$\textsuperscript{th} row and $j$\textsuperscript{th} column of $P_{\mathbf{x}^*}$.
\end{restatable}

\Cref{thm:main} is weaker than the Weight Distribution Hypothesis.
Once more, we do not show that all full-covariance weight distributions can be exactly replicated by a deeper factorized network.
We now have non-linear networks which give richer functions, potentially allowing richer covariance, but the non-linearities have introduced analytical complications.
However, it illustrates the way in which deep factorized networks can emulate rich covariance in a shallower network.

\begin{remark}
\Cref{thm:main} is restricted to activations that are non-zero everywhere although we believe that in practice it will hold for activations that can be zero, like ReLU.
If the activation can be zero then, for some $\mathbf{x}^*$, enough activations could be `switched off' such that the effective depth is less than three.
This seems unlikely in a trained network, since it amounts to throwing away most of the network's capacity, but we cannot rule it out theoretically.
In \cref{a:proof_linearized_product_matrix} we empirically corroborate that activations are rarely all `switched off' in multiple entire layers.
\end{remark}

\subsubsection{Numerical Simulation}\label{exp_heatmap}
We confirm empirically that the local product matrix develops complex off-diagonal correlations using a neural network with Leaky ReLU activations trained on FashionMNIST using mean-field variational inference.
We estimate the covariance matrix using 10,000 samples of a trained model (\cref{fig:cov_heatmap-d}--\subref{fig:cov_heatmap-e}).
Just like in the linear case (\cref{fig:cov_heatmap-a}--\subref{fig:cov_heatmap-c}), as the model gets deeper the induced distribution on the product matrix shows complex off-diagonal covariance.
There are additional correlations between elements of the product matrix based on which activation pattern is predominantly present at that point in input-space.
See \cref{a:heatmap} for further experimental details.

\subsection{Does Mean-field Approximate Inference Work Empirically?}\label{evaluation}
Here, we compare the performance of Bayesian neural networks with complex posterior approximations to those with mean-field approximations.
We show that over a spectrum of model sizes, examining both variational inference and amortized stochastic gradient Markov chain Monte Carlo, performance does not seem to be greatly determined by the approximation.

\begin{figure}
\begin{minipage}{0.48\textwidth}
  \centering
  \includegraphics[width=\columnwidth]{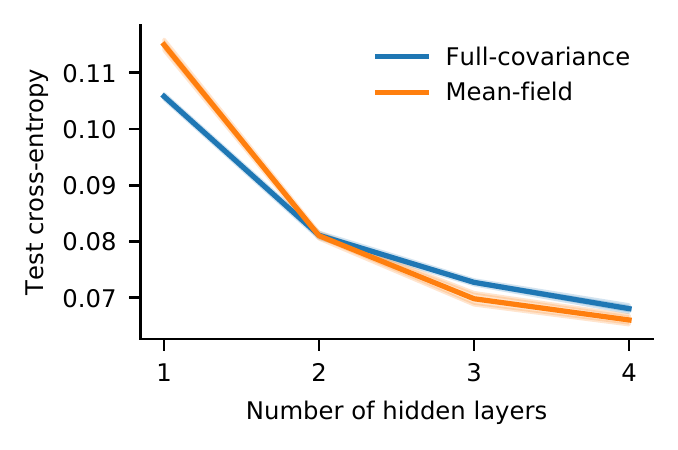}
  \vspace{-6mm}
  \caption[Mean-field matches full-covariance after two layers on Iris]{Full- vs diagonal-covariance. After two hidden layers mean-field matches full-covariance. Iris dataset.}
  \label{fig:iris}
  \vspace{-3mm}
\end{minipage}
\hfill
\begin{minipage}{0.48\textwidth}
  \centering
  \includegraphics[width=\textwidth]{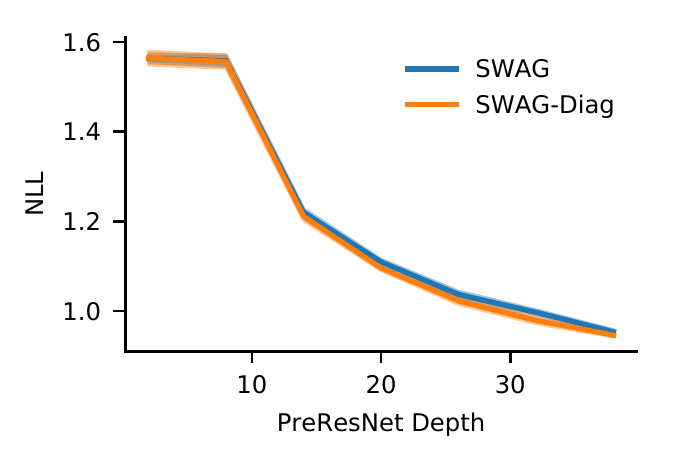}
  \vspace{-6mm}
  \caption[CIFAR-10. Diagonal and low-rank SWAG show similar performance]{CIFAR-100. Diagonal- and Structured-SWAG show similar log-likelihood in PresNets of varying depth.}\label{fig:swag}
\end{minipage}
\centering
\begin{minipage}{0.48\textwidth}
      \includegraphics[width=\textwidth]{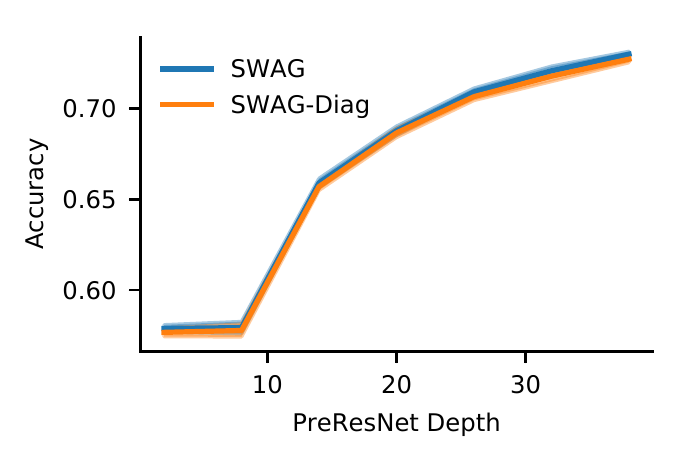}
    \caption[CIFAR-100. Diagonal and low-rank SWAG show similar performance]{CIFAR-100. Accuracy for diagonal and low-rank covariance SWAG. Like log-likelihood, there is no clear difference in performance between these models, all of which are above the depth threshold implied by our work.}\label{fig:swag_acc}
\end{minipage}
\end{figure}

\paragraph{Depth in Full- and Diagonal-covariance Variational Inference.}
Training with full-covariance variational inference is intractable, except for very small models, because of optimization difficulties.
In \cref{fig:iris}, we show the test cross-entropy of small models of varying depths on the Iris dataset from the UCI repository.
With one layer of hidden units the full-covariance posterior achieves lower cross-entropy.
For deeper models, however, the mean field network matches the full-covariance one.
Full details of the experiment can be found in \cref{a:iris}.

\definecolor{diag}{gray}{0.93}

\paragraph{Structured- and Diagonal-covariance Uncertainty on CIFAR-100.}
Although we cannot compute samples from the true posterior in larger models, we attempt an approximate investigation using SWAG \citep{maddoxSimple2019}.
This involves fitting a Gaussian distribution to approximate SG-MCMC samples on CIFAR-100.
SWAG approximates the Gaussian distribution with a low rank empirical covariance matrix, while SWAG-Diag uses a factorized Gaussian.
We show in \cref{fig:swag} that there is no observable difference in negative log-likelihood or accuracy between the diagonal and low-rank approximation.
All of the models considered have more than two layers of hidden units (the minimum size of a PresNet).
This suggests that there is a mode of the true posterior over weights for these deeper models that is sufficiently mean-field that a structured approximation provides little or no benefit.
It also suggests that past a threshold of two hidden layers, further depth is not essential.

  \begin{table}[]
    \resizebox{\textwidth}{!}{
    \begin{tabular}{@{}llllll@{}}
    \toprule
    Architecture        & Method      & Covariance      & Accuracy            & NLL               & ECE                 \\ \midrule
    \rowcolor[HTML]{EFEFEF} 
    ResNet-18    & VOGN$^\ddagger$        & Diagonal        & 67.4\% $\pm$ 0.263 & 1.37 $\pm$ 0.010   & 0.029  $\pm$ 0.001  \\
    ResNet-18    & Noisy K-FAC$^{\dagger\dagger}$ & MVG             & 66.4\% $\pm$ n.d.  & 1.44 $\pm$ n.d.   & 0.080  $\pm$ n.d.   \\
    \rowcolor[HTML]{EFEFEF} 
    DenseNet-161 & SWAG-Diag$^{\dagger}$   & Diagonal        & 78.6\% $\pm$  0.000 & 0.86 $\pm$  0.000 & 0.046  $\pm$ 0.000  \\
    DenseNet-161 & SWAG$^{\dagger}$        & Low-rank + Diag & 78.6\% $\pm$  0.000 & 0.83 $\pm$  0.000 & 0.020  $\pm$  0.000 \\
    \rowcolor[HTML]{EFEFEF} 
    ResNet-152   & SWAG-Diag$^{\dagger}$   & Diagonal        & 80.0\% $\pm$  0.000 & 0.86 $\pm$  0.000 & 0.057  $\pm$  0.000 \\
    ResNet-152   & SWAG$^{\dagger}$        & Low-rank + Diag & 79.1\% $\pm$  0.000 & 0.82 $\pm$  0.000 & 0.028  $\pm$  0.000 \\ \bottomrule
    \end{tabular}
    }
    \caption[Imagenet. Mean-field and structured-covariance methods show similar performance.]{Imagenet. Comparison of diagonal-covariance/mean-field (in grey) and structured-covariance methods on Imagenet. The differences on a given architecture between comparable methods is small and has inconsistent sign.
    $^\dagger$ \citep{maddoxSimple2019}. $^\ddagger$ \citep{osawaPractical2019}. $^{\dagger\dagger}$ \citep{zhangNoisy2018} as reported by \citet{osawaPractical2019}.}
    \label{tbl:imagenet_full}
  \end{table}

  \begin{table}[]
    \resizebox{\textwidth}{!}{
    \begin{tabular}{@{}llllll@{}}
    \toprule
    Architecture    & Method                                          & Covariance      & Accuracy           & NLL               & ECE               \\ \midrule
    \rowcolor[HTML]{EFEFEF} 
    VGG-16          & SWAG-Diag$^\dagger$                             & Diagonal        & 93.7\% $\pm$ 0.15 & 0.220 $\pm$ 0.008 & 0.027 $\pm$ 0.003 \\
    VGG-16          & SWAG$^\dagger$                                  & Low-rank + Diag & 93.6\% $\pm$ 0.10    & 0.202 $\pm$ 0.003 & 0.016 $\pm$ 0.003 \\
    \rowcolor[HTML]{EFEFEF} 
    VGG-16          & Noisy Adam$^{\ddagger\ddagger}$                 & Diagonal        & 88.2\% $\pm$ n.d.  & n.d.              & n.d.              \\
    \rowcolor[HTML]{EFEFEF} 
    VGG-16          & BBB$^{\ddagger\ddagger}$                        & Diagonal        & 88.3\% $\pm$ n.d.  & n.d.              & n.d.              \\
    VGG-16          & Noisy KFAC$^{\ddagger\ddagger}$                 & MVG             & 89.4\% $\pm$ n.d.  & n.d.              & n.d.              \\
    \rowcolor[HTML]{EFEFEF} 
    PreResNet-164   & SWAG-Diag$^\dagger$                             & Diagonal        & 96.0\% $\pm$ 0.10  & 0.125 $\pm$ 0.003 & 0.008 $\pm$ 0.001 \\
    PreResNet-164   & SWAG$^\dagger$                                  & Low-rank + Diag & 96.0\% $\pm$ 0.02  & 0.123 $\pm$ 0.002 & 0.005 $\pm$ 0.000 \\
    \rowcolor[HTML]{EFEFEF} 
    WideResNet28x10 & SWAG-Diag$^\dagger$                             & Diagonal        & 96.4\% $\pm$ 0.08  & 0.108 $\pm$ 0.001 & 0.005 $\pm$ 0.001 \\
    WideResNet28x10 & SWAG$^\dagger$                                  & Low-rank + Diag & 96.3\% $\pm$ 0.08 & 0.112 $\pm$ 0.001 & 0.009 $\pm$ 0.001 \\
    \rowcolor[HTML]{EFEFEF} 
    ResNet-18       & VOGN$^\ddagger$                                 & Diagonal        & 84.3\% $\pm$ 0.20 & 0.477 $\pm$ 0.006 & 0.040 $\pm$ 0.002 \\
    \rowcolor[HTML]{EFEFEF} 
    AlexNet         & VOGN$\ddagger$ & Diagonal        & 75.5\% $\pm$ 0.48 & 0.703 $\pm$ 0.006 & 0.016 $\pm$ 0.001 \\
     \bottomrule
    \end{tabular}
    }
    \caption[CIFAR-10. Mean-field and structured-covariance methods show similar performance.]{CIFAR-10. For a given architecture, it does not seem that mean-field (grey) methods systematically perform worse than methods with structured covariance, although there is some difference in the results reported by different authors.
    $^\dagger$ \citep{maddoxSimple2019}. $^\ddagger$ \citep{osawaPractical2019}. $^{\ddagger\ddagger}$ \citep{zhangNoisy2018}.}
    \label{tbl:cifar10}
  \end{table}

\paragraph{Large-model Mean-field Approximations on Imagenet.}
The performance of mean-field and structured-covariance methods on large-scale tasks can give some sense of how restrictive the mean-field approximation is.
Mean-field methods have been shown to perform comparably to structured methods in large scale settings like Imagenet, both in accuracy and measures of uncertainty like log-likelihood and expected calibration error (ECE) (see \cref{tbl:imagenet_full}).
For VOGN \citep{osawaPractical2019} which explicitly optimizes for a mean-field variational posterior, the mean-field model is marginally better in all three measures.
For SWAG, the accuracy is marginally better and log-likelihood and ECE marginally worse for the diagonal approximation.
This is consistent with the idea that there are some modes of large models that are approximately mean-field (which VOGN searches for but SWAG does not) but that not all modes are.
These findings offer some evidence that the importance of structured covariance is at least greatly diminished in large-scale models, and may not be worth the additional computational expense and modelling complexity.
\footnote{
  The standard deviations, of course, underestimate the true variability of the method in question on Imagenet as they only consider difference in random seed with the training configuration otherwise identical.
  Fuller descriptions of the experimental settings used by the authors are provided in the cited papers.}

A similar evaluation for CIFAR-10 shows a similar lack of clear advantage to using a richer covariance for deeper models \cref{tbl:cifar10}.
Within the same architecture, there is little evidence of systematic differences between mean-field and structured-covariance methods and any differences which do appear are marginal.
Note that \citet{zhangNoisy2018} report difficulty applying batch normalization to mean-field methods, but \citet{osawaPractical2019} report no difficulties applying batch normalization for their mean-field variant of Noisy Adam.
For this reason, we report the version of Noisy KFAC run without batch normalization to make it comparable with the results shown for Bayes-by-Backprop (BBB) and Noisy Adam.
With batch normalization, Noisy KFAC gains some accuracy, reaching 92.0\%, but this seems to be because of the additional regularization, not a property of the approximate posterior family.

\section{Discussion}\label{discussion}
Researchers have a longstanding intuition that the mean-field approximation for Bayesian neural networks is a severe one.
However, this impression has not been based on a rigorous notion of severity because of the difficulty of understanding what makes a `good' approximate posterior in the first place.
In this chapter, we have examined three underlying questions:
\begin{itemize}
  \item Does the \textit{true} posterior distribution have strong correlations between the parameters?
  \item Do there exist approximate posterior distributions that are `good' even if they are mean-field?
  \item Do actual methods for approximate Bayesian inference uncover these `good' approximations?
\end{itemize}
For the first, we have shown the way in which the \textit{architecture} of the parametric model influences characteristics of the \textit{true} posterior.
This has an important and underappreciated significance: good approximate Bayesian inference depends heavily on \textit{matching} the architecture to the approximation scheme and dataset because different architectures have different posteriors on different datasets which will sometimes suit some approximation schemes better than others.
In particular, we have argued that mean-field approximating distributions are more likely to be a good fit in deep neural networks than shallow ones for MLPs and probably LSTMs.
At the same, observational investigation of HMC samples for large residual CNNs suggests that the true posterior may have off-diagonal components, especially in the middle and later layers.
This suggests once again that the choice of architecture has significant implications for the appropriate approximating distribution.

For the second, we have proven under mild assumptions that there exist approximating distributions over Bayesian neural networks of sufficient depth and width which are arbitrarily close to the true posterior predictive.

For the third, we have provided empirical evidence that schemes like variational inference and stochastic gradient Markov chain Monte Carlo do in fact uncover these modes.
We have provided a motivating construction for how this might happen because of the ability of blocks of several diagonal-covariance layers to `simulate' shallower layers with full-covariance.

Despite this, it remains the case that mean-field variational inference is rarely used in practice because of poor results.
In the next chapter, we consider ways in which the approximating distribution interacts not just with the data and architecture but also the \textit{optimization process}.
We will show how the standard Gaussian approximating posterior has bad sampling properties that affect optimization, and provide an alternative distribution which fixes these properties.
\ifSubfilesClassLoaded{
\bibliographystyle{plainnat}
\bibliography{thesis_references}}{}
\end{document}

\chapter{Approximation Assumptions Affect Optimization}
\label{chp:optimization}


\begin{figure}[t]
\centering
\includegraphics{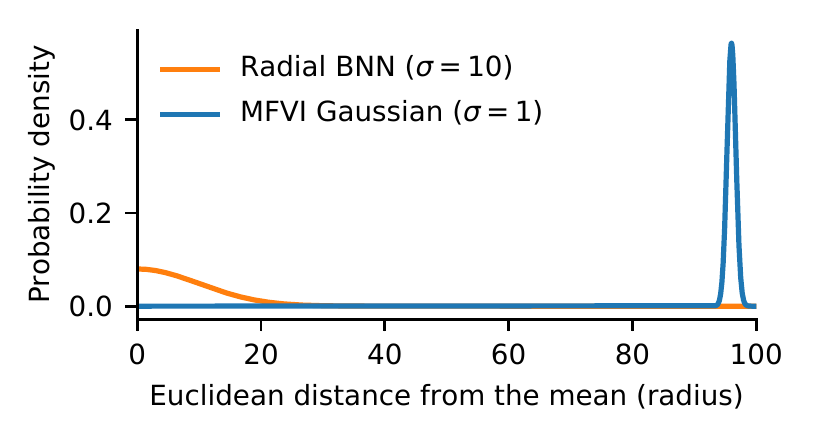}
\caption[Comparison of the radial densities of high-dimensional multivariate Gaussians and our distribution]{
MFVI uses a multivariate Gaussian approximate posterior whose probability mass is tightly clustered at a fixed radius from the mean depending on the number of dimensions---the `soap-bubble'.
In our Radial BNN, samples from the approximate posterior are more reflective of the mean.
This helps training by reducing gradient variance.
The plot shows the p.d.f.\ based on dimensionality of a 3$\times$3 convolutional layer with 64 channels.}
\label{fig:pdf_plot}
\end{figure}

Approximate Bayesian inference is often performed using optimization techniques.
This is most marked for variational inference (VI) which converts the inference problem into an optimization problem on the evidence lower bound (ELBO).\footnote{Stochastic gradient Markov chain Monte Carlo methods also use optimization methods, but in a very different way. It is possible that the approach considered in this chapter would influence the mixing properties of SG-MCMC, but it is out of scope for this work.}
A topic which has received little attention is the ways in which the choice of approximating distribution interact with the procedure used to optimize the ELBO.
In this chapter, we show how the standard multivariate Gaussian approximating distribution has pathological high-dimensional properties: typical points are far from both each other and the mean, resulting in high variance estimates.
As a result, we propose an alternative sampling distribution with more desirable sampling properties.
We show how this reduces gradient variance and can lead to `better' approximate posteriors.
This is an example of the ways in which careful use of approximations can influence the performance of approximate Bayesian inference in ways that are not connected to the quality of the underlying probabilistic model.

In the previous chapter, we argued that the mean-field approximation in approximate Bayesian inference for sufficiently deep neural networks is not as severe an approximation as people often assume.
Nevertheless, researchers have difficulties in employing mean-field methods in practice.
To make mean-field variational inference (MFVI) work, researchers often resort to ad-hoc tweaks to the loss or optimization process which side-step the variational inference arguments that motivate the approach in the first place!
Our analysis of optimization helps us understand why MFVI struggles despite the fact that the approximating distribution is capable of expressing sufficiently rich predictive distributions.

\section{Understanding the Soap Bubble}
\label{s:optimization:soap_bubble}
The `soap-bubble' is a well-known property in multi-variate Gaussian distributions as the number of dimensions increases (e.g., see \citet{bishopIntroduction2006, betancourtConceptual2018}).
Although the highest probability density is near the mean, because there is just so much more volume further from the mean in high-dimensional spaces most of the probability mass is located in a narrow band---a `soap-bubble'.
This band is where the opposing pressures of growing space and shrinking density intersect.
For high-dimensional spaces, the band is both vanishingly narrow and far away from the mean.

One way to understand this is to examine the probability density function of the multivariate Gaussian along its radius (following the derivation of \citep{bishopIntroduction2006}).
Consider a $D$-dimensional isotropic Gaussian from which we sample $\vw \sim \gaussian(\vmu, \vsigma^2)$.
Take a thin shell with thickness $\eta$ at radius $r$ away from the mean $\vmu$.
As $\eta$ tends to zero, the probability density function over the radius is given by:
\begin{align}
    &\lim_{\eta\to0} p(r - \eta < \norm{\vw - \boldsymbol{\mu}} < r + \eta) \\
    &= \underbrace{\frac{S_D}{(2\pi\sigma^2)^{D/2}}}_{\text{Normalizing constant}} \cdot \underbrace{\vphantom{\frac{S_D}{(2\pi\sigma^2)^{D/2}}}r^{D-1}}_{\text{Growing volume}} \cdot \underbrace{\vphantom{\frac{S_D}{(2\pi\sigma^2)^{D/2}}}e^{-\frac{r^2}{2\sigma^2}}}_{\text{Shrinking density}}
    \label{eq:soap_bubble}
\end{align}
where $S_D$ is the surface area of a hypersphere in a $D$-dimensional space.

\begin{figure}
    \centering
    \includegraphics{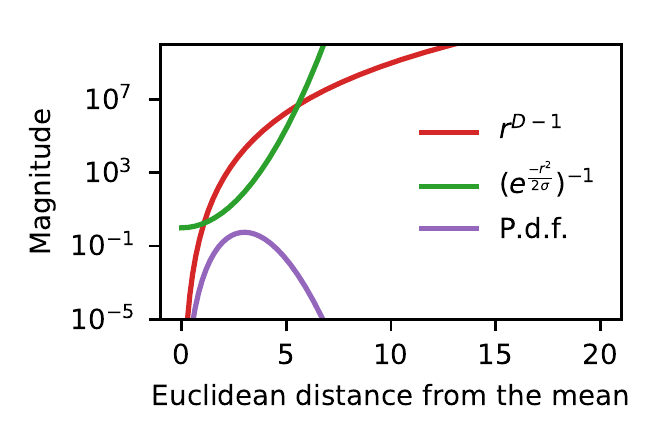}
    \caption[Visualizing the soap bubble]{We can understand the `soap-bubble' by looking at components of the p.d.f. in eq. \ref{eq:soap_bubble}. For small $r$ the volume term (red) dominates and the normalized p.d.f. is very small. For big $r$ the Gaussian density term (inverse shown in green) dominates and the p.d.f. is small again. Almost all the probability mass is in the intermediate region where neither term dominates: the `soap bubble'. The intermediate region becomes narrower and further from the mean as $D$ is bigger. Here we show $D=10$.}
    \label{fig:soap_bubble}
\end{figure}

The two non-normalization terms are shown in \cref{fig:soap_bubble}.
The inverted density (in green) grows quadratically on a log-scale while the volume (in red) logarithmically.
To begin, the low volume dominates and the p.d.f.\ over the radius is small.
At the end, the low density dominates and the p.d.f.\ is small again.
Most of the mass is in a band where the two overlap---and the position and narrowness of this band depends on the dimensionality of $\rvw$.

\subsection{The Soap Bubble In Variational Inference}
A powerful way to optimize the variational posterior for a neural network is using Monte Carlo expectations to compute the ELBO \citep{gravesPractical2011,blundellWeight2015}.
The ELBO for a variational approximate posterior distribution $q(\rvw)$ over the weights of our neural network $\rvw$ relative to a true posterior distribution $p(\rvw \mid \data)$ can be written
\begin{align}
    \textrm{ELBO}(q,p) &= \E{q(\rvw)}{\log p(\data \mid \rvw)} - \KLdiv{q(\rvw)}{p(\rvw)}\\
    &= \E{q(\rvw)}{\log p(\data \mid \rvw)} - \E{q(\rvw)}{\log \frac{q(\rvw)}{p(\rvw)}}.
\end{align}
Estimating the ELBO, therefore, involves estimating two expectations over the variational distribution.
The variance of Monte Carlo expectations is a significant problem which has received much attention, for example through the use of importance sampling \citep{kahnMethods1953}.
Of fundamental importance, however, are the sampling properties of $q(\rvw)$.

The soap-bubble has the consequence that almost all samples from the distribution are very distant from the mean.
Typical samples from $\rvw$ are a distance from the mean proportional to $\sigma\sqrt{D}$, for standard deviation parameter $\sigma$ and the number of parameters $D$.\footnote{For simplicity, to calculate the distance of typical samples from the mean we imagine an isotropic posterior. The approximate posteriors are not isotropic, which means the true expression is slightly more complicated, but the pattern is similar.}
This effect is therefore more extreme in larger neural networks.

In addition, because of the geometry of hyper-spheres in high-dimensional spaces, with high probability each sample is far from all the other samples.
Similarly, this distance is proportional to $\sigma \sqrt{D}$ and the probability distribution of distances becomes increasingly tight as $D$ grows.

All else equal, we might expect this to lead to predictions and losses from multiple samples of the weights which are less correlated with each other than if the samples were near to each other in weight-space.
We anticipate (and demonstrate in \S\ref{s:understanding}) that the distance between samples from the MFVI approximate posterior makes the gradient estimator of the log-likelihood term of the loss in \cref{eq:MFVI_unit_prior} have a large variance, which makes optimization more difficult.

\begin{tcolorbox}[title=Soap bubbles in related settings, colback=white, parbox=false]
In Bayesian optimization, \citet{ohBOCK2018} consider the difficulties posed by soap-bubbles.
\citet{ohRadial2019} use hyperspherical coordinate system for variational inference in a way that removes the soap-bubble.
They motivate their posterior as a way to model correlations between weights to remove the restrictiveness of the mean-field assumption (c.f., \ref{chp:parameterization}).
As a result, although their approach is superficially similar to ours in some ways, their approximate posterior is radial over each row in their weight matrix, rather than the whole layer, and they introduce an expensive posterior distribution (von Mises-Fisher) to explicitly model weight correlations within rows, whereas we do \textit{not} seek to learn any correlations between parameters in the hyperspherical space.
That is, their method uses a different technique to solve a different problem.
\end{tcolorbox}
\section{Overcoming the Soap Bubble}\label{s:optimization:method}
To address this, we propose an alternative approximate posterior distribution without a `soap-bubble'.
Our strategy is to use a simple approximate posterior distribution in a hyperspherical space corresponding to each layer, which has been chosen to avoid a soap-bubble, and then transform this distribution into the coordinate system of the weights.
We show that the Radial BNN can be sampled efficiently in weight-space, without needing explicit coordinate transformations, and derive an analytic expression for the loss that makes training as fast and as easy to implement as MFVI.

A `soap-bubble' arises when, for large $D$, the probability density function over the radius from the mean is sharply peaked at a large distance from the mean (see \cref{fig:pdf_plot}).
Therefore, we pick a probability distribution which cannot have this property.
We can easily write down a probability density function which cannot have a `soap-bubble' by explicitly modelling the radius from the mean.
The hyperspherical coordinate system suits our needs: the first dimension is the radius and the remaining dimensions are angles.
We pick the simplest practical distribution in hyperspherical coordinates with no soap bubble:
\begin{itemize}
    \item In the radial dimension: $r=|\tilde{r}|$ for $\tilde{r} \sim \mathcal{N}(0,1)$.\footnote{The absolute value rules out negative radii. In practice, we can neglect the absolute value because of the rotational symmetry in angular dimensions.}
    \item In the angular dimensions: uniform distribution over the hypersphere---all directions equally likely.
\end{itemize}

A critical property is that it is easy to sample this distribution \textit{in the weight-space coordinate system}---we wish to avoid the expense of explicit coordinate transformations when sampling from the approximate posterior.
Instead of sampling the posterior distribution directly, we use the local reparameterization trick \cite{rezendeStochastic2014, kingmaAutoEncoding2014}, and sample the noise distribution instead.
This is similar to \citet{gravesPractical2011} and \citet{blundellWeight2015} who sample their weights
\begin{equation}
    \textbf{w} \coloneqq \boldsymbol{\mu} + \boldsymbol{\sigma} \odot \boldsymbol{\epsilon}_{\text{MFVI}}, \label{eq:reparameterization}
\end{equation}
where $\boldsymbol{\epsilon}_{\text{MFVI}} \sim \mathcal{N}(0, \mathbf{I})$ and $\cdot$ is an element-wise multiplication.
In order to sample from the Radial BNN posterior we make a small modification:
\begin{equation}
    \textbf{w}_{\text{radial}} \coloneqq  \boldsymbol{\mu} + \boldsymbol{\sigma} \odot \frac{\boldsymbol{\epsilon}_{\text{MFVI}}}{\norm{\boldsymbol{\epsilon}_{\text{MFVI}}}} \cdot \rr ,
\end{equation}
which works because dividing a multi-variate Gaussian random variable by its norm provides samples from a direction uniformly selected from the unit hypersphere \citep{mullerNote1959,marsagliaChoosing1972}.
As a result, sampling from our posterior is nearly as cheap as sampling from the MFVI posterior.
The only extra steps are to normalize the noise, and multiply by a scalar Gaussian random variable.

In fact, we can speed this up even further by noting that in high-dimensional space the norm $\norm{\bm{\epsilon}_{\text{MFVI}}}$ can be calculated analytically with high probability (precisely because of the soap-bubble!).
This saves us the step of normalizing the noise, making each additional sample from the approximate posterior only more expensive than the standard MFVI samples by scalar multiplication and division.

In our implementations we sample from the Radial distribution separately for each layer, and indeed we sample weights and biases separately.

\begin{tcolorbox}[title=Visualizing the Radial posterior distribution, colback=white, parbox=false,sidebyside align=top, lower separated=false]
Avoiding the soap-bubble requires lighter tails than a Gaussian distribution.
Here we show the marginal distribution over a single weight in a layer with 10 dimensions.
It is much more sharply peaked than an equivalent Gaussian.
\tcblower
\centering
    \includegraphics[width=0.5\columnwidth]{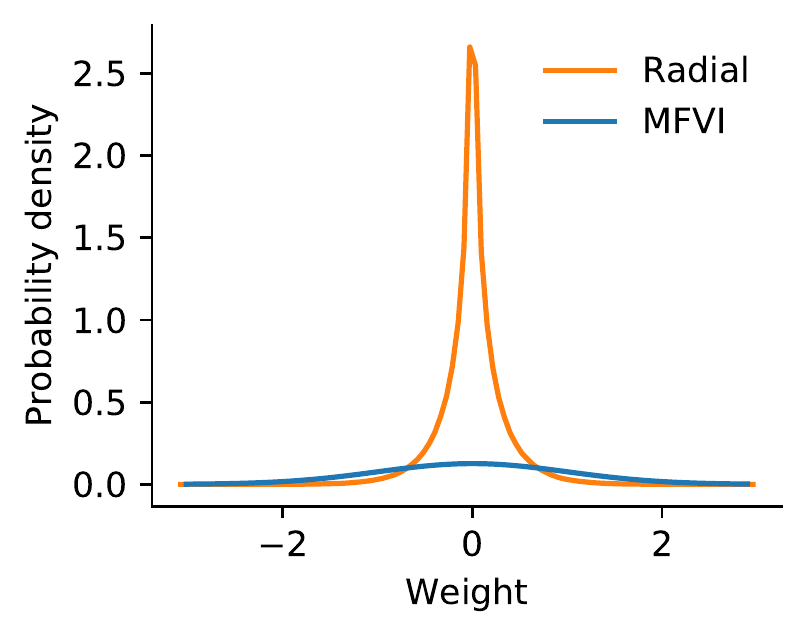}
    \label{fig:marginal}
    \captionof{figure}{The Radial Gaussian distribution is much more sharply peaked and has lighter tails than a Gaussian.}
\end{tcolorbox}

\subsection{Evaluating the Objective}
\renewcommand{\wr}{\vw^{(r)}}
\newcommand{\wx}{\vw^{(x)}}
\newcommand{\er}{\boldsymbol{\epsilon}^{(r)}}
\newcommand{\ex}{\boldsymbol{\epsilon}^{(x)}}
\newcommand{\eri}{\epsilon^{(r)}}
\newcommand{\exi}{\epsilon^{(x)}}
\newcommand{\bmu}{\boldsymbol{\mu}}
\newcommand{\bsi}{\boldsymbol{\sigma}}
\newcommand{\bth}{\boldsymbol{\theta}}
To use our approximate posterior for variational inference we must be able to estimate the ELBO loss.
The Radial BNN posterior does not change how the expected log-likelihood is estimated, using mini-batches of data points and MC integration.

The KL divergence between the approximate posterior and prior can be written:
\begin{align}
    \KLdiv{q(\vw)}{p(\vw)} &= \int q(\textbf{w}) \log [q(\textbf{w})] d\textbf{w} - \int q(\textbf{w}) \log \big[ p(\textbf{w})\big] d\textbf{w} \nonumber \\ &= \mathcal{L}_{\text{entropy}} - \mathcal{L}_{\text{cross-entropy}}.
\end{align}
We estimate the cross-entropy term using MC integration, just by taking samples from the posterior and averaging their log probability under the prior.
We find that this is low-variance in practice, and is often done for MFVI as well \citep{blundellWeight2015}.

We can evaluate the entropy of the posterior analytically.
We derive the entropy term in Appendix \ref{a:derivation_of_entropy}:
\begin{align}
    \mathcal{L}_{\text{entropy}} = - \sum_i\log[\sigma_i] + \text{const.}
\end{align}
where $i$ sums over the weights.
This is, up to a constant, the same as when using an ordinary multivariate Gaussian in MFVI.\footnote{In the common case of a unit multivariate Gaussian prior and an approximate posterior $\mathcal{N}(\vmu, \vsigma^2)$ over the weights $\rvw$, the negative evidence lower bound (ELBO) objective is:
\begin{align}
    \mathcal{L}_{\text{MFVI}} = &\overbrace{\rule{0pt}{3ex}\sum_i \frac{1}{2}\big[\sigma^2_i + \mu^2_i\big]}^{\text{prior cross-entropy}} - \overbrace{\rule{0pt}{3ex}\sum_i \log[\sigma_i]}^{\mathclap{\substack{\text{approximate-} \\ \text{posterior entropy}}}} - \overbrace{\rule{0pt}{3ex}\E{\vw \sim q(\vw)}{\log p(\mathbf{y}|\vw,\mathbf{X})}}^{\text{data likelihood}}. \label{eq:MFVI_unit_prior}
\end{align}}
Although they are not needed for optimization, for sake of completeness, we also derive the constant terms in the Appendix.

In Appendix \ref{a:derivation_of_prior}, we also provide a derivation of the cross-entropy loss term in the case where the prior is a Radial BNN.
This is useful in continual learning (see \cref{s:evaluation:continual_learning}) where we use the posterior from training one model as a prior when training another but is more computationally expensive than using a Gaussian prior because it requires an explicit change of variables.

Code implementing Radial BNNs can be found at \url{https://github.com/SebFar/radial_bnn}.
Training Radial BNNs has the same computational complexity as MFVI---$\mathcal{O}(D)$, where $D$ is the number of weights in the model.

\section{Diabetic Retinopathy Pre-screening}\label{s:optimization:experiments}

\begin{figure}
    \centering
    \minipage{0.24\linewidth}
      \includegraphics[width=\linewidth]{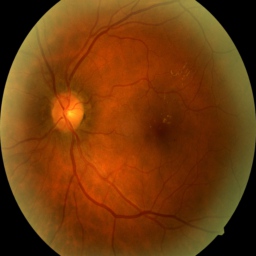}
    \endminipage \
    \minipage{0.24\linewidth}
      \includegraphics[width=\linewidth]{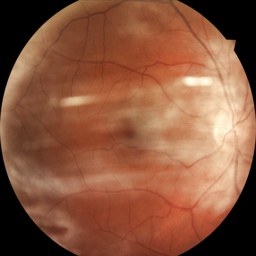}
    \endminipage \
    \minipage{0.24\linewidth}
      \includegraphics[width=\linewidth]{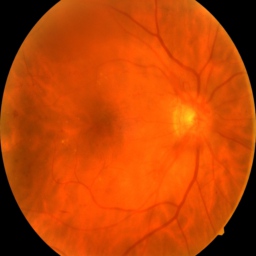}
    \endminipage \
    \minipage{0.24\linewidth}%
      \includegraphics[width=\linewidth]{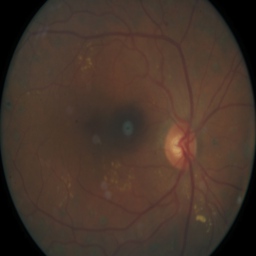}
    \endminipage
    \caption[Example retinopathy images]{Examples from retinopathy dataset.
    In order: healthy eye; healthy eye with camera artefacts;diseased eye; diseased eye.
    The chance that bad images cause misdiagnosis makes uncertainty-aware models vital.}
    \label{fig:eye_examples}
\end{figure}

Here, we establish the robustness and performance of Radial BNNs using a Bayesian medical imaging task identifying diabetic retinopathy in `fundus' eye images \citep{leibigLeveraging2017}, using models with $\sim$15M parameters and inputs with $\sim$260,000 dimensions, in \cref{s:retinopathy} (see \cref{fig:eye_examples}).
In later chapters, we return to the Radial BNN posterior when demonstrating evaluations based on active and continual learning.

In our experiments, Radial BNNs are more robust to hyperparameter choice than MFVI and that Radial BNNs outperform the current state-of-the-art Monte-Carlo (MC) dropout and deep ensembles on this task.\footnote{In some recent work including \citet{nadoUncertainty2021} Radial BNNs perform worse than MFVI or deep ensemble methods.
There are several possible explanations.
First, many implementations use a variety of implicit `hacks' for MFVI that have evolved through countless hours of researcher time.
It may be that similar tuning would let Radial BNNs to reach higher levels of performance.
Second, our experiments might overestimate the performance of Radial BNNs.
Although we made every effort to tune the hyperparameters of all models using the same number of seeds and Bayesian optimization, I wanted my method to perform better and this may have subtly influenced experimental decisions.
Third, details of the experiments and architecture varied between settings.
For example, we used multiple variational samples in the forward pass and pretrained the means using maximum likelihood loss for several epochs before performing variational inference, and used VGG-style architectures.}
Our work is focused on large datasets and big models, which is where the most exciting application for deep learning are.
That is where complicated variational inference methods that try to learn weight covariances become intractable, and where the `soap-bubble' pathology emerges.
We show that on a large-scale diabetic retinopathy diagnosis image classification task: our radial posterior is more accurate, has better calibrated uncertainty, and is more robust to hyperparameters than MFVI with a multivariate Gaussian and therefore requires fewer iterations and less experimenter time.

\subsubsection{Dataset description}
\label{s:retinopathy}

We perform classification on a dataset of `fundus' images taken of the back of retinas in order to diagnose diabetic retinopathy \citep{kaggleDiabetic2015} building on \citet{leibigLeveraging2017} and \citet{filosBenchmarking2019}.\footnote{The precise dataset and augmentations which we used came from an early version of \citet{filosBenchmarking2019}, which is now different from the publicly available dataset. We describe the data augmentations, which differ from the current benchmark, in detail in \cref{a:experimental_settings} so that the experiments could be reproduced if desired.}
Diabetic retinopathy is graded in five stages, where 0 is healthy and 4 is the worst.
Following \citet{leibigLeveraging2017}, we distinguish the healthy (classes 0 and 1) from those that require medical observation and attention (2, 3, and 4).
Images (512x512) include left and right eyes separately, which are not considered as a pair by the models, and come from two different camera technologies in many different physical locations.
Model uncertainty is used to identify badly-taken or confusing images which could be used to refer affected patients to experts for more detailed examination.

The diabetic retinopathy dataset has several properties which differ from standard image classification datasets but are similar to many real-world settings.
First, there is substantial class imbalance, with many more healthy than unhealthy images.
Second, the consequences of misclassification are much greater in the case of a false negative (a sick person does not receive treatment) than a false positive (a healthy person receives an unnecessary but non-invasive follow-up examination).
Third, the dataset contains a mixture of left- and right- eyes which appear slightly differently as well as images taken using two different camera orientations (coded using a notch at the top and bottom right of the image).
These make data-augmentation using reflective symmetries particularly important for training.

\subsubsection{Performance and Calibration}
\begin{figure}
\centering
\end{figure}
In \cref{tab:auc} we compare the classification area under the curve (AUC) of the receiver operating characteristic of predicted classes (higher is better).\footnote{We follow prior work which selected AUC as a metric because classes are unbalanced (mostly healthy) which makes the AUC more reflective of model performance than something like accuracy. Plausibly better would be to construct a loss which takes the relative value of different kinds of misclassification into account.}
We consider the model performance under different thresholds for referring data to experts.
At 0\%, the model makes predictions about all data.
At 30\%, the 30\% of images about which the model is least confident are referred to experts and do not get scored for the model---the AUC should therefore become higher if the uncertainties are well-calibrated.
We show that our Radial BNN outperforms MFVI by a wide margin, and even outperforms MC dropout.
While the deep ensemble is better at estimating uncertainty than a single Radial BNN, it has three times as many parameters.
An ensemble of Radial BNNs outperforms deep ensembles at all levels of uncertainty.
Radial BNN models trained on this dataset also show empirical calibration that is closer to optimal than other methods (see \cref{fig:calibration}).

The model hyperparameters were all selected individually by Bayesian optimization using ten runs.
Full hyperparameters and search strategy, preprocessing, and architecture are provided in \cref{a:dr_hypers}.
We include both the original MC dropout results from \citet{leibigLeveraging2017} as well as our reimplementation using the same model architecture as our Radial BNN model.
The only difference between the MC dropout and Radial BNN/MFVI architectures is that we use more channels for MC dropout, so that the number of parameters is the same in all models.
We pick the VGG-16 architecture for comparability with prior work, but note that \citet{nadoUncertainty2021} evaluate performance on a ResNet.
As discussed above, they find that a method similar to our `MFVI w/ tweaks' outperforms Radial BNNs, which might mean that the behaviours described in \cref{s:understanding} are alleviated by the presence of residual connections.
We estimate the standard error of the AUC using bootstrapping.

\begin{tcolorbox}[title=The difficulty of `naive' mean-field VI, colback=white, parbox=false,breakable=true]
    \label{s:tweaks}
In practice, training BNNs with MFVI is difficult and people adopt unprincipled `hacks'.
For example, in order to produce variational inference methods that work on larger datasets and architectures, \citep{nadoUncertainty2021} use a number of `hacks' including adjusting the `prior' to always have the same mean as the current approximate posterior and scaling the KL-divergence term of the loss between 0.5\% and 45\% of its true value during training.
Indeed, because they use learning-rate decay schedules, even if they had increased the KL-divergence scaling term all the way to $1$ by the end of training, one could not expect that the approximate posterior was even close to the optimum of the ELBO.
As a result, their implementation is far from performing principled variational inference.
They are not an outlier in this regard, but are selected only because they have a prominent, public, and well-documented code-base.
These tricks are widely used.
Some further tricks include performing early-stopping alongside a KL-annealing schedule or a tiny initialized variance (e.g., \citet{nguyenVariational2018}) or simply performing `variational inference' using the log-likelihood loss only \citep{fortunatoNoisy2018}.

Others have attempted to scale mean-field variational inference to larger settings, such as \cite{osawaPractical2019} who use variational online Gauss-Newton methods \citep{khanFast2018}.
However, this method relies on several significant approximations to make estimates of the Hessian tractable and the performance of the method lags significantly behind deep ensembles.

From an alternative perspective, \citet{wuDeterministic2019} argue that mean-field VI is sensitive to initialization and priors.
They see the variance of ELBO estimates as the problem and introduce a deterministic alternative.
We agree that this is a crucial problem, but offer a simpler and cheaper alternative solution which does not require extra assumptions about the distribution of activations.
\end{tcolorbox}

\subsubsection{Robustness}\label{s:robustness}
\begin{table*}
        \centering
        \resizebox{\textwidth}{!}{\begin{tabular}{llcccccc}
            \toprule
             \multirow{2}{*}{Method} &
             \multirow{2}{*}{Architecture} &
             \multirow{2}{*}{\shortstack{\# \\ Params}} &
             \multirow{2}{*}{\shortstack{Epoch Train \\ Time (m)}} &
             \multicolumn{4}{c}{ROC-AUC for different percent data referred to experts}\\
             &&&&0\% & 10\% & 20\% & 30\%
             \\
             \midrule
             MC-dropout & [Leibig et al., 2017] & $\sim$21M & -& 92.7$\pm$0.3\%&93.8$\pm$0.3\%&94.7$\pm$0.3\%&95.6$\pm$0.3\%\\
             MC-dropout & VGG-16 & $\sim$15M & 5.6 & 93.0$\pm$0.04\% &94.1$\pm$0.05\% &94.5$\pm$0.05\% &95.1$\pm$0.07\%\\
             MFVI & VGG-16* & $\sim$15M & 16.0 & 63.6$\pm$0.13\% &63.5$\pm$0.09\% &63.5$\pm$0.09\% &62.6$\pm$0.10\% \\
             MFVI w/ tweaks & VGG-16* & $\sim$15M & 16.0 & 93.9$\pm$0.04\% &94.4$\pm$0.05\% &95.4$\pm$0.04\% &96.4$\pm$0.05\%\\
             \textbf{Radial BNN} & VGG-16* & $\sim$15M & 16.2 & \textbf{94.3$\pm$0.04\%} &\textbf{95.3$\pm$0.06\%} &\textbf{96.1$\pm$0.06\%} &\textbf{96.8$\pm$0.04\%} \\
             \midrule
             Deep Ensemble & 3xVGG-16 & $\sim$45M & 16.8$\dagger$ & 93.9$\pm$0.04\% & 96.0$\pm$0.05\% & 96.6$\pm$0.04\% &97.2$\pm$0.04\% \\
             \textbf{Radial Ensemble} & 3xVGG-16* & $\sim$45M & 48.6$\dagger$ & \textbf{94.5$\pm$0.05\%} &\textbf{97.9$\pm$0.04\%} &\textbf{98.0$\pm$0.03\%} &\textbf{98.1$\pm$0.03\%}\\
            \bottomrule
        \end{tabular}}
        \caption[Diabetic retinopathy prescreening performance evaluation]{Diabetic Retinopathy Prescreening: Our Radial BNN outperforms SOTA MC-dropout and is able to scale to model sizes that MFVI cannot handle without ad-hoc tweaks (see \cref{s:tweaks}). Even with tweaks, Radial BNN still outperforms. Deep Ensembles outperform a single Radial BNN at estimating uncertainty, but are worse than an ensemble of Radial BNNs with the same number of parameters. $\pm$ indicates bootstrapped standard error from 100 resamples of the test data.
        We use an adapted VGG-16* model for our Bayesian deep learning methods which have fewer channels so that \# of parameters is the same as non-Bayesian model.
        This architecture was chosen to match prior work \citep{leibigLeveraging2017, filosBenchmarking2019}.
        \citet{nadoUncertainty2021} compare Radial BNNs to other variational methods on residual networks rather than the VGG architecture, and find the performance of their other baselines (most comparable to MFVI w/ tweaks in design) is slightly better. This may be because residual connections reduce the effects we observed in this paper.
        $\dagger$ three times the train time of a single model, though this could be parallelized.}
        \label{tab:auc}
\end{table*}
\begin{figure}
\centering
\begin{subfigure}[t]{0.48\textwidth}
    \includegraphics[width=\textwidth]{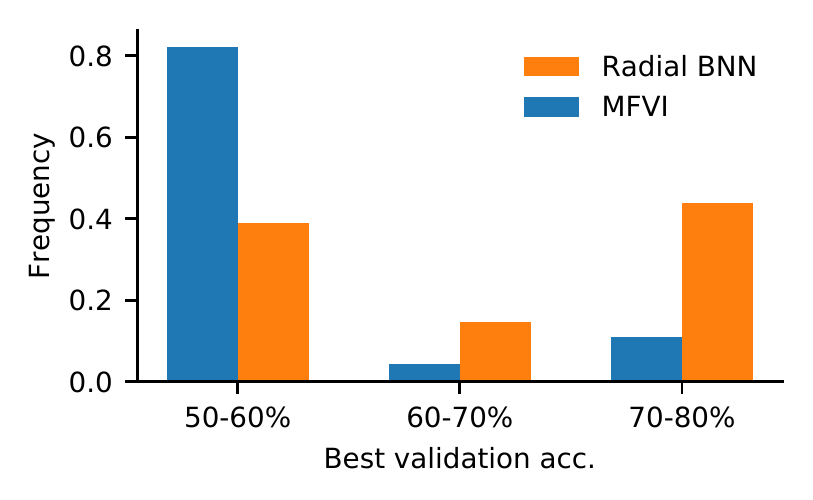}
    \caption{}
    \label{fig:dr_robust}
\end{subfigure}
\begin{subfigure}[t]{0.48\textwidth}
    \includegraphics[width=\textwidth]{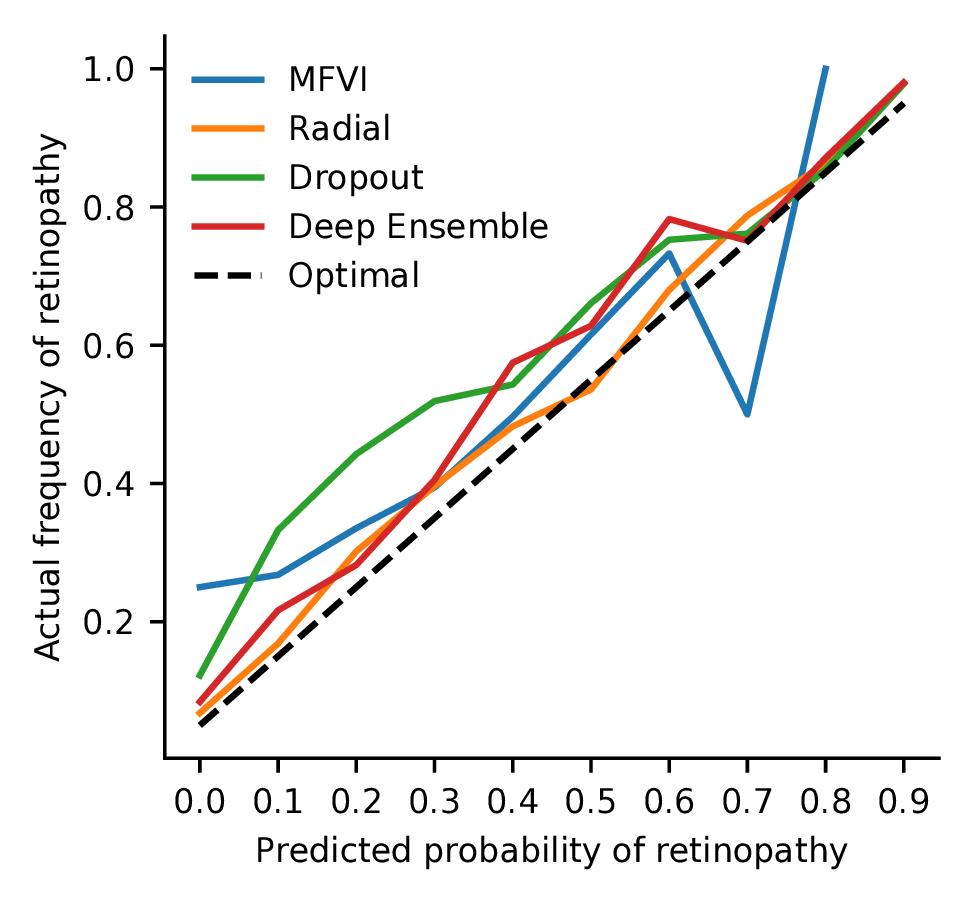}
    \caption{}
    \label{fig:calibration}
\end{subfigure}
  \caption[Hyperparameter robustness on diabetic retinopathy]{
      (a) The Radial BNN posterior is more robust to hyperparameters on a downsampled version of the retinopathy dataset.
      Over 80\% of configurations for the MFVI baseline learned almost nothing.
      4 times more Radial BNNs had good accuracies than MFVI models.
      (b) Radial BNN is almost perfectly calibrated, compared with MC dropout and deep ensembles (overconfident) and ordinary MFVI without ad-hoc tweaks which is not well calibrated.
      X-axis labels are the lower-bound of each range (e.g., 0.0 is 0.0-0.1).}
\end{figure}
The radial posterior was more robust to hyperparameter variation (Figure \ref{fig:dr_robust}).
We assess robustness on a downsampled version of the diabetic retinopathy dataset (256x256) using a smaller model with a similar architecture to VGG-16, but which trained to convergence in about a tenth the time and had only $\sim$1.3M parameters.
We randomly selected 86 different runs from plausible optimizer, learning rate, learning rate decay, batch size, number of variational samples per forward pass, and initial variance.
82\% of hyperparameters tried for the MFVI baseline resulted in barely any improvement over randomly guessing, compared with 39\% for the radial posterior.
44\% of configurations for our radial posterior reached good AUCs, compared with only 11\% for MFVI.
This is despite the fact that we did allow models to pre-train the means using a negative log-likelihood loss for one epoch before beginning ELBO training, a common tweak to improve MFVI.

\begin{tcolorbox}[title=UCI experiments and Bayesian deep learning, colback=white, parbox=false,breakable]
These retinopathy experiments are orders of magnitude larger than most other VI work at the time, which was heavily influenced by experiments on UCI datasets from \citet{hernandez-lobatoProbabilistic2015} with between 4 and 16 input dimensions and using fewer than 2000 parameters.

Radial BNNs, like MFVI, perform worse than more expensive methods on the UCI datasets.
This is expected---our method is designed for models with high-dimensional weight-space, not for the artificial constraints of the experimental settings used on the UCI evaluations.
An over-reliance on UCI focuses on settings where Bayesian deep learning is not particularly suitable because the advantages of neural networks are not critical.
Recently, researchers have moved towards more large-scale evaluations for uncertainty.

Nevertheless, completely abandoning UCI may also be a mistake, because it is important to understand the shortcomings that Bayesian deep learning can show on regression problems.
Most recent Bayesian deep learning work (including this thesis) overly emphasises computer vision classification.
These are interesting, important, and a relevant application where deep learning has been phenomenally successful.
Nevertheless, regression is an important category of prediction problem where we have specific reasons to expect many current methods to perform less well \citep{foongExpressiveness2020}.
Most importantly, it is possible to be extraordinarily wrong about regression in a way that is impossible with classification (assuming a constant misclassification utility).
If we do stop using UCI datasets to evaluate Bayesian deep learning, we must be careful to also incorporate other regression problems into standard evaluations.

\vspace{6mm}
{\centering
\resizebox{\columnwidth}{!}{%
\begin{tabular}{lccccccccc}
    
 \textbf{Dataset} &  \multicolumn{1}{c}{\textbf{MFVI}} & \multicolumn{1}{c}{\textbf{Radial}} & \multicolumn{1}{c}{\textbf{Dropout}} & \multicolumn{1}{c}{\textbf{VMG}}& \multicolumn{1}{c}{\textbf{FBNN}} & \multicolumn{1}{c}{\textbf{PBP\_MV}} & \multicolumn{1}{c}{\textbf{DVI}}\\\hline
 & \multicolumn{7}{c}{Avg. Test LL and Std. Errors}\\ \hline
Boston & -2.58$\pm$0.06 & -2.58$\pm$ 0.05 & -2.46$\pm$0.25 & -2.46$\pm$0.09 & -2.30$\pm$0.04 & 02.54$\pm$0.08 & -2.41$\pm$0.02 & \\
        Concrete &-5.08$\pm$0.01 &-5.08$\pm$0.01 & -3.04$\pm$0.09&-3.01$\pm$0.03& -3.10$\pm$0.01 & -3.04$\pm$0.03 & -3.06$\pm$0.01\\
        Energy & -1.05$\pm$0.01& -0.91$\pm$0.03 & -1.99$\pm$0.09 & -1.06$\pm$0.03 & \textbf{-0.68$\pm$0.02} & -1.01$\pm$0.01 & -1.01 $\pm$ 0.06\\
        Kin8nm  & 1.08$\pm$0.01 & \textbf{1.35$\pm$0.00} & 0.95$\pm$0.03 &1.10$\pm$0.01 & - & 1.28$\pm$0.01 & 1.13$\pm$0.00\\
        Naval  &-1.57$\pm$0.01& -1.58$\pm$0.01 & 3.80$\pm$0.05& 2.46$\pm$0.00 & \textbf{7.13$\pm$0.02}& 4.85$\pm$0.06 & 6.29$\pm$0.04\\
        Pow. Plant  & -7.54$\pm$0.00 & -7.54$\pm$0.00 & -2.80$\pm$0.05 &-2.82$\pm$0.01 & - & -2.78$\pm$0.01 & -2.80$\pm$0.00\\
       Protein &-3.67$\pm$0.00&-3.66$\pm$0.00&-2.89$\pm$0.01 &-2.84$\pm$0.00 & -2.89$\pm$0.00 & -2.77$\pm$0.01 & -2.85$\pm$0.01\\
        Wine &-3.15$\pm$0.01 & -3.15$\pm$0.01 & -0.93$\pm$0.06 & -0.95$\pm$0.01 & -1.04$\pm$0.01 & -0.97$\pm$0.01 & \textbf{-0.90$\pm$0.01}\\
        Yacht &-4.20$\pm$0.05 &-4.20$\pm$0.05& -1.55$\pm$0.12&-1.30$\pm$0.02 & -1.03$\pm$0.03 & -1.64$\pm$0.02 & \textbf{-0.47$\pm$0.03}\\ \hline
        & \multicolumn{7}{c}{Avg. Test RMSE and Std. Errors} \\\hline
Boston & 3.42$\pm$0.23 & 3.36$\pm$0.23 & 2.97$\pm$0.85 &2.70$\pm$0.13 & 2.38$\pm$0.10 & 3.11$\pm$0.15 & -\\
        Concrete &5.71$\pm$0.15& 5.62$\pm$0.14 & 5.23$\pm$ 0.53 &4.89$\pm$0.12 & 4.94$\pm$0.18& 5.08$\pm$0.14& -\\
        Energy & 0.81$\pm$0.08& 0.66$\pm$0.03 & 1.66$\pm$0.19 & 0.54$\pm$0.02 & 0.41$\pm$0.20 &0.45$\pm$0.01 & -\\
        Kin8nm  & 0.37$\pm$0.00 & 0.16$\pm$0.00& 0.10$\pm$0.00&0.08$\pm$0.00 & - & \textbf{0.07$\pm$0.00} &-\\
        Naval & 0.01$\pm$0.00 & 0.01$\pm$0.00 & 0.01$\pm$0.00 &0.00$\pm$0.00 & 0.00$\pm$0.00 & 0.00$\pm$0.00&-\\
        Pow. Plant  &4.02$\pm$0.04 & 4.04$\pm$0.04 & 4.02$\pm$0.18 & 4.04$\pm$0.04 & - & 3.91$\pm$0.14&-\\
        Protein & 4.40$\pm$0.02 & 4.34$\pm$0.03 & 4.36$\pm$0.04 & 4.13$\pm$0.02& 4.33$\pm$0.03& 3.94$\pm$0.02&-\\
        Wine & 0.65$\pm$0.01 & 0.64$\pm$0.01 & 0.62$\pm$0.04 & 0.63$\pm$0.01 & 0.67$\pm$0.01 & 0.64$\pm$0.01&-\\
        Yacht &1.75$\pm$0.42& 1.86$\pm$0.37&1.11$\pm$0.38&0.71$\pm$0.05& \textbf{0.61$\pm$0.07} & 0.81$\pm$0.06&-\\
    \end{tabular}%
   }
    \captionof{table}[Comparison of mean-field VI and Radial Bayesian neural networks on UCI datasets]{Avg.\ test RMSE, predictive log-likelihood and s.e.\ for UCI regression datasets. Bold is where one model is better than the next best $\pm$ their standard error.
    Results are from multiple papers and hyperparameter search is not necessarily consistent.
    MFVI and Radial are our implementations of standard MFVI and our proposed model respectively.
    Dropout results are from \citet{galUncertainty2016}.
    Variational Matrix Gaussian (VMG) results are from \citet{louizosStructured2016}.
    Functional Bayesian Neural Networks (FBNN) results are from \citet{sunFunctional2019}.
    Probabilistic Backpropagation Matrix Variate Gaussian (PBP\_MV) results are from \citet{sunLearning2017}.
    Deterministic VI (DVI) results are from \citet{wuDeterministic2019}.}}
\end{tcolorbox}

\section{Discussion and analysis}\label{s:understanding}
In \cref{s:optimization:method} we observed that the multivariate Gaussian distribution typically used in MFVI features a `soap-bubble'---almost all of the probability mass is clustered at a radius proportional to $\sigma\sqrt{D}$ from the mean in the large $D$ limit (illustrated in Figure \ref{fig:pdf_plot}).
This has two consequences in larger models.
First, unless the weight variances are very small, a typical sample from the posterior has a high $L_2$ distance from the means.
Second, because the mass is distributed uniformly over the hypersphere that the `soap-bubble' clusters around, each sample from the multivariate Gaussian has a high expected $L_2$ distance from every other sample (similarly proportional to $\sigma\sqrt{D}$).
This means that as $\sigma$ and $D$ grow, samples from the posterior are very different from each other, which we might expect to result in high gradient variance.

In contrast, in Radial BNNs the expected distance between samples from the posterior is independent of $D$ for the dimensionality typical of neural networks.
The expected $L_2$ distance between samples from a unit hypersphere rapidly tends to $\sqrt{2}$ as the number of dimensions increases.
Since the radial dimension is also independent of $D$, the expected $L_2$ distance between samples from the Radial BNN is independent of $D$.
This means that, even in large networks, samples from the Radial BNN will tend to be more representative of each other.

As a result, we might expect that the gradient variance is less of a problem.
We consider this experimentally by examining several hypotheses:
\begin{description}
    \item[MFVI variance] The variance of gradient estimates for MFVI explodes as the standard deviation of the approximate posterior grows.
    \item[Radial variance] Meanwhile, the variance of gradient estimates for Radial BNNs does not explode as the standard deviation of the approximate posterior grows.
    \item[MFVI problem] The exploding variance of gradient estimates is harmful for MFVI training.
    \item[MFVI mechanism] The mechanism for the failure of MFVI is that the approximate posterior overfits to the loss reflecting the cross-entropy to the prior rather than the log-likelihood because it is low-variance even for large $\sigma$.
\end{description}

\subsection{MFVI and Radial variance}
\begin{figure}
\begin{subfigure}[b]{0.48\textwidth}
    \includegraphics[width=\textwidth]{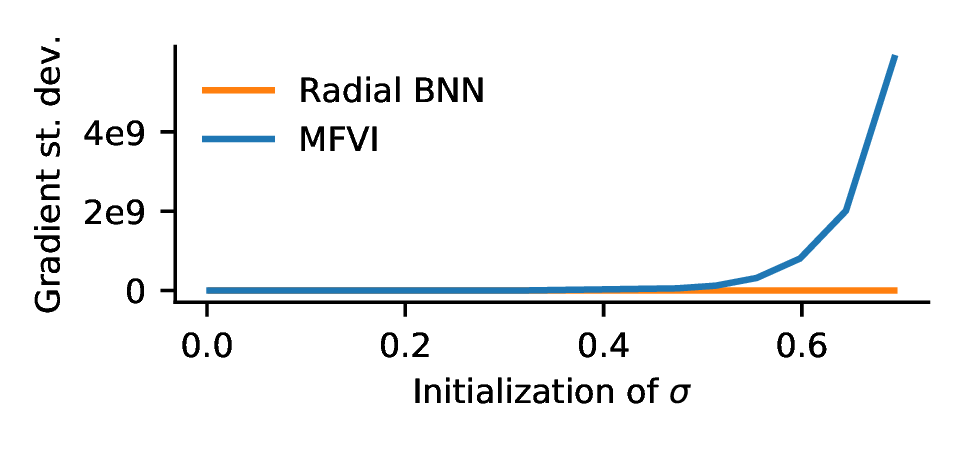}
    \caption{}\label{fig:gradients}
\end{subfigure}
\begin{subfigure}[b]{0.48\textwidth}
    \includegraphics[width=\textwidth]{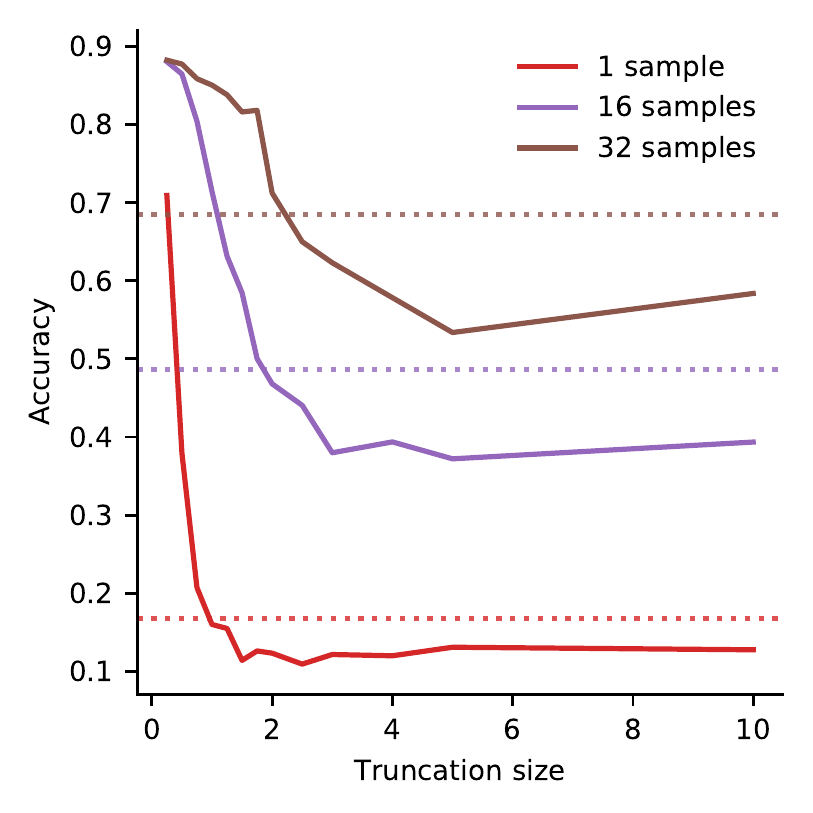}
    \caption{}\label{fig:truncation}
\end{subfigure}
\caption[Diagnosing high variance estimators of the optimization target using a truncated proxy]{(a) The variance of gradient estimates in the standard MFVI posterior explodes as the weight variance parameter grows.
(b) Dotted lines show untruncated Gaussian performance. Highly truncated Gaussians \textit{improve} MFVI. This effect is most significant when small numbers of samples from the posterior are used to estimate the gradient. We conclude that despite bias, the low variance offered by truncation improves gradient estimates. Results averaged over 10 initial seeds for each truncation size.} 
\end{figure}

We consider the first two hypotheses in \cref{fig:gradients}.
We show that for the standard MFVI posterior, the variance of initial gradients explodes after the weight standard deviation exceeds roughly $0.3$.
Meanwhile, for Radial BNNs this does not occur.
This figure shows a 3x3 conv layer with 512 channels, but we observe a similar pattern regardless of the specifics of the architecture.
This matters because, for MFVI with a unit Gaussian prior, the KL-divergence term of the loss is minimized by $\sigma_i=1$---well within the region where gradient noise has exploded.

\subsection{MFVI problem}
We can show that the gradient variance is hurting training for MFVI by using an alternative scheme that uses the same approximate posterior but has a low-variance and biased estimator of the gradient.
We do this by truncating the sampling distribution using rejection sampling, selecting only samples from a Gaussian distribution which fall under a threshold.
Our new estimate of the loss is biased (because we are not sampling from the distribution used to compute the KL divergence) but has lower variance (because only samples near the mean are used).

In \cref{fig:truncation} we show that the truncated models to outperform `correct' MFVI with standard deviations initialized at $\sigma=0.12$.
Moreover, if our hypothesis is correct, smaller the number of variational samples should result in a higher variance and therefore a larger effect size.
This is what we observe.

\subsection{MFVI mechanism}
\begin{figure}
    \centering
\includegraphics{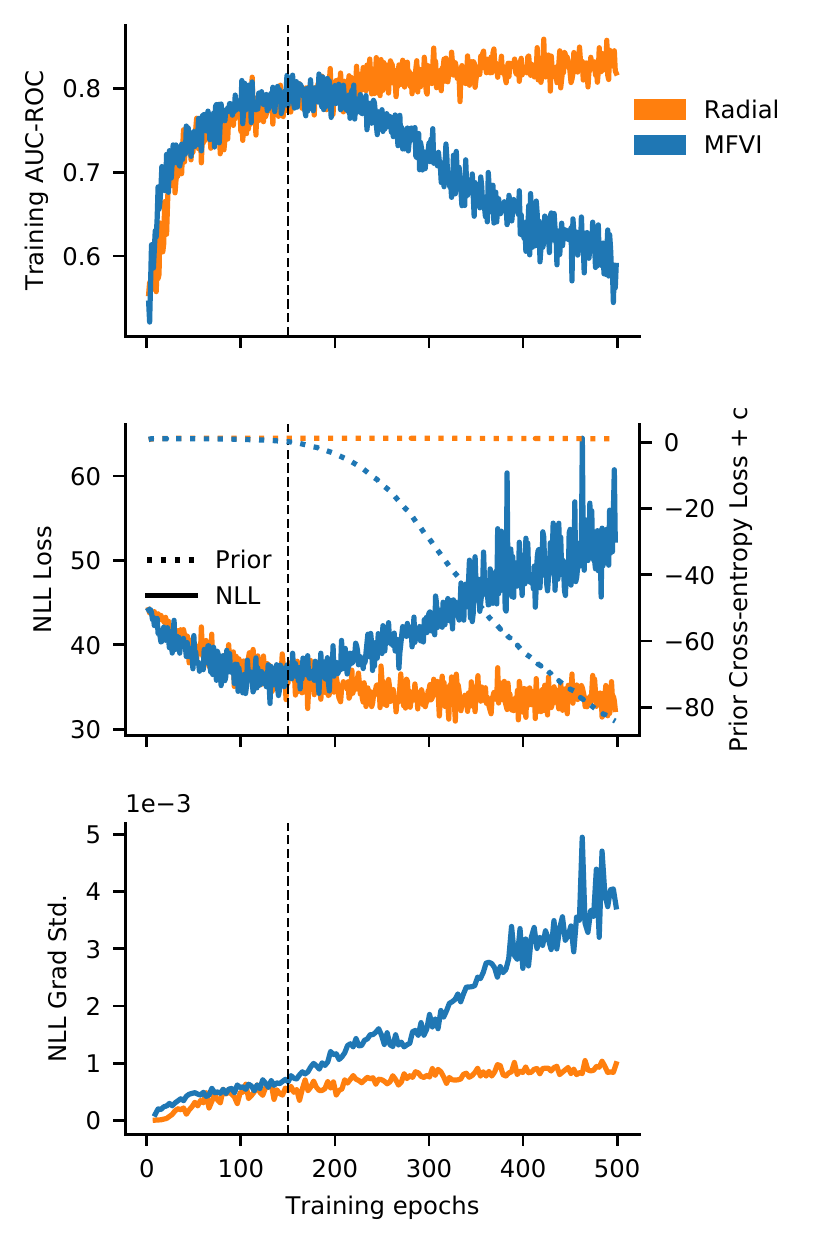}
\caption[Diagnosing optimization failures for multivariate Gaussians]{We can track the deterioration of the MFVI training dynamics. \textbf{Top}: After $\sim$150 epochs (dashed line) \textit{training} set performance degrades for MFVI while Radial continues to improve. \textbf{Middle}: for MFVI, the NLL term of the loss \textit{increases} during training, but the prior cross-entropy term falls faster so the overall loss continues to fall. \textbf{Bottom}: The standard deviation of the NLL gradient estimator grows sharply for MFVI after about 150 epochs. This coincides with the moment where the loss is optimized by minimizing the prior cross-entropy and sacrificing the NLL.} \label{fig:training_loss}
\end{figure}

Last, we examine the specific mechanism by which MFVI is failing due to the high variance gradient estimates.
In Figure \ref{fig:training_loss} we show a sample training run on the down-sampled version of the diabetic retinopathy dataset using an MFVI and Radial BNN with the same hyperparameters.
Pathologically, the \textit{training} accuracy falls for MFVI after about 150 epochs (top graph).
The critical moment corresponds to the point where the training process begins to optimize the prior cross-entropy term of the loss, sacrificing the negative log-likelihood term (middle graph).
We can further show that this corresponds to the point where the standard deviation of the negative log-likelihood term of the gradient begins to sharply increase for MFVI.
Meanwhile, the prior cross-entropy term is computed analytically, so its variance does not grow as the values of $\sigma$ increase during training from their tiny initializations.

This suggests that, MFVI fails in this setting because the high variance of the negative log-likelihood (NLL) term of the loss causes the optimizer to improve the cross-entropy term at the expense of the NLL term.
For Radial BNNs, however, the NLL gradient variance stays low throughout training.
Of course, it is possible that in other settings (architecture and data) the behaviour is different.
The fact that \citet{nadoUncertainty2021} do not find a benefit to using Radial BNNs in ResNets relative to their `tweaked' MFVI might mean that the benefit in ResNets is smaller (although they do not compare to a vanilla implementation of MFVI).

In addition, it is possible that where an MFVI approximate posterior \textit{is} able to optimize successfully, the requirement that samples from it behave somewhat similarly in function-space might result in especially flat minima, possibly resulting in better generalization.

\subsection{Avoiding the Pathology in MFVI}
Based on this analysis, we can see why Radial BNNs fix a sampling problem in MFVI.
But this also helps explain why the ad-hoc tweaks which researchers have been using for MFVI have been successful.
These tweaks chiefly serve to keep the weight variance low.
Researchers initialize with small variances \citep{blundellWeight2015, fortunatoBayesian2017,nguyenVariational2018}.
Sometimes they adapt the loss function to remove or reduce the weight of the KL-divergence term, which reduces the pressure on weight variances to grow \citep{fortunatoNoisy2018}.
Other times researchers pre-train the means with just the NLL loss, which makes it possible to stop training after relatively little training on the ELBO loss, which stops the variances from growing too much \citep{nguyenVariational2018}.
Another approach, which we have not seen tried, would be to use a very tight zero-centred prior, effectively enforcing the desire to have a basically deterministic network (a prior inversely proportional to $\sqrt{D}$ would balance the `soap-bubble' variance).
However, this sort of very tight prior is not compatible with the use of data-dependent priors in sequential learning.
\Citet{nadoUncertainty2021} instead centre the `prior' at the current mean of the approximate posterior and move the `prior' during training, which they find performs well although it departs from the standard motivation for priors.

For most of these tweaks, the resulting network is not fully optimizing the ELBO.
This does not necessarily make the resulting network useless---after all, the ELBO is only a bound on the actual model evidence.
However, if we have a theoretically principled way to fix our sampling problems without resorting to ad-hoc tweaks, then we should prefer that.
Radial BNNs offer exactly that theoretically principled fix.

\section{Conclusion}
In this chapter, we have demonstrated how the choice of approximating distribution can interact with the optimization procedure for approximate Bayesian methods that employ optimization.
We demonstrated an interaction in mean-field variational inference caused by the interaction between optimization and approximation, which are independent of the hypothesized restrictiveness of the mean-field approximation discussed in the previous chapter.
By examining this interaction, we have proposed an alternative approximating distribution whose optimization and prediction are computationally equivalent to the multivariate Gaussian approximate posterior.
In different architectures and datasets, it is possible that we would not observe the same interactions that we observed here.
In that case, the proposed fix would be less likely to work---but it would nevertheless demonstrate the same core point: that architecture, data, and approximation scheme can interact in surprising ways that affect the quality of the Bayesian approximation.

\ifSubfilesClassLoaded{
\bibliographystyle{plainnat}
\bibliography{thesis_references}}{}
\end{document}

\chapter{Evaluating Bayesian Deep Learning}
\label{chp:evaluation}

Evaluating Bayesian deep learning is difficult.
In some sense, the ideal evaluation could be a comparison between an approximate and true posterior distribution but, as we discuss in \cref{chp:challenges}, establishing a satisfactory distance measure is challenging (c.f. \ref{s:challenges:conceptual_challenges_for_approximate_bayes}).
Even setting this challenge aside, HMC remains our best tool for sampling from the true posterior and it is difficult and expensive for modern architectures although high-quality reference samples exist for, for example, ResNets and have been used for benchmarking evaluations \citep{wilsonEvaluating2021}.

In the context of model misspecification, \citet{keyBayesian1999} propose casting model evaluation as a Bayesian decision problem and evaluating the expected utility of the posterior distribution on a `generic' utility function, such as the log-likelihood.
In fact, this can be seen as a solution to the approximate inference conundrum as well.
The trouble is that different applications interact with nuances of the approximations, which makes a generic utility less appropriate.

In this chapter, we consider two archetypal use cases for Bayesian deep learning which test different aspects of the `Bayesian' procedure.
Both of these build off what I see as the key characteristic of approximate Bayesian methods: that the parameter distributions of our models reflect our beliefs about states of affairs.
First, we consider continual learning as a test of the ability of an approximate posterior to be used as part of a sequential Bayesian updating scheme.
Second, we consider active learning as a test of the ability of an approximate posterior to represent its subjective epistemic state.
Both of these are key properties associated with the Bayesian approach.

A natural way to evaluate approximate Bayesian inference, therefore, is to evaluate the use a Bayesian method to accomplish these two tasks.
For continual learning, this might mean adopting a specific Bayesian approximation, using it for sequential inference, and evaluating the methods using the performance of the final model over a sequence of tasks.
For active learning, it might mean estimating the expected information gain under a Bayesian approximation of the posterior to acquire data and measuring the performance of the final model.

Unfortunately, neither of these should be regarded as an effective evaluation of approximate inference.
Specifically, we show that most existing evaluations of continual learning `abuse' the evaluation structure to \textit{bypass} the Bayesian approximation.
We identify one effective evaluation setting, but note that existing posterior approximations perform extremely badly on it.
We further show that active learning performance depends significantly on implicit biases introduced by the acquisition procedure, and that the extent of this bias is different for different models in a manner which does not depend on the quality of the approximate posterior.
This recalls the results of \cref{chp:parameterization} and \cref{chp:optimization} which showed how the action of the approximation scheme depends on the model architecture; here even the action of the \textit{evaluation} scheme depends on the architecture.
Removing the bias at least removes one source of differing performance between models that is unrelated to Bayesian inference.
However, we do not guarantee that active learning with the bias removed might not suffer from some \textit{other} confounding factor when used as an evaluation of approximate Bayesian inference.

\section{Continual Learning}
\label{s:evaluation:continual_learning}
Bayesians do not all agree on what exactly `Bayesianism' entails.
One consistent element, which is especially clear in the work of \citet{jaynesProbability2003} and \citet{coxAlgebra1961}, is the requirement that a Bayesian's subjective beliefs ought to follow the laws of probability as further observations are obtained.
At the same time, an explicit axiom in some cases, a Bayesian's subjective beliefs ought to depend only on their prior and the observations and not on the order in which the data were processed.

All of these things are put under strain in continual learning, a challenge which poses significant difficulties for Bayesian inference in neural networks.

\subsection{The Continual Learning Challenge}
Continual learning requires that a model is trained on multiple datasets arriving in series, each of which is then forgotten before the next dataset arrives \citep{kirkpatrickOvercoming2017,zenkeContinual2017}.
These datasets are often called \textit{tasks}---although this term is loaded in a way that is not enforced in the formalism.
The goal is that the model learns to perform as well as it would have had it trained on the entire dataset at the same time, which involves both avoiding forgetting information learned on earlier as well as transferring information from early tasks forwards \citep{chaudhryRiemannian2018}.
The test dataset is therefore generally, but not always, treated as a balanced mixture of the individual task distributions.

Continual learning is related to a number of other interesting machine learning problems.
\begin{description}
    \item[Online learning] \citep{opperBayesian1999} differs in that data arrive in sequence one-by-one.
    This is equivalent to continual learning in the special case that each task is a single datapoint. 
    \item[Meta learning] \citep{rendellLayered1987} differs in that the multiple datasets are available at the same time and that the test distribution is generally separate from the training tasks in some way.
    \item[Multi-task learning] \citep{caruanaMultitask1997} differs in that the multiple datasets are available at the same time.
    \item[Curriculum learning] \citep{bengioCurriculum2009}  differs in that the goal of curriculum learning is to actively construct the tasks/datasets so that the performance is \textit{better} than it would have been had the model been trained on all the data together. (Note that curriculum learning therefore generally \textit{relies} on the failure of neural network training to be indifferent to data-ordering.)
\end{description}
Continual learning can therefore be thought of as a form of softly online multi-task learning.

Although it is widely contended that continual learning is of practical value, in fact this is not very clear.
Foundational work on the subject \citep{zenkeContinual2017,kirkpatrickOvercoming2017} was motivated more by curiosity about the phenomenon of catastrophic forgetting \citep{robinsCatastrophic1995,choyNeural2006,goodfellowEmpirical2013} which occurs when neural networks are trained on first one dataset and then another.
(In fact, `catastrophic forgetting' is in some sense the obvious consequence of optimizing the parameters with respect to a different objective and the shocking and interesting phenomenon ought to be considered to be \textit{magical remembering}, which is a consequence of imperfect local optimization.)

Continual learning has been proposed as valuable for adapting to changing environments over time \citep{lomonacoCVPR2020}, or as useful for robotics and industrial control.
In fact these settings are often more cleanly online multitask learning because real robotic applications rarely provide data in clearly demarcated sequential sets.
The difficulty of being precise about what exactly is needed in the `continual' learning setting causes difficulties for evaluating progress in the field.
One case where regulatory constraints might demand continual learning is when privacy laws require data to be deleted but allow the storage of models derived from the data, which has motivated differentially private continual learning \citep{farquharDifferentially2018}.

Partly because of the lack of clear agreement about the intended use case for continual learning, there is considerable variation in the evaluations which are employed and the constraints which are enforced.
Our goal in this chapter is not to arbitrate which constraints and assumptions are most useful for some application or another.
Rather, our goal is to identify a set of constraints that provides the most interesting measure of the quality of an approximate posterior.

Formally, in a typical supervised learning setting, we aim to learn parameters $\vw$ using an independently and identically distributed (i.i.d.) labelled training dataset $\data \equiv \{(\vx_i, y_i)\}$ to accurately predict $p(\pred_i \mid \vx_i; \vw)$.
Instead, for continual learning we split the data into $T$ disjoint subsets $\data^t \equiv \{(\vx^t_i, y^t_i)\}$ which are individually i.i.d.\ and represent a task.

Researchers often make several further assumptions, but not consistently.
Some of these are discussed in more detail in later sections, but amongst the most common decisions are:
\begin{itemize}
    \item Is the model architecture is fixed? Or are additional parameters available for each new task?
    \item Is the task identity is known during training, during testing, or for both?
    \item Can a small amount of data (a coreset) be carried forwards? Carried backwards? Or must everything be lost?
    \item Can further data products (e.g., gradients) be carried between tasks?
\end{itemize}
The answers to these questions are absolutely pivotal, and many continual learning methods rely entirely on the ability to exploit one or more of these `loopholes' in the set-up.
Further assumptions are often made about the relationship between the tasks, and these will be considered in more detail below.

\subsection{Continual Learning as Bayesian Sequential Learning}
By considering the weights of our neural network as subjective belief distributions, we can frame continual learning as a Bayesian updating problem, following \citep{nguyenVariational2018}.
The posterior given two tasks is, using Bayes' theorem
\begin{align}
    p(\rvw \mid \data^1, \data^2) &= \frac{p(\data^1, \data^2 \mid \rvw) p(\rvw)}{p(\data^1, \data^2)},
    \intertext{and assuming that the task data are conditionally independent of each other,}
    &= \frac{p(\data^2 \mid \rvw)p(\data^1 \mid \rvw) p(\rvw)}{p(\data^1, \data^2)},
    \intertext{and lastly applying Bayes' theorem a second time}
    &=\frac{p(\data^2 \mid \rvw)p(\rvw \mid \data^1)}{p(\data^2 \mid \data^1)}.\label{eq:bayesian_continual_learning}
\end{align}
This is identical to Bayes' theorem for $p(\rvw \mid \data^2)$ except that the `prior' distribution for $\rvw$ is given by the posterior on the first dataset, $p(\rvw \mid \data^1)$, and that the normalization term is now conditioned on the prior datasets.
That is to say, the posterior given both tasks is the same as the posterior given the second task where we already use the first task's posterior as the prior.
\citet{nguyenVariational2018} propose using approximate inference for this latest Bayesian updating step, and further observed that the decomposition can be applied recursively for arbitrarily many tasks.
They call their method Variational Continual Learning (VCL).
This generalizes the method of \citet{kirkpatrickOvercoming2017} which, when combined with the comment by \citet{huszarNote2018}, can be interpreted as a Laplace approximation for Bayesian sequential learning.
\citet{nguyenVariational2018} further use coresets to improve their method's performance, which turn out to be essential, but when we discuss ``VCL'' we mean to refer to the version of the method without coresets, and we will be explicit when coresets are additionally used.

The relationship is exact.
If we had access to the exact posterior, it would be precisely the same to compute the posterior step-wise rather than all at one time.
Indeed, it would be identical regardless of the order in which we performed the inference.

Note that we can choose to adopt a Bayesian perspective about both the parameter distribution and the data distribution \citep{farquharUnifying2018}.
In what follows, we will assume that we are trying to evaluate the quality of a predictive parametric distribution, which makes a framework like VCL appropriate.
If the goal were to evaluate the quality of a generative model, an alternative continual learning algorithm would be more appropriate.

\subsection{Inadequate Evaluations of Continual Learning}
\label{s:evaluations:inadequate}
Several evaluations are widely used in the literature despite being extremely limited as evaluations for continual learning algorithms, and which are also poor evaluations for Bayesian deep learning as a solution to continual learning.
In this section, we discuss two of the most common flawed evaluations: permuted MNIST, and multi-headed split evaluations.
These are flawed evaluations of continual learning generally, and when continual learning is set up as an evaluation of Bayesian sequential updating they become flawed evaluations for approximate Bayesian inference.
In recent years, research practice has often recognized the shortcomings identified in this section, which were novel at the time of publication.

\subsubsection{Permuted MNIST}
The Permuted MNIST experiment was introduced by \citet{goodfellowEmpirical2013}.
In their experiment, a model is trained on MNIST as $\mathcal{D}_1$.
Each later $\data^t$ for $1 < t \leq 10$ is constructed from the MNIST data but with the pixels of each digit randomly permuted.
A fresh permutation is drawn for each task and applied to all images in the same way for that task.
After training on each dataset, one evaluates the model on each of the previous datasets as well as the current.
\citet{goodfellowEmpirical2013} used this experiment to investigate feature extraction, but it has since become widely used for continual learning evaluation \citep{zenkeContinual2017,kirkpatrickOvercoming2017,leeOvercoming2017,lopez-pazGradient2017,nguyenVariational2018,ritterScalable2018,huOvercoming2019,chaudhryRiemannian2018}.

Unfortunately, permuted MNIST represents an unrealistic best case scenario for continual learning---although it satisfies the literal definition of continual learning.
The positions of the pixels are fully randomized---which suited its original purpose.
\citet{goodfellowEmpirical2013} investigated whether neural networks have `high-level concepts' that get re-mapped to pixel positions when the input space is permuted, and found evidence they did not.
Now, however, the experiment has been repurposed for continual learning, where it is not suitable.

An image from each permuted dataset is practically unrecognizable given previous datasets.
The actual world is almost never structured like this---new situations look confusingly similar to old ones.
A model presented with a new task in Permuted MNIST will be uncertain, while in settings that are not deliberately randomized a model will tend to make confident but false predictions.
This makes Permuted MNIST significantly different from real-world settings in a way that directly affects how new tasks are learned.
As an illustration of this, we find that the average entropy of predictions for the second task of permuted MNIST is $0.45$ while for a slightly more realistic task (split MNIST, discussed in \cref{s:evaluation:split_mnist}) the entropy is $0.003$.

Moreover, from the perspective of Bayesian inference, permuted MNIST represents an atypical inference task.
Because each task is so distinct, the likelihood term of the ELBO loss is similar in magnitude for all datapoints, resulting in a strong contribution from the prior term.
In contrast, for more natural data distributions, the likelihood term is very high in early training steps because of confident but incorrect predictions.
This reduces the regularization of the prior early in training, leading the approximate posterior to leave the local optimum for the previous task and resulting in a less suitable approximation to the joint posterior.

\begin{figure}
    \begin{subfigure}{0.48\textwidth}
        \includegraphics[height=1.2in]{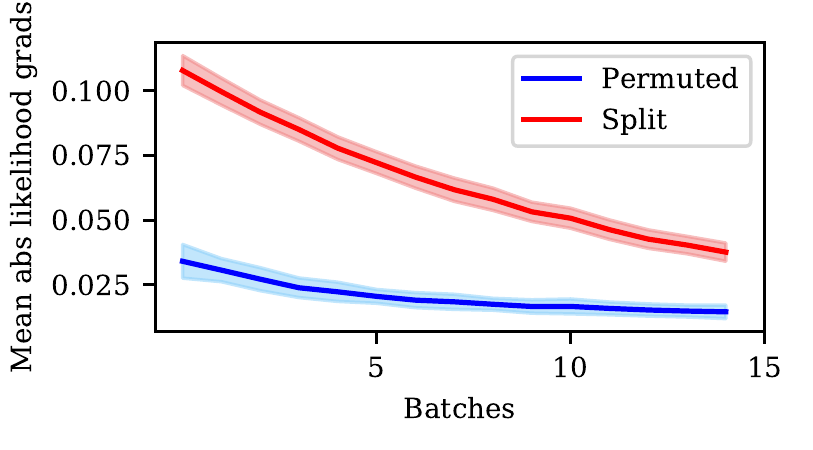}
        \label{fig:perm_prob}
        \caption{}
    \end{subfigure}
    \begin{subfigure}{0.48\textwidth}
        \includegraphics[height=1.2in]{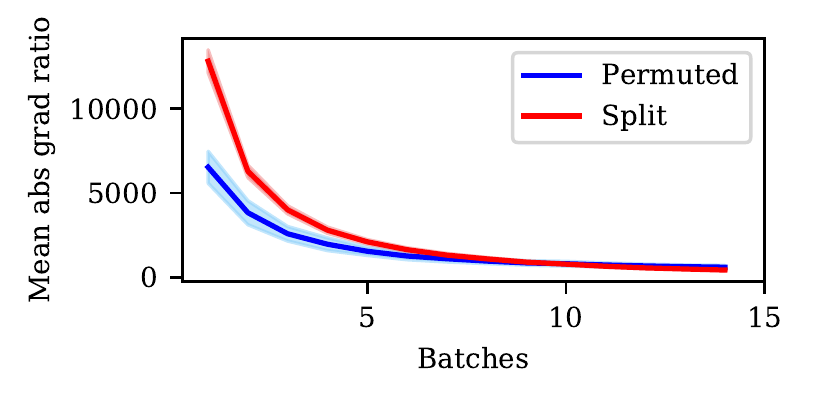}
        \caption{}
        \label{fig:perm_prob_2}
    \end{subfigure}
\caption[Permuted MNIST is an unrealistically dissimilar continuation]{Early in the second task, because permuted MNIST tasks are so disjoint, (a) the likelihood term of gradients on the final layer is unusually low (b) relative to the prior term.
This makes continual learning unrealistically easy in this evaluation.
Averaged over 100 runs, shading is one standard deviation.}
\end{figure}

\subsubsection{`Multi-headed' Continual Learning}
\label{s:evaluation:split_mnist}

The Split MNIST experiment was introduced by \citet{zenkeContinual2017} in a \textit{multi-headed} form and used by other authors including \citet{shinContinual2017,nguyenVariational2018, chaudhryRiemannian2018} (\citet{ritterScalable2018} use a two-task variant).
A number of variants have later emerged which use other datasets such as FashionMNIST, CIFAR, and Core50.
However, although they increase the difficulty of function approximation, these variants do not significantly affect the dynamics we describe here.
The experiment constructs a series of five related tasks.
The first task is to distinguish the digits (0, 1), then (2, 3) etc.

Most papers use a multi-headed variant the model prediction is constrained to be only from the two classes represented in each task.
For example, when evaluating the performance on the first task, the model only needs to predict probabilities for zero versus one.
In some cases, multi-heading is taken even further and training is only done on the head governing the specific classes included in the task \citep{zenkeContinual2017,nguyenVariational2018,ritterScalable2018}.
This effectively means that each task has a dedicated task-specific component of the parametric model.

A \textit{single-headed} version does not limit predictions during either training or testing.
As \citet{chaudhryRiemannian2018} note, the multi-headed variant is much easier to solve, but it requires knowledge of the task and which classes are represented in each task.
Multi-heading is often used in similar non-MNIST evaluations, for example, in \citet{zenkeContinual2017,nguyenVariational2018,ritterScalable2018}.
\citet{chaudhryRiemannian2018} use both a single- and multi-headed version of Split MNIST.
\citet{huOvercoming2019} use a single-headed set-up, but their method effectively learns a different head for each task.

Unfortunately, Bayesian methods that focus on parameter distributions of predictive functions for continual learning tend to look much better in multi-headed evaluations than single-headed ones.
Suppose some model has already been trained on the first four tasks of Split MNIST and is then tested on the digit `1'.
In the single-headed variant, when shown a `1' it may incorrectly predict the label is seven, which was seen more recently.
In the multi-headed variant, we \textit{knowingly assume} that the label comes from [0:1].
Because the model only needs to decide between 0 and 1 (and not even consider if the image is a 7), a multi-headed model could correctly predict the label is 1 even though the same approach will completely fail in a single-headed experiment setup.
That is to say, solving the multi-headed split MNIST problem does not require actually learning the correct posterior over the weights conditioned on the joint dataset, $p(\rvw \mid \data^0, \dots, \data^t)$, or even the slightly weaker requirement of learning the predictive posterior, $p(y_i^t \mid \vx_i^t, \data^0, \dots, \data^t)$.
Instead, it requires the much weaker condition of learning a predictive distribution with an additional task label $p(y_i^t \mid \vx_i^t, t, \data^0, \dots, \data^t)$.

The problem is, in fact, even deeper.
If we again consider \cref{eq:bayesian_continual_learning}, in which we derived the Bayesian recurrence relation which allows Bayesian continual learning, a key step was to assume that the quantity $\rvw$ was \textit{the same} in our two probabilistic expressions!
However, in the multi-headed setting, we are in fact learning \textit{multiple functions} with overlapping parameterization.
Multi-headed VCL learns $f^1$ parameterized by a shared $\rvw$ as well as task-specific $\rvw^1$ and then learns $f^2$ with $\rvw$ and $\rvw^2$.
So beginning as we did before we have
\begin{align}
    p(\rvw, \rvw^1, \rvw^2 \mid \data^1, \data^2) &= \frac{p(\data^1, \data^2 \mid \rvw, \rvw^1, \rvw^2) p(\rvw, \rvw^1, \rvw^2)}{p(\data^1, \data^2)},
    \intertext{and with the same conditional independence assumption,}
    &= \frac{p(\data^2 \mid \rvw, \rvw^1, \rvw^2)p(\data^1 \mid \rvw, \rvw^1, \rvw^2) p(\rvw, \rvw^1, \rvw^2)}{p(\data^1, \data^2)}.
   \intertext{And now we would like to show, using the `reverse' Bayes rule argument from before, that this is proportional to}
   &\propto p(\data^2\mid \rvw, \rvw^2)p(\rvw, \rvw^1 \mid \data^1).
\end{align}
This would follow if $\data^i$ and $\data^j$ were conditionally independent given $\rvw^j$ for $i \neq j$ and vice versa.
But that is not true, and should not in general be expected to be approximately true for datasets where we expect tasks to be related to each other.
That is to say, if multi-headed continual learning is to be used as an evaluation for Bayesian approximate inference, it must be with respect to a much more sophisticated probabilistic model than any current methods use.

Nevertheless, in practice, good multi-headed performance is significantly easier to achieve than single-headed performance, as we show below, which has led many papers to continue to evaluate themselves in this way.
A multi-headed evaluation can therefore make it seem as if an approach has solved a continual learning problem when it has not.

\subsection{Good Evaluations of Bayesian Deep Learning with Continual Learning}
A good evaluation of Bayesian inference with continual learning ought to test the Bayesian performance of the part of the scheme which does the `heavy lifting'.
As an extreme example, a `coreset' method which stored the entirety of all past datasets in order to perform variational inference would not demonstrate the effectiveness of the Bayesian approximation.

The single-headed split MNIST evaluation is a passable test of the performance of an approximate Bayesian posterior when the method employed does not use a coreset.
We show below, however, that using a coreset allows almost perfect performance \textit{even without anything else at all} and that past evaluations of methods like VCL have inadvertently disguised their true effectiveness by relying on coresets.

We use three variants of VCL.
First, we use pure VCL without a coreset, exactly as described by \citet{nguyenVariational2018}.
This represents a pure prior-focused approach.
Second, we use a small coreset of 40 datapoints extracted from each dataset (we use their k-center coreset approach, rather than a randomly selected one, but as they note this does not have a large effect).
The second method is the same as pure VCL except that, at the end of training on each task, the model is trained on the coresets before testing, as described in their work.
This reflects a hybrid approach.
Third, we use a `coreset only' approach.
It is exactly like the second variant except that the prior used for variational inference is the initial prior each time---it is not updated after each task.\footnote{This is not the coreset only algorithm used in \citet{nguyenVariational2018}. Theirs is seeing \textit{only} coresets---much less data---which is why it performs badly on even the first task.}

We use two further baselines to show that the effects we find are not an artifact of VCL specifically.
EWC \citep{kirkpatrickOvercoming2017} is used as a Bayes-inspired alternative to VCL which can be interpreted as Bayesian continual learning using a Laplace-approximation rather than variational inference.
Last, we compare to a more powerful `data-space' approach called VGR \citep{farquharUnifying2018} which in fact side-steps the problem of learning the posterior distribution over the parameters of the neural network by relying on generative models of the data to estimate the data-likelihood term of the ELBO loss.
VGR, VCL and its variants use a Bayesian neural network (BNN) and variational inference, while EWC does not, so performance comparisons between these architectures should be interpreted carefully.

\begin{figure}
    \begin{subfigure}{0.48\textwidth}
        \includegraphics[width=\columnwidth]{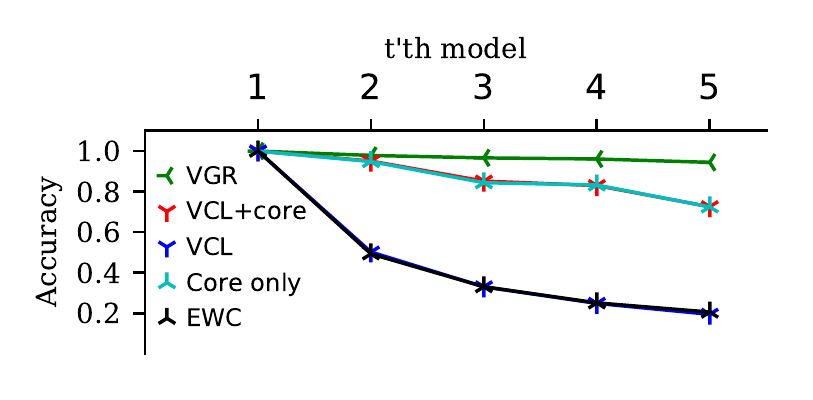}
        \caption{Single-headed Split MNIST}
        \label{fig:split_single_all}
    \end{subfigure}
    \begin{subfigure}{0.48\textwidth}
        \includegraphics[width=\columnwidth]{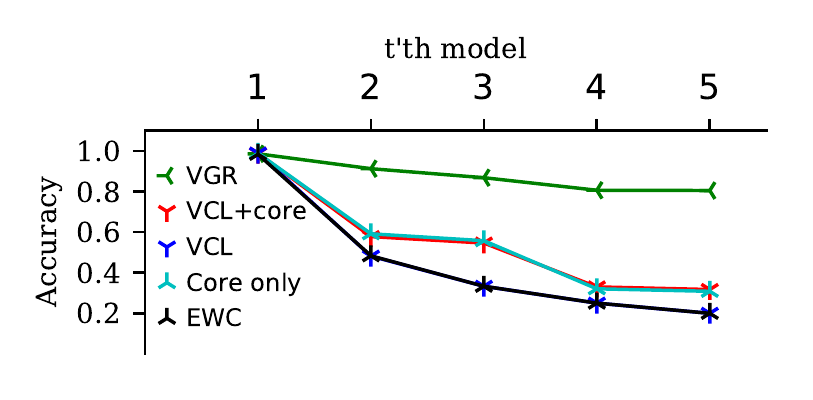}
        \caption{Single-headed Split FashionMNIST}
        \label{fig:fashion_single}
    \end{subfigure}
    \caption{Single-headed Split set-ups reveal the failure of Bayesian sequential updating to prevent catastrophic forgetting.}
\end{figure}

The performance of VCL with coreset appears to be entirely driven by the presence of the coresets (see \cref{fig:split_single_all} and \cref{fig:fashion_single}).
When coresets are removed, VCL alone completely forgets old tasks---its accuracy comes from correctly classifying the most recent task only.
(By chance, a model with good accuracy on only the latest task will get roughly $100\%$, then $50\%$, $33\%$ etc.\ which is exactly what we see.)
EWC performs exactly like VCL.
This suggests that neither mean-field variational inference with a multi-variate Gaussian posterior nor the Laplace approximation are succeeding in capturing enough of the posterior to allow sequential Bayesian updating.
Moreover, it suggests that neither is getting anywhere close.

Using FashionMNIST rather than MNIST is harder and a worthwhile additional test, but does not reveal a radically different story.
VCL performs much worse on FashionMNIST than VGR even though both use the same model.

\subsection{Contrast to Inadequate Evaluations}
\begin{figure}
   \centering
    \begin{subfigure}{0.48\textwidth}
        \includegraphics[width=\textwidth]{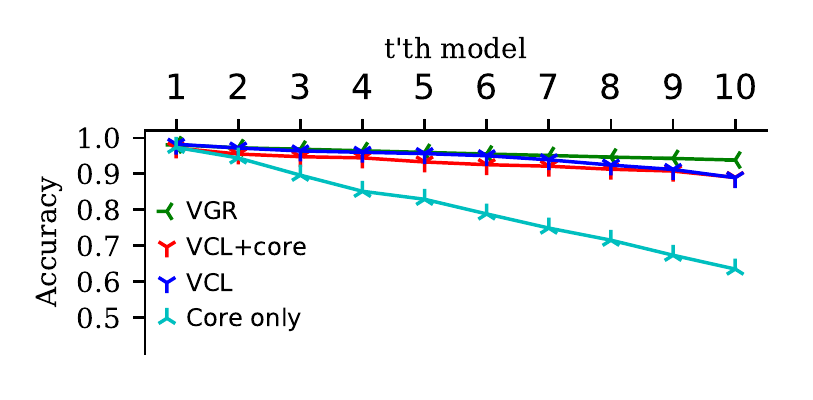}
        \caption{Permuted MNIST}
    \end{subfigure}
    \begin{subfigure}{0.48\textwidth}
        \includegraphics[width=\textwidth]{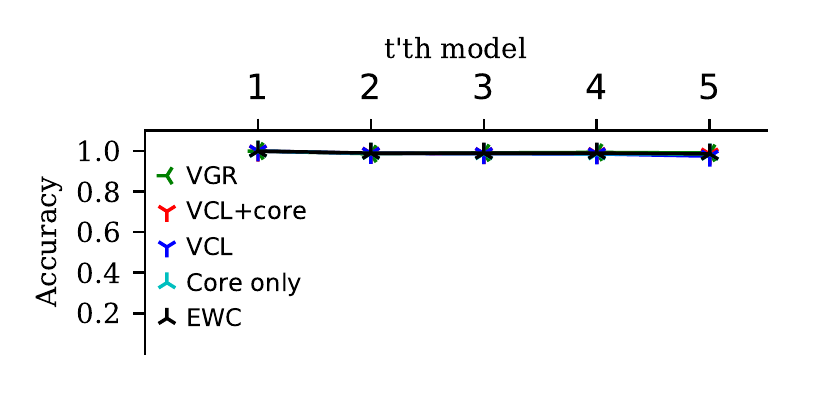}
        \caption{Multi-headed MNIST}
    \end{subfigure}
    \begin{subfigure}{0.48\textwidth}
        \includegraphics[width=\textwidth]{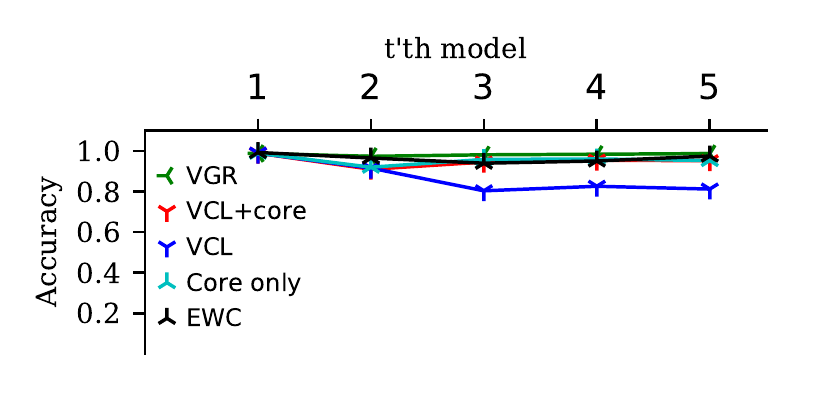}
        \caption{Multi-headed FashionMNIST}
    \end{subfigure}
    \begin{subfigure}{0.48\textwidth}
        \includegraphics[width=\textwidth]{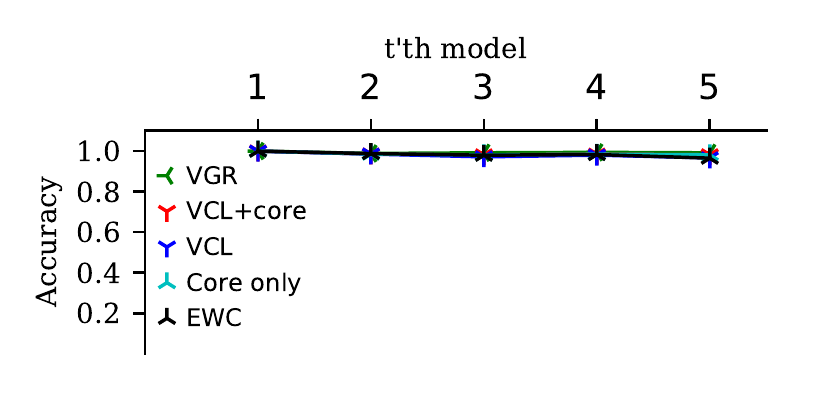}
        \caption{Test-time Knowledge MNIST}
    \end{subfigure}
    \begin{subfigure}{0.48\textwidth}
        \includegraphics[width=\textwidth]{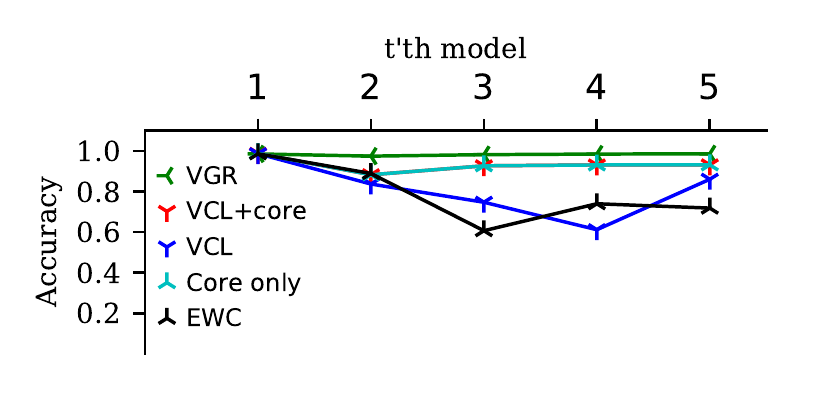}
        \caption{Test-time Knowledge FashionMNIST}
    \end{subfigure}
    \caption[]{The inadequate evaluations discussed in \cref{s:evaluations:inadequate} do not succeed in distinguishing the over-reliance of `Bayesian' methods on coresets.}
    \label{fig:inadequate_evaluations}
\end{figure}

To demonstrate the failure of inadequate evaluations to identify shortcomings of the approximate inference procedure, we show the performance of these methods on permuted MNIST/FashionMNIST and multi-headed split MNIST/FashionMNIST which assumes access to labels during either training and testing or just testing.
In \cref{fig:inadequate_evaluations} we show that none of these evaluations succeeds in distinguishing the over-reliance of `Bayesian' methods on coresets.
The `multi-headed' versions are trained and tested using only the `active' head, assuming task knowledge.
In contrast, the `test-time knowledge' versions are trained in the same way as a single-headed network but assume knowledge of the task label at test-time.

\clearpage

\section{Active Learning}
\newcommand{\nn}{f_\vtheta}
\newcommand{\ed}{\hat{p}(\rvx, \ry)}
\newcommand{\sed}{\tilde{p}(\rvx, \ry)}
\newcommand{\pr}{r}
\newcommand{\er}{\hat{R}}
\newcommand{\Rt}{\tilde{R}}
\newcommand{\Rs}{\tilde{R}_{\text{LURE}}}
\newcommand{\Rp}{\tilde{R}_{\text{PURE}}}
\newcommand{\Dpool}{\data_{\textup{pool}}}
\newcommand{\Dtrain}{\data_{\textup{train}}}
\newcommand{\ssu}{\sigma_{\textup{LURE}}}
\newcommand{\rhat}{\hat{r}}
\newcommand{\var}{\Var}
\newcommand{\muD}{\mu_{m|i,\data}}
\newcommand{\mugD}{\mu_{|\data}}
\newcommand{\mumgD}{\mu_{m|\data}}
\newcommand{\mukgD}{\mu_{k|\data}}

\ifSubfilesClassLoaded{
\newtheorem{theorem}{Theorem}
\newtheorem{lemma}{Lemma}
\newtheorem{remark}{Remark}}{}
\newenvironment{proofsketch}{%
   \renewcommand{\proofname}{Proof Sketch}\proof}{\endproof}
\textit{\textbf{Statement of contribution:} the work in this section draws heavily on \citep{farquharStatistical2021} which was written with equal contribution by Tom Rainforth. In particular, Tom was chiefly responsible for the proofs of the unbiasedness and consistency of $\Rp$ and $\Rs$ and the analysis of their variance. For this reason, rather than restating those results in this thesis, I just cite the original paper for their properties.
}
\vspace{5mm}

   Active learning aims to improve learning label-efficiency by selecting informative points to label \citep{atlasTraining1990,settlesActive2010}.
Although not all approaches to active learning are grounded in information theory (e.g., \citep{senerActive2018}), a number of promising approaches have taken a Bayesian approach and interpreted the problem of active learning as estimating the expected information gain about the subjective belief distributions over the network's parameters when new labels are acquired \citep{mackayInformationBased1992, houlsbyBayesian2011}.
This suggests that the effectiveness of an approximate posterior at supporting active learning might be a good test of its approximation of the true posterior.
These methods have been applied to deep learning, especially in computer vision \citep{galDeep2017,wangCostEffective2017}

A naive approach to evaluating approximate Bayesian inference methods would work as follows:
\begin{enumerate}
   \item Define some approximate Bayesian inference scheme, a dataset, and a probabilistic model.
   \item Perform approximate inference given an initial dataset.
   \item  Select new points to acquire in order to expand the dataset using Bayesian Active Learning by Disagreement (BALD) \citep{houlsbyBayesian2011}, which estimates the expected information gain about parameter distributions from acquiring newly labelled points \citep{mackayInformationBased1992}.
   \item Repeat the procedure some number of times, and then repeat the whole thing for several approximate inference schemes.
   \item Compare the performance of the resulting models. We then evaluate the performance of our approximate inference scheme based on the method that produced the best problem-specific outcome.
\end{enumerate}

Unfortunately, this will not work as a reliable evaluation of approximate inference in Bayesian neural networks.
In particular, we show that the performance of active learning relies partly on the presence of an implicit bias which is introduced by the active sampling scheme which has nothing to do with the posterior approximation.
A `bad' posterior approximation can guide the model towards useful data-points by representing implicit `prior knowledge' which is not encoded directly in the Bayesian problem, but gradually emerges from natural-selection of research.

We demonstrate this by showing how to remove the bias, and show that this is harmful in some cases while it is helpful in others.
We argue that the reasons for this are related to the presence of over-fitting and interpolation, rather than the quality of the Bayesian inference approximations.
In fact, we show that removing the bias can change the ordering of which approximation scheme appears superior, without having any effect on the underlying posterior approximation.

Removing this confounding factor may allow \textit{active testing} \citep{kossenActive2021}, rather than active learning, to be a more effective evaluation of the quality of approximate inference.
However, this conclusion can only be tentative, because there may be other shortcomings of active testing as an evaluation scheme which have not yet been identified.

\subsection{Bias in Active Learning}\label{s:problem_statement}
Active learning introduces a bias to the Monte Carlo estimation of the risk or of the loss used during training a neural network or Bayesian neural network.
Intuitively, this comes from the fact that the data distribution is no longer identically and independently distributed (i.i.d.) following the population distribution, and that the data distribution is not independent of the parameter distribution.
This sampling bias has been noted and discussed by, for example, \citet{mackayInformationBased1992} who dismissed it from a Bayesian perspective based on the `likelihood principle'---that, given a statistical model, the likelihood is a complete summary of the evidence, regardless of how the data was acquired.
However, the principle remains controversial \citep{rainforthAutomating2017} because there do seem to be situations where the manner in which the data are gathered ought to influence inferences and in this case applying the likelihood principle would assume a well-specified model.
As a result, the likelihood is arguably \textit{not} a complete summary of the evidence.
Sampling bias has also been considered by \citet{dasguptaHierarchical2008, beygelzimerImportance2009, chuUnbiased2011, gantiUPAL}.

Let us characterize the bias introduced by active learning more formally.
In supervised learning, generally, we aim to find a decision rule $\nn$ corresponding to inputs, $\rvx$, and outputs, $\ry$, drawn from a population data distribution $\pdata(\rvx, \ry)$ which, given a loss function $\Ls(\ry, \nn(\rvx))$, minimizes the population risk:
\begin{equation}
   \pr = \E{\rvx, \ry \sim \pdata}{\Ls(\ry, \nn(\rvx))}.
\end{equation}
The population risk cannot be found exactly, so instead we consider the empirical distribution for some dataset of $N$ points drawn from the population.
This gives the empirical risk: an unbiased and consistent estimator of $\pr$ when the data are drawn i.i.d from $\pdata$ and are independent of $\rvtheta$,
\begin{equation}
   \er = \frac{1}{N} \sum_{n=1}^{N}\nolimits \Ls(\ry_n, \nn(\rvx_n)).
\end{equation}
In pool-based active learning \citep{lewisSequential1994,settlesActive2010}, we begin with a large unlabeled dataset, known as the pool dataset $\Dpool \equiv \{\rvx_n|1\leq n \leq N\}$, and sequentially pick the most useful points for which to acquire labels.
The lack of most labels means we cannot evaluate $\er$ directly, so we use the \emph{sub-sample empirical risk} evaluated using the $M$ actively sampled labelled points:
\begin{equation}
   \tilde{R} = \frac{1}{M}\sum_{m=1}^{M}\nolimits \Ls(\ry_m, \nn(\rvx_m)).\label{eq:subsample_empirical_estimator}
\end{equation}
Though almost all active learning research uses this estimator (see Appendix \ref{a:active_learning_practice}), it is not an unbiased estimator of either $\er$ or $\pr$ when the $M$ points are actively sampled.
Under active---i.e.~non--uniform---sampling the $M$ datapoints are not drawn from the population distribution, resulting in a bias.

Note an important distinction between what we will call ``statistical bias'' and ``overfitting bias.''
The bias from active learning above is a statistical bias in the sense that using $\Rt$ biases our estimation of $r$, regardless of $\rvtheta$.
As such, optimizing $\rvtheta$ with respect to $\Rt$ induces bias into our optimization of $\rvtheta$.
In turn, this breaks any consistency guarantees for our learning process: if we keep $M/N$ fixed, take $M\to\infty$, and optimize for $\rvtheta$, we no longer get the optimal $\rvtheta$ that minimizes $r$.
Almost all work on active learning for neural networks currently ignores the issue of statistical bias.

However, even without this statistical bias, indeed even if we use $\er$ directly, the training process itself also creates an overfitting bias: evaluating the risk using training data induces a dependency between the data and $\rvtheta$.
This is why we usually evaluate the risk on held-out test data when doing model selection.
Dealing with overfitting bias is beyond the scope of our work as this would equate to solving the problem of generalization.
The small amount of prior work which does consider statistical bias in active learning entirely ignores this overfitting bias without commenting on it.

We mostly focus on statistical bias in active learning, so that we can produce estimators that are valid and consistent, and let us optimize the intended objective, not so they can miraculously close the train--test gap.
From a more formal perspective, our results all assume that $\rvtheta$ is chosen independently of the training data; an assumption that is almost always (implicitly) made in the literature.
This ensures our estimators form valid objectives, but also has important implications that are typically overlooked.
We return to this in \S\ref{s:overfitting}, examining the interaction between statistical and overfitting bias.

\begin{figure}
    \centering
    \begin{subfigure}[b]{0.9\textwidth}
      \includegraphics[width=\textwidth]{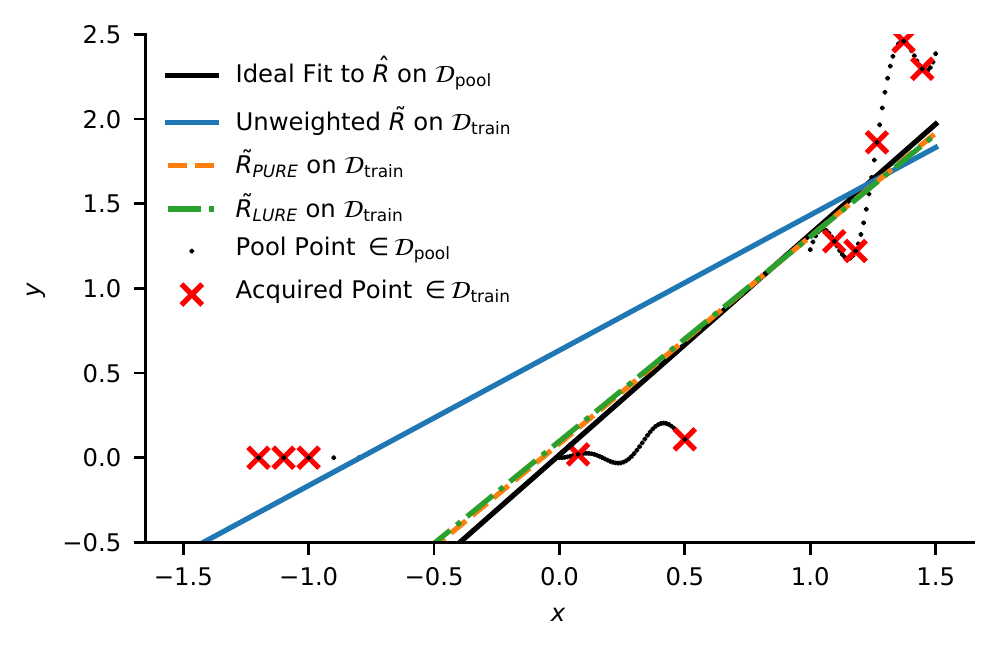}
      \caption{Illustration with toy data.}
      \label{fig:linear_plot}
    \end{subfigure}
    \begin{subfigure}[b]{0.9\textwidth}
      \includegraphics[width=\textwidth]{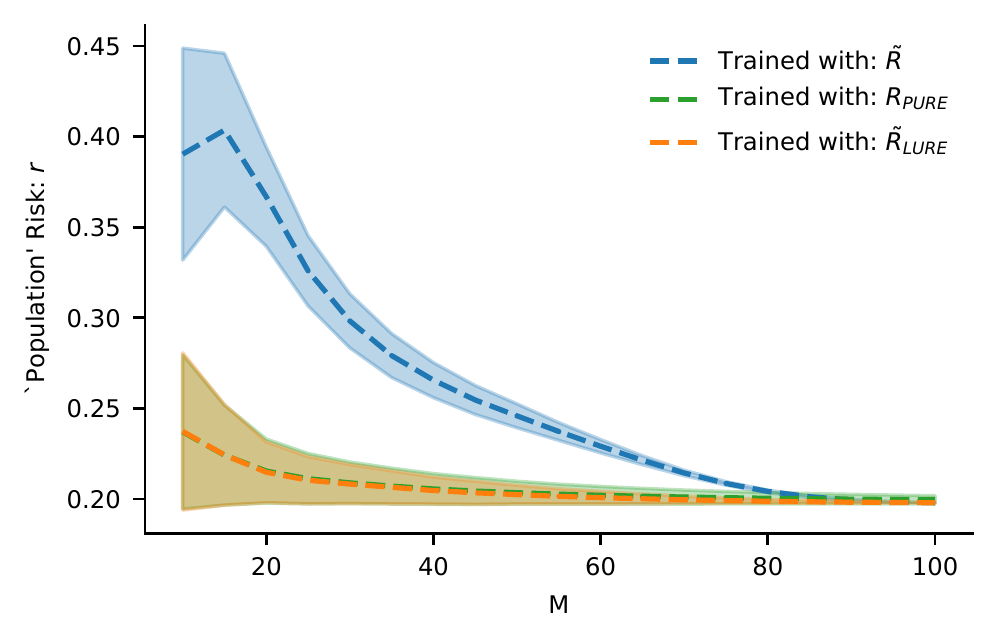}
      \caption{Test MSE. Linear regression}
      \label{fig:linear_error}
    \end{subfigure}
   \caption[Illustration of active learning in an unbiased way]{\textbf{(a)} Active learning deliberately over-samples unusual points (red x's) which no longer match the population (black dots). Common practice uses the biased unweighted estimator $\tilde{R}$ which puts too much emphasis on unusual points. Our unbiased estimators $\Rp$ and $\Rs$ fix this, learning a function using only $\Dtrain$ nearly equal to the ideal you would get if you had labels for the whole of $\Dpool$, despite only using a few points.
   \textbf{(b)} Removing the bias greatly improves the error of linear models fit to the data, with near-optimal performance using very few examples.}
   \label{fig:linear}
\end{figure}

\newpage
\subsection{Active Learning Is An Ineffective Posterior Evaluation}
In this thesis, we will help ourselves to two unbiased risk estimators, $\Rp$ and $\Rs$, as an alternative to $\tilde{R}$.
These estimators are introduced by \citet{farquharStatistical2021}, who prove that they are unbiased and consistent and that, under mild assumptions, they have lower variance than $\tilde{R}$.
Using these estimators, we will show that the statistical bias introduced by active-learning sampling has an impact on performance evaluations for active learning despite not directly representing the quality of the approximate inference scheme.

\subsubsection{Empirical Investigation of Removing Bias During Active Learning}
Intuitively, removing bias in training while also reducing the variance ought to improve the downstream task objective: test loss and accuracy.
To investigate this, we train models using $\Rt$, $\Rs$, and $\Rp$ with actively sampled data and measure the population risk of each model.

On a toy linear regression task (Figure \ref{fig:linear}), our estimators improve the test loss---even with small numbers of acquired points we have nearly optimal test loss.
Here, therefore, removing active learning bias greatly improves downstream performance despite having \textit{no} effect on the inference, which is analytical.

But we further investigate FashionMNIST and MNIST using a Bayesian neural network trained either with the Radial approximating distribution or Monte Carlo dropout (MCDO) \citep{galDropout2015}.
This demonstrates that removing the bias does not always improve active learning performance.
We use an unbalanced variant of both datasets which should allow active learning to shine, though note that we check results on the standard set as well in the appendix.
Following \citet{kirschBatchBALD2019}, we use a `consistent' version of MCDO which reduces score variance.

\begin{figure}
   \centering
   \begin{subfigure}[b]{0.9\textwidth}
      \includegraphics[width=\textwidth]{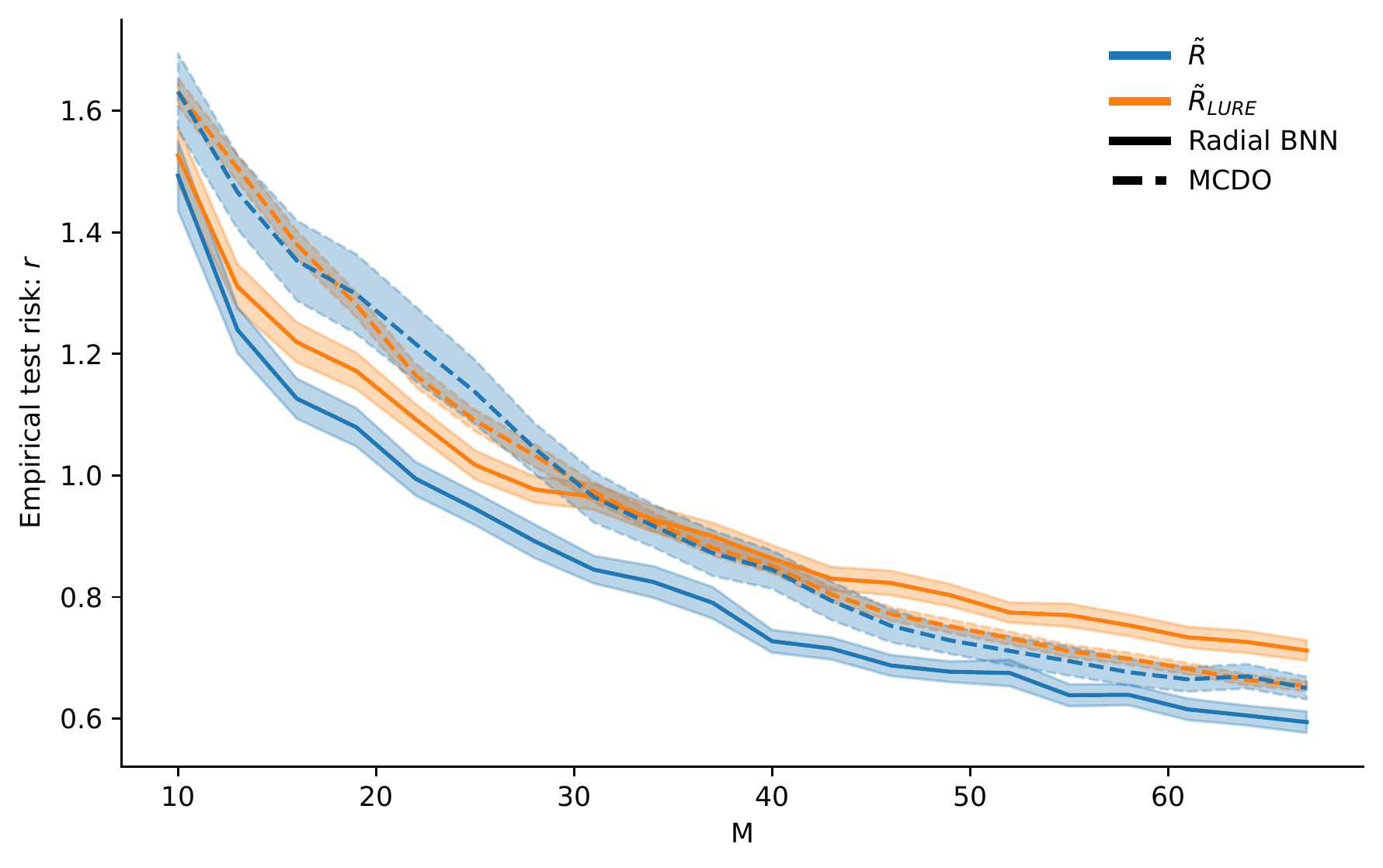}
      \caption{Test loss.}
      \label{fig:mnist_al_eval_loss}
   \end{subfigure}
   \begin{subfigure}[b]{0.9\textwidth}
      \includegraphics[width=\textwidth]{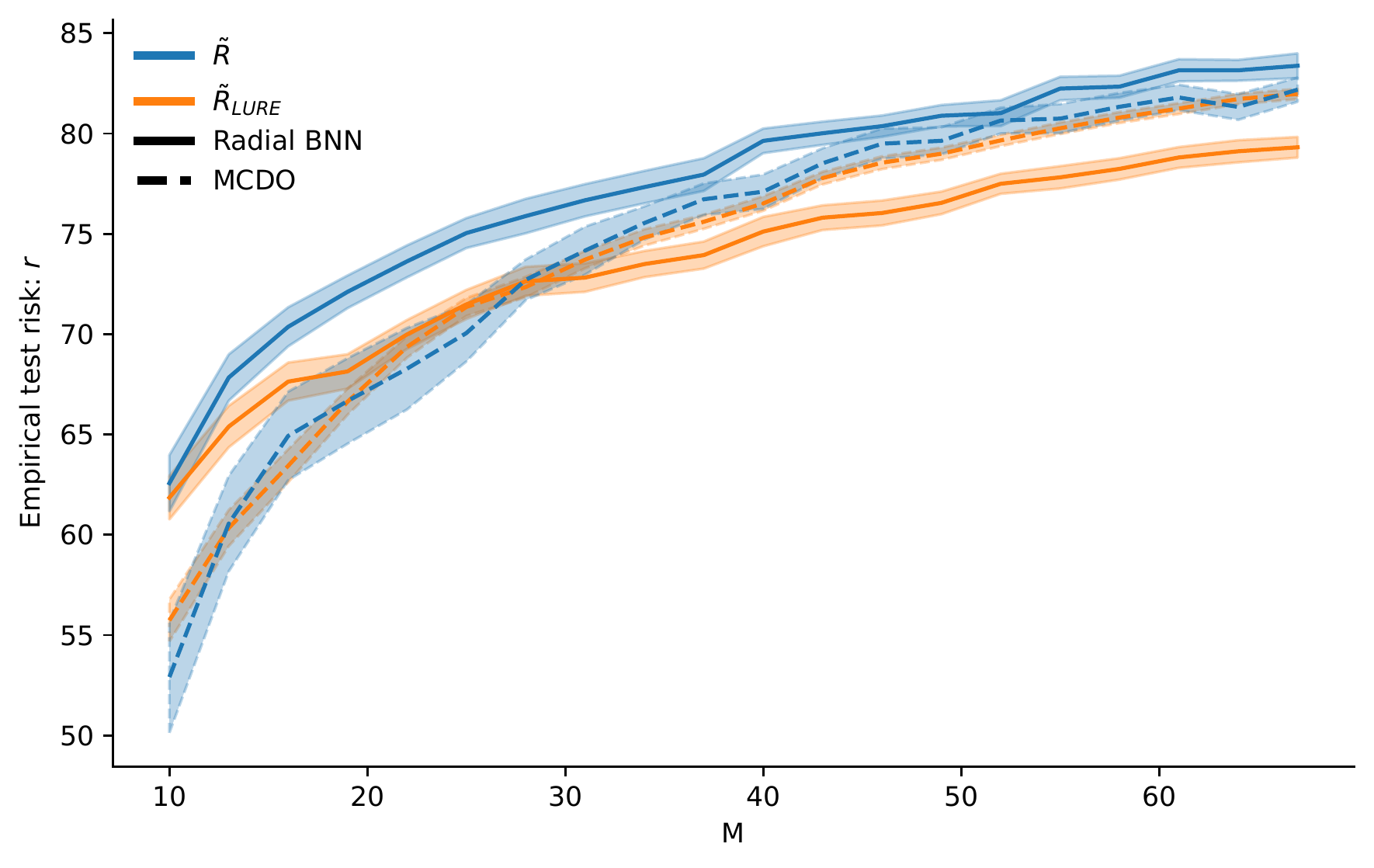}
      \caption{Test accuracy.}
      \label{fig:mnist_al_eval_accuracy}
   \end{subfigure}
   \caption[Active learning is an ineffective evaluation of posterior quality]{\textbf{MNIST.} Training with an unbiased ELBO flips the ordering of active learning performance for Monte Carlo dropout and Radial BNNs. Radial BNNs (solid line) perform best when a biased estimator is used (blue) but worst with the unbiased estimator (orange).
   MCDO (dashed line) performs in the middle with both biased and unbiased estimators.
   Because the ordering of the performance evaluations of methods is sensitive to de-biasing, active learning is not a reliable way to evaluate the approximate inference.
   Shading represents standard error. $\Rp$ is omitted for visual clarity but is very similar to the result for $\Rs$ and is shown in \cref{fig:bias_with_fitting}.}
\end{figure}
In all cases there is a small but significant negative impact on the full test dataset loss of training with $\Rs$ or $\Rp$ and a slightly larger negative impact on test accuracy (\cref{fig:bias_with_fitting}).
That is, we get a better model by training using a biased estimator with higher variance!
The reasons for this are discussed in \cref{s:overfitting}.

This effect is strong for Radial BNNs and comparatively weak for MCDO, especially as more data is acquired.
For MNIST, the biased version of Radial BNNs performs slightly better on both accuracy and negative-log-likelihood than the biased version of MCDO, while the unbiased version of MCDO performs noticeably better than the unbiased version of the Radial BNN.
That is to say, removing the bias in label-acquisition changes the ordering of which of these two approximate inference schemes appears better, despite having no direct effect on the quality of the posterior approximation.
This demonstrates the failure of active learning to serve as an effective evaluation of posterior approximation.

At the same time, for Fashion MNIST, the differences between the two approximate posteriors are large enough that the effect of de-biasing does not change the ordering.
This highlights that Bayesian active learning performance is not insensitive to the choice of posterior approximation, it just does not offer a definitive ordering of methods.

\begin{figure}
   \centering
   \begin{subfigure}[b]{0.9\textwidth}
      \includegraphics[width=\textwidth]{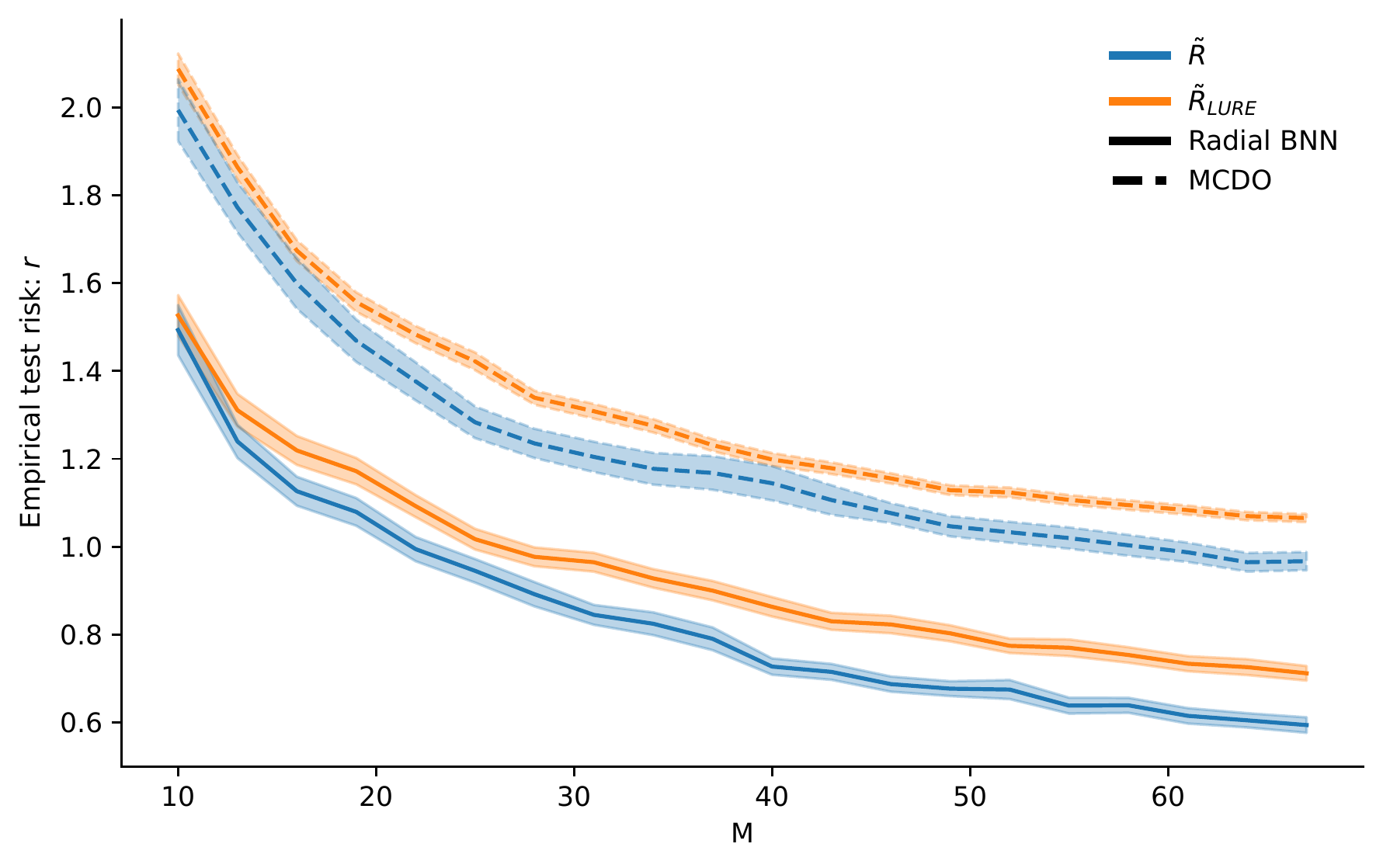}
      \caption{Test loss.}
      \label{fig:mnist_ae_eval_loss}
   \end{subfigure}
   \begin{subfigure}[b]{0.9\textwidth}
      \includegraphics[width=\textwidth]{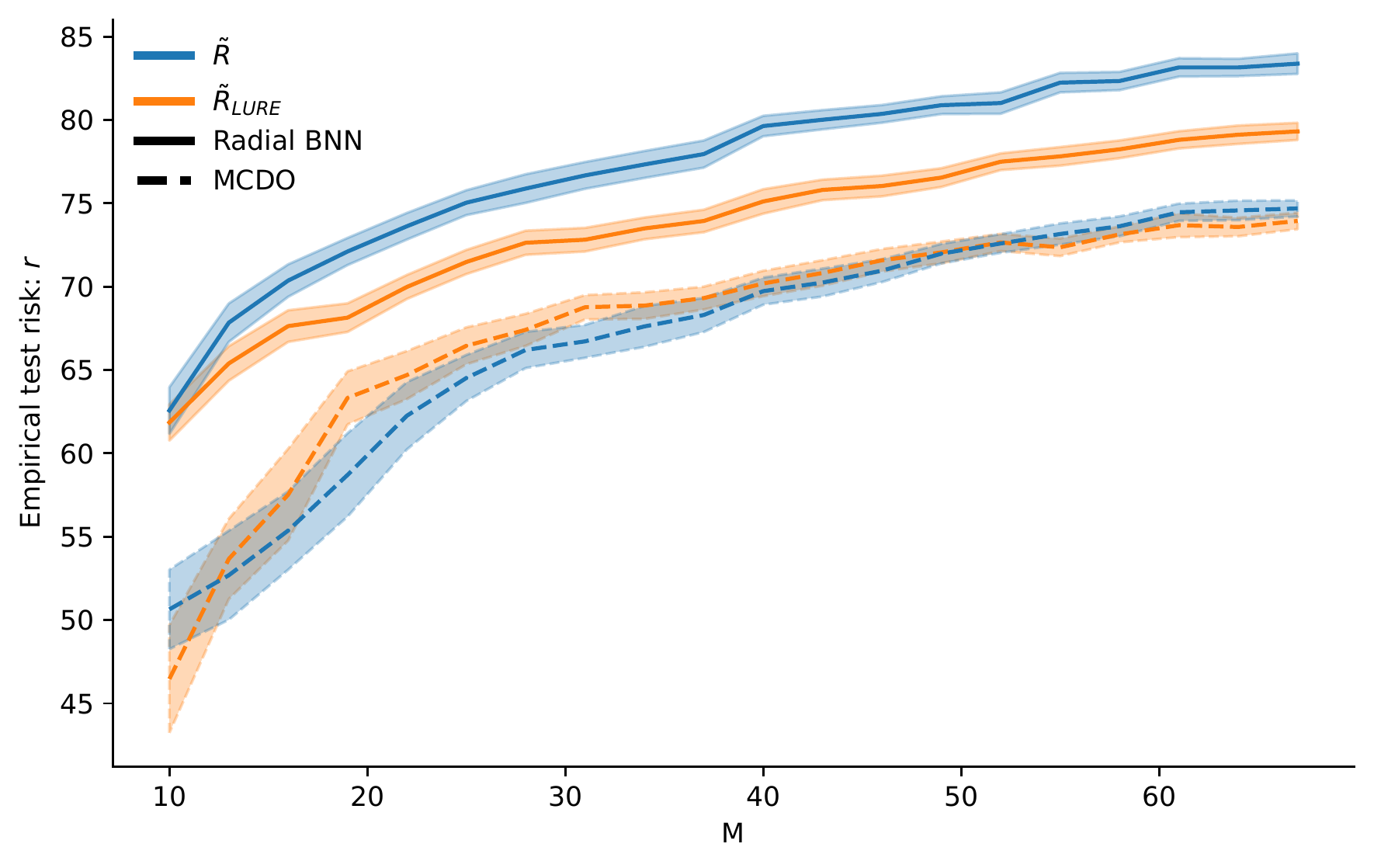}
      \caption{Test accuracy.}
      \label{fig:mnist_ae_eval_accuracy}
   \end{subfigure}
   \caption[In some cases, the effect of debiasing is small enough to allow active learning to reveal performance differences.]{\textbf{FashionMNIST.} In some cases, the effect of debiasing is small enough to allow active learning to reveal performance differences. Here, MCDO performs sufficiently worse than Radial BNNs that even though the debiasing step hurts active learning performance more for Radial BNNs, the ordering is preserved.}
\end{figure}

\subsubsection{Empirical Investigation of Bias in Active Model Evaluation}
This is in contrast with active model evaluation---in which we attempt to estimate the test loss of a \textit{fixed} model.
We take a fixed and fully trained model and then actively acquire labels from the test dataset.
Our goal is to estimate the full test risk with as few datapoints as possible.
The overall approach is discussed in more detail by \citet{kossenActive2021}.

In this case, it is clear that removing the bias is the only sensible way to interpret the task---the goal of model evaluation is to discover the actual risk of the model, the evaluation is not being used to implicitly guide the model towards informative data.
Now, the only impact of an improved approximate posterior distribution is to better estimate the mutual information between the parameters and data labels.
The more quickly the variance of the test risk estimator reduces, the better the approximate posterior distribution.

We can measure performance using, for example, the mean-squared error between the actively-sampled risk estimate and the full test-pool risk (recall that we do not expect to be able to fix the gap between the full test pool and the true population risk using active sampling alone).

\begin{figure}
   \centering
   \begin{subfigure}[b]{0.9\textwidth}
      \includegraphics[width=\textwidth]{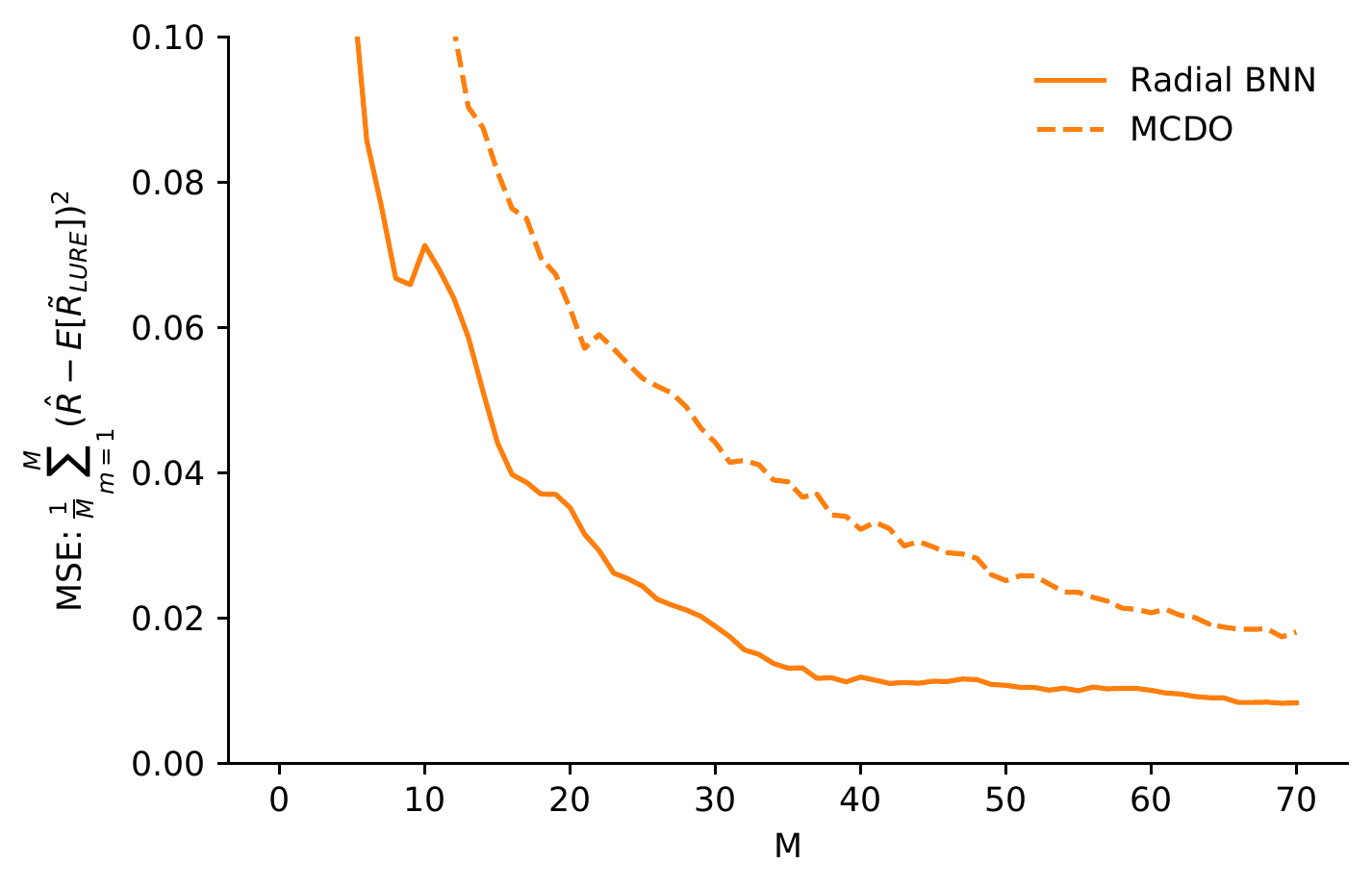}
      \caption{MNIST.}
   \end{subfigure}
   \begin{subfigure}[b]{0.9\textwidth}
      \includegraphics[width=\textwidth]{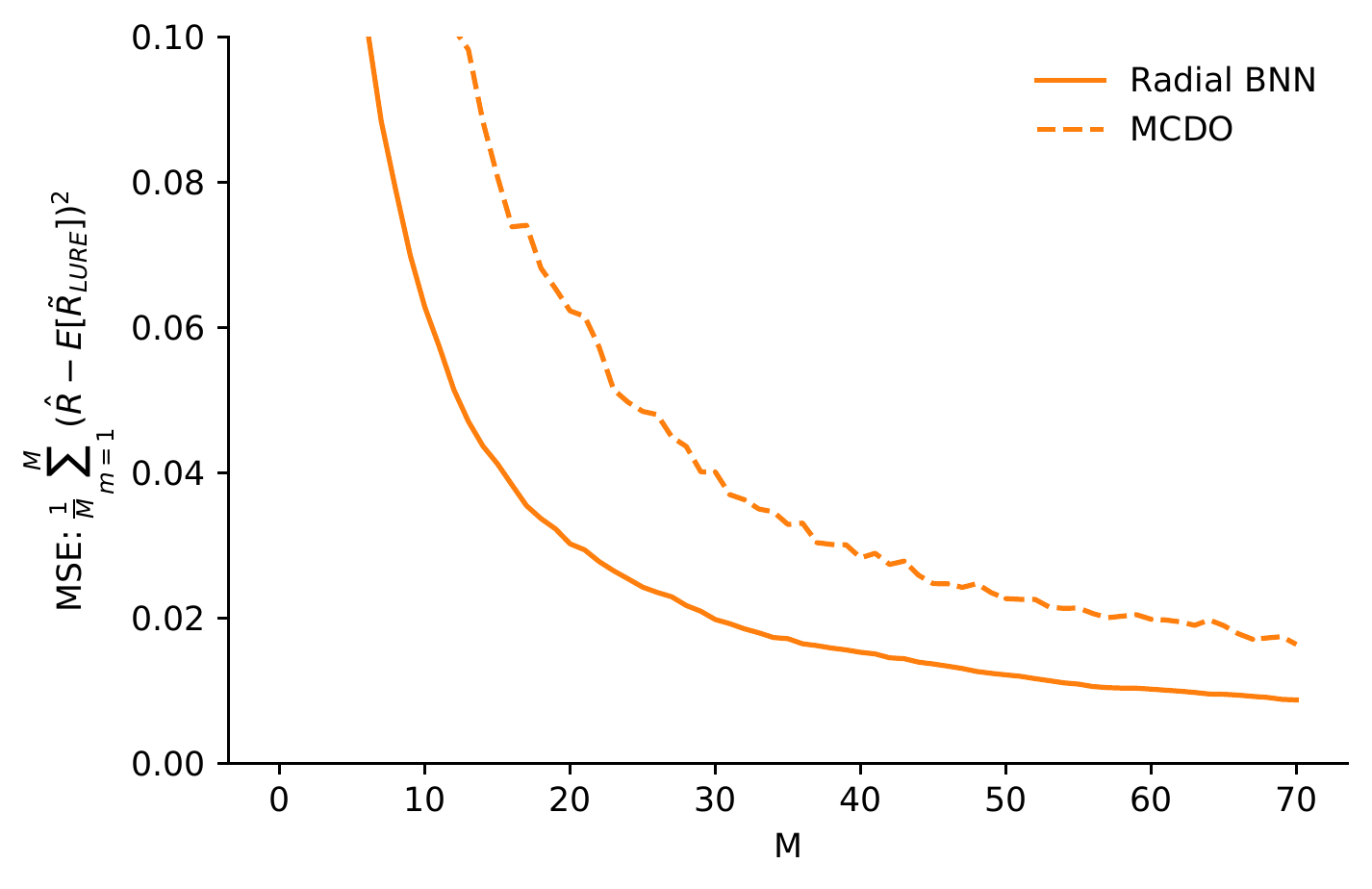}
      \caption{FashionMNIST.}
   \end{subfigure}
   \caption[Active evaluation can be used as a way to evaluate approximate inference performance.]{\textbf{Active Evaluation Mean-squared Error.} We compute the error between the actively evaluated risk and the `true' population risk (as evaluated on the full test dataset). Unlike active \textit{learning}, the unbiased risk estimate is genuinely what we value, not a means to an end, which makes this a more effective evaluation of the approximate posterior.
   These specific evaluations suggest that the Radial BNN is serving as a more effective approximate posterior than MCDO in this case.}
   \label{fig:active_evaluation}
\end{figure}

In \cref{fig:active_evaluation}, we show the mean-squared error of the actively sampled risk estimator for both MNIST and FashionMNIST as a way of comparing Radial BNNs and Monte Carlo dropout.
In both cases, Radial BNNs appear to be providing much better guidance as to the expected information gain provided by test labels.
Although we only examine small image datasets here, \citep{kossenActive2021} explores active testing in much larger settings, including CIFAR-100, in work which extends beyond the scope of this thesis.\footnote{The author of this thesis is a joint-first-author of \citet{kossenActive2021} but anticipates that his co-author will submit some or all of that paper in their own thesis.}

I also want to emphasise that although active testing avoids \textit{one} of the problems of using active learning as a measure of approximate posterior performance, I have not shown that it does not suffer from other problems of its own.

\begin{figure}
   \vspace{-10mm}
   \centering
   \begin{subfigure}[b]{0.7\textwidth}
      \centering
      \includegraphics[width=\textwidth]{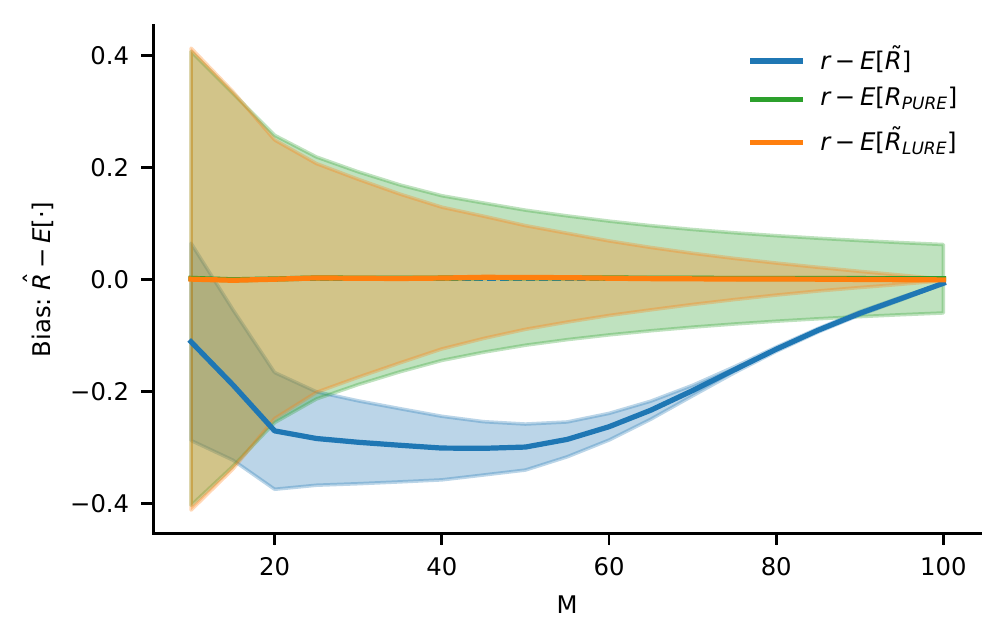}
      \caption{Linear regression}
      \label{fig:bias_linear_no_fit}
   \end{subfigure}\hfill
   
   \begin{subfigure}[b]{0.7\textwidth}
      \centering
      \includegraphics[width=\textwidth]{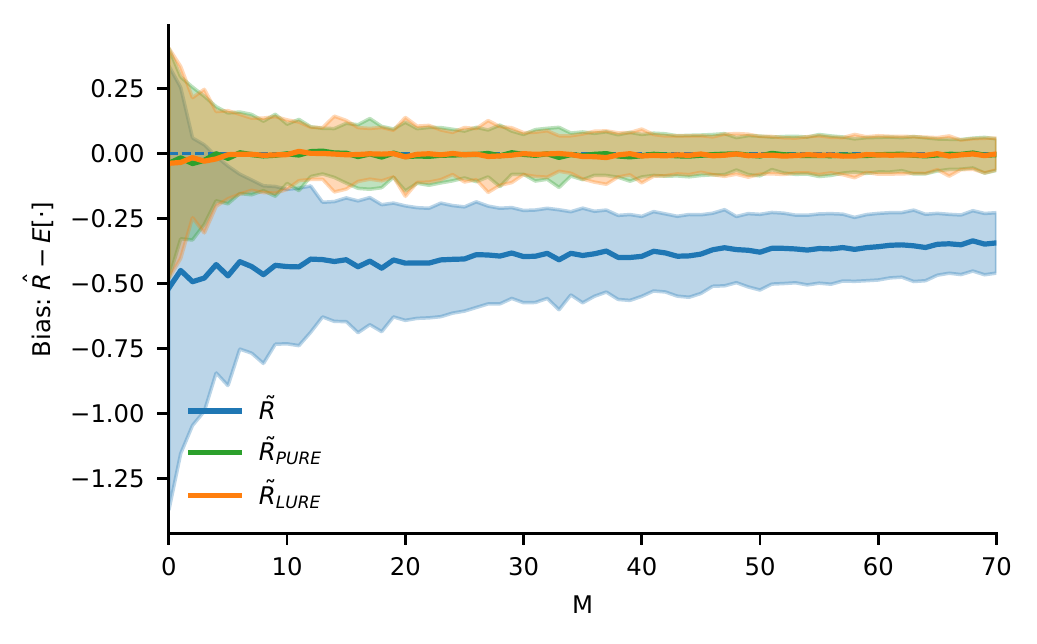}
      \caption{Radial: MNIST}
      \label{fig:bias_mnist_no_fit}
   \end{subfigure}\hfill

   \begin{subfigure}[b]{0.7\textwidth}
      \centering
      \includegraphics[width=\textwidth]{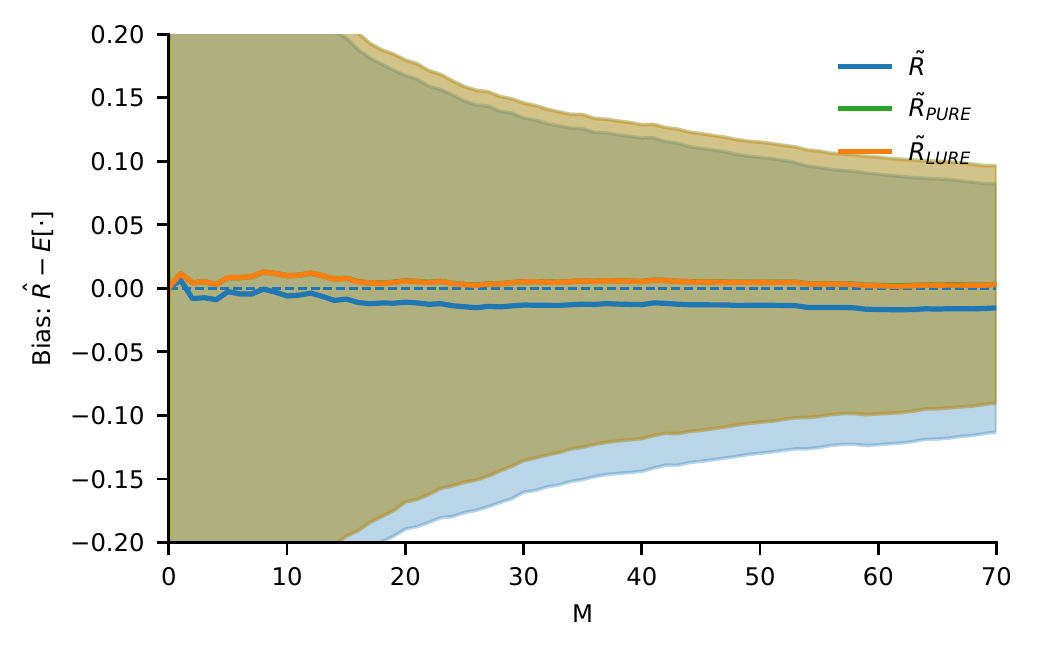}
      \caption{Radial: Fashion}
      \label{fig:bias_fashion_mnist_no_fit}
   \end{subfigure}

~~~~~~~~
   \caption[Unbiased active learning evaluation]{$\Rp$ and $\Rs$ remove a bias introduced by active learning, while unweighted $\tilde{R}$, which most active learning work uses, is biased. Note the sign: $\tilde{R}$ \emph{overestimates} risk because active learning samples the hardest points. Variance for $\Rp$ and $\Rs$ depends on the acquisition distribution placing high weight on high-expected-loss points, which makes this an evaluation of the approximate posterior distribution. Shading is $\pm1$ standard deviation.}
   \label{fig:bias_no_fit}
\end{figure}

\subsection{Active Learning Bias in the Context of Overall Bias}
\label{s:overfitting}
In the previous sections, we relied on the observation that removing the active learning bias can hurt active learning performance.
This is counterintuitive, so it would help to have a plausible explanation of this phenomenon in order to reassure ourselves that we understand roughly what is happening.
We hypothesise that the finding that $\Rs$ hurts training for the BNN makes sense in the context of the bias introduced by overfitting.
That is to say, we need to examine the effect of removing statistical bias in the context of \emph{overall} bias---training would ordinarily induce an overfitting bias (OFB) even if we had not used active learning.

If we optimize parameters $\rvtheta$ according to $\er$, then $\E{}{\er(\rvtheta^*)} \neq r$, because the optimized parameters $\rvtheta^*$ tend to explain training data better than unseen data.
Using $\Rs$, which removes \emph{statistical bias}, we can isolate OFB in an active learning setting.
More formally, supposing we are optimizing any of the discussed risk estimators (which we will write using $\tilde{R}_{(\cdot)}$ as a placeholder to stand for any of them) we define the OFB as:
\begin{equation}
   B_{\textup{OFB}}(\tilde{R}_{(\cdot)}) = r - \Rs(\rvtheta^*) \quad \text{where } \quad \rvtheta^* = \argmin_{\rvtheta}\nolimits (\tilde{R}_{(\cdot)})
\end{equation}
$B_{\textup{OFB}}(\tilde{R}_{(\cdot)})$ depends on the details of the optimization algorithm and the dataset.
Understanding it fully means understanding generalization in machine learning and is outside our scope.
We can still gain insight into the interaction of active learning bias (ALB) and OFB.
Consider the possible relationships between the magnitudes of ALB and OFB:
\begin{description}
    \item[\textbf{ALB $>>$ OFB}] Removing ALB reduces overall bias and is most likely to occur when $f_{\theta}$ is not very expressive such that there is little chance of overfitting.
    \item[\textbf{ALB $<<$ OFB}] Removing ALB is irrelevant as model has massively overfit regardless.
    \item[\textbf{ALB $\approx$ OFB}] Here \emph{sign} is critical. If ALB and OFB have opposite signs and similar scale, they will tend to cancel each other out. Indeed, they usually have opposite signs.
\end{description}
$B_{\textup{OFB}}$ is usually positive: $\rvtheta^*$ fits the training data better than unseen data.
ALB is generally negative: we actively choose unusual, surprising, or informative points which are harder to fit than typical points.

Therefore, when significant overfitting is possible, unless ALB is also large, removing ALB will have little effect and can \emph{even be harmful}.
This hypothesis would explain the observations in \S\ref{s:overfitting} if we were to show that $B_{\textup{OFB}}$ was small for linear regression but had a similar magnitude and opposite sign to ALB for the BNN.
This is exactly what we show in Figure \ref{fig:overfitting_bias}.

Specifically, we see that for linear regression, the $B_{\textup{OFB}}$ for models trained with $\tilde{R}$, $\Rp$, and $\Rs$ are all small (Figure \ref{fig:bias_linear_fit}) when contrasted to the ALB shown in Figure \ref{fig:bias_linear_no_fit}.  Here ALB $>>$ OFB; removing ALB matters.
For BNNs we instead see that the OFB has opposite sign to the ALB but is either similar in scale for MNIST (Figures \ref{fig:bias_mnist_no_fit} and \ref{fig:bias_mnist_fit}), or the OFB is much larger than ALB for Fashion MNIST (Figures \ref{fig:bias_fashion_mnist_fit} and \ref{fig:bias_fashion_mnist_no_fit}).
The two sources of bias thus (partially) cancel out.
Essentially, using active learning can be treated (quite instrumentally) as an ad hoc form of regularization.
This explains why removing ALB can hurt active learning with neural networks.

\begin{figure}
   \vspace{-20mm}
    \centering
   \begin{subfigure}[b]{0.7\textwidth}
      \centering
      \includegraphics[width=\textwidth]{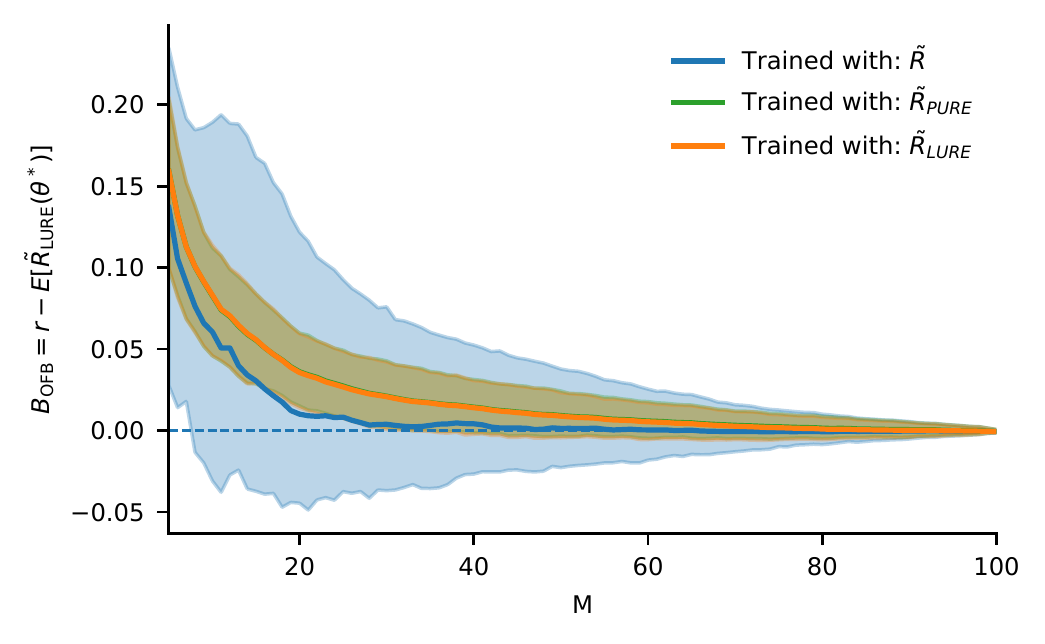}
      \caption{Linear regression}
      \label{fig:bias_linear_fit}
   \end{subfigure}
   \begin{subfigure}[b]{0.7\textwidth}
      \centering
      \includegraphics[width=\textwidth]{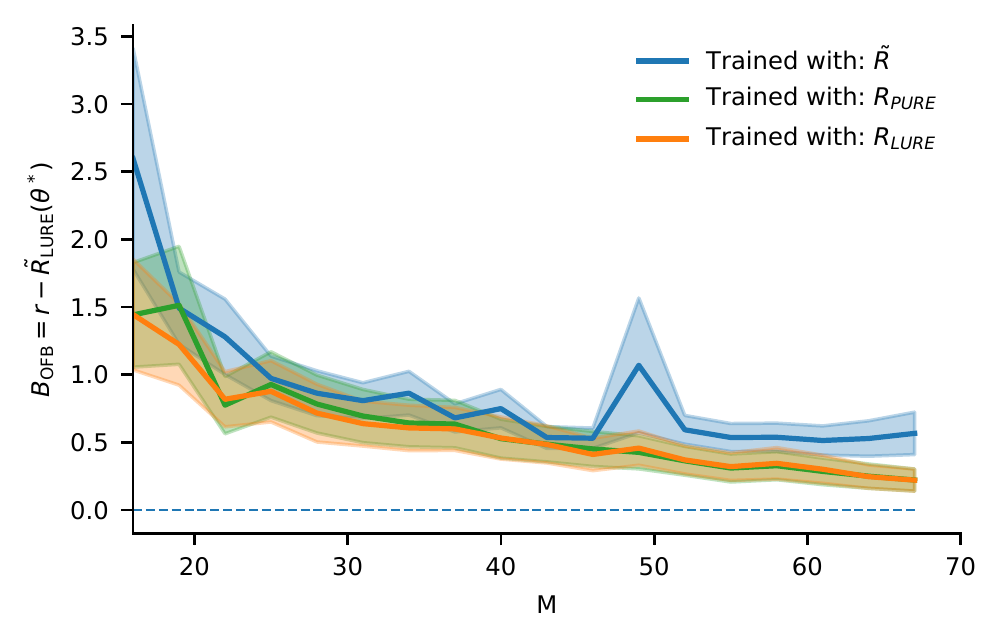}
      \caption{Radial: MNIST}
      \label{fig:bias_mnist_fit}
   \end{subfigure}
   \begin{subfigure}[b]{0.7\textwidth}
      \centering
      \includegraphics[width=\textwidth]{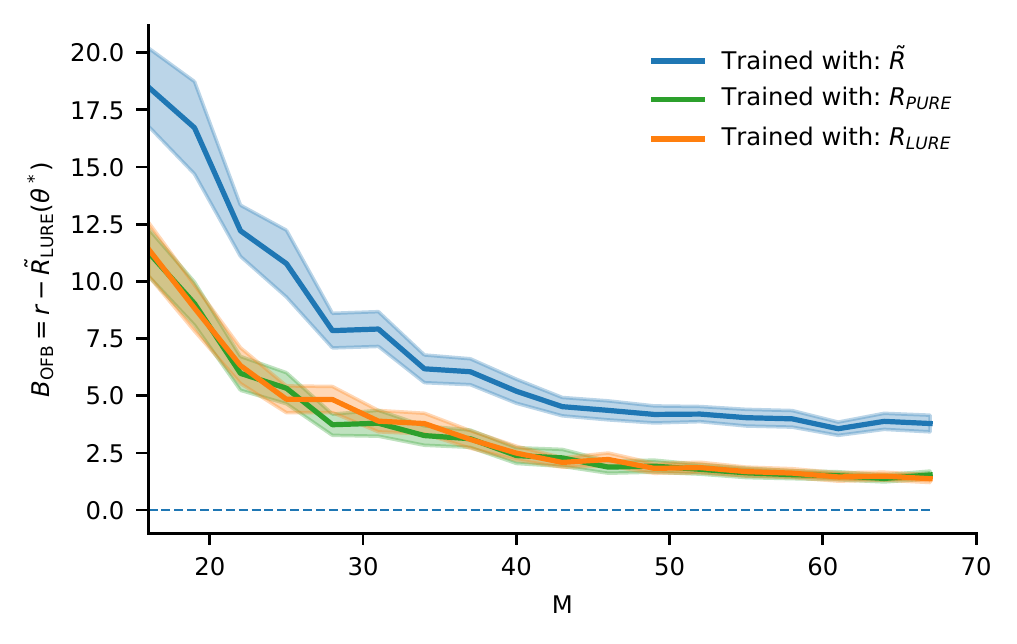}
      \caption{Radial: Fashion}
      \label{fig:bias_fashion_mnist_fit}
   \end{subfigure}

~~~~~~~~
   \caption[Overfitting bias can cancel with the bias introduced by active learning]{Overfitting bias---$B_{\textup{OFB}}$---for models trained using the three objectives. (a) Linear regression, $B_{\textup{OFB}}$ is small compared to ALB (c.f.\ Figure \ref{fig:bias_linear_no_fit}). Shading IQR. 1000 trajectories. (b) Radial BNN, $B_{\textup{OFB}}$ is similar scale and opposite magnitude to ALB (c.f.\ Figure \ref{fig:bias_mnist_no_fit}). (A single outlier 2 orders of magnitude greater than the rest caused the spike at 50 in this figure.)
    (c) Radial BNN on FashionMNIST, OFB is somewhat larger than with MNIST, particularly for $\tilde{R}$ (i.e.~our approaches reduce overfitting) and dominates active learning bias (c.f.\ Figure \ref{fig:bias_fashion_mnist_no_fit}).
     Shading $\pm1$ standard error. 150 trajectories.}
   \label{fig:overfitting_bias}
\end{figure}

\subsection{Related Work on Unbiased Active Learning}
There have been some attempts to address active learning bias, but these have generally required fundamental changes to the active learning approach and only apply to particular setups.
\citet{beygelzimerImportance2009}, \citet{chuUnbiased2011}, and \citep{cortesActive2019} apply importance-sampling corrections \citep{sugiyamaActive2006,bachActive2006} to \emph{online} active learning.
Unlike pool-based active learning, this involves deciding whether or not to sample a new point as it arrives from an infinite distribution.
This makes importance-sampling estimators much easier to develop, but as \citet{settlesActive2010} notes, ``\emph{the pool-based scenario appears to be much more common among application papers}.''

\citet{gantiUPAL} address unbiased active learning in a pool-based setting by sampling from the pool \emph{with replacement}.
This effectively converts pool-based learning into a stationary online learning setting, although it overweights data that happens to be sampled early.
Sampling with replacement is unwanted in active learning because it requires retraining the model on duplicate data which is either impossible or wasteful depending on details of the setting.
Moreover, they only prove the consistency of their estimator under very strong assumptions (well-specified linear models with noiseless labels and a mean-squared-error loss).
\citet{imbergOptimal2020} consider optimal proposal distributions in an importance-sampling setting.
Outside the context of active learning, \citet{byrdWhat2019} question the value of importance-weighting for deep learning, which aligns with our findings below.

\section{Conclusion}
Continual and active learning can both be used to test interesting parts of Bayesian approximate posterior distributions.
In both cases, there are perfectly valid non-Bayesian approaches to solving the problem.
But when a Bayesian approach is taken, performance on the underlying task can be used as an evaluation of the degree to which the Bayesian approximation is succeeding in its goals of updating sequentially or representing uncertainty.

However, we have shown that continual learning evaluations must be carefully constructed to show that it is the Bayesian sequential updating which is doing the `heavy lifting', rather than relying on cached information.
When constructed in a rigorous way many standard approximate inference schemes struggle greatly on even very simple tasks.

Active learning, in contrast, may just be the wrong problem.
We are accidentally evaluating the ability of the approximate posterior to inject implicit biases into the data-gathering which represents prior assumptions which we neither understand nor can specify.
Applying a de-biasing trick can reverse the ordering of model performance indicated by doing active learning.
Perhaps, active model evaluation may be a more suitable evaluation for approximate posterior distributions because it avoids some unintended interactions between model parameters and data which are present in active learning.

However, a major limitation of this entire approach is that it is hard to know how general any lessons learned from these evaluations can be.
If we find that one approximate inference technique performs very well on a well-constructed continual learning evaluation with some dataset, should we expect it to perform well on a different dataset?
On a differently constructed continual learning evaluation?
Of course, we know that predicting this is a hard problem, which receives enormous amounts of attention in the field of machine learning.

Using this evaluation strategy, then, have we really said anything about the underlying approximate inference?
Or are we only able to make much more limited claims about specific problems?

I think the truth is somewhere in between.
The more generalizable the good performance is (across a wide range of specific but similar evaluations) the more confident we can be that it is something about the underlying technique which is producing those results in a generally effective way.
But we will never be able to say with complete confidence, on this approach, that one Bayesian approximation is superior to another (in the way that we might have hoped following the intuitions of Dutch books, for example).

\ifSubfilesClassLoaded{
\bibliographystyle{plainnat}
\bibliography{thesis_references}}{}
\end{document}

\chapter{Conclusion and Discussion}
\label{chp:conclusion}

We set out hoping to motivate a principled automatic system for reasoning about beliefs given evidence.
Bayesian motivations for the fundamental role of belief distributions led us towards intractable Bayesian inference \citep{ramseyTruth1926,definettiForesight1937,coxAlgebra1961,jaynesProbability2003}.
Approximations have a long history of success in science as a way of overcoming intractable but principled calculations.
But the original, historical, motivations for Bayesian inference were binary.
Under the overly-demanding requirements set by those early motivations for Bayesian inference, a set of belief distributions is either `rational' or `irrational' depending on whether or not they follow the rules of probability.
Anything that is approximately Bayesian is, within this framework, more properly \textit{not} Bayesian, and irrational.
This includes both approximations in model choice and computational approximations.
As a result, it appears that no physically instantiable reasoning system (and certainly no person) is `rational' under these overly-demanding requirements.

This leaves us with the central organizing question of this thesis: is there a good way of deciding when an approximately Bayesian method is doing a good job of acting like a Bayesian tool?
In short: how can we evaluate the success of a Bayesian approximation?

One solution which is clearly not good enough is to use the same evaluations that every other method uses.
In principle, Bayesian methods ought to unlock entirely new \textit{machine learning behaviours}.
While Bayesian methods might (or might not) allow better generalization on standard predictive tasks, that should not be their main attraction.
Being able to treat your machine learning platform as a distribution, not just a single estimate, allows you to adapt your entire pipeline in new ways, for example by actively guiding labelling or exploration activity.

We briefly considered using a distance metric or divergence between an approximate and `true' posterior to establish an ordering of more and less Bayesian methods.
Investigating this quickly revealed a problem: not only was it challenging to pick a preferred metric, but plausible candidate metrics disagreed about ordering.
Moreover, we were able to show that the `cost' imposed by various approximations is intimately tied up with details of the situation.
The same approximation algorithm can have very different results depending on the parametric function, approximating family, dataset, and optimization algorithm.
For example, we showed in \cref{chp:parameterization} that the commonly-used mean-field assumption can be extremely restrictive for small models, but may be much less of a constraint for some purposes in larger parametric models whose connections can re-introduce the desired complexity in predictive distributions.
Similarly, we showed in \cref{chp:optimization} that the choice of approximating distribution can interact with the use of gradient-based optimizers by affecting the sampling properties of stochastic gradient estimators.

It also appears to be difficult to adapt the basic structure of the historical, more `binary', arguments motivating Bayesian inference to a measure of `Bayesianness' or degree-of-rationality.
Arguments based on axioms in the style of \citet{coxAlgebra1961} and \citet{jaynesProbability2003} are difficult to express in a `soft' form.
Axioms are either tight enough to constrain us to require our beliefs to be manipulated like probabilities or they are not.
Without a deep conceptual breakthrough of the kind I was not able to discover, it is not clear what sort of pseudo-probability would be strong enough that in the limit of infinite resources becomes perfectly rational and which corresponds to a measure of degrees-of-rationality that allows us to compare two inferences to decide which we ought to prefer.

Similarly, methods that are based on Dutch-books seem to me to be resistant to being made `soft'.
The moment you allow \textit{some} departure from the rational ideal, one's belief distribution is definitely exploitable.
To turn a Dutch-book argument into something that can measure degrees-of-rationality we would need to add concepts for degree-of-exploitability.
My best effort to define these degrees returns again and again to \textit{rates} of loss.
But the actual expected rate of loss depends not on the difference between your own approximation and the true posterior---rather it depends on the interaction between your approximation, the \textit{actual data-distribution}, and your utility function.
At the same time, your maximal rate of loss depends just on the one point in belief-space where your mistake causes the largest possible penalty (in turn this depends on your utility function).
Maximal rate of loss is not sensitive to the rest of your distribution, nor does the Bayesian ideal minimize the maximum rate of loss.
In short, my attempts to soften Dutch-book arguments to define degrees-of-exploitability wound up in the same pseudo-frequentist space as the expected utility measures which I ended up considering.

It seems hard, therefore, to rescue our evaluation of degrees-of-Bayesiannness by adapting the original binary motivations for Bayesian inference.
There are, however, attempts to do something similar from the fields of bounded rationality and probabilistic numerics.

Bounded rationality-style approaches \citep{wheelerBounded2020} offer some mechanisms for understanding the necessity of `imperfect' reasoning due to the importance of approximations and heuristics.
Bounded rationality research has also explored how these sorts of approximations occur in humans both within psychology and economics.
However, less emphasis has been placed on principled reasoning under bounded modelling and computation, and a key question which remains unresolved is how to know which of two approximations is `better' in some context.
It seems unlikely to me that any such answer can be grounded without reference to expected data-distributions and decision-problems which is why I turn to expected utility of approximate distributions.

Relatedly, probabilistic numerics \citep{hennigProbabilistic2022} frames computational approximation as an inference problem in its own right.
When we are trying to estimate some quantity, we can compute it probabilistically and understand the computation as attempting to minimize our uncertainty about the answer given some degree of effort.
However, there are two reasons why this approach does not quite solve our problems.
First, probabilistic numerics broadly describes how to apply probabilistic tools to the problem of numerical estimation, but that does not mean that it provides a way of framing arbitrary numerical methods as probabilistic ones.
That is to say, given some apparently successful numerical method (or heuristic algorithm) it is generally not obvious what probabilistic assumptions correspond to that heuristic.
An example of this can be seen in \cref{chp:parameterization} and \cref{chp:optimization}.
In those chapters, we explored how variational methods relied on pragmatic heuristics that made their Bayesian interpretation very complicated indeed.
This means that probabilistic numerics is almost solving the inverse problem to the one which this thesis focuses on: using Bayesian methods to do numerics rather than understanding what numerical tricks pragmatically-Bayesian methods are actually depending on.
Second, existing approaches to probabilistic numerics broadly do not answer the issue of which modelling assumptions and inference approximations to use as part of the computation effort.
By applying the Bayesian toolkit to the computational procedure we give ourselves all the same tools for model selection that we originally had for Bayesian inference, but no new ones.
In this way, the question of evaluation (which method is `better') is unchanged by the use of probabilistic numerics.
We still either need some `soft' but purely Bayesian criterion of `correct reasoning' or we need to resort to semi-frequentist tools like expected utility.

This inherent case-by-case nature of the quality of Bayesian approximations led us to a more concrete way of understanding approximation quality.
Within a particular \textit{decision-problem}---in the context of a given data-distribution and utility function---we can evaluate the expected utility of an approximate posterior distribution.

At its simplest, we could imagine picking a single universal decision-problem and using it as a proxy for quality generally \citep{keyBayesian1999}.
Although this might seem like a hopelessly naive approach, it is important to remember that much of current machine learning more-or-less does exactly this!
The procedure of comparing computer-vision architectures based on their top-$k$ ImageNet accuracy, for example, is precisely identical with that approach.
We can make this very naive approach richer by increasing the range of performance measures which are considered as well as the variety of decision-problems.
At the furthest opposite extreme, we could say that the notion of general quality is nonsense, and each approximating distribution needs to be judged entirely fresh on each new problem.

This thesis can be thought of as taking some steps to flesh out two possible families of decision-problems as possible evaluations.
A good evaluation should be:
\begin{itemize}
    \item Relevant: good performance on an exemplar of the decision-problem family ought to be useful.
    \item Representative: good performance on an exemplar of the decision-problem family ought to correlate with good performance on all members of the family.
    \item Well-posed: the boundaries of the decision-problem family should be sufficiently clearly specified that practitioners can generally spot when the appropriate evaluation is being used.
    \item Isolated: the decision-problem ought, as much as possible, to test an isolated property of the model being investigated without muddling several mechanisms of action together so that it is hard to know which is causing success or failure.
\end{itemize}

We can select both continual and active learning as decision-problem families that might be suitable for evaluations of approximate Bayesian inference.
Bayesianism characterized by the idea that beliefs about the world ought to be represented by probability distributions.
Continual learning is an example of a decision-problem that checks if the sequential updating process in an approximate Bayesian method is working.
If a method is trying to perform sequential updating but is not succeeding at continual learning, then it is not actually performing sequential probabilistic inference.
Meanwhile, active learning is an example of a decision-problem that checks if the distributions represent beliefs about the world.
If active learning methods based on the expected information gain about those belief distributions are working well then it shows that the belief distributions are reflecting the actual epistemic state with respect to the data and model.

In both cases, I believe that the decision-problems are relevant.
They are fairly representative, but not perfectly.
In both cases, good performance in one setting can be indicative of good general performance on that style of decision-problem.
Continual learning methods' performances can depend on the dataset (witness, for example, the special features of the permuted MNIST dataset for continual learning which are discussed in \cref{chp:evaluation}).
Meanwhile, active learning methods can work well on specific kinds of data without performing well on others (for example, the failures of entropy-based methods to work well for data with heteroskedastic aleatoric noise).

Both active and continual learning are relatively well-posed, at least sufficiently for sub-fields to arise to address each (even if, by my recollections at least, the panel of the 2018 continual learning workshop at ICML agreed to disagree on what the aim of the field was!).
I also chose these two decision-problem families partly because I think they are relatively isolated.
It is entirely possible to build an approximate inference method that performs extremely well on some measures of fitting the underlying dataset but does very badly at either active or continual learning.
For example, a Bayesian neural network fit with variational inference can trivially get near-perfect accuracy on MNIST but can still not carry out single-headed Split MNIST with better than chance catastrophic forgetting.
This suggests that there is something separable about the continual learning and prediction tasks.
Evaluations based on this characteristic behaviour can therefore start to tease apart different sorts of failures and successes of Bayesian inference.

We must be clear that performing well on these problems is \textit{not}, on its own, evidence that a method is doing good Bayesian inference.
On the contrary, it is possible to construct heuristic approaches to these problems which do not rely on any Bayesian fundamentals at all.
For example, one possible solution to continual learning is simply to allow oneself extra parameters for each new task and then to stitch things together at the end \citep{schwarzProgress2018}.
Rather, if a method is constructed to perform approximate Bayesian inference as a solution to these problems but a well-designed evaluation shows that they are performing badly then it is evidence that the approximate inference is not working as intended.

As part of this investigation, however, we came across some ways in which evaluations based on continual or active learning are not optimal.
Continual learning evaluations can be made too easy to solve, and multi-headed versions of the problem add extra parameters that lie outside the motivating Bayesian formalism.
In addition, several methods that are explicitly Bayesian in fact rely heavily on non-Bayesian components that smuggle information from the past via side-channels which lie outside of the approximate inference step.
At the same time, sufficiently rigorous evaluations reveal the simple fact that existing Bayesian approximations in parameter-space are not doing a good enough job capturing the parameter distribution to allow sequential updating.

Active learning, meanwhile, is maybe not quite isolated enough.
We showed that the goal of picking informative points was interacting with some effective heuristics which act as implicit regularization.
While there is nothing wrong with performing implicit regularization, because it lies outside the explicit probabilistic formalism of the problem definition, the importance of the implicit inductive bias means that active learning is an unreliable evaluation for the effectiveness of the underlying approximate inference.

Instead of either of these, we propose two scenarios that appear to be good evaluations of important parts of Bayesian inference.
We see approximate Bayesian inference as chiefly characterized by the idea that distributions over parameters reflect subjectively uncertain beliefs about the states they represent.
In this context, continual learning and active learning are especially relevant.
The former tests the ability of the current belief representation to encode the evidence accumulated so far.
The latter tests the ability of the current belief representation to reflect missing information.

We considered whether active \textit{testing} \citep{kossenActive2021} might be an alternative evaluation for approximate inference.
In the case of active testing, because the model-performance itself does not depend on the implicit bias, removing the bias might be just the `right' thing to do, providing a clearer evaluation metric.
In fact, however, it is not necessarily the case that this is better.
After all, a biased but lower-variance testing regime might be better for some utility functions, making even active testing a possibly non-isolated evaluation for approximate Bayesian inference.

Despite the challenges facing both continual and active learning as evaluations of Bayesian inference, I nevertheless believe that these are an excellent place to start.
Future work ought to develop more, and more careful, evaluations that test the characteristic behaviours of Bayesian inference in a variety of ways that collectively allow us to build a picture of the successes and failures of different approximation methods in different settings.

We should also always bear in mind that these evaluations cannot be blindly applied, because testing the quality of the Bayesian approximation depends on the evaluation being set up in a way that \textit{actually} tests that.
It would be easy, for example, to design a so-called Bayesian approximation with the problem of active learning specifically in mind which smuggled some geometric heuristics under the hood and performed very well on some example problem where those heuristics happened to be effective.
At the same time, an approximation which performs extremely well for one dataset or architecture might genuinely perform badly on another because of unexpected interactions between the underlying components of the approximation and learning procedure.

An evaluation is not quite the same thing as a benchmark, and I suspect that the best ways to identify the ways in which Bayesian approximations are and are not working might not lend themselves very well to benchmark construction because they might be easy to fake.
As scientists, however, we can rise to the challenge of probing our models fairly and carefully and trying to genuinely understand which ones are working and why.

\ifSubfilesClassLoaded{
\bibliographystyle{plainnat}
\bibliography{thesis_references}}{}
\end{document}

\bibliographystyle{plainnat}
\bibliography{thesis_references}

\appendix

\ifSubfilesClassLoaded{
  \newtheorem{lemma}{Lemma}
  \newtheorem{theorem}{Theorem}
  \newtheorem{proposition}{Proposition}
  \newtheorem{remark}{Remark}
}{}
\newcommand{\red}[1]{\textcolor{red}{#1}}
\chapter{Appendix to Chapter 3}
\label{app:parameterization}

  \begin{table}[]
    \begin{tabular}{@{}ll@{}}
    \toprule
    Hyperparameter & Setting description                                             \\ \midrule
    Architecture                                     & MLP                                                             \\
    Number of hidden layers                          & 3                                                            \\
    Layer Width                                      & 100                                                              \\
    Activation                                       & Leaky ReLU                          \\
    Approximate Inference Algorithm                  & Mean-field VI (Flipout \citep{wenFlipout2018})                 \\
    Optimization algorithm                           & Amsgrad \citep{reddiConvergence2018}       \\
    Learning rate                                    & $10^{-3}$                                      \\
    Batch size                                       & 250                                                              \\
    Variational training samples                     & 1                                                              \\
    Variational test samples                         & 1                                                              \\
    Temperature & 65 \\
    Noise scale & 0.05 \\
    Epochs & 6000\\
    Variational Posterior Initialization               & Tensorflow Probability default                         \\
    Prior                                       & $\mathcal{N}(0, 1.0^2)$                                                               \\
    Dataset                                          & Toy (see text) \\
    Number of training runs & 1 \\
    Number of evaluation runs & 1 \\
    Measures of central tendency & n.a. \\
    Runtime per result & $<5$m\\
    Computing Infrastructure & Nvidia GTX 1060 \\
    \bottomrule
    \end{tabular}
    \caption[Hyperparameters for toy regression visualization]{Experimental Setting---Toy Regression Visualization. Note that for these visualizations we are purely demonstrating the possibility of in-between uncertainty. As a result, a single training/evaluation run suffices to make an existence claim, so we do not do multiple runs in order to calculate a measure of central tendency.}
    \label{tbl:hypers-toy}
    \end{table}
  
  \section{Experimental Details}
  \subsection{Full Description of Covariance Visualization}\label{a:heatmap}
  Here we provide details on the method used to produce \cref{fig:cov_heatmap}.
  The linear version of the visualization is discussed in \cref{ss:parameterization:linear} and the piecewise-linear version is discussed in \cref{ss:parameterization:piecewise}.

  In all cases, we train a neural network using mean-field variational inference in order to visualize the covariance of the product matrix.
  The details of training are provided in \cref{tbl:hypers-heatmap}.
  The product matrix is calculated from the weight matrices of an $L$-layer network.
  In the linear case, this is just the matrix product of the $L$ layers.
  In the piecewise-linear case the definition of the product matrix is described in more detail in \cref{a:lpm_def}.
  All covariances are calculated using 10,000 samples from the converged approximate posterior.
  Note that for $L$ weight matrtices there are $L-1$ layers of hidden units.
\begin{table}[]
  \begin{tabular}{@{}ll@{}}
  \toprule
  Hyperparameter & Setting description                                             \\ \midrule
  Architecture                                     & MLP                                                             \\
  Number of hidden layers                          & 0-9                                                            \\
  Layer Width                                      & 16                                                              \\
  Activation                                       & Linear or Leaky Relu with $\alpha=0.1$                          \\
  Approximate Inference Algorithm                  & Mean-field Variational Inference                                \\
  Optimization algorithm                           & Amsgrad \citep{reddiConvergence2018}       \\
  Learning rate                                    & $10^{-3}$                                      \\
  Batch size                                       & 64                                                              \\
  Variational training samples                     & 16                                                              \\
  Variational test samples                         & 16                                                              \\
  Epochs & 10\\
  Variational Posterior Initial Mean               & \citet{heDeep2016}                         \\
  Variational Posterior Initial Standard Deviation & $\log[1 + e^{-3}]$          \\
  Prior                                       & $\mathcal{N}(0, 0.23^2)$                                                               \\
  Dataset                                          & FashionMNIST \citep{xiaoFashionMNIST2017} \\
  Preprocessing                                    & Data normalized $\mu=0$, $\sigma=1$                             \\
  Validation Split                                 & 90\% train - 10\% validation\\
  Number of training runs & 1 \\
  Number of evaluation runs & 1 \\
  Measures of central tendency & n.a. \\
  Runtime per result & $<3$m\\
  Computing Infrastructure & Nvidia GTX 1080 \\
  \bottomrule
  \end{tabular}
  \caption[Hyperparameters for covariance visualization]{Experimental Setting---Covariance Visualization. Note that for these visualizations we are purely demonstrating the possibility of off-diagonal covariance. As a result, a single training/evaluation run suffices to make an existence claim, so we do not do multiple runs in order to calculate a measure of central tendency.}
  \label{tbl:hypers-heatmap}
  \end{table}
  
  We compare these learned product matrices, in \cref{fig:no_block}, to a randomly sampled product matrix.
  To do so, we sample weight layers whose entries are distributed normally.
  Each weight is sampled with standard deviation 0.3 and with a mean 0.01 and each weight matrix is 16x16.
  This visualization is with a linear product matrix of 5 layers.

  \subsection{Effect of Depth Measured on Iris Experimental Settings}\label{a:iris}
  \begin{table}[]
    \begin{tabular}{@{}ll@{}}
    \toprule
    Hyperparameter & Setting description                                             \\ \midrule
    Architecture                                     & MLP                                                             \\
    Number of hidden layers                          & 1-4                                                            \\
    Layer Width                                      & 4                                                              \\
    Activation                                       & Leaky Relu                          \\
    Approximate Inference Algorithm                  & Variational Inference                                \\
    Optimization algorithm                           & Amsgrad \citep{reddiConvergence2018}       \\
    Learning rate                                    & $10^{-3}$                                      \\
    Batch size                                       & 16                                                              \\
    Variational training samples                     & 1                                                              \\
    Variational test samples                         & 1                                                              \\
    Epochs & 1000 (early stopping patience=30)\\
    Variational Posterior Initial Mean               & \citet{heDeep2016}                         \\
    Variational Posterior Initial Standard Deviation & $\log[1 + e^{-6}]$          \\
    Prior                                       & $\mathcal{N}(0, 1.0^2)$                                                               \\
    Dataset                                          & Iris \citep{xiaoFashionMNIST2017} \\
    Preprocessing                                    & None.                             \\
    Validation Split                                 & 100 train - 50 test\\
    Number of training runs & 100 \\
    Number of evaluation runs & 100 \\
    Measures of central tendency & (See text.) \\
    Runtime per result & $<5$m\\
    Computing Infrastructure & Nvidia GTX 1080 \\
    \bottomrule
    \end{tabular}
    \caption[Hyperparameters for full-covariance experiment]{Experimental Setting---Full Covariance.}
    \label{tbl:hypers-iris}
    \end{table}
  We describe the full- and diagonal-covariance experiment settings in \cref{tbl:hypers-iris}.
  We use a very small model on a small dataset because full-covariance variational inference is unstable, requiring a matrix inversion of a $K^4$ matrix for hidden unit width $K$.
  Unfortunately, for deeper models the initializations still resulted in failed training for some seeds.
  To avoid this issue, we selected the 10 best seeds out of 100 training runs, and report the mean and standard error for these.
  Because we treat full- and diagonal-covariance in the same way, the resulting graph is a fair reflection of their relative best-case merits, but not intended as anything resembling a `real-world' performance benchmark.
  
  Readers may consider the Iris dataset to be unhelpfully small, however this was a necessary choice.
  We note that the small number of training points creates a broad posterior, which is the best-case scenario for a full-covariance approximate posterior.

  \subsection{HMC Experimental Settings}\label{a:parameterization:hmc}

  \begin{table}[]
    \resizebox{\textwidth}{!}{
    \begin{tabular}{@{}l
    >{\columncolor[HTML]{EFEFEF}}l 
    >{\columncolor[HTML]{EFEFEF}}l ll
    >{\columncolor[HTML]{EFEFEF}}l 
    >{\columncolor[HTML]{EFEFEF}}l ll@{}}
    \toprule
                            & \multicolumn{2}{c}{\cellcolor[HTML]{EFEFEF}ReLU} & \multicolumn{2}{c}{Leaky ReLU 0.5} & \multicolumn{2}{c}{\cellcolor[HTML]{EFEFEF}Leaky ReLU 0.95} & \multicolumn{2}{c}{Linear}      \\ 
    \# Hidden Layers & Test Acc.          & Acceptance         & Test Acc.   & Acceptance  & Test Acc.               & Acceptance               & Test Acc. & Acceptance \\\midrule
    1                       & 99.1\%                 & 84.8\%                  & 98.4\%          & 84.9\%           & 91.9\%                      & 85.4\%                        & 83.9\%        & 78.0\%          \\
    2                       & 99.7\%                 & 77.0\%                  & 99.5\%          & 73.9\%           & 96.4\%                      & 76.3\%                        & 84.2\%        & 44.4\%          \\
    3                       & 99.1\%                 & 58.0\%                  & 99.6\%          & 46.3\%           & 97.2\%                      & 74.5\%                        & 84.4\%        & 37.0\%          \\
    4                       & 99.5\%                 & 62.2\%                  & 99.6\%          & 50.9\%           & 95.8\%                      & 68.2\%                        & 84.4\%        & 43.2\%          \\
    5                       & 98.1\%                 & 61.8\%                  & 99.5\%          & 53.8\%           & 98.4\%                      & 62.4\%                        & 84.3\%        & 35.2\%          \\
    6                       & 95.4\%                 & 78.5\%                  & 99.6\%          & 51.0\%           & 98.0\%                      & 62.6\%                        & 84.1\%        & 33.7\%          \\
    7                       & 92.7\%                 & 68.1\%                  & 99.7\%          & 54.6\%           & 97.5\%                      & 59.7\%                        & 84.0\%        & 33.0\%          \\
    8                       & 87.8\%                 & 68.3\%                  & 99.6\%          & 49.7\%           & 98.0\%                      & 62.5\%                        & 83.8\%        & 36.4\%          \\
    9                       & 80.6\%                 & 73.9\%                  & 99.6\%          & 46.3\%           & 97.4\%                      & 60.2\%                        & 83.9\%        & 36.5\%          \\
    10                      & 74.6\%                 & 74.9\%                  & 99.5\%          & 45.7\%           & 97.1\%                      & 61.8\%                        & 83.8\%        & 40.4\%          \\ \bottomrule
    \end{tabular}}
    \vspace{2mm}
    \caption[HMC diagnostics for ReLU networks]{HMC samples for ReLU networks are most accurate for smaller numbers of layers, the samples from deeper models may therefore be slightly less reliable. Acceptance rates tend to be with 10-20 percentage points of 65\%, regarded as a good balance of exploration to avoiding unnecessary resampling. The more linear models are less accurate, as one would expect for a dataset that is not linearly separable.}
    \label{tbl:hmc}
\end{table}
  We begin by sampling from the true posterior using HMC.
  
  We use the simple two-dimensional binary classification `make moons' task.
  We use 500 training points (generated using $\mathrm{random\_state=0}$).\footnote{\url{https://scikit-learn.org/stable/modules/generated/sklearn.datasets.make_moons.html\#sklearn.datasets.make_moons}}
  Using \citet{cobbIntroducing2019}, we apply the No-U-turn Sampling scheme \citep{hoffmanNoUTurn2014} with an initial step size of 0.01.
  We use a burn-in phase with 10,000 steps targetting a rejection rate of 0.8.
  We then sample until we collect 1,000 samples from the true posterior, taking 100 leapfrog steps in between every sample used in order to ensure samples are less correlated.
  For each result, we recalculate the HMC samples 20 times with a different random seed.
  All chains are initialized at the optimum of a mean-field variational inference model in order to help HMC rapidly find a mode of the true posterior.
  We use a prior precision, normalizing constant, and $\tau$ of 1.0.
  The model is designed to have as close to 1000 non-bias parameters each time as possible, adjusting the width given the depth of the model.
  We observe that the accuracies for the ReLU network fall for the deeper models, suggesting that after about 7 layers the posterior estimate may become slightly less reliable (see \cref{tbl:hmc}).
  Acceptance rates are broadly in a sensible region for most of the chains.
  
  Using these samples, we find a Gaussian fit.
  For each model we fit a Gaussian mixture model with between 1 and 4 components and pick the one with the best Bayesian information criterion (see \cref{fig:mode_selection}).
  We then find the best diagonal fit to this distribution, which is a Gaussian distribution with the same mean and with a precision matrix equal to the inverse of the diagonal precision of the full-covariance Gaussian.
  We do this because variational inference uses the mode-seeking KL-divergence, so we are interested in the properties of a single Gaussian mode.
  The overall empirical covariance would lead to a mode-covering distribution, while optimizing the mode-seeking KL to the empirical distribution would of course result in a point-sized distribution centred at one of the HMC samples.
  Using one mode of a mixture of Gaussians is therefore the closest we can come to finding a single mode of the true posterior of the sort that VI might uncover.
  
  Finally, we calculate the KL divergence between these two distributions.
  The graph reports the mean and shading reflects the standard error of the mean, though note that because all runs are initialized from the same point, this underestimates the overall standard error.
  All experiments in this an other sections were run on a desktop workstation with an Nvida 1080 GPU.

  For the Wasserstein divergence, we estimate the distance between the empirical distributions formed by the HMC samples and samples from the full- and diagonal-covariance posterior approximation.
  We used the Python Optimal Transport package to estimate the divergence.
  
  \subsection{HMC Investigation of CIFAR10}
  \label{a:cifar10_hmc}

  Our investigation of ResNet-20-FRN in CIFAR10 shows broadly similar results to that of CIFAR100 but are shown here for completeness.

\begin{figure}
  \centering
  \begin{subfigure}[b]{\textwidth}
    \includegraphics[width=0.7\textwidth]{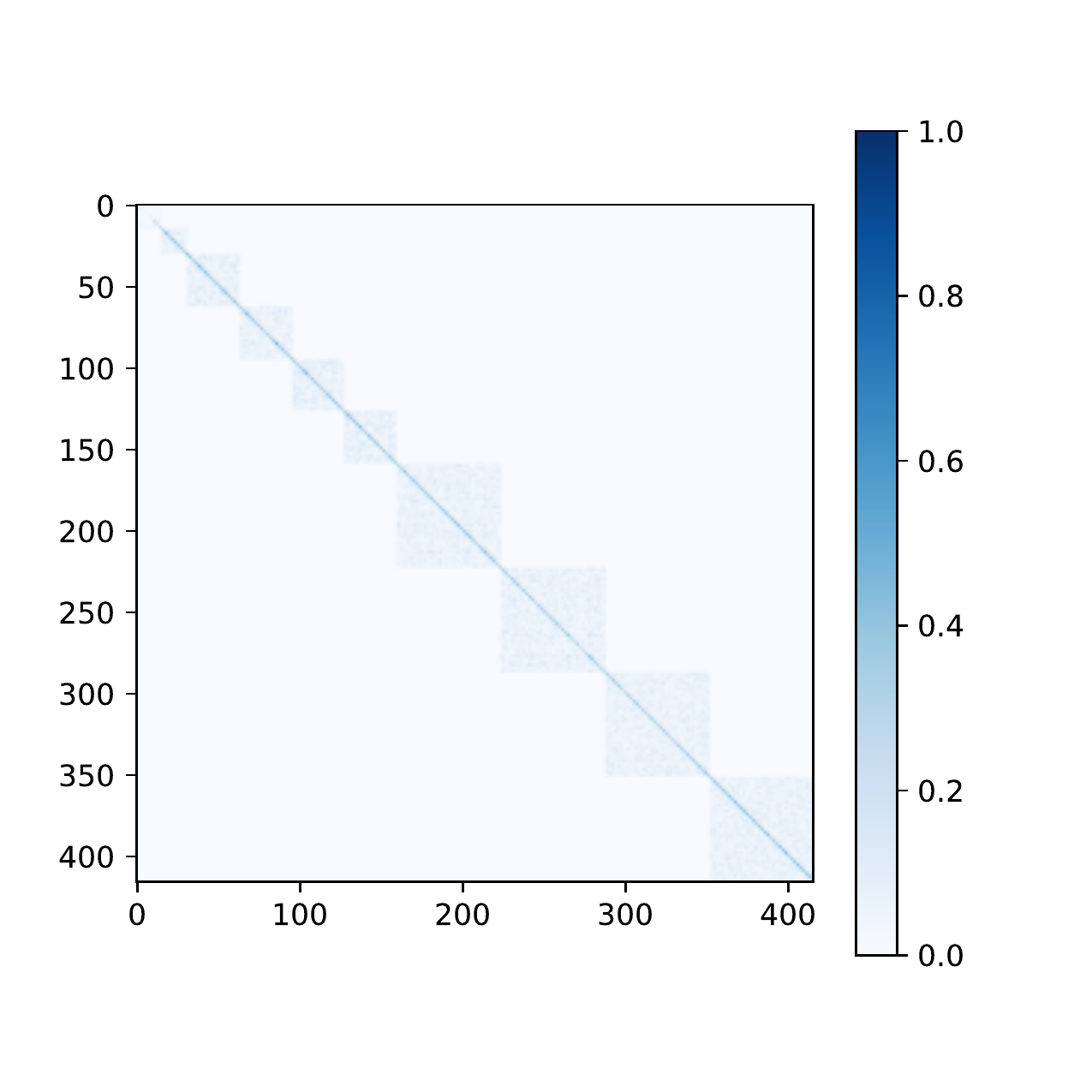}
    \caption{}
  \end{subfigure} \hfill
  \begin{subfigure}[b]{\textwidth}
    \includegraphics[width=0.7\textwidth]{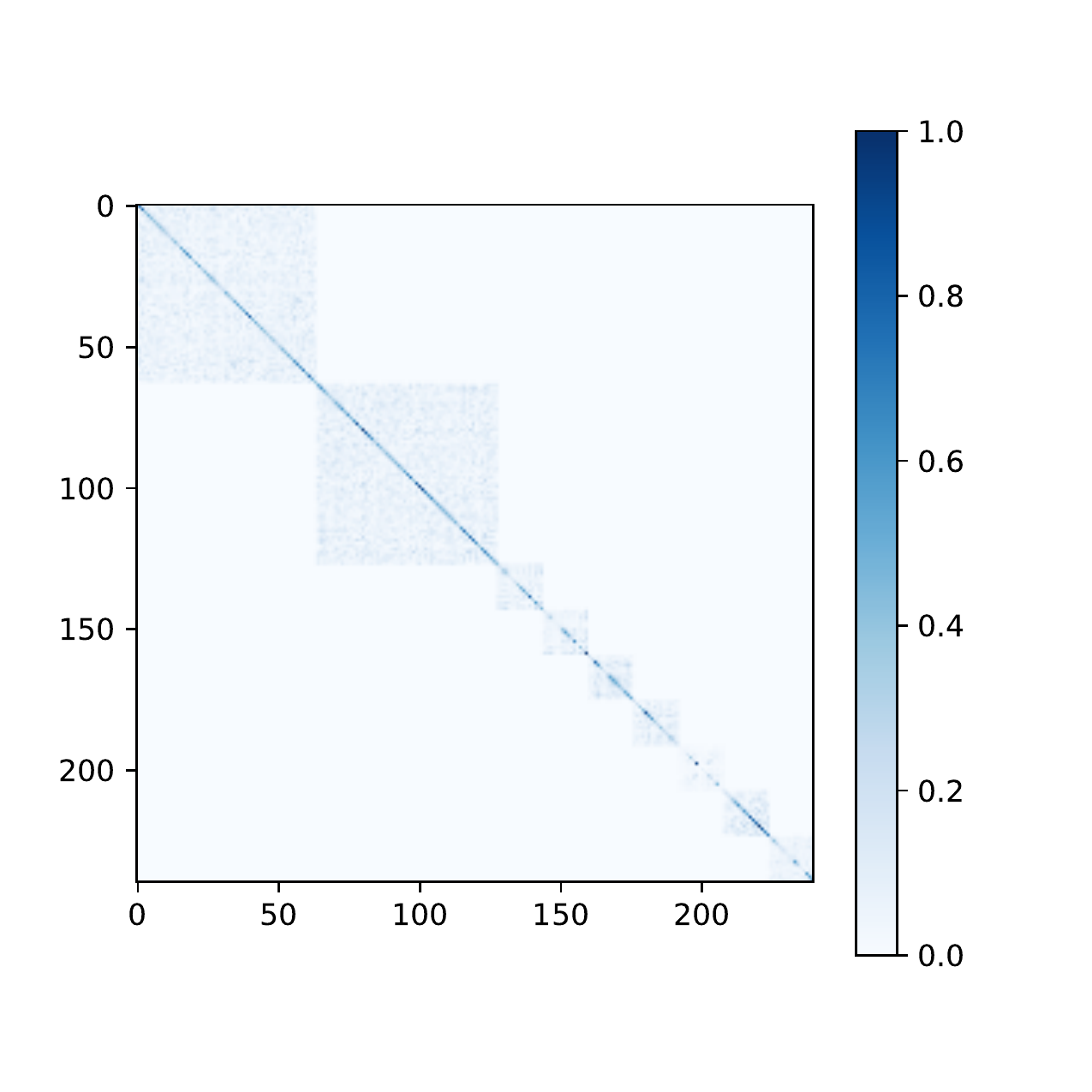}
    \caption{}
  \end{subfigure}
\end{figure}
\begin{figure}
  \ContinuedFloat
  \centering
  \begin{subfigure}[b]{\textwidth}
    \includegraphics[width=0.7\textwidth]{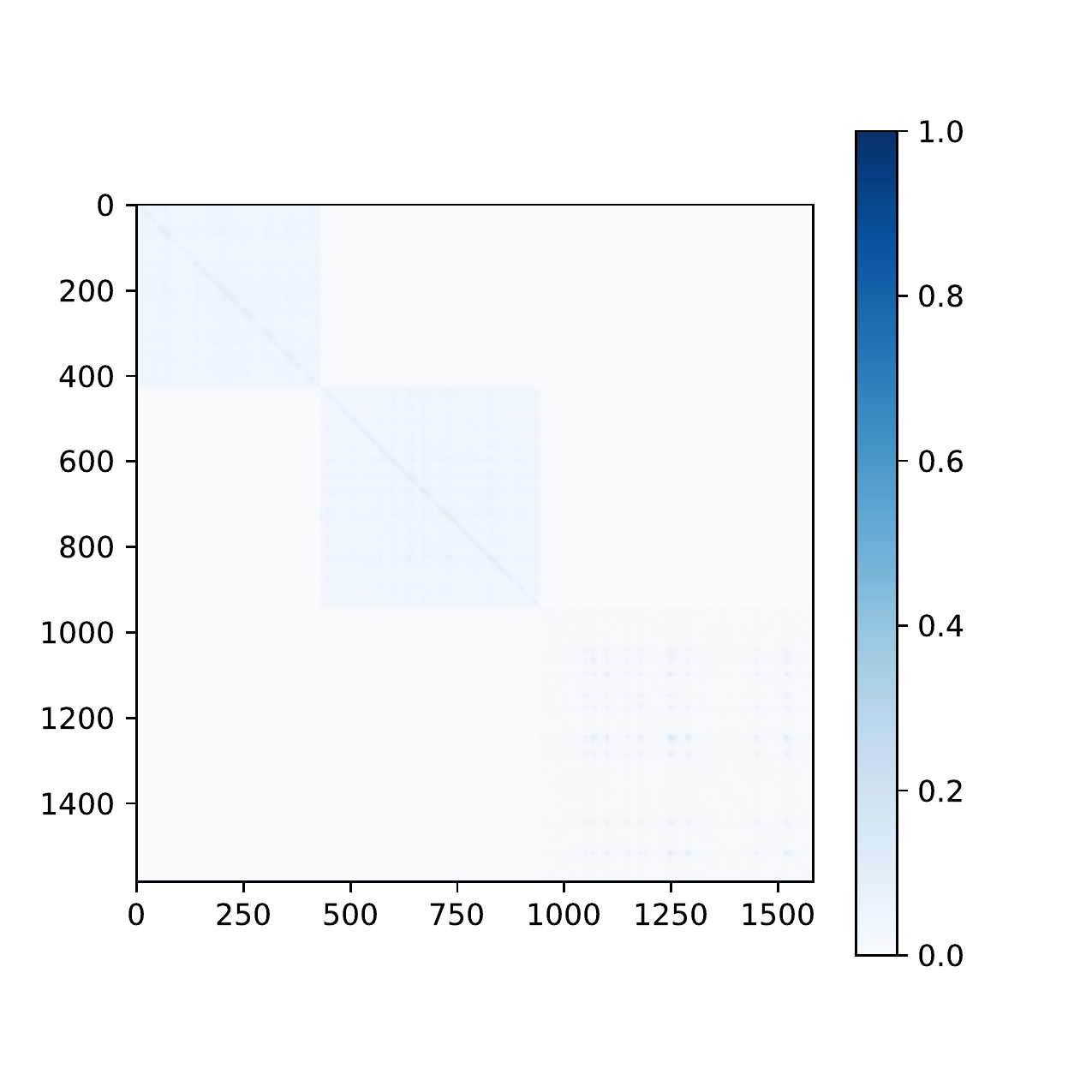}
    \caption{}
  \end{subfigure} \hfill
  \caption[CIFAR10, ResNet-20-FRN]{CIFAR10, ResNet-20-FRN: We show the HMC posterior sample covariance. Results are similar to CIFAR100. (a) Early in the network (first ten layers with fewer than 100 parameters), convolutional layers have mostly diagonal covariances. (b) In the middle of the network (61st to 70th layers with fewer than 100 parameters), there is significant off-diagonal covariance. (c) Larger layers (between 100 and 1000 parameters) also seem to show more off-diagonal covariance. However, estimation error is greater for these layers.}
  \label{fig:cifar10_hmc}
\end{figure}

\subsubsection{Figures for other chains}
Here we provide correlation plots for the second and third chains provided by \citet{izmailovWhat2021}.
\begin{figure}
  \centering
  \begin{subfigure}[b]{\textwidth}
    \includegraphics[width=0.7\textwidth]{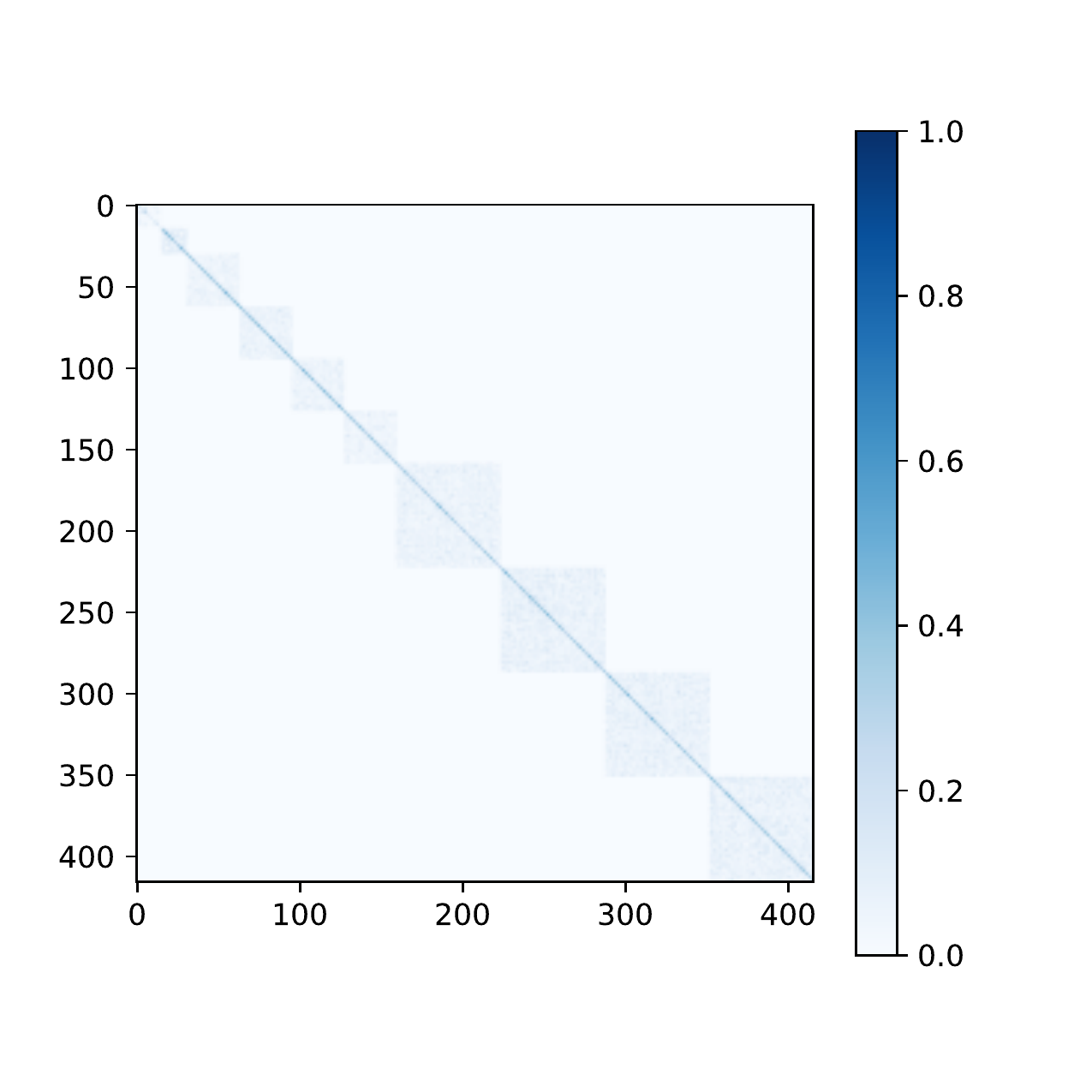}
    \caption{}
  \end{subfigure} \hfill
  \begin{subfigure}[b]{\textwidth}
    \includegraphics[width=0.7\textwidth]{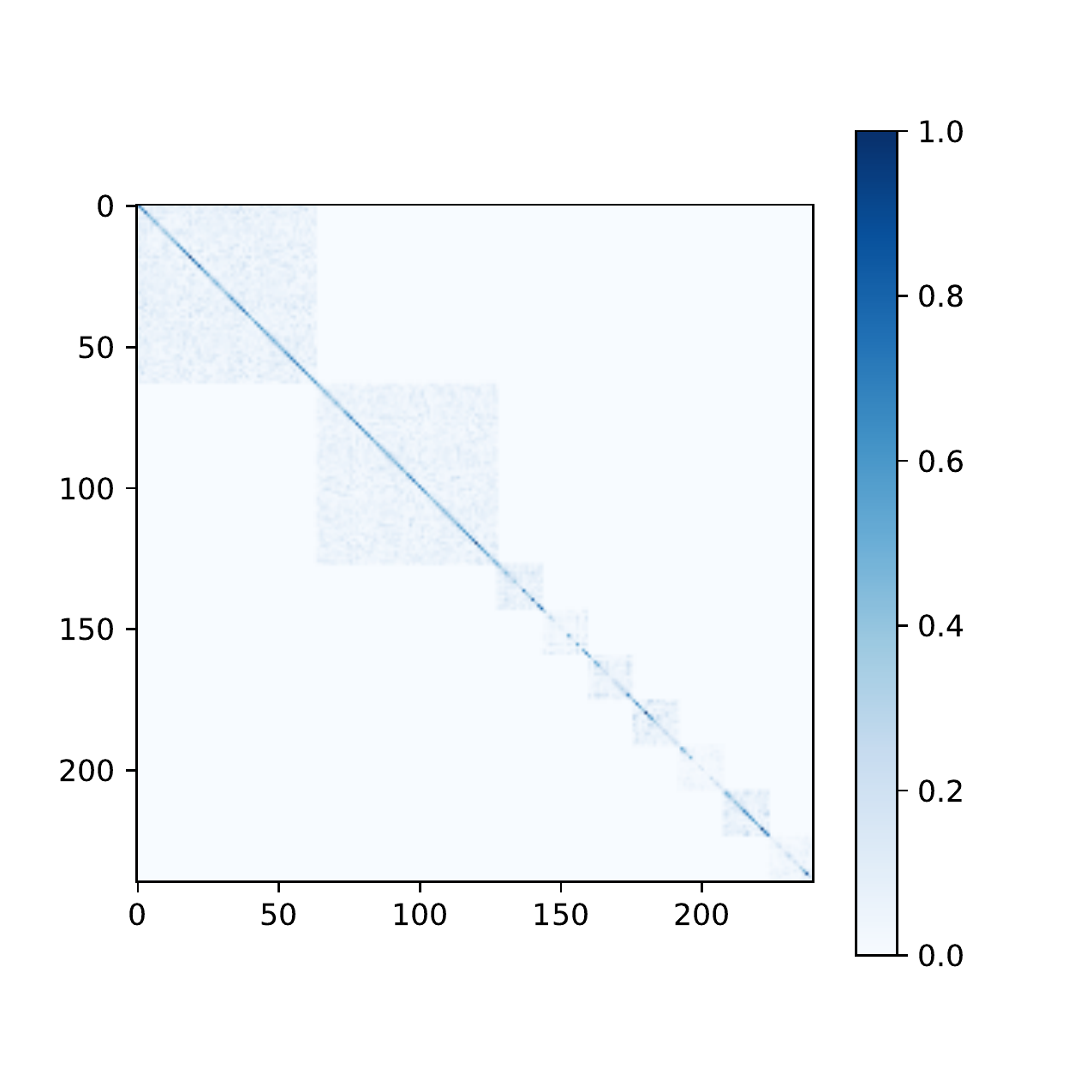}
    \caption{}
  \end{subfigure}
\end{figure}
\begin{figure}
  \ContinuedFloat
  \centering
  \begin{subfigure}[b]{\textwidth}
    \includegraphics[width=0.7\textwidth]{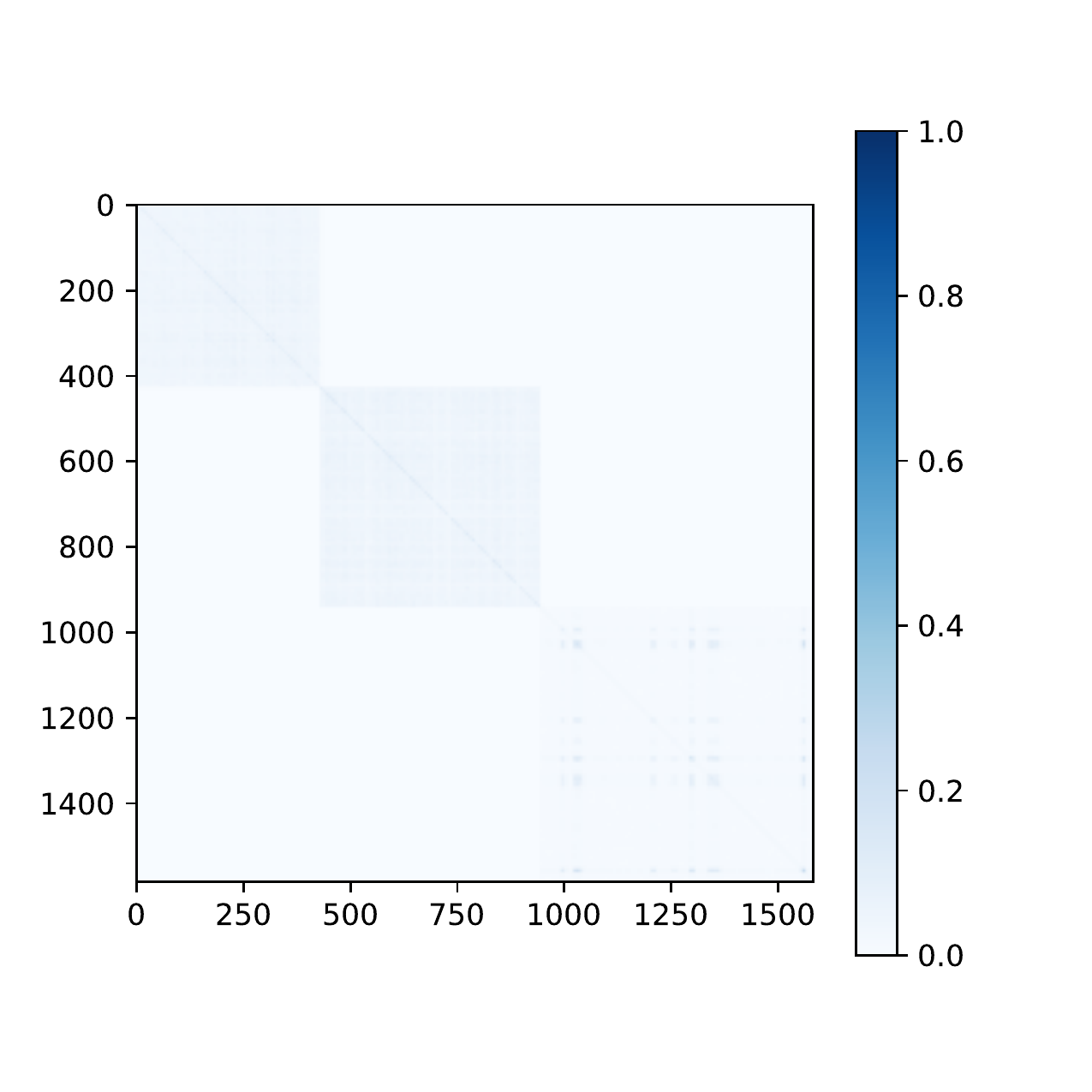}
    \caption{}
  \end{subfigure} \hfill
  \caption[CIFAR10, ResNet-20-FRN with second chain.]{We show the posterior sample covariance for CIFAR10, ResNet-20-FRN on the second chain. (a) First ten layers with fewer than 100 parameters. (b) 61st through 70th layers with fewer than 100 parameters. (c) Layers with 100-1000 parameters.}
  \label{fig:cifar10_hmc_1}
\end{figure}
\begin{figure}
  \centering
  \begin{subfigure}[b]{\textwidth}
    \includegraphics[width=0.7\textwidth]{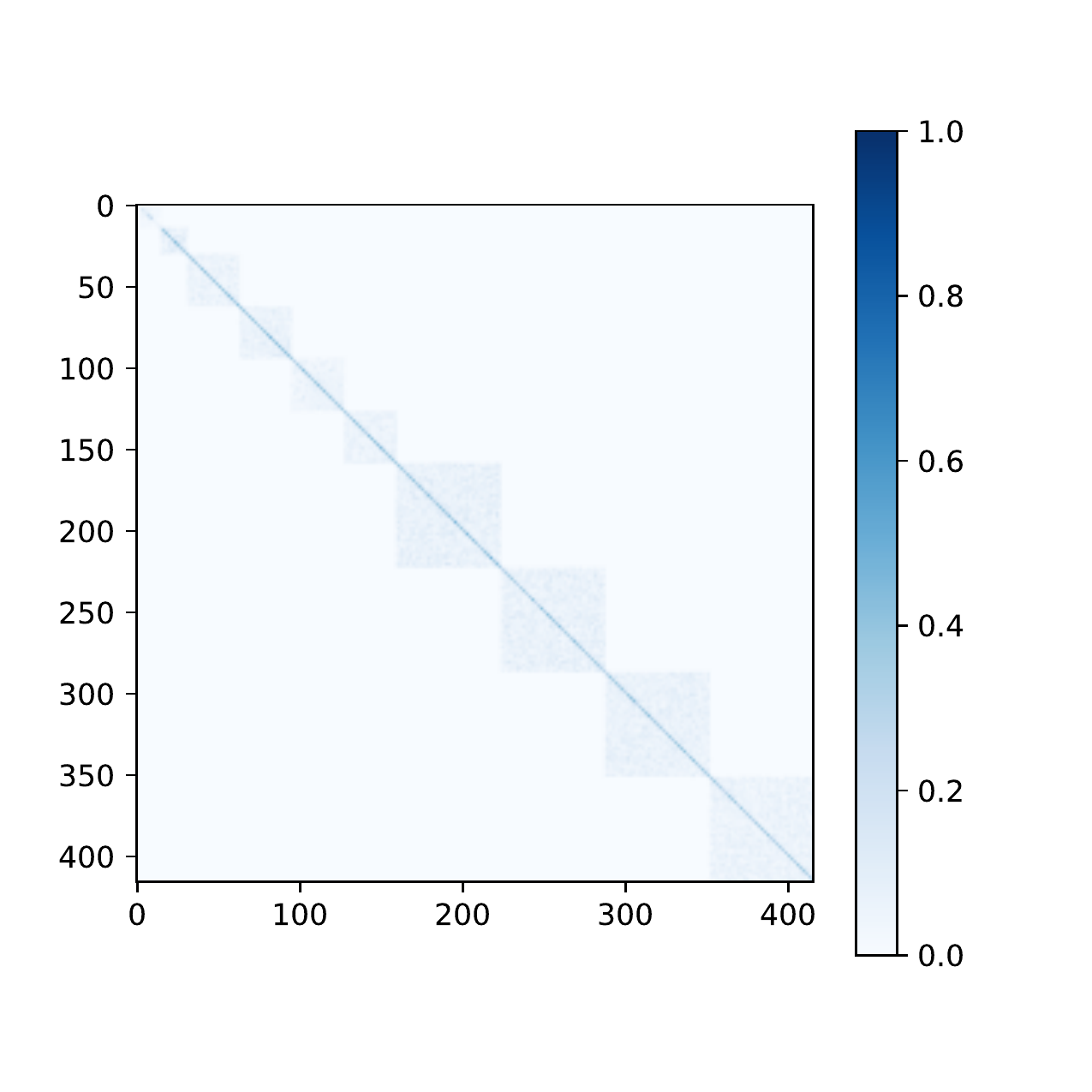}
    \caption{}
  \end{subfigure} \hfill
  \begin{subfigure}[b]{\textwidth}
    \includegraphics[width=0.7\textwidth]{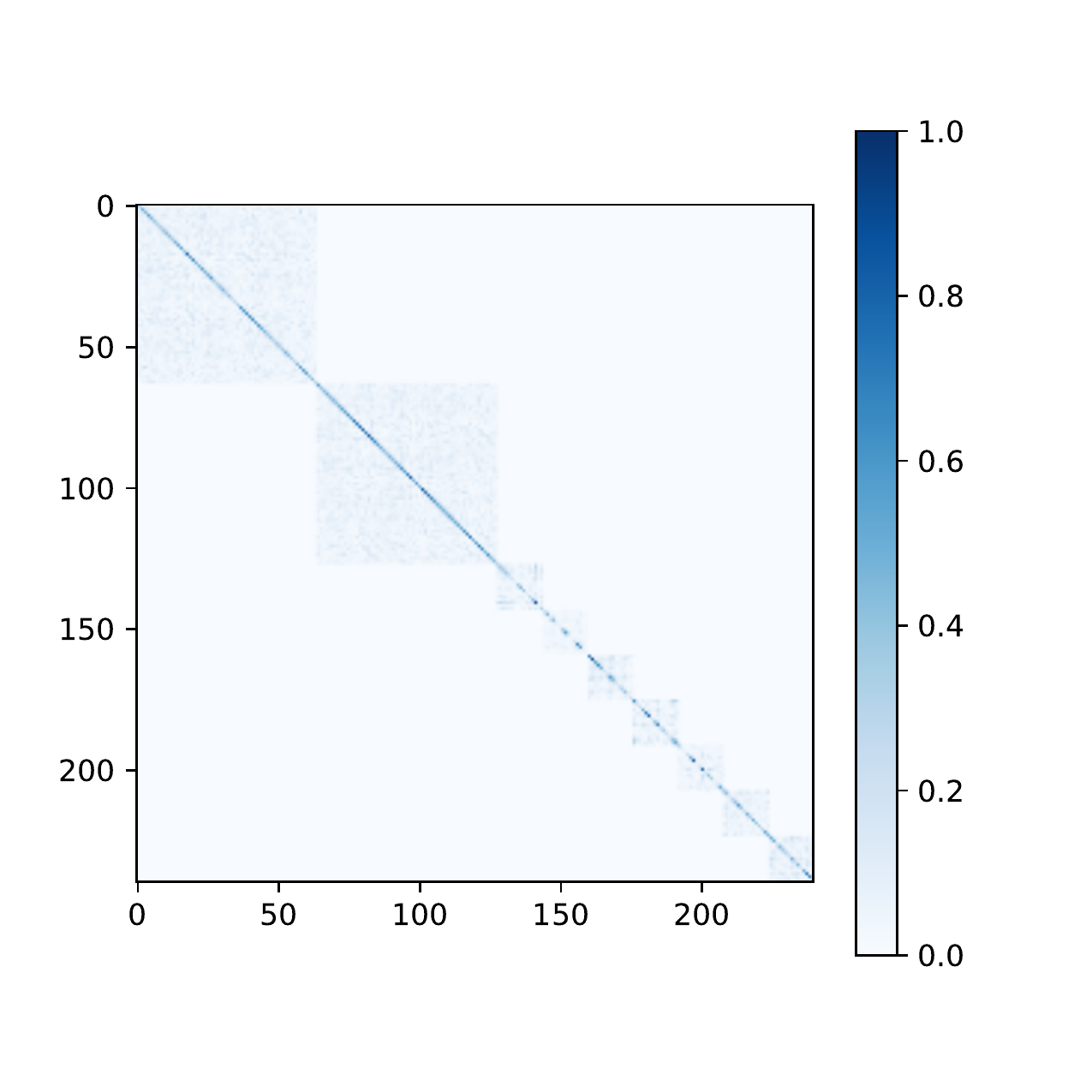}
    \caption{}
  \end{subfigure}
\end{figure}
\begin{figure}
  \ContinuedFloat
  \centering
  \begin{subfigure}[b]{\textwidth}
    \includegraphics[width=0.7\textwidth]{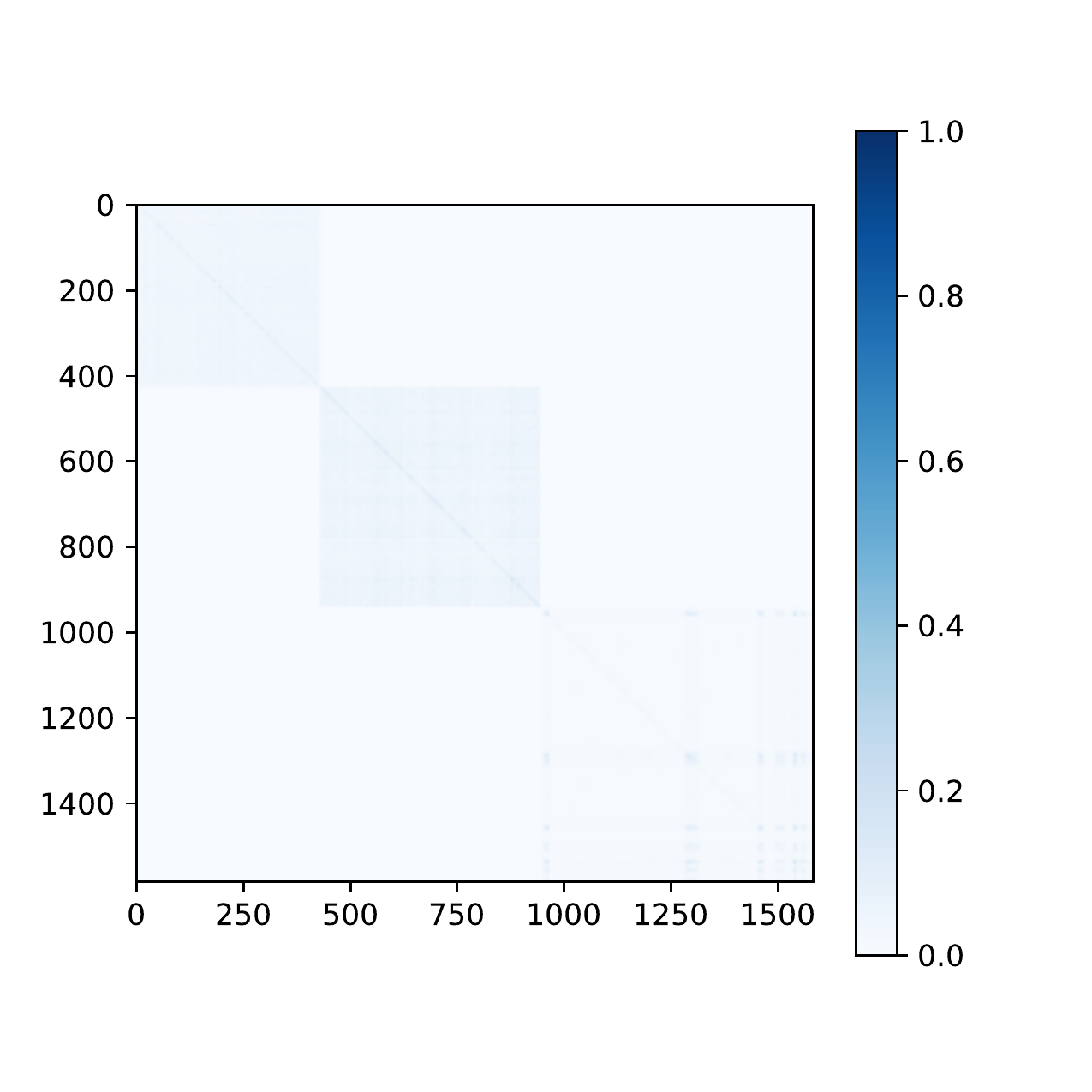}
    \caption{}
  \end{subfigure} \hfill
  \caption[CIFAR10, ResNet-20-FRN with third chain.]{We show the HMC posterior sample covariance for CIFAR10, ResNet-20-FRN on the third chain. (a) First ten layers with fewer than 100 parameters. (b) 61st through 70th layers with fewer than 100 parameters. (c) Layers with 100-1000 parameters.}
  \label{fig:cifar10_hmc_2}
\end{figure}
\clearpage
\begin{figure}
  \centering
  \begin{subfigure}[b]{\textwidth}
    \includegraphics[width=0.7\textwidth]{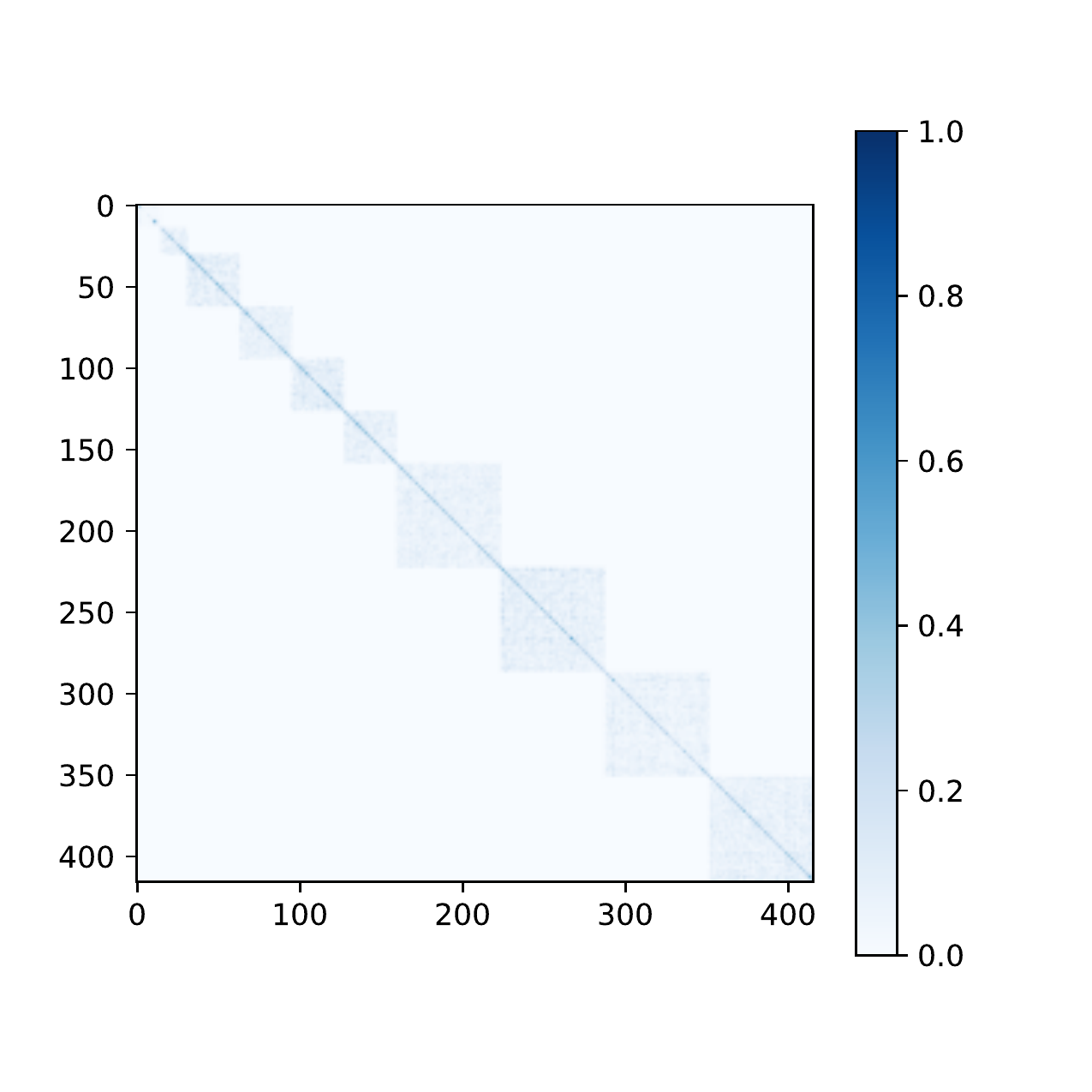}
    \caption{}
  \end{subfigure} \hfill
  \begin{subfigure}[b]{\textwidth}
    \includegraphics[width=0.7\textwidth]{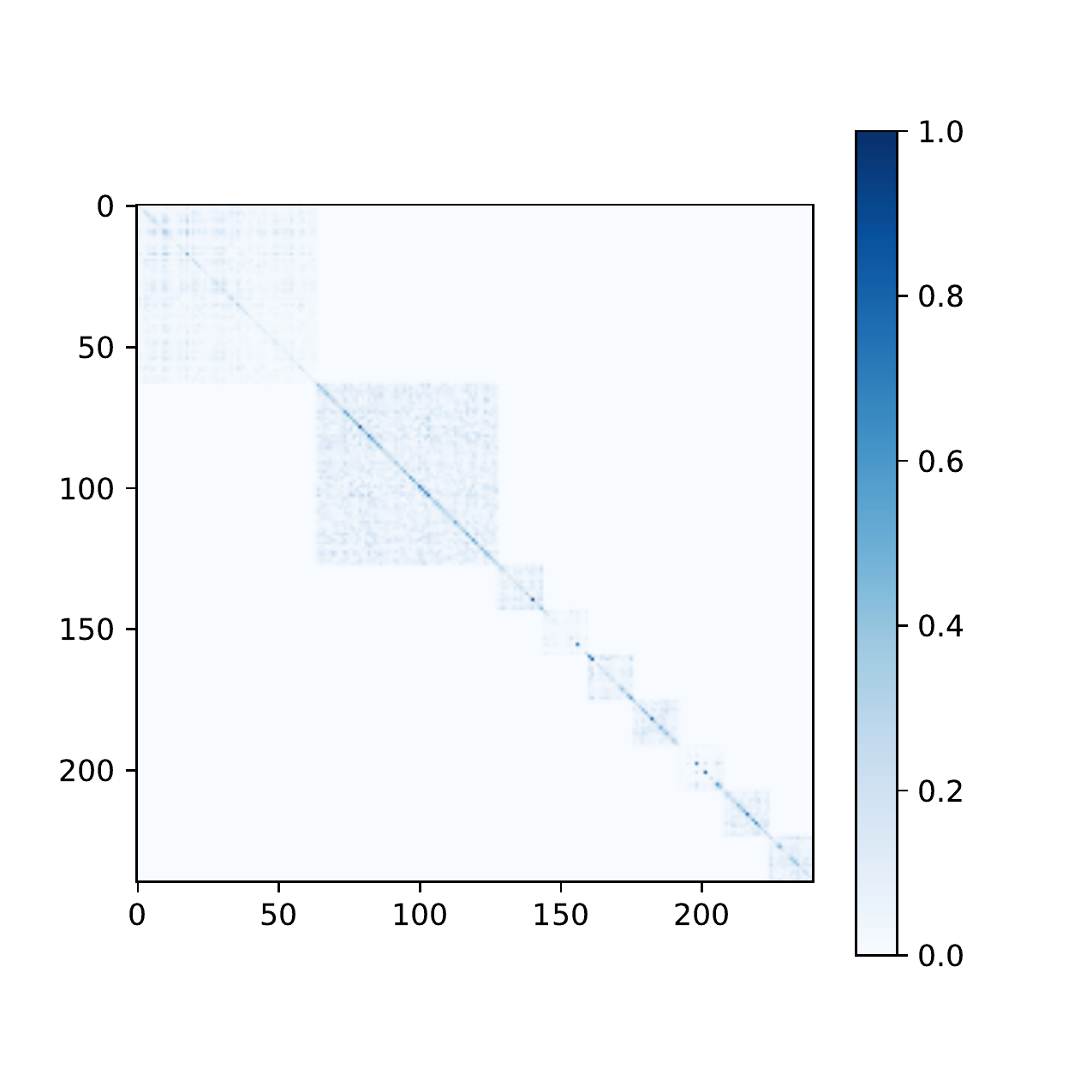}
    \caption{}
  \end{subfigure}
\end{figure}
\begin{figure}
  \ContinuedFloat
  \centering
  \begin{subfigure}[b]{\textwidth}
    \includegraphics[width=0.7\textwidth]{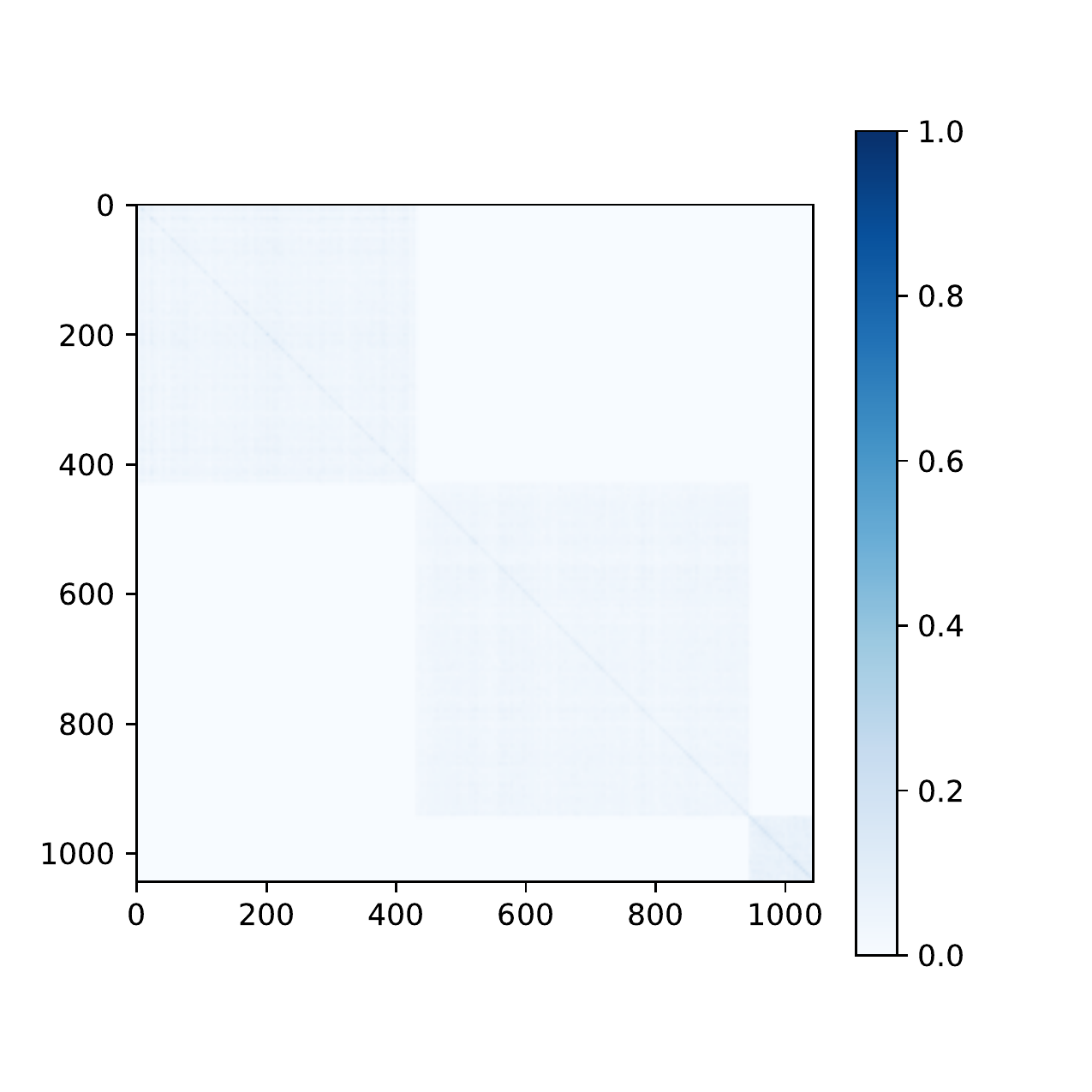}
    \caption{}
  \end{subfigure} \hfill
  \caption[CIFAR100, ResNet-20-FN with second chain]{We show the posterior sample covariance for CIFAR100, ResNet-20-FRN on the second chain.(a) First ten layers with fewer than 100 parameters. (b) 61st through 70th layers with fewer than 100 parameters. (c) Layers with 100-1000 parameters.}
  \label{fig:cifar100_hmc_1}
\end{figure}
\begin{figure}
  \centering
  \begin{subfigure}[b]{\textwidth}
    \includegraphics[width=0.7\textwidth]{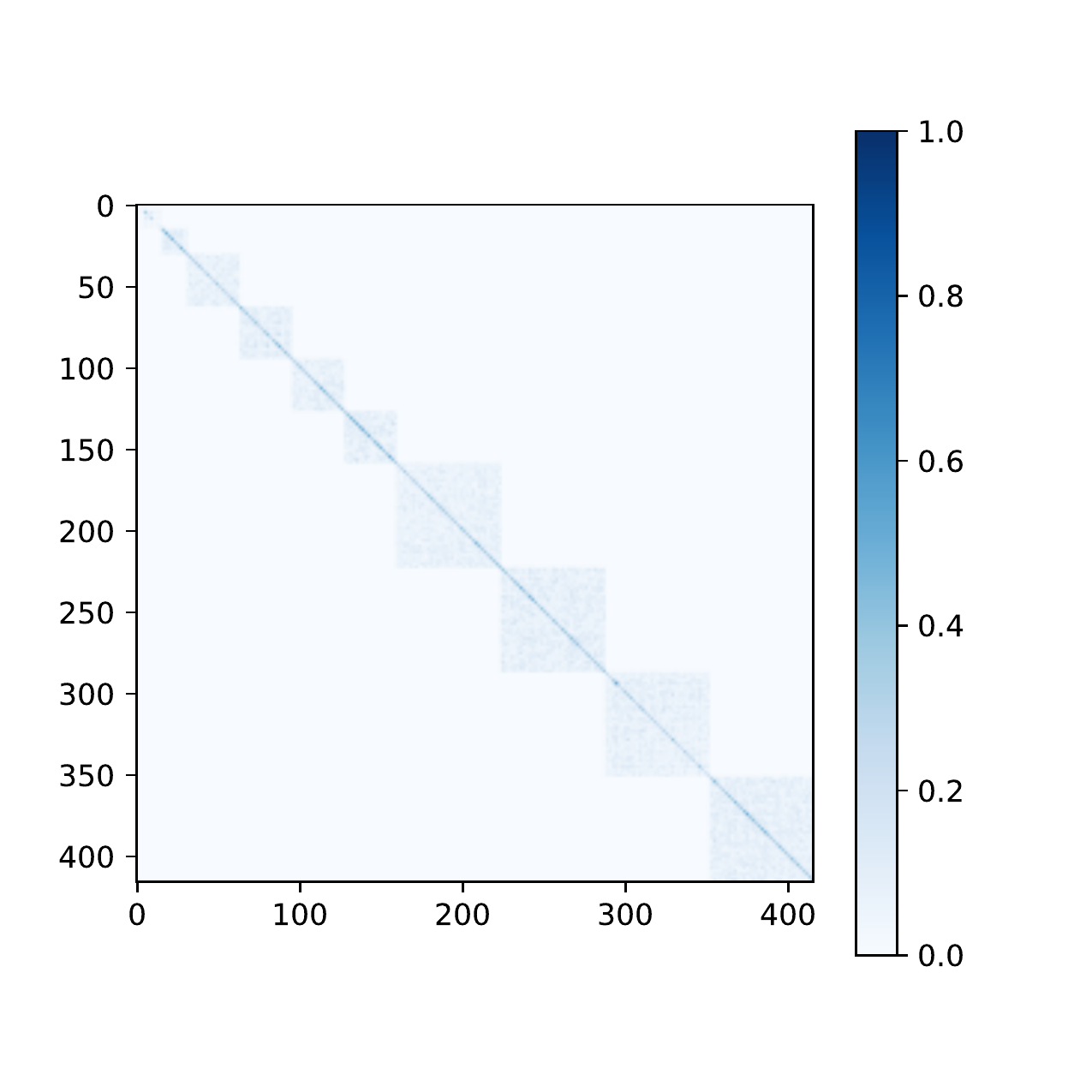}
    \caption{}
  \end{subfigure} \hfill
  \begin{subfigure}[b]{\textwidth}
    \includegraphics[width=0.7\textwidth]{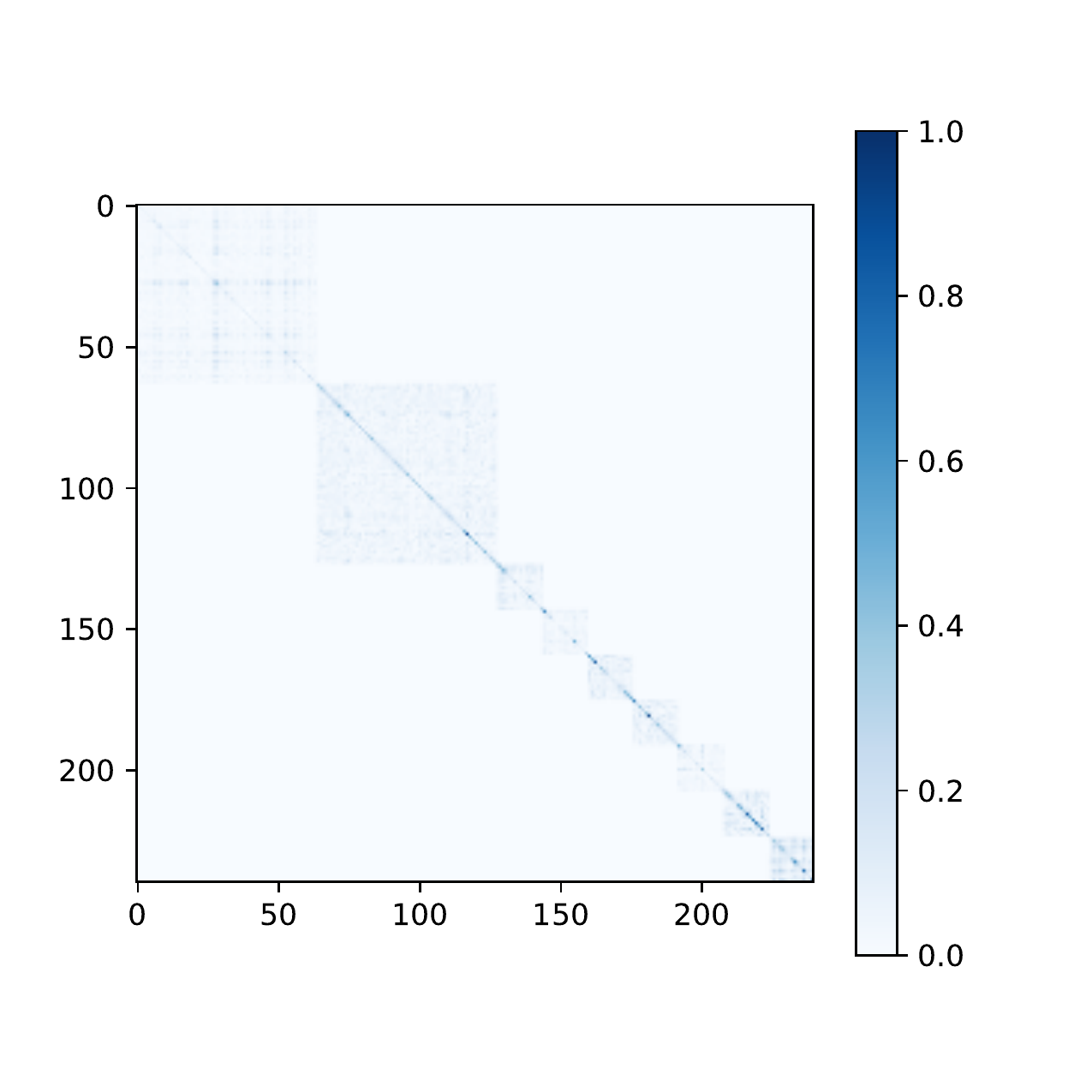}
    \caption{}
  \end{subfigure}
\end{figure}
\begin{figure}
  \ContinuedFloat
  \centering
  \begin{subfigure}[b]{\textwidth}
    \includegraphics[width=0.7\textwidth]{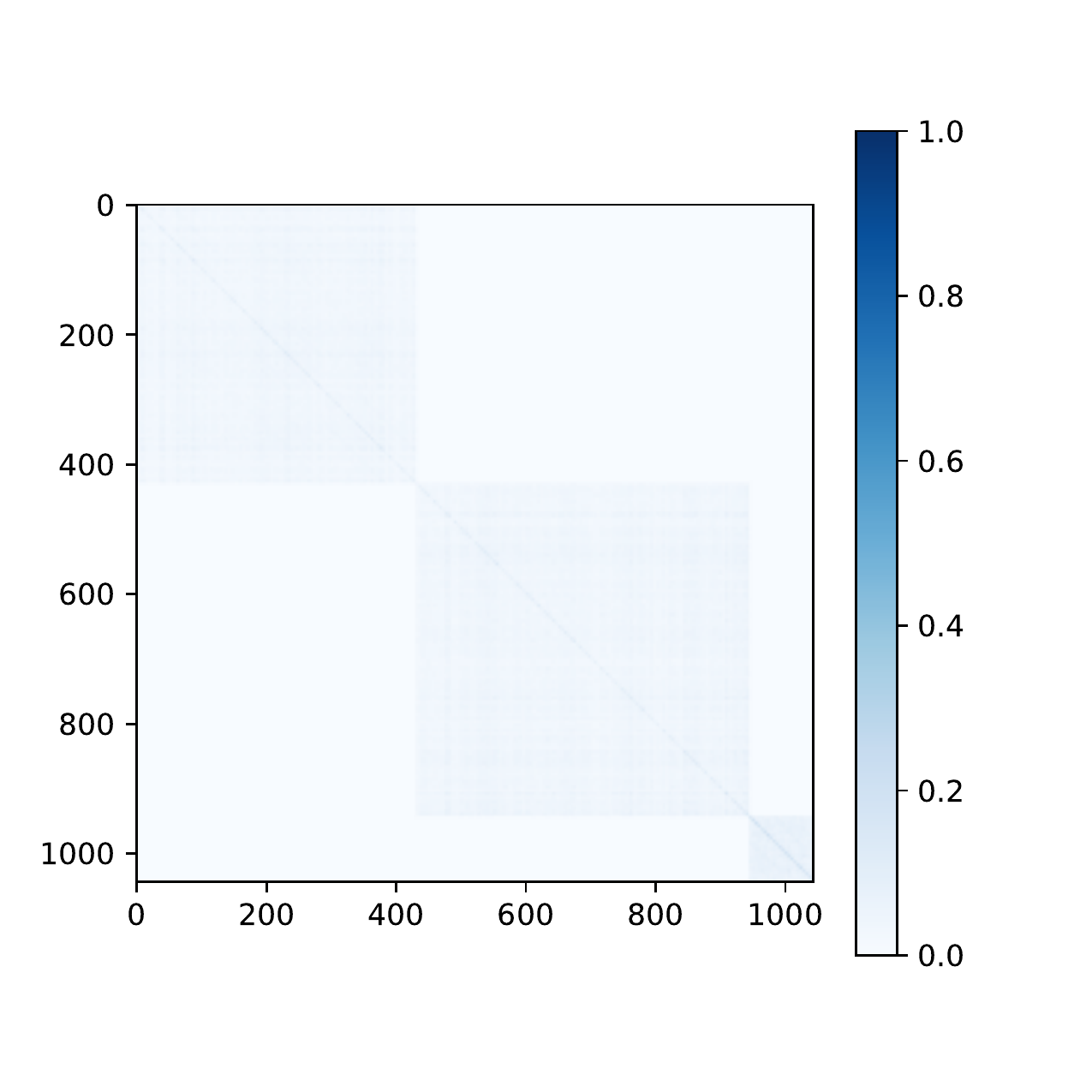}
    \caption{}
  \end{subfigure} \hfill
  \caption[CIFAR100, ResNet-20-FRN with third chain]{We show the posterior sample covariance for CIFAR100, ResNet-20-FRN on the third chain. (a) First ten layers with fewer than 100 parameters. (b) 61st through 70th layers with fewer than 100 parameters. (c) Layers with 100-1000 parameters.}
  \label{fig:cifar100_hmc_3}
\end{figure}
\begin{figure}
  \centering
  \begin{subfigure}[b]{\textwidth}
    \includegraphics[width=0.7\textwidth]{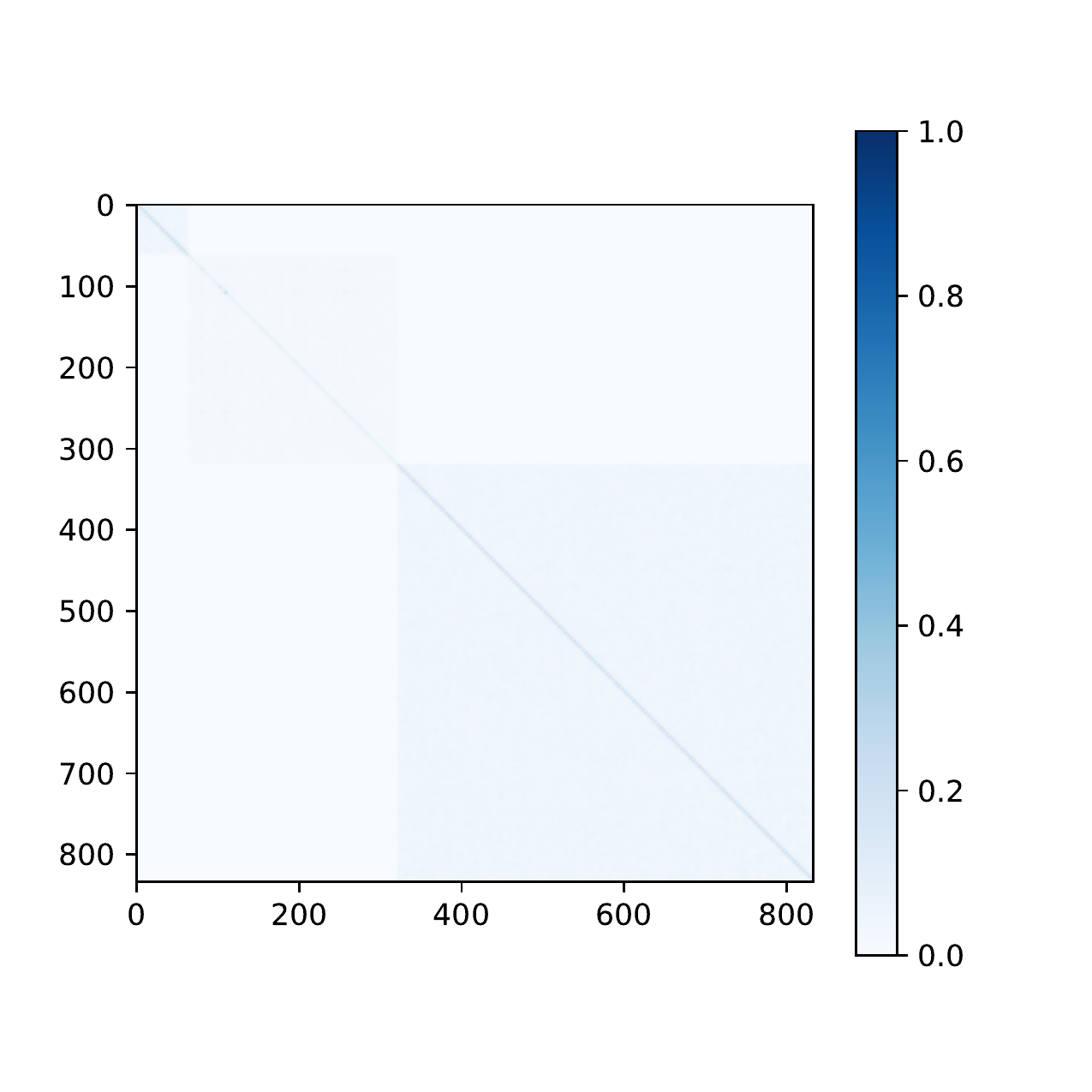}
    \caption{Second chain.}
  \end{subfigure} \hfill
  \begin{subfigure}[b]{\textwidth}
    \includegraphics[width=0.7\textwidth]{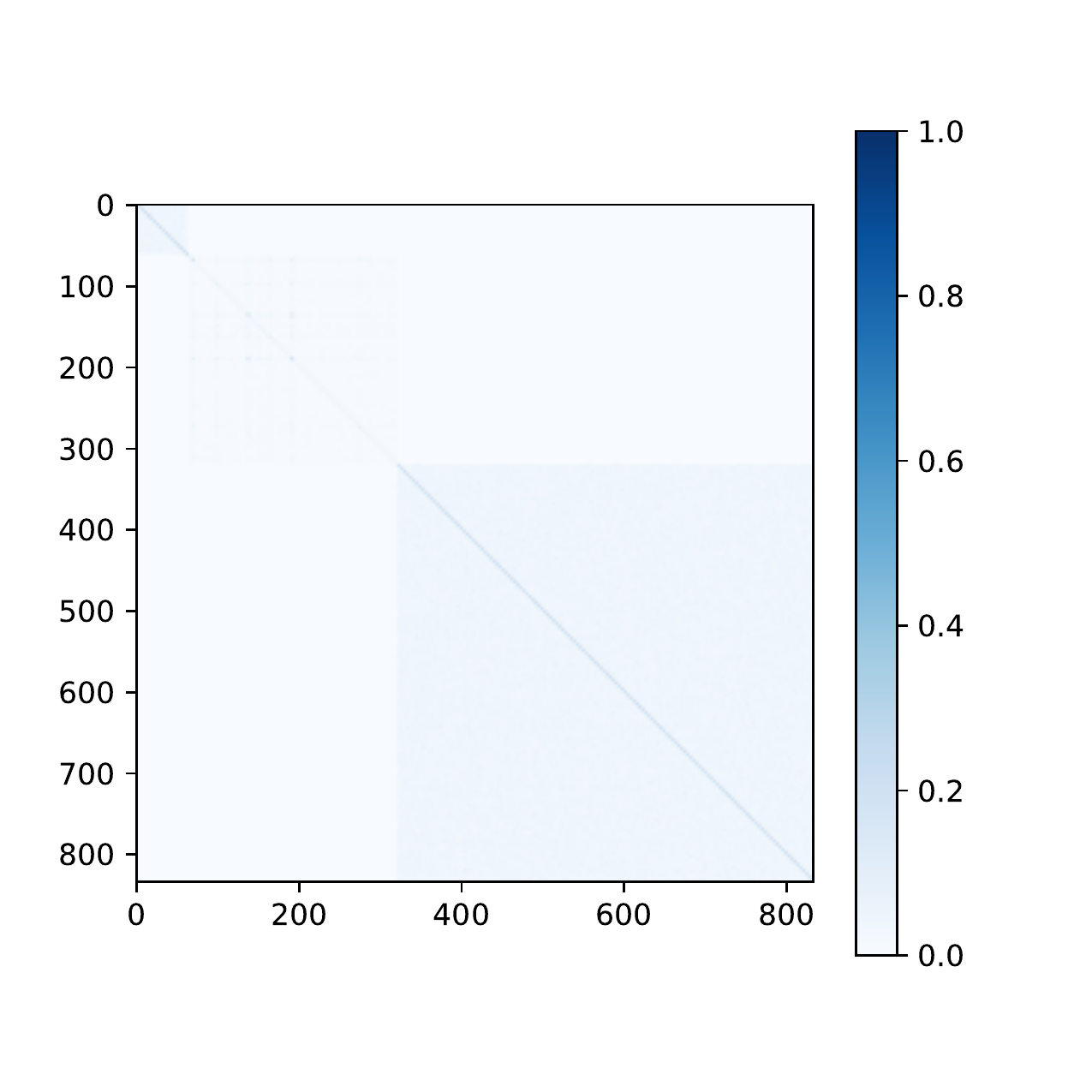}
    \caption{Third chain.}
  \end{subfigure}
  \caption[IMDB LSTM posterior sample covariance chains two and three.]{We show the HMC posterior sample covariance for IMDB LSTM on the second and third chains on layers with fewer than 2000 parameters.}
\end{figure}
\FloatBarrier
  
  \subsection{Diagonal- and Structured-SWAG at Varying Depths}\label{a:swag}
  We use the implementation of SWAG avaliable publicly at \url{https://github.com/wjmaddox/swa_gaussian}. 
  We adapt their code to vary the depth of the PreResNet architecture for the values ${2, 8, 14, 20, 26, 32, 38}$.
  We use the hyperparameter settings used by \citet{maddoxSimple2019} for PreResNet154 on CIFAR100 on all datasets to train the models.
  We use 10 seeds to generate the error bars, which are plotted with one standard deviation.
  We use the same SWAG run to fit both the full and diagonal approximations, and use 30 samples in the forward pass.

  \section{Proofs}
  \subsection{Full Derivation of the Product Matrix Covariance}\label{a:derivation}
  \ifSubfilesClassLoaded{
    \begin{lemma}{lemma}
        For $L\geq3$, the product matrix $M^{(L)}$ of factorized weight matrices can have non-zero covariance between any and all pairs of elements. That is, there exists a set of mean-field weight matrices $\{W^{(l)} | 1 \leq l < L\}$ such that $M^{(L)} = \prod W^{(l)}$ and the covariance between any possible pair of elements of the product matrix:
        \begin{equation}
          \textup{Cov}(m^{(L)}_{ab}, m^{(L)}_{cd}) \neq 0,
        \end{equation}
        where $m^{(L)}_{ij}$ are elements of the product matrix in the $i$\textsuperscript{th} row and $j$\textsuperscript{th} column, and for any possible indexes $a$, $b$, $c$, and $d$.
  \end{lemma}}
  {\lemgreaterthanzero*}

\begin{proof}
    We begin by explicitly deriving the covariance between elements of the product matrix.

    Consider the product matrix, $M^{(L)}$, which is the matrix product of an arbitrary weight matrix, $W^{(L)}$, with a mean field distribution over it's entries, and the product matrix with one fewer layers, $M^{(L-1)}$.
    Expressed in terms of the elements of each matrix in row-column notation this matrix multiplication can be written:
  \begin{equation}
      m^{(L)}_{ab} = \sum_{i=1}^{K_{L-1}} w^{(L)}_{ai}m^{(L-1)}_{ib}.\label{eq:product_matrix_definition}
  \end{equation}
  We make no assumption about $K_{L-1}$ except that it is non-zero and hence the weights can be any rectangular matrix.\footnote{We set aside bias parameters, as they complicate the algebra, but adding them only strengthens the result because each bias term affects an entire row.}
  The weight matrix $W^{(L)}$ is assumed to have a mean-field distribution (the covariance matrix is zero for all off diagonal elements) with arbitrary means:
  \begin{align}\label{eq:mean_field}
       \mathrm{Cov}\big(w^{(L)}_{ac}, w^{(L)}_{bd}\big) &= \Sigma^{(L)}_{abcd} = \delta_{ac}\delta_{bd}\sigma^{(L)}_{ab}; \nonumber\\
       \E{}{w^{(L)}}_{ab} &= \mu_{ab}^{(L)}.
  \end{align}
  $\delta$ are the Kronecker delta. Note that the weight matrix is 2-dimensional, but the covariance matrix is defined between every element of the weight matrix.
  While it can be helpful to regard it as 2-dimensional also, we index it with the four indices that define a pair of elements of the weight matrix.
  
  We begin by deriving the expression for the covariance of the $L$-layer product matrix $\mathrm{Cov}(m^{(L)}_{ab}, m^{(L)}_{cd})$.
  Using the definition of the product matrix in \cref{eq:product_matrix_definition}:
  \begin{align}
      \mathrm{Cov}\big(m^{(L)}_{ab}, m^{(L)}_{cd}\big) &= \mathrm{Cov}\big(\sum_{i} w^{(L)}_{ai}m^{(L-1)}_{ib}, \sum_{j} w^{(L)}_{cj}m^{(L-1)}_{jd}\big).
  \end{align}
  We then simplify this using the linearity of covariance (for brevity call the covariance of the product matrix $\hat{\Sigma}^{(L)}_{abcd}$):
  \begin{align}
      \hat{\Sigma}^{(L)}_{abcd} &= \sum_{ij}\mathrm{Cov}\big(w^{(L)}_{ai}m^{(L-1)}_{ib}, w^{(L)}_{cj}m^{(L-1)}_{jd}\big),
      \intertext{rewriting using the definition of covariance in terms of a difference of expectations:}
       &= \sum_{ij} \E{}{\big[w^{(L)}_{ai}m^{(L-1)}_{ib}w^{(L)}_{cj}m^{(L-1)}_{jd}\big]}\nonumber\\
       &\qquad - \E{}{\big[w^{(L)}_{ai}m^{(L-1)}_{ib}\big]}\E{}{\big[w^{(L)}_{cj}m^{(L-1)}_{jd}\big]},
    \intertext{using the fact that by assumption the new layer is independent of the previous product matrix:}
       &= \sum_{ij} \E{}{\big[w^{(L)}_{ai}w^{(L)}_{cj}\big]}\E{}{\big[m^{(L-1)}_{ib}m^{(L-1)}_{jd}\big]}\nonumber\\
       &\qquad - \E{}{\big[w^{(L)}_{ai}\big]}\E{}{\big[w^{(L)}_{cj}\big]}\E{}{\big[m^{(L-1)}_{ib}\big]}\E{}{\big[m^{(L-1)}_{jd}\big]},
    \intertext{and rewriting to expose the dependence on the covariance of $M^{(L-1)}$:}
      &= \sum_{ij} \Big(\E{}{\big[w^{(L)}_{ai}w^{(L)}_{cj}\big]} - \E{}{\big[w^{(L)}_{ai}\big]}\E{}{\big[w^{(L)}_{cj}\big]}\Big) \nonumber\\ 
      & \qquad\qquad\cdot \Big(\E{}{\big[m^{(L-1)}_{ib}m^{(L-1)}_{jd}\big]}- \E{}{\big[m^{(L-1)}_{ib}\big]}\E{}{\big[m^{(L-1)}_{jd}\big]}\Big)\nonumber\\
      & \qquad +\E{}{\big[w^{(L)}_{ai}\big]}\E{}{\big[w^{(L)}_{cj}\big]}\Big(\E{}{\big[m^{(L-1)}_{ib}m^{(L-1)}_{jd}\big]} -\E{}{\big[m^{(L-1)}_{ib}\big]}\E{}{\big[m^{(L-1)}_{jd}\big]}\Big) \nonumber \\
      & \qquad +\E{}{\big[m^{(L-1)}_{ib}\big]}\E{}{\big[m^{(L-1)}_{jd}\big]} \Big(\E{}{\big[w^{(L)}_{ai}w^{(L)}_{cj}\big]} - \E{}{\big[w^{(L)}_{ai}\big]}\E{}{\big[w^{(L)}_{cj}\big]}\Big),
      \intertext{substituting the covariance:}
      &=\sum_{ij} \mathrm{Cov}\Big(w^{(L)}_{ai}, w^{(L)}_{cj}\Big)\cdot \mathrm{Cov}\Big(m^{(L-1)}_{ib}, m^{(L-1)}_{jd}\Big) \nonumber \\
      & \qquad+\E{}{\big[w^{(L)}_{ai}\big]}\E{}{\big[w^{(L)}_{cj}\big]}\mathrm{Cov}\Big(m^{(L-1)}_{ib}, m^{(L-1)}_{jd}\Big) \nonumber \\
      & \qquad+ \E{}{\big[m^{(L-1)}_{ib}\big]}\E{}{\big[m^{(L-1)}_{jd}\big]} \mathrm{Cov}\Big(w^{(L)}_{ai},w^{(L)}_{cj}\Big). \label{eq:recursive}
  \end{align}
  This gives us a recursive expression for the covariance of the product matrix.
  
  It is straightforward to substitute in our expressions for mean and variance in a mean-field network provided in \cref{eq:mean_field}, where we use the fact that the initial $M^{(1)}$ product matrix is just a single mean-field layer.
  
  In this way, we show that:
  \begin{align}
      \hat{\Sigma}^{(2)}_{abcd} &= \sum_{ij} \Big(\delta_{ac}\delta_{ij}\sigma^{(2)}_{ai}\Big)\cdot \Big(\delta_{ij}\delta_{bd}\sigma^{(1)}_{ib}\Big) \nonumber \\
      & \phantom{= \sum_{ij} }+\mu^{(2)}_{ai}\mu^{(2)}_{cj}\Big(\delta_{ij}\delta_{bd}\sigma^{(1)}_{ib}\Big)+ \E{}{\big[m^{(1)}_{ib}\big]}\E{}{\big[m^{(1)}_{jd}\big]} \Big(\delta_{ac}\delta_{ij}\sigma^{(2)}_{ai}\Big)\\
      & = \sum_{i} \delta_{ac}\delta_{bd}\sigma^{(2)}_{ai}\sigma^{(1)}_{ib}+\delta_{bd}\mu^{(2)}_{ai}\mu^{(2)}_{ci}\sigma^{(1)}_{ib} + \delta_{ac}\mu^{(1)}_{ib}\mu^{(1)}_{id}\sigma^{(2)}_{ai}. \label{eq:two_layer}
  \end{align}
  The first term of \cref{eq:two_layer} has the Kronecker deltas $\delta_{ac}\delta_{bd}$ meaning that it contains diagonal entries in the covariance matrix.
  The second term has only $\delta_{bd}$ meaning it contains entries for the covariance between weights that share a column.
  The third term has only $\delta_{ac}$ meaning it contains entries for the covariance between weights that share a row.
  
  This covariance of the product matrix already has some off-diagonal terms, but it does not yet contain non-zero covariance for weights that share neither a row nor a column.
  
  But we can repeat the process and find $\hat{\Sigma}^{(3)}_{abcd}$ using \cref{eq:recursive} and our expression for $\hat{\Sigma}^{(2)}_{ibjd}$:
  \begin{align}
      \hat{\Sigma}^{(3)}_{abcd} &= \sum_{ij} \Big(\delta_{ac}\delta_{ij}\sigma^{(3)}_{ai}\Big)\cdot \hat{\Sigma}^{(2)}_{ibjd}+\mu^{(3)}_{ai}\mu^{(3)}_{cj}\hat{\Sigma}^{(2)}_{ibjd}\nonumber\\
      &\qquad+ \E{}{\big[m^{(2)}_{ib}\big]}\E{}{\big[m^{(2)}_{jd}\big]} \Big(\delta_{ac}\delta_{ij}\sigma^{(3)}_{ai}\Big)\\
      &= \sum_{ij} \Big(\delta_{ac}\delta_{ij}\sigma^{(3)}_{ai}\Big)\cdot \sum_k \Big(\delta_{ij}\delta_{bd}\sigma^{(2)}_{ik}\sigma^{(1)}_{kb} + \delta_{bd}\mu^{(2)}_{ik}\mu^{(2)}_{jk}\sigma^{(1)}_{kb} \nonumber + \delta_{ij}\mu^{(1)}_{kb}\mu^{(1)}_{kd}\sigma^{(2)}_{ik}\Big) \nonumber \\
      & \qquad+\mu^{(3)}_{ai}\mu^{(3)}_{cj}\cdot \sum_k \Big(\delta_{ij}\delta_{bd}\sigma^{(2)}_{ik}\sigma^{(1)}_{kb} + \delta_{bd}\mu^{(2)}_{ik}\mu^{(2)}_{jk}\sigma^{(1)}_{kb} + \red{\delta_{ij}\mu^{(1)}_{kb}\mu^{(1)}_{kd}\sigma^{(2)}_{ik}}\Big) \nonumber \\
      & \qquad+ \mu^{(2)}_{ib}\mu^{(2)}_{jd} \Big(\delta_{ac}\delta_{ij}\sigma^{(3)}_{ai}\Big).
  \end{align}
  It is the term in red which has no factors of Kronecker deltas in any of the indices a, b, c, or d.
  It is therefore present in all elements of the covariance matrix of the product matrix, regardless of whether they share one or both index.
  This shows that, so long as the distributional parameters themselves are non-zero, the product matrix can have a fully non-zero covariance matrix for $L = 3$.

  We note that there are many weight matrices for which the resulting covariance is non-zero everywhere---we think this is actually typical.
  Indeed, empirically, we found that for any network we cared to construct, we were unable to find covariances that \emph{were} zero anywhere.
  However, for our existance proof, we simply note that for any matrix in which all the means are positive each term of the resulting expression is positive (the standard deviation parameters may be taken as positive without loss of generality).
  In that case, it is impossible that any term cancels with any other, so the resulting covariance is positive everywhere.

  Last, we examine the recurrence relationship in \cref{eq:recursive}.
  Once it is the case that $\mathrm{Cov}(m_{ib}^{(L-1)}, m_{jd}^{(L-1)}) \neq 0$ for all possible indices, the covariances between elements of $M^{(L)}$ may also be non-zero.
  Observe simply that if the means of the top weight matrix are positive, then each of the terms in \cref{eq:recursive} are positive, so it is impossible for any term to cancel out with any other.
  The fact that the elements of $M^{(L-1)}$ have non-zero covariances everywhere therefore entails that there is a weight matrix $W^{(L)}$ such that $M^{(L)}$ has non-zero covariance between all of its elements also, as required.

  \begin{remark}
    Here, we show only an existance proof, and therefore we restrict ourselves to positive means and standard deviations to simplify the proof. In fact, we believe that non-zero covariance is the norm, rather than a special case, and found this in all our numerical simulations for both trained and randomly sampled models.
    However, we do not believe that (in the linear case) \emph{any} covariance matrix can be created from a deep mean-field product.
  \end{remark}
\end{proof}

  \subsection{Matrix Variate Gaussian as a Special Case of Three-Layer Product Matrix}\label{a:mvg}

  We can gain insight into the richness of the possible covariances by considering the limited case of the product matrix $M^{(3)} = ABC$ where $B$ is a matrix whose elements are independent Gaussian random variables and $A$ and $C$ are deterministic.
  We note that this is a highly constrained setting, and that the covariances which can be induced with $A$ and $C$ as random variables have the more complex form shown in \cref{a:derivation}.
  We can show the following:

  \ifSubfilesClassLoaded{
    \begin{proposition}{proposition}
    The Matrix Variate Gaussian (Kronecker-factored) distribution is a special case of the distribution over elements of the product matrix.
    In particular, for $M^{(3)} = ABC$, $M^{(3)}$ is distributed as an MVG random variable when $A$ and $C$ are deterministic and $B$ has its elements distributed as fully factorized Gaussians with unit variance.
    \end{proposition}
  }
{\thmmatrixnormal*}
\newcommand{\flatten}[1]{\mathrm{vec}(#1)} 

\begin{proof}
  Consider the product matrix $M^{(3)} = ABC$. where $B$ is a matrix whose elements are independent Gaussian random variables and $A$ and $C$ are deterministic.
  The elements of $B$ are distributed with mean $\mu_B$ and have a diagonal covariance matrix $\Sigma_B$.

	We begin by recalling the property of the Kronecker product that:
	\begin{align}
		\flatten{ABC} = (C^\top \otimes A) \flatten{B}. \label{eq:kvec}
	\end{align}
  By definition $\flatten{M^{(3)}} = \flatten{ABC} = (C^\top \otimes A) \flatten{B}$.
  Because $C^\top \otimes A$ is deterministic, it follows from a basic property of the covariance that the covariance of the product matrix $\Sigma_{M^{(3)}}$ is given by:
	\begin{align}
    \Sigma_{M^{(3)}} = (C^\top \otimes A) \Sigma_B (C^\top \otimes A)^\top.
	\end{align}
	Using the fact that the transpose is distributive over the Kronecker product, this is equivalent to:
	\begin{align}
    \Sigma_{M^{(3)}} = (C^\top \otimes A) \Sigma_B (C \otimes A^\top).
	\end{align}
	
  Because we only want to establish that the family of distributions expressible contains the matrix variate Gaussians, we do not need to use all the possible freedom, and we can set $\Sigma_B = I$.
  In this special case:
	\begin{align}
    \Sigma_{M^{(3)}} = (C^\top \otimes A) (C \otimes A^\top).
	\end{align}
	Using the mixed-product property, this is equivalent to:
	\begin{align}
    \Sigma_{M^{(3)}} = (C^\top C) \otimes (A A^\top).
	\end{align}	
	Now, we note that any positive semi-definite matrix can be written in the form $A = M^\top M$, so this implies that, defining the positive semi-definite matrices $V = C^\top C$ and $U = A A^\top$, we have that the covariance $\Sigma_{M^{(3)}}$ is of the form,
	\begin{align}
    \Sigma_{M^{(3)}} = V \otimes U. \label{eq:covariance}
  \end{align}
  
  Similarly, we can consider the mean of the product matrix $\mu_{M^{(3)}}$.
  From \cref{eq:kvec}, we can see that:
  \begin{align}
    \mu_{M^{(3)}} = (C^\top \otimes A) \flatten{\mu_B}. \label{eq:mean}
  \end{align}

  But since we have not yet constrained $\mu_B$, it is clear that this allows us to set any $\mu_{M^{(3)}}$ we desire by choosing $\mu_B = (C^\top \otimes A)^{-1} \mu_{M^{(3)}}$.

  So far, we have only discussed the first- and second-moments, and the proof has made no assumptions about specific distributions.
  However, we now observe that a random variable $X$ is distributed according to the Matrix Variate Gaussian distribution according to some mean $\mu_X$ and with scale matrices $U$ and $V$ if and only if $\flatten{X}$ is a multivariate Gaussian with mean $\vec{\mu_X}$ and covariance $U \otimes V$.
  
  Therefore, given \cref{eq:mean} and \cref{eq:covariance}, the special case of $M^{(3)}$ where the first and last matrices are deterministic and the middle layer has a fully-factorized Gaussian distribution over the weights with unit variance is a Matrix Variate Gaussian distribution where:
  \begin{align}
    &\flatten{\mu_X} = (C^\top \otimes A) \flatten{\mu_B};\\
    & V = C^\top C;\\
    & U = A^\top A.
  \end{align}
  \end{proof}

  \subsection{Proof of Linearized Product Matrix Covariance}\label{a:proof_linearized_product_matrix}
  \subsubsection{Proof of Local Linearity}\label{a:local_linearity}
  We consider local linearity in the case of piecewise-linear activations like ReLU.
  \ifSubfilesClassLoaded{
    \begin{lemma}{lemma}
        Consider an input point $\mathbf{x}^* \in \mathcal{D}$. Consider a realization of the model weights $\vtheta$. Then, for any $\mathbf{x}^*$, the neural network function $f_{\vtheta}$ is linear over some compact set $\mathcal{A}_{\vtheta} \subset \mathcal{D}$ containing $\mathbf{x}^*$. Moreover, $\mathcal{A}_{\vtheta}$ has non-zero measure for almost all $\mathbf{x}^*$ w.r.t. the Lebesgue measure.
    \end{lemma}
  }{
  \linear*}
  \begin{proof}
  Neural networks with finitely many piecewise-linear activations are themselves piecewise-linear.
  Therefore, for a finite neural network, we can decompose the input domain $\mathcal{D}$ into regions $\mathcal{D}_i \subseteq \mathcal{D}$ such that
  \begin{enumerate}
      \item $\cup \mathcal{D}_i = \mathcal{D}$,
      \item $\mathcal{D}_i \cap \mathcal{D}_j = \varnothing \quad \forall i\neq j$,
      \item $f_{\thet}$ is a linear function on points in $\mathcal{D}_i$ for each i.
  \end{enumerate}
  
  For a finite neural network, there are at most finitely many regions $\mathcal{D}_i$.
  In particular, with hidden layer widths $n_i$ in the $i$'th layer, with an input domain $\mathcal{D}$ with dimension $n_0$, \citet{montufarNumber2014} show that the network can define maximally a number of regions in input space bounded above by:
  \begin{equation}
      \left( \prod_{i=1}^{L-1} \left\lfloor \frac{n_i}{n_0}
            \right\rfloor^{n_0} \right)
            \sum_{j=0}^{n_0} { n_L \choose j }. 
  \end{equation}
  Except in the trivial case where the input domain has measure zero, this along with (1) and (2) jointly entail that at least one of the regions $\mathcal{D}_i$ has non-zero measure.
  This, with (3) entails that only a set of input points of zero measure do \emph{not} fall in a linear region of non-zero measure.
  These points correspond to inputs that lie directly on the inflection points of the ReLU activations.
  \end{proof}

  We visualize $\mathcal{A}_{\thet_i}$ in \cref{fig:linear_regions_1_sample}.
  This shows a two-dimensional input space (from the two moons dataset).
  Parts of the space within which a neural network function is linear are shown in one color.
  The regions are typically smallest where the most detail is required in the trained function.
  
  \begin{figure}
    \centering
    \begin{subfigure}{0.48\textwidth}
      \includegraphics[width=\textwidth]{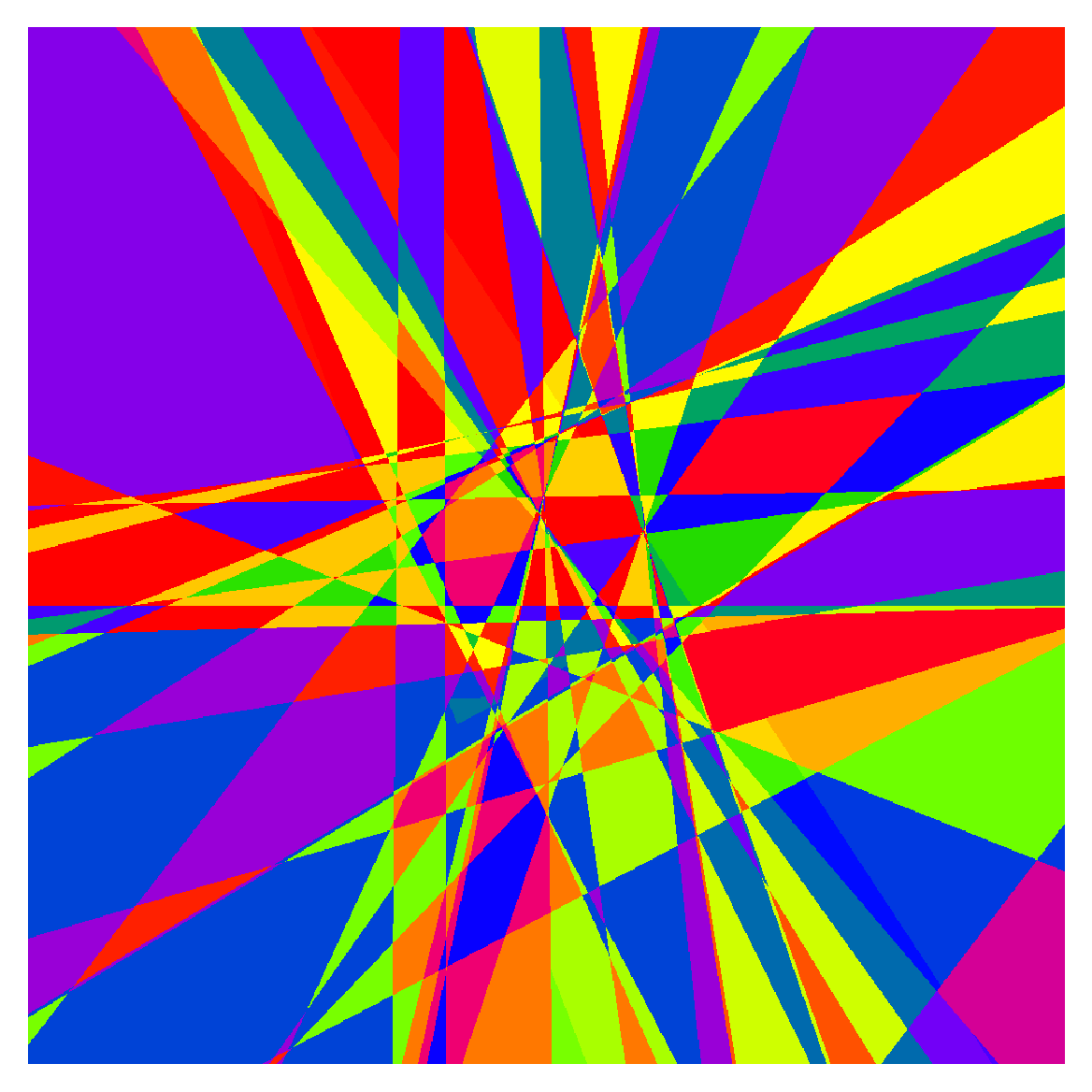}
      \caption{1 sample: $\mathcal{A}_{\thet_i}$}
      \label{fig:linear_regions_1_sample}
    \end{subfigure}
    \begin{subfigure}{0.48\textwidth}
      \includegraphics[width=\textwidth]{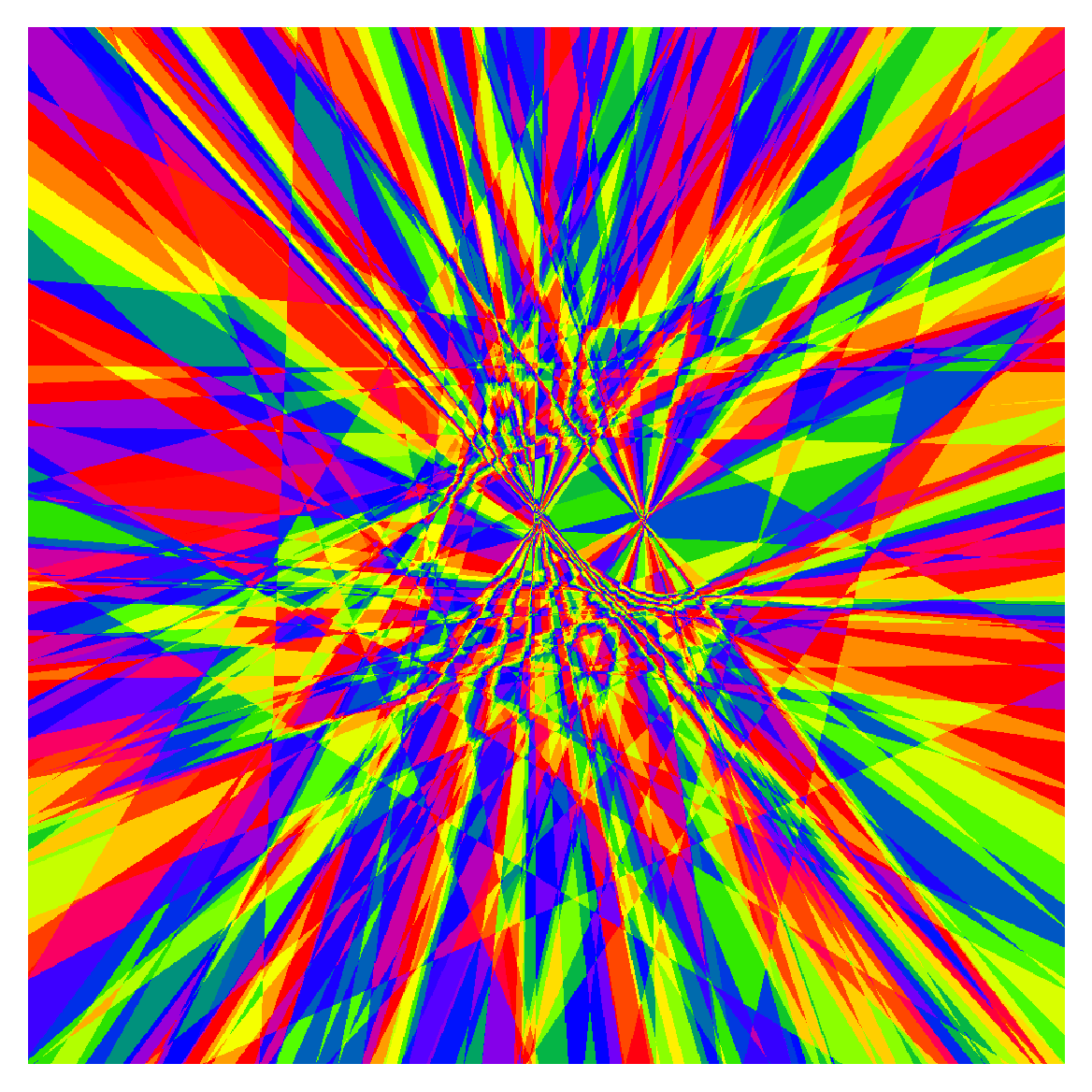}
      \caption{5 samples: $\mathcal{A} = \bigcap_{0 \leq i < 5}\mathcal{A}_{\thet_i}$}
      \label{fig:linear_regions_5_sample}
    \end{subfigure}
    \caption[Visualizing local product matrices]{Visualizaton of the linear regions in input-space for a two-dimensional binary classification problem (two moons). Colored regions show contiguous areas within which a neural network function is linear. We use an abitrary numerical encoding of these regions (we interpret the sign pattern of activated relus as an integer in base 2) and a cylic colour scheme for visualisation, so the color of each region is arbitrary, and two non-contiguous regions with the same color are not the same region. The neural network has one hidden layer with 100 units and is trained for 1000 epochs on 500 datapoints from scipy's two moons using Adam. (a) a single model has fairly large linear regions, with the most detail clustered near the region of interest. (b) The regions within which all samples are linear (the intersection set $\mathcal{A}$) are smaller, but finite. The local product matrix is valid within one of these regions for any input point.}
    \label{fig:linear_regions}
    
  \end{figure}

  \subsubsection{Defining the Local Product Matrix}\label{a:lpm_def}
  We define a random variate representing the local product matrix, for an input point $\mathbf{x}^*$, using the following procedure.

  To draw a finite $N$ samples of the random variate, we sample $N$ realizations of the weight parameters $\Theta = \{\thet_i \text{ for } 1\leq i \leq N\}$.
  For each $\thet_i$, given $\mathbf{x}^*$ there is a compact set $\mathcal{A}_{\thet_i} \subset \mathcal{D}$ within which $f_{\thet_i}$ is linear (and $\mathbf{x}^* \in \mathcal{A}_i$) by \cref{lemma:linearity}.
  Therefore, all samples of the neural network function are linear in the intersection region $\mathcal{A} = \bigcap_i \mathcal{A}_{\thet_i}$.
  We note that $\mathcal{A}$ at least contains $\mathbf{x}^*$.
  Moreover, so long as $\mathcal{D}$ is a compact subset of the reals, $\mathcal{A}$ has non-zero measure.\footnote{Intuitively, we know that $\mathbf{x}^*$ is in all $\mathcal{A}_{\thet_i}$, so when we add a new sample we know that there is either overlap around $\mathbf{x}^*$ or the point $\mathbf{x}^*$ is on the boundary of the new subset, which means we could equally well pick a different set that has $\mathbf{x}^*$ on its boundary and \textit{does} have non-zero-measure overlap with the previous sets.
  
  More formally, consider some compact set $\mathcal{A}_{\thet_0} \subset \mathcal{D}$ with non-zero measure such that $\mathbf{x}^* \in \mathcal{A}_{\thet_0}$.
  Take some new compact set $\mathcal{A}_{\thet_1} \subset \mathcal{D}$ with non-zero measure also such that $\mathbf{x}^* \in \mathcal{A}_{\thet_1}$.
  Define the intersection between those sets $\mathcal{B} = \mathcal{A}_{\thet_0} \cap \mathcal{A}_{\thet_1}$.
  Suppose that $\mathcal{B}$ has zero measure.
  But both $\mathcal{A}_{\thet_0}$ and $\mathcal{A}_{\thet_1}$ contain $\mathbf{x}^*$, so the only way that $\mathcal{B}$ could have zero measure is if $\mathbf{x}^*$ is an element in the boundary of both sets.
  But if $\mathcal{A}_{\thet_1}$ has $\mathbf{x}^*$ on its boundary, then, by the continuity of the real space, there is at least one other compact set $\mathcal{A}_{\thet_1}'$, different to $\mathcal{A}_{\thet_1}$, such that $\mathbf{x}^*$ is on its boundary.
  But, since by hypothesis $\mathcal{A}_{\thet_0}$ has non-zero measure, there exists such a set $\mathcal{A}_{\thet_1}'$ which has a non-zero-measure intersection with $\mathcal{A}_{\thet_0}$.
  We can therefore select $\mathcal{A}_{\thet_1}'$ instead of $\mathcal{A}_{\thet_1}$ when building $\mathcal{A}$, such that the intersection with $\mathcal{A}_{\thet_0}$ has non-zero measure.
  By repeated application of this argument, we can guarantee that for any finite $\Theta$ we are able to find a set of $\mathcal{A}_{\thet_i} \subset \mathcal{D}$ such that $\forall i:\mathbf{x}^* \in \mathcal{A}_{\thet_i}$ and $\mathcal{A}$ has non-zero measure.
  This argument does not guarantee that the measure of $\mathcal{A}$ in the limit as $N$ tends to infinity is non-zero.
  }
  \Cref{fig:linear_regions_5_sample} shows a visualization of $\mathcal{A}$ with 5 samples.
  The linear regions are smaller, because there is a discontinuity if any of the models is discontinuous.
  Nevertheless, the space is composed of regions of finite size within which the neural network function is linear.

  For each $\thet_i$ we can compute a local product matrix within $\mathcal{A}$.
  Ordinarily, setting aside the bias term for simplicity, a neural network hidden layer $\mathbf{h}_{l+1}$ can be written in terms of the hidden layer before it, a weight matrix $W_l$, and an activation function. 
  \begin{align}
    \mathbf{h}_{l+1} = \sigma(W_l\mathbf{h}_{l})
  \end{align}
  We observe that within $\mathcal{A}$ the activation function becomes linear.
  This allows us to define an activation vector $\mathbf{a}_{\mathbf{x}^*}$ within $\mathcal{A}$ such that the equation can be written:
  \begin{align}
    \mathbf{h}_{l+1} = \mathbf{a}_{\mathbf{x}^*} \cdot (W_l\mathbf{h}_{l}).
  \end{align}
  The activation vector can be easily calculated by calculating $W_l\mathbf{h}_l$, seeing which side of the (Leaky) ReLU the activation is on within that linear region for each hidden unit, and selecting the correct scalar (0 or 1 for a ReLU, or $\alpha$ or 1 for a Leaky ReLU).

  This allows us to straightforwardly construct a product matrix for each $\thet_i$ which takes the activation function into account (in the linear case, we effectively always set $\mathbf{a}_{\mathbf{x}^*}$ to equal the unit vector).
  The random variate $P_{\mathbf{x}^*}$ is constructed with these product matrices for realizations of the weight distribution.
  
  Samples from the resulting random variate $P_{\mathbf{x}^*}$ are therefore distributed such that samples from $P_{\mathbf{x}^*} \mathbf{x}^*$ have the same distribution as samples of the predictive posterior $y$ given $\mathbf{x}^*$ within $\mathcal{A}$.
  
  \subsubsection{Proof that the Local Product Matrix has Non-zero Off-diagonal Covariance}
  \ifSubfilesClassLoaded{
    \begin{proposition}{proposition}
      \label{thm:main}
  Given a mean-field distribution over the weights of neural network $f$ with piecewise linear activations, $f$ can be written in terms of the local product matrix $P_{\mathbf{x}^*}$ within $\mathcal{A}$. 
  
  For $L \geq 3$, for activation functions which are non-zero everywhere, there exists a set of weight matrices $\{W^{(l)}|1 \leq l < L\}$ such that all elements of the local product matrix have non-zero off-diagonal covariance:
  \begin{equation}
    \textup{Cov}(p^{\mathbf{x}^*}_{ab}, p^{\mathbf{x}^*}_{cd}) \neq 0,
  \end{equation}
  where $p^{\mathbf{x}^*}_{ij}$ is the element at the $i$\textsuperscript{th} row and $j$\textsuperscript{th} column of $P_{\mathbf{x}^*}$.
    \end{proposition}
  }
  {
  \maintheorem*
  }
  
  \begin{proof}
  First, we show that the covariance between arbitrary entries  of each realization of the product matrix of linearized functions can be non-zero.
  Afterwards, we will show that this implies that the covariance between arbitrary entries of the product matrix random variate, $P_{\mathbf{x}^*}$ can be non-zero.
  
  Consider a local product matrix constructed as above.
  Then for each realization of the weight matrices, the product matrix realization $M^{(L)}_i$, defined in the region around $\mathbf{x}^*$ following \cref{lemma:linearity}.
  We can derive the covariance between elements of this product matrix within that region in the same way as in \cref{lemma:covariance}, finding similarly that:
  \begin{align}
    \hat{\Sigma}^{(3)}_{abcd} 
    &= \alpha_{abcd} \sum_{ij} \Big(\delta_{ac}\delta_{ij}\sigma^{(3)}_{ai}\Big)\cdot \sum_k \Big(\delta_{ij}\delta_{bd}\sigma^{(2)}_{ik}\sigma^{(1)}_{kb} \nonumber \\
    &\phantom{\sum_{ij} \Big(\delta_{ac}\delta_{ij}\sigma^{(3)}_{ai}\Big)\cdot \sum_k \Big(} + \delta_{bd}\mu^{(2)}_{ik}\mu^{(2)}_{jk}\sigma^{(1)}_{kb} \nonumber \\
    &\phantom{\sum_{ij} \Big(\delta_{ac}\delta_{ij}\sigma^{(3)}_{ai}\Big)\cdot \sum_k \Big(} + \delta_{ij}\mu^{(1)}_{kb}\mu^{(1)}_{kd}\sigma^{(2)}_{ik}\Big) \nonumber \\
    & \phantom{= \sum_{ij} }+\mu^{(3)}_{ai}\mu^{(3)}_{cj}\cdot \sum_k \Big(\delta_{ij}\delta_{bd}\sigma^{(2)}_{ik}\sigma^{(1)}_{kb} \nonumber \\
    &\phantom{\sum_{ij} \Big(\delta_{ac}\delta_{ij}\sigma^{(3)}_{ai}\Big)\cdot \sum_k \Big(} + \delta_{bd}\mu^{(2)}_{ik}\mu^{(2)}_{jk}\sigma^{(1)}_{kb} \nonumber \\
    &\phantom{\sum_{ij} \Big(\delta_{ac}\delta_{ij}\sigma^{(3)}_{ai}\Big)\cdot \sum_k \Big(} + \delta_{ij}\mu^{(1)}_{kb}\mu^{(1)}_{kd}\sigma^{(2)}_{ik}\Big) \nonumber \\
    & \phantom{= \sum_{ij} }+ \mu^{(2)}_{ib}\mu^{(2)}_{jd} \Big(\delta_{ac}\delta_{ij}\sigma^{(3)}_{ai}\Big),
\end{align}
where $\alpha$ is a constant determined by the piecewise-linearity in the linear region we are considering.
Note that we must assume here that $\alpha_{ij} \neq 0$ except in a region of zero measure, for example a LeakyReLU, otherwise it is possible that the constant introduced by the activation could eliminate non-zero covariances.
We discuss this point further below.
  
  Now note that the covariance of the sum of independent random variables is the sum of their covariances.
  Therefore the covariance of $I$ realizations of $P$ (suppressing the notation $(L)$) is:
  \begin{align}
      \mathrm{Cov}\big(P_{ab}, P_{cd}\big) = \frac{1}{I}\sum_{i=1}^{I} \hat{\Sigma}^i_{abcd}\label{eq:sum_independent}
  \end{align}
  
  As before, consider the case of positive means and standard deviations.
  Just like before, this results in a positive entry in the covariance between any two elements for each realization of the product matrix, and by \cref{eq:sum_independent} the entry for any element of the local product matrix remains positive as well.
  This suffices to prove the theorem for the case of $L=3$.
  Just as \cref{lemma:covariance} extends to all larger $L$, this result does also.
  
  \begin{remark}
    The above proof assumes that $\alpha_{ij} \neq 0$.
    In fact, for common piecewise non-linearities like ReLU, $\alpha$ may indeed be zero.
    This means that the non-linearity can, in principle, `disconnect' regions of the network such that the `effective depth' falls below 3 and there is not a covariance between every element.
  
    We cannot rule this out theoretically, as it depends on the data and learned function.
    In practice, we find it very unlikely that a trained neural network will turn off all its activations for any typical input, nor that enough activations will be zero that the product matrix does not have shared elements after some depth.
    
    However, we do observe that for at least some network structures and datasets it is uncommon in practice that all the activations in several layers are `switched off'.
    We show in \cref{fig:cov_heatmap} an example of a local product matrix covariance which does not suffer from this problem. We find that for a model trained with mean-field VI on the FashionMNIST test dataset the number of activations switched on is on average 48.5\% with standard deviation 4.7\%. There were only four sampled models out of 100 samples on each of 10,000 test points where an entire row of activations was `switched off', reducing the effective depth by one, and this never occurred in more than one row. Indeed, \citep{goldblumTruth2019} describe settings with all activations switched off as a pathological case where SGD fails.
  \end{remark}
  \end{proof}

  \subsection{Existence Proof of a Two-Hidden-Layer Mean-field Approximate Posterior Inducing the True Posterior Predictive}\label{a:parameterization:uat}
  In this section we prove that:

  \ifSubfilesClassLoaded{
    \begin{proposition}{proposition}
  Let $p(y|\rvx, \data)$ be some fixed posterior predictive distribution for univariate $y$.
  Let $\epsilon>0$ and $\delta>0$.
  Then, there exists a sufficiently wide BNN with two hidden layers and for which there exists a mean-field Gaussian approximate posterior distribution $q(\vtheta)$ over its weights such that the probability density function over the outputs $p_{\vtheta}(\hat{y}; \mathbf{x})$ approximates the posterior predictive probability density function in that:
  \begin{equation}
    \forall \mathbf{x}: \textup{Pr}\big(\lvert p_{\vtheta}(\hat{y}|\mathbf{x}) - p(y|\mathbf{x},\mathcal{D})\rvert > \epsilon\big) < \delta,i
  \end{equation}
  provided that the cumulative density function of the posterior predictive is montonically increasing and continuous.
    \end{proposition}
  }{
  \theoremuat*}
  \newcommand{\xin}{\mathbf{x}}
  \newcommand{\yout}{\mathbf{y}}
  \newcommand{\Yout}{\mathbf{Y}}
  \newcommand{\Wa}{\thet_0}
  \newcommand{\Wb}{\thet_1}
  \newcommand{\Wc}{\thet_2}
  \newcommand{\ba}{\mathbf{b}_0}
  \newcommand{\bb}{\mathbf{b}_1}
  \newcommand{\bc}{\mathbf{b}_2}
  \newcommand{\hb}{\mathbf{h}_1}
  \newcommand{\hc}{\mathbf{h}_2}
  \newcommand{\tpdf}{f_{Y|\xin'}}
  \newcommand{\apdf}{f_{\hat{Y}|\xin'}}
  \newcommand{\tcdf}{F_{Y|\xin'}}
  \newcommand{\acdf}{F_{\hat{Y}|\xin'}}
  \newcommand{\itcdf}{F_{Y|\xin'}^{-1}}
  \newcommand{\ifcdf}{G_{\xin'}^{-1}}
  \newcommand{\iafcdf}{\hat{G}^{-1}}
  \begin{proof}
    
  We extend an informal construction by \citet{galUncertainty2016} which aimed to show that a sufficiently deep network with a unimodal approximate posterior could induce a multi-modal posterior predictive by learning the inverse cumulative distribution function (c.d.f.) of the multi-modal distribution.
  In our case, we are not chiefly interested in the number of modes, but more generally the expressive power of the mean-field distribution in a BNN of sufficient width and depth.
  First, we outline a simplified version of the proof that highlights the main mechanisms involved but is not constructed with a Bayesian neural network.
  Later, we prove the full result for Bayesian neural networks.

  \subsubsection{Simplified Construction}
  \tikzstyle{default}=[fill=white, draw=black, shape=circle]
  \tikzstyle{none}=[]
  \tikzstyle{pointer}=[->, draw=black]
  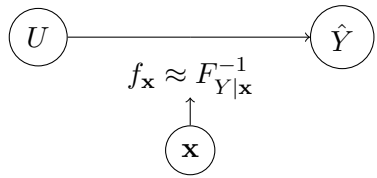
\begin{figure}[!h]
    \centering
    \begin{tikzpicture}
      \begin{pgfonlayer}{nodelayer}
        \node [style=default] (1) at (-4, 4) {$U$};
        \node [style=none, label={below:$f_{\mathbf{x}} \approx F^{-1}_{Y|\xin}$}] (2) at (-2, 4) {};
        \node [style=default] (3) at (-2, 2.5) {$\mathbf{x}$};
        \node [style=default] (4) at (0, 4) {$\hat{Y}$};
      \end{pgfonlayer}
      \begin{pgfonlayer}{edgelayer}
        \draw (1) to (2.center);
        \draw [style=pointer] (3) to (-2, 3.2);
        \draw [style=pointer] (2.center) to (4);
      \end{pgfonlayer}
    \end{tikzpicture}
    \caption[Structure of the UAT argument]{The random variable whose probability density function is the true predictive posterior $p(y|\mathbf{x}, \mathcal{D})$ can be written $Y|\mathbf{x}$. If it has an inverse cumulative density function (c.d.f.), $F^{-1}_{Y|\mathbf{x}}$, we can transform a uniform random variable $U$ onto it. We can approximate this inverse c.d.f.\ with $f_{\mathbf{x}}$ indexed by $\mathbf{x}$. The random variable given by $\hat{Y} \coloneqq f_{\mathbf{x}}(U)$ can be constructed to be `similar' to $Y|\mathbf{x}$ as we show.}
    \label{fig:simple_proof}
  \end{figure}

  \begin{lemma}\label{lem:simplified_construction}
    Let $p(y=Y|\mathbf{x}, \mathcal{D})$ be the probability density function for the posterior predictive distribution of any given univariate regression function.
    Let $U$ be a uniformly distributed random variable.
    Let $f(\cdot)$ be a deterministic neural network with a single hidden layer of arbitrary width and invertible non-polynomial activations.
    Let $\hat{Y}$ be the random variable defined by $f(U, \mathbf{x})$.
    Then, for any $\epsilon > 0$, there exists a set of parameters defining a sufficiently wide $f$ such that the absolute value of the difference in probability densities for any point is bounded:
    \begin{equation}
      \forall y, \mathbf{x}: \quad \abs{p(y=\hat{Y}) - p(y=Y|\mathbf{x}, \mathcal{D})} < \epsilon, \label{eq:simplified_result}
    \end{equation}
    so long as the cumulative distribution function of the posterior predictive is continuous in output-space and the probability density function is non-zero everywhere.
  \end{lemma}
  \begin{proof}
    An outline of the proof for the simplified case is shown in Figure \ref{fig:simple_proof}.

    Suppose there is a true posterior distribution over function outputs whose probability density function (p.d.f) is given by $p(y=Y|\mathbf{x}, \mathcal{D})$.
    These define a random variable that we will denote $Y | \mathbf{x}$.

    We also have some approximation that takes $\mathbf{x}$ as an input and returns some $y$ in the output space.
    Later, this will be our Bayesian neural network, but for now we simplify.
    Instead, our procedure is to have some deterministic neural network which accepts as an input a realization of a uniformly distributed random variable, $U$, and the input point $\mathbf{x}$.
    We define a random variable $\hat{Y} \coloneqq f(U, \mathbf{x})$.

    We would like to show that it is possible to construct a neural network $f$ such that the result in equation \eqref{eq:simplified_result} holds---that $\hat{Y}$ is suitably similar to the true predictive posterior random variable $Y|\mathbf{x}$.

    First, note that if $Y|\mathbf{x}$ has an inverse cumulative density function (c.d.f.) then, by the universality of the uniform, transforming a uniform random variable by this function creates a random variable distributed as $Y|\mathbf{x}$.
    As a result, if there is such an invertible cumulative density function, there is also a function mapping $U$ and $\mathbf{x}$ onto $Y|\mathbf{x}$.

    Second, consider the conditions under which $Y|\mathbf{x}$ has an invertible c.d.f.
    We must assume that the c.d.f.\ is continuous in output space and that the probability density function is non-zero everywhere. 
    The first is reasonable for most normal problems, we often make a stronger assumption of Lipschitz continuity.
    The second is also relatively mild, corresponding to non-dogmatic certainty (a posterior distribution that puts zero probability density on some output given some input can never update away from that in light of new information).
    Given these mild assumptions, therefore, we know that there exists a continuous function $F_{Y|\mathbf{x}}^{-1}$ which is the inverse of the c.d.f.\ of $Y|\mathbf{x}$.
  
    Third, we consider how we might approximate this function.
    Here, we invoke the universal approximation theorem (UAT) \citep{leshnoMultilayer1993}.
    This states that for any continuous function $g$, arbitrary fixed error, $\epsilon$, and compact subset $\mathcal{A}$ of $\mathbb{R}^D$, there exists a deterministic neural network with an arbitrarily wide single layer of hidden units and a non-polynomial activation, $f$ such that:
    \begin{equation}
      \forall \mathbf{a} \in \mathcal{A}: \lvert f(\mathbf{a}) - g(\mathbf{a}) \rvert < \epsilon. 
    \end{equation}
    By setting the arbitrary continuous function as the inverse c.d.f.\ of $Y|\mathbf{x}$, that is, $g(U,\mathbf{x}) = F^{-1}_{Y|\mathbf{x}}(U)$ (which we have already assumed is continuous) it follows that:
    \begin{equation}
      \forall u, \mathbf{x} \in \mathcal{A}: \lvert f(u, \mathbf{x}) - F^{-1}_{Y|\mathbf{x}}(u) \rvert < \epsilon, \quad u \sim U. \label{eq:uat_2}
    \end{equation}
    Fourth, we convert this bound on the inverse c.d.f.\ into a bound on the c.d.f.\ and then a bound on the p.d.f.
    We rewrite the function represented by the neural network to make explicit that we are using $\mathbf{x}$ to index a function from $u$ to $y$: $y = f(u, \mathbf{x}) = f_{\mathbf{x}}(u)$.

    For this step, we will need to be able to invert the approximation to the inverse CDF such that $u = f^{-1}_{\mathbf{x}}(y)$ (for fixed $\xin$).
    In general for neural network functions this is not true. As a result, we employ a construction which breaks apart the line over which $y$ runs into subsegments within which the network is invertible. For non-periodic activation functions which have only zero-measure non-monotonic regions (e.g., ReLU) there will be finitely many of these segments given a finite number of hidden units. Let us index over these subregions with $i$, noting that we can think of the distribution over $y$ as a weighted mixture distribution whose members have zero-density outside of the subregions. The approximate inverse CDF of each of these sub-region mixture members can be written as $f_{\xin}^{i}(u)$ such that $f_{\xin}(u) = \sum_i f_{\xin}^{i}(u)$. Each of these $f_{\xin}^{i}(u)$ is invertible.
    We can therefore rewrite equation \eqref{eq:uat_2} as:
    \begin{align}
       \forall y, \mathbf{x} \in \mathcal{A}: \quad &\abs{\sum_i f_{\mathbf{x}}({f^i_{\mathbf{x}}}^{-1}(y)) - F^{-1}_{Y|\mathbf{x}}({f^i_{\mathbf{x}}}^{-1}(y)) } < \epsilon. \label{eq:uat_cdf}
      \intertext{which implies that:}
      \forall y, \mathbf{x} \in \mathcal{A}: \quad &\lvert \sum_i y - F^{-1}_{Y|\mathbf{x}}({f^{i}_{\xin}}^{-1}(y)) \rvert < \epsilon. \label{eq:uat_simple}
    \end{align}

    Remember that we assumed above that the c.d.f. of $Y|\xin$ is uniformly continuous, which means that for any $y'$ and $y''$, and any $\epsilon > 0$ there exists a $\delta$ such that if $\abs{y' - y''} < \delta$ then $\abs{F_{Y|\mathbf{x}}(y') - F_{Y|\mathbf{x}}(y'')} < \epsilon$.
    Alongside equation \eqref{eq:uat_simple}, and canceling the c.d.f.\ with the inverse c.d.f.\ this entails that:
    \begin{equation}
      \forall y, \mathbf{x} \in \mathcal{A}: \quad \lvert \sum_i F_{Y|\mathbf{x}}(y) - {f^{i}_{\xin}}^{-1}(y)) \rvert < \epsilon. \label{eq:uat_full_cdf}
    \end{equation}
    But since ${f^{i}_{\xin}}^{-1}(y)$ is zero by construction outside of its subregion, this results in a bound on the overall c.d.f. of the random variable.
    That is to say, the bound on the inverse c.d.f.\ implies a bound on the c.d.f.

    Finally, we remember that the cumulative density is the integral of the probability density function.
    Therefore, by Theorem 7.17 of \citet{fedorovTheory1972} and the uniform convergence in the c.d.f.s, it follows that:
    \begin{equation}
      \forall y, \xin \in \mathcal{A}: \quad \abs{p(y=Y|\mathbf{x}, \mathcal{D}) - p(y=\hat{Y})} < \epsilon
    \end{equation}
    introducing a bound in the probability density functions of the random variable of true posterior outputs and the outputs of the approximation $\hat{Y} = f(U, \xin)$.
  \end{proof}

  \subsubsection{Full Construction}
  The full construction extends the result above in the following ways:
  \begin{itemize}
    \item Rather than separately introducing $U$, we show how the first layer of a Bayesian neural network can map $\xin \to \xin', Z$, where $Z$ is a unit Gaussian random variable and $\xin'$ is a noised version of $\xin$.
    \item Rather than using a deterministic neural network for the universal approximation theorem, we apply the stochastic adaptation introduced by \citet{foongExpressiveness2020}.
    \item Rather than a univariate regression, we consider multivariate regression.
  \end{itemize}

  Like \citet{leshnoMultilayer1993} we note that the extension from univariate to multivariate regression follows trivially from the existence of a mapping from $\mathbb{R} \to \mathbb{R}^K$.\footnote{A slightly complication is added by the continuity requirements. However, we note that the assumption that the p.d.f.\ is finite everywhere guarantees that there is a continuous function over $y$ which contains continous segments for each of $K$ dimensions, even if those individual segments are not continuous with each other.}

  We first give some intuition as to how the proof works.
  The first weight layer serves to map $\mathbf{x} \to \mathbf{x}', Z$.
  This sets us up in a similar situation to the proof in the previous section, where we began with $\mathbf{x}, U$.
  This requires two small adjustments to the proof above.
  The first is that the random variable we introduce is now Gaussian, rather than Uniform.
  The second is that all our results will be in terms of $\xin'$, rather than $\xin$, and an additional step will be required to convert a probabilistic bound in one to the other (noting that we can freely set the weights in the first layer to have arbitrarily small variance).

  The second weight layer will play the role of the neural network in the simplified proof. This also will require a small modification, because earlier we assumed that the neural network was deterministic, but it is now stochastic. This means that the final result becomes a probabilistic
  bound.
  
  As a result of all of these changes, the proof becomes considerably more complicated, though nothing important changes in the intuition behind the construction.

  \textbf{Step 1: Mapping $\xin \to \xin', Z$}
  
  Consider inputs $\mathbf{x} \in \mathbb{R}^D$.
  We define a two-hidden-layer neural network with an invertible non-polynomial activation function $\phi: \mathbb{R} \mapsto \mathbb{R}$.
  The first component of the network is a single weight matrix mapping onto a vector of hidden units: $\hb = \phi(\Wa \xin + \ba)$.
  The second component is a neural network with a layer of hidden units defined relative to the first layer of units $\hc = \phi(\Wb \hb + \bb)$, and outputs $\yout = \Wc \hc + \bc$.
  The distribution over the outputs $\yout$ defines the random variable $\hat{\Yout}$.
  Here, $\Wa$, $\Wb$, and $\Wc$ are matrices of independent Gaussian random variables and $\ba$, $\bb$, and $\bc$ are vectors of independent Gaussian random variables.
  Given some dataset $\mathcal{D}$ the predictive posterior distribution over outputs is $p(\yout|\mathbf{x}, \mathcal{D})$ which we associate with the random variable $\Yout$.
  This is our (intractable) target.

  Consider only the first component of the neural network, which maps $\xin$ onto $\hb$.
  We can construct simple constraints on $\Wa$ and $\ba$ such that:
  \begin{equation}
    \hb = \begin{pmatrix}
      \phi(z) \\
      \phi(\xin')
    \end{pmatrix},
  \end{equation}
  where $z \sim Z$, a unit Gaussian random variable, and $\xin' \in \mathbb{R}^D$, such that $\textup{Pr}\left(\norm{\xin' - \xin} > \epsilon_1\right) < \delta_1$.
  In particular, suppose that for $\Wa \in \mathbb{R}^{D \times D+1}$ and $\ba \in \mathbb{R}^{D+1}$:
  \begin{align}
    \Wa &= \mathcal{N}\left(M_{\Wa}, \Sigma_{\Wa}\right) \text{ where } M_{\Wa} = (\mathbf{0}_D, \mathbb{I}_{D\times D}); \Sigma_{\Wa} = \sigma^2\mathbb{I}_{D\times D+1};\\
    \ba &= \mathcal{N}\left(\boldsymbol{\mu}_{\ba}, \boldsymbol{\sigma}_{\ba}\right) \text{ where } \boldsymbol{\mu}_{\ba} = \mathbf{0}_{D+1}, \boldsymbol{\sigma}_{\ba} = \begin{pmatrix}1& \sigma& \dots& \sigma\end{pmatrix}.
  \end{align}
  By multiplication, straightforwardly $\xin' = \mathcal{N}\left(\xin, 2 \sigma^2 \mathbf{1}\right)$.
  It follows trivially that for any $\epsilon_1$ and $\delta_1$ there exists some $\sigma$ such that the bound holds.
  We will apply this bound at the end of the proof to convert a bound in $\xin'$ to one in $\xin$.
  
  Here, we introduce a distinction between the weights which determine $\xin'$ and those that create $Z$. The only weights which determine $Z$ are the first element of $\Wa$ and the first element of $\ba$. Call these $\thet_{Z}$. We then define the remainder of the weight distributions as $\thet_{\textup{Pr}} \coloneqq \{\Wa, \Wb, \Wc, \ba, \bb, \bc\} \setminus \thet_{Z}$.
  This distinction is important, because the probabilistic bound in the proof will be over $\thet_{\textup{Pr}}$ while the distribution over $\thet_{Z}$ will induce the random variable $\hat{\mathbf{Y}}$.

  \textbf{Step 2: Invoking a Result Similar to the Simplified Construction}
  We show that there is a function which maps $Z$ and $\xin'$ onto $\Yout$, under reasonable assumptions similar to those of the simplified construction.
  For brevity, we denote the probability density function (p.d.f.) of the true posterior predictive distribtion of the random variable $\Yout$ conditioned on $\xin'$ and $\mathcal{D}$ as $\tpdf \equiv p(Y=y | \xin', \mathcal{D})$ and the cumulative density function (c.d.f.) as $\tcdf$.
  We similarly write the inverse c.d.f. as $\itcdf$.

  We must adapt the simplified construction to account for the fact that rather than simply approximating the inverse of the c.d.f.\ we now need to also transform the Gaussian random variable onto a Uniform one and invert the activation.
  We show below:

  \begin{restatable}{lemma}{leminvcdf}\label{lemma:invcdf}
     There exists continous function $\ifcdf = \itcdf \cdot F_{Z} \cdot \phi^{-1} $, where $F_{Z}$ is the c.d.f. of the unit Gaussian and $\phi^{-1}$ is the inverse of the activation function, such that the random variable $\ifcdf(\phi(Z))$ is equal in distribution to $\Yout|\mathbf{x}$ , if the p.d.f.\ of the posterior predictive is non-zero everywhere and the c.d.f.\ is continuous.
  \end{restatable}
  The limitation to p.d.f.s is modest as before.
  The naming of the function $\ifcdf$ is suggestive, and indeed its inverse exists if $\tcdf$ is invertible, since the c.d.f.\ of a unit Gaussian is invertible (though this function cannot be easily expressed).

  Whereas in the simplified construction we showed that the neural network could approximate the inverse c.d.f., here we show that the second hidden layer of our larger Bayesian neural network can approximate the more complicated function required by lemma \ref{lemma:invcdf}.
  This allows the second hidden layer of the neural network to transform $\xin', Z$ onto $\hat{\Yout}$ such that $\hat{\Yout}$ is appropriately similar to $\Yout$.
  First we show that we can approximate the function $\ifcdf$, generating a probabilistic bound because the weights of the neural network are now Gaussian random variables:

  \begin{restatable}{lemma}{lemfmatching}\label{lemma:fmatching}
    For a uniformly continous function $\ifcdf(z): z, \xin' \mapsto \yout$, for any $\epsilon, \delta>0$ and compact subset $\mathcal{A}$ of $\mathbb{R}^D$, there exist fully-factorized Gaussian approximating distributions $q(\Wb)$, $q(\Wc)$, $q(\bb)$, and $q(\bc)$, and a function over the outputs of the later part of the neural network: $\iafcdf(\xin', z) \equiv \Wc(\sigma(\Wb\hb) + \bb) +\bc$ (remembering that $\hb \equiv \phi(z, \xin')$), such that:
    \begin{equation}
        \textup{Pr}\Big(\big\lvert \iafcdf(\xin', z) - \ifcdf(z)\big\rvert > \epsilon \Big) < \delta, \quad \forall \xin', z, \mathcal{A}.
      \end{equation}
    The probability measure is over the weight distributions of $\Wb, \Wc, \bb, \bc$.
  \end{restatable}

 Having shown that the second component can approximate the inverse c.d.f.\ to within a bound, we as before we further show that the random variable created by this transformation has a p.d.f.\ within a bound of the p.d.f.\ of $\Yout$, which suffices to prove the desired result.
  
  For this, we show this below for the transformed variable $\xin'$:
  \begin{restatable}{lemma}{lempdf}\label{lemma:pdf}
    For any $\epsilon > 0$ and $\delta > 0$ there exists a mean-field weight distribution $q(\Wb,\Wc,\bb,\bc)$ such that the probability density functions are bounded:
    \begin{equation}
      \textup{Pr}\Big(\big\lvert p(y_i = \hat{\Yout}_i) - p(y_i = \Yout_i|\xin', \mathcal{D})\big\rvert > \epsilon \Big) < \delta, \quad \forall \xin', y_i, \mathcal{A}.
    \end{equation}
  \end{restatable}
  We then move the bounds onto an expression in the original features, $\xin$.
  Recall from before that because the variance of the weights in the first layer can be arbitrarily small, that for any $\epsilon$ there is a $\delta$:
\begin{equation}
  \textup{Pr}\left(\norm{\xin' - \xin} > \epsilon\right) < \delta \quad \forall \xin, \xin' \in \mathcal{A},
\end{equation}
where the probability measure is over $\thet_{\textup{Pr}}$.
Moreover, since we have assumed that the probability density function is continuous, this bound alongside the previous bound on the probability density functions jointly entail that:
\begin{equation}
   \textup{Pr}\Big(\big\lvert p(y_i = \hat{\Yout}_i) - p(y_i = \Yout_i|\xin, \mathcal{D})\big\rvert > \epsilon \Big) < \delta, \quad \forall \xin \in \mathcal{A}, y_i.
\end{equation}
where the probability measure is over $\thet_{\textup{Pr}}$.
\end{proof}

Below, we prove the lemmas required in the proposition above.
  
  \leminvcdf*
  \begin{proof}
      Trivially $\phi^{-1}(\phi(Z)) = Z$.

      Let $F_Z$ be the cummulative distribution function (c.d.f.) of the unit Gaussian random variable Z.
      By the Universality of the Uniform, $U = F_Z(Z)$ has a standard uniform distribution.

      Let $F_{Y|\xin}$ be the c.d.f.\ of $\Yout$ conditioned on $\mathbf{x}$, and $\mathcal{D}$.
      Suppose that the posterior predictive is non-zero everywhere (that is, you cannot rule out that there's even the remotest chance of any $y_i$ given some input $\mathbf{x}$, however small).
      Then, since the c.d.f.\ is continuous by assumption, $F_{Y|\xin}$ is invertible.

      Again, by the Universality of the Uniform $U' = F_{Y|\xin}(y)$ has a standard uniform distribution. So $\forall u: p(U = u) = p(U' = u)$.
      Moreover $F_{Y|\xin}$ is invertible.
      So $p(Y=y) = F_{Y|\xin}^{-1}(F_Z(Z))$.

      It follows that there exists a continuous function as required.

  \end{proof}

  \lemfmatching*
  \begin{proof}
    The Universal Approximation Theorem (UAT) states that for any continuous function $f$, and an arbitrary fixed error, $e$, and compact subset $\mathcal{A}$ of $\mathbb{R}^D$, there exists a deterministic neural network with an arbitrarily wide single layer of hidden units and a non-polynomial activation, $\sigma$:
    \begin{equation}
      \forall \mathbf{x} \in \mathcal{A}: \lvert \sigma(\mathbf{w}_2(\sigma(\mathbf{w}_1\mathbf{x}) + \mathbf{b}_1) +\mathbf{b}_2) - f(\mathbf{x}) \rvert < e. \label{eq:uat}
    \end{equation}
    
    In addition, we make use of Lemma 7 of \citep{foongExpressiveness2020}. This states that for any $e', \delta_2 > 0$, for some fixed means $\boldsymbol{\mu}_1$, $\boldsymbol{\mu}_2$, $\boldsymbol{\mu}_{b_1}$, $\boldsymbol{\mu}_{b_2}$ of $q(\thet_1)$, $q(\thet_2)$, $q(\mathbf{b}_1)$, and $q(\mathbf{b}_2)$ respectively, there exists some standard deviation $s'>0$ for all those approximate posteriors such that for all $s < s'$, for any $\mathbf{h}_1 \equiv (z, \xin') \in \mathbb{R}^{N+1}$
    \begin{equation}
      \text{Pr}\Big(\big\lvert \sigma(\thet_2(\sigma(\thet_1\mathbf{h}_1) + \mathbf{b}_1) +\mathbf{b}_2) - \sigma(\boldsymbol{\mu}_2(\sigma(\boldsymbol{\mu}_1\mathbf{h}_1) + \boldsymbol{\mu}_{b_1}) +\boldsymbol{\mu}_{b_2})\big\rvert > e' \Big) < d.
  \end{equation}
  Note that the deterministic weights of \cref{eq:uat} can just be these means.
  As a result:
  \begin{equation}
    \text{Pr}\Big(\big\lvert \sigma(\thet_2(\sigma(\thet_1\mathbf{h}_1) + \mathbf{b}_1) +\mathbf{b}_2) - f(\hb)\big\rvert > e + e' \Big) < \delta.
\end{equation}
We note that we define $\iafcdf(\xin', z) \equiv \sigma(\Wc(\sigma(\Wb\hb) + \bb) +\bc)$ as above, and that $f(\hb)$ may be $G_{\xin'}^{-1}(z)$, which is assumed to be uniformly continuous.
It follows, allowing $\epsilon = e + e'$:
\begin{equation}
    \text{Pr}\Big(\big\lvert \iafcdf(\xin', z) - G_{\xin'}^{-1}(z)\big\rvert > \epsilon \Big) < \delta.
\end{equation}
as required.

\lempdf*
In this lemma, we show that a bound on the inverse c.d.f.\ used to map $Z$ onto our target implies a bound in the p.d.f.\ of that constructed random variable to the p.d.f.\ of our target.
This lemma follows the argument of the simplified construction, with some additional complexity of notation introduced by the requirement that the input random variable was Gaussian rather than Uniform. Here, we complete the proof steps with a univariate $y$ to simplify notation, noting that because we can map $\mathbb{R} \to \mathbb{R}^K$ the multivariate regression follows trivially from the univariate result.

We first note that the result of \cref{lemma:fmatching} can be applied to $z' = \hat{G}(\xin', y)$ using an inverted version of our network function such that:
  \begin{equation}
    \textup{Pr}\left(\abs*{\iafcdf(\xin', \hat{G}(\xin', y)) - \ifcdf(\hat{G}(\xin', y))} > \epsilon_2 \right) < \delta_2, \quad \forall \xin' \in \mathcal{A}, y.
  \end{equation}
  We further note that by the triangle inequality:
  \begin{align}
    \abs*{\ifcdf(G_{\xin'}(y)) - \ifcdf(\hat{G}(\xin', y))} &\leq \abs*{\ifcdf(G_{\xin'}(y)) - \iafcdf(\xin', \hat{G}(\xin', y))} \\ &\quad+ \abs*{\iafcdf(\xin', \hat{G}(\xin', y)) - \ifcdf(\hat{G}(\xin', y))},
    \intertext{and since $\ifcdf(G_{\xin'}(y)) = \iafcdf(\xin', \hat{G}(\xin', y)) = \xin'$:}
    &\leq \abs*{\iafcdf(\xin', \hat{G}(\xin', y)) - \ifcdf(\hat{G}(\xin', y))}.
  \end{align}
  Inserting this inequality into the result of \cref{lemma:fmatching}, we have that:
  \begin{equation}
    \textup{Pr}\Big(\abs*{\ifcdf(G_{\xin'}(y)) - \ifcdf(\hat{G}(\xin', y))} > \epsilon_2 \Big) < \delta_2, \quad \forall \xin' \in \mathcal{A}, y.\label{eq:reworked_bound}
\end{equation}
But we further note that we have assumed that the c.d.f.\ $G_{\xin'}$ is uniformly continous so for any $y'$ and $y''$, and for any $\epsilon'>0$ there is an $\epsilon''>0$ and vice versa, such that if:
\begin{equation}
  \abs*{y' - y''} < \epsilon',
\end{equation}
then:
\begin{equation}
  \abs*{G_{\xin'}(y') - G_{\xin'}(y'')} < \epsilon''.
\end{equation}
It follows that for any $\epsilon$ and $\delta$ there is a $q(\thet)$ such that:
\begin{equation}
  \textup{Pr}\Big(\abs*{G_{\xin'}(\ifcdf(G_{\xin'}(y))) - G_{\xin'}(\ifcdf(\hat{G}(\xin', y)))} > \epsilon \Big) < \delta, \quad \forall \xin' \in \mathcal{A}, y,
\end{equation}
and therefore:
\begin{equation}
  \textup{Pr}\Big(\abs*{(G_{\xin'}(y) - \hat{G}(\xin', y)} > \epsilon \Big) < \delta, \quad \forall \xin' \in \mathcal{A}, y.
\end{equation}
Next, we remember that $G_{\xin'} = \phi \cdot F^{-1}_Z \cdot \tcdf$, and that $\phi$ and $F^{-1}_{Z}$ are continuous, and therefore:
\begin{equation}
  \textup{Pr}\Big(\abs*{(\tcdf(y) - \acdf(y)} > \epsilon \Big) < \delta, \quad \forall \xin' \in \mathcal{A}, y,
\end{equation}
where $\acdf(y) = F_Z(\phi^{-1}(\hat{G}(\xin', y)))$.

As a final step, we remember that the cumulative density is the integral of the probability density function.
Therefore, by Theorem 7.17 of \citet{fedorovTheory1972} and the uniform convergence in the c.d.f.s, it follows that there for any bounds there exists $q(\thet)$ such that:
\begin{equation}
  \textup{Pr}\Big(\big\lvert \apdf - \tpdf\big\rvert > \epsilon \Big) < \delta, \quad \forall \xin' \in \mathcal{A},
\end{equation}
Writing out the probability density functions fully and mapping the univariate function the multivariate we have:
\begin{equation}
  \textup{Pr}\Big(\big\lvert p(y_i = \hat{\Yout}_i) - p(y_i = \Yout_i|\xin', \mathcal{D})\big\rvert > \epsilon \Big) < \delta, \quad \forall \xin' \in \mathcal{A}, y_i.
\end{equation}

  \end{proof}

\ifSubfilesClassLoaded{
\bibliographystyle{plainnat}
\bibliography{thesis_references}}{}
\end{document}

\renewcommand{\wr}{\vw^{(r)}}
\newcommand{\wx}{\vw^{(x)}}
\newcommand{\er}{\boldsymbol{\epsilon}^{(r)}}
\newcommand{\ex}{\boldsymbol{\epsilon}^{(x)}}
\newcommand{\eri}{\epsilon^{(r)}}
\newcommand{\exi}{\epsilon^{(x)}}
\newcommand{\bmu}{\boldsymbol{\mu}}
\newcommand{\bsi}{\boldsymbol{\sigma}}
\newcommand{\bth}{\boldsymbol{\theta}}

\chapter{Appendix to Approximation Assumptions Affect Optimization}
\section{Derivation of the Entropy Term of the KL-divergence}\label{a:derivation_of_entropy}
In this section, we show that the component of KL-divergence term of the loss which is the entropy of the posterior distribution over the weights $q(\wx)$ can be estimated as:

\begin{align}
    \mathcal{L}_{\text{entropy}} \coloneqq& \int q(\wx) \log [q(\wx)] d\wx \label{eq:entropy}\\
    =&- \sum_i\log[\sigma_i^{(x)}] + \text{const}
\end{align}
where $i$ is an index over the weights of the model.

Throughout this section we use a superscript indicates the basis---an $(x)$ means we are in the Cartesian coordinate system tied to the weight-space while $(r)$ is the hyperspherical coordinate system (the letter is the canonical `first' coordinate of that coordinate system).

We begin by applying the reparameterization trick \citep{kingmaAutoEncoding2014,rezendeStochastic2014}.
Following the auxiliary variable formulation of \citet{galUncertainty2016}, we express the probability density function of $q(\wx)$ with an auxiliary variable.
\begin{align}
    q(\wx) &= \int q(\wx, \er)d\er\\
    &= \int q(\wx|\er)q(\er)d\er\\
    &= \int \delta(\wx - g(\mu, \sigma, \er)q(\er)d\er. \label{auxilliary}
\end{align}

In equation (\ref{auxilliary}), we have used a reparameterization trick transformation:
\begin{align}
    g(\mu, \sigma, \er) = \bmu + \bsi \odot \mathbf{T}_{rx}(\er)
\end{align}
where $\bmu$ and $\bsi$ are parameters of the model and where $\mathbf{T}_{rx}$ is the standard transformation from hyperspherical into Cartesian coordinates.

Substituting equation (\ref{auxilliary}) into the definition of the entropy loss term in equation (\ref{eq:entropy}), and applying the definition of the Kronecker delta we can eliminate dependence on $\wx$:

\begin{align}
\mathcal{L}_{\text{entropy}} &= \int q(\wx) \log [q(\wx)] d\wx\\    
    &= \int \bigg(\int \delta(\wx - g(\bmu, \bsi, \er) q(\er)d\er\bigg) \log [q(\wx)] d\wx\\
    &= \int q(\er) \log [q(g(\bmu, \bsi, \er))] d\er.
\end{align}

Then, we perform a coordinate transformation from $g(\bmu, \bsi, \er)$ to $\er$ using the Jacobian of the transformation and simplify.
\begin{align}
    &= \int q(\er) \log \Bigg[q(\er) \bigg|\frac{\partial g(\bmu, \bsi, \er)}{\partial \er}\bigg|^{-1}\Bigg]d\er\\
    &= \int q(\er) \log \Bigg[q(\er) \bigg|\prod_i \sigma_i^{(x)} \frac{\partial \exi_i}{\partial \eri_j}\bigg|^{-1}\Bigg]d\er\\
    &= \int q(\er) \log \Bigg[q(\er) \bigg|\text{diag}(\bsi) \frac{\partial \exi_i}{\partial \eri_j}\bigg|^{-1}\Bigg]d\er\\
    &= \int q(\er) \log \Bigg[\frac{q(\er)}{\prod_i \sigma_i^{(x)}} \bigg|\frac{\partial \exi_i}{\partial \eri_j}\bigg|^{-1}\Bigg]d\er \label{eq:unworkedout_entropy}
\end{align}
In the last line we have used the fact that $\forall{i}:$ $\sigma_i^{(x)} \geq 0$ allowing us to pull the determinant of this diagonal matrix out.
$\bigg|\frac{\partial \exi_i}{\partial \eri_j}\bigg|$ is the determinant of the Jacobian for the transformation from Cartesian to hyperspherical coordinates for which we use the result by \citet{muleshkovEasy2016}:
\begin{align}
    \bigg|\frac{\partial \ex_i}{\partial \er_j}\bigg| &= \text{abs}\Bigg(\big(-1)^{D-1}\big(\epsilon_0^{(r)}\big)^{D-1} \prod_{i = 2}^{D-1}\big(\sin(\epsilon_i^{(r)})\big)^{i-1}\Bigg).
\end{align}

We know that $\epsilon_0^{(r)} \geq 0$ because the radial dimension in hyperspherical coordinates can be assumed positive without loss of generality.
We also know $0 \leq \epsilon_i^{(r)} \leq \pi$ for $2 \leq i \leq D - 1$ for the hyperspherical coordinate system.
So we can simplify the signs:
\begin{align}
    = \big(\epsilon_0^{(r)}\big)^{D-1} \prod_{i = 2}^{D-1}\big(\sin(\epsilon_i^{(r)})\big)^{i-1}.\label{eq:coord_conversion}
\end{align}

Therefore, plugging equation (\ref{eq:coord_conversion}) into (\ref{eq:unworkedout_entropy}):
\begin{align}
    \mathcal{L}_{\text{entropy}} &=  \int q(\er) \log \Bigg[\frac{q(\er)}{\text{abs}(\prod_i \sigma_i^{(x)})} \bigg|\frac{\partial \exi_i}{\partial \eri_j}\bigg|^{-1}\Bigg]d\er\\
    &=  \int q(\er) \log [q(\er)] \nonumber \\ &\qquad- \log[\text{abs}(\prod_i \sigma_i^{(x)})] \nonumber \\ &\qquad -\log \bigg[ \big(\epsilon_0^{(r)}\big)^{D-1} \prod_{i = 2}^{D - 1}\big(\sin(\epsilon_i^{(r)})\big)^{i-1}\bigg] d\er.\label{eq:3_terms}
\end{align}

Only the middle term depends on the parameters, and we must therefore only compute this term in order to compute gradients.
For sake of completeness, we address the other integrals below, in case one wants to have the full value of the loss (though since it is a lower bound in any case, the full value is not very useful).

The probability density function of the noise variable is separable into independent distributions. The distribution of $\epsilon^{(r)}_0$ is a unit Gaussian. The angular dimensions are distributed so that sampling is uniform over the hypersphere. However, this does \textbf{not} mean that the distribution over each angle is uniform, as this would lead to bunching near the n-dimensional generalization of the poles. (Intuitively, there is more surface area per unit of angle near the equator, as is familiar from cartography.) Instead, we use the fact that the area element over the hypersphere is:
\begin{align}
    dA = \left(\eri_{0}\right)^{D-1}\prod_{i=1}^{D-1}\sin(\eri_{i})^{i-1}d\eri_{i}
\end{align}
where we remember that $\eri_{1}$ is between $-\pi$ and $\pi$, and the rest of the angular elements of $\er$ are between $0$ and $\pi$. The resulting probability density function is:
\begin{align}
    q(\boldsymbol{\epsilon}^{(r)}) &= \prod_{i=0}^{D} q(\epsilon_i^{(r)}) = \left(\eri_0\right)^{D-1}\frac{1}{\sqrt{2\pi}}e^{-\frac{\epsilon_0^2}{2}} \cdot \prod_{i=2}^{D-1}\sin(\eri_{i})^{i-1}. \label{eq:pdf}
\end{align}

As a result, equation (\ref{eq:3_terms}) becomes analytically tractable.
Using equation (\ref{eq:pdf}) we have that
\begin{align}
    \log\left[q(\eri)\right] - \log\left[\left(\eri_0\right)^{D-1}\prod_{i=2}^{D-1}\left(\sin(\eri)\right)^{i-1}\right]= -\frac{1}{2}\log 2\pi - \frac{(\eri_0)^2}{2}
\end{align}
where the second two terms cancel with the third term in equation (\ref{eq:3_terms}).
That is, the first and third terms of equation (\ref{eq:3_terms}) mostly cancel.
That is,
\begin{align}
    \mathcal{L}_{\text{entropy}} &= - \int q(\eri)\left(\log[\abs{\prod_i \sigma_i^{(x)}}] + \frac{1}{2}\log 2\pi + \frac{(\eri_0)^2}{2}\right) d\er.\\
    &= - \log[\abs{\prod_i \sigma_i^{(x)}}] - \frac{1}{2}\log 2\pi - \int_0^\infty q(\eri_0) \frac{(\eri_0)^2}{2} d\eri_0 \\
    \intertext{which, remembering that the radial dimension is distributed as a doubled unit normal (adjusting for being strictly positive) leaves us with}
    &= - \sum_i \log[\sigma_i^{(x)}] - \frac{1}{2}\log 2\pi - \frac{1}{2}.\\
    \intertext{And because we are optimizing the loss and can generally neglect constant terms}
    \mathcal{L}_{\text{entropy}} &= - \sum_i\log[\sigma_i^{(x)}] + \text{const}.
\end{align}

\section{Setting a Radial Prior}\label{a:derivation_of_prior}
In most of our experiments, we use a typical multivariate Gaussian unit prior in order to ensure comparability with prior work.
However, in some settings, such as the Variational Continual Learning setting, it is useful to use the radial posterior as a prior.
In these cases, the entropy term is identical to the term for a Gaussian prior.
The cross-entropy term can, in principle, be found using Monte Carlo estimation just like the Gaussian case.
However, this requires us to compute the expectation of the log of the probability density function.
While the probability density function is easy to write down in hyperspherical coordinates, it is not trivial to state for Cartesian coordinates.
This means that we must complete a change of variables in order to compute the probability density function in hyperspherical coordinates

We begin similarly to the previous derivation, with all unchanged expect that we are estimating
\begin{align}
    \mathcal{L}_{\text{cross-entropy}} &= \int q(\wx) \log [p(\wx)] d\wx.
\end{align}
In addition to our coordinate transformation $g(\vmu, \vsigma, \er)$, we also define the probability density function over the prior with a similar transformation
\begin{equation}
    g_\text{prior}(\vmu_{\text{prior}}, \vsigma_\text{prior}, \er) = \vmu_\text{prior} + \vsigma_\text{prior} \cdot \mathbf{T}_\text{rx}(\er) \text{ with } \er \sim p_\epsilon(\er) = q_\epsilon(\er).
\end{equation}
We begin similarly to the derivation for the entropy loss above
\begin{align}
    \mathcal{L}_{\text{cross-entropy}} &= \int q(\wx) \log [p(\wx)] d\wx\\    
    &= \int \bigg(\int \delta(\wx - g(\bmu, \bsi, \er) q(\er)d\er\bigg) \log [p(\wx)] d\wx\\
    &= \int q_\epsilon(\er) \log [p(g(\bmu, \bsi, \er))] d\er.\label{eq:radial_cross_entropy}
\end{align}
Then, we perform a coordinate transformation which is, unlike the previous case, from $g_\text{prior}(\vmu, \vsigma, \er)$ to $\er$.
By the change of variables formula
\begin{align}
    p(\wx) &= p_\epsilon(g^{-1}_\text{prior}(\wx))\abs*{\det\left[\frac{\partial g^{-1}_\text{prior}(\wx)}{\partial \wx}\right]},\\
    \intertext{and then by the inverse function theorem}
    &= p_\epsilon(g^{-1}_\text{prior}(\wx)) \abs*{\det\left[\frac{\partial g_\text{prior}(\vmu_\text{prior}, \vsigma_\text{prior}, \er)}{\partial\er}\right]^{-1}},\\
    \intertext{and by the chain rule}
    &= p_\epsilon(g^{-1}_\text{prior}(\wx)) \abs*{\det\left[\frac{\partial g_\text{prior}(\vmu_\text{prior}, \vsigma_\text{prior}, \er)}{\partial\ex}\frac{\partial \er}{\partial \ex}\right]^{-1}}.\\
    \intertext{Using our earlier results from \cref{eq:coord_conversion} and \cref{eq:unworkedout_entropy} this is equal to}
    &= p_\epsilon(g^{-1}_\text{prior}(\wx)) \abs*{\big(\epsilon_0^{(r)}\big)^{D-1} \prod_{i = 2}^{D}\big(\sin(\epsilon_i^{(r)})\big)^{i-1}\prod_j\sigma_{\text{prior},j}}^{-1}.
\end{align}
We can then substitute this expression back into \cref{eq:radial_cross_entropy} resulting in
\begin{align}
    \Ls_\text{cross-entropy} &= \int q_\epsilon(\er) \log \frac{q_\epsilon(g^{-1}_\text{prior}(g(\vmu, \vsigma, \er)))}{\abs*{\big(\epsilon_0^{(r)}\big)^{D-1} \prod_{i = 2}^{D}\big(\sin(\epsilon_i^{(r)})\big)^{i-1}\prod_j\sigma_{\text{prior},j}}}.\\
    \intertext{Because the denominator is independent of the optimized variables, we can neglect the constant and focus on}
    &= \int q_\epsilon(\er) \log q_\epsilon(g^{-1}_\text{prior}(g(\vmu, \vsigma, \er))),\\
    \intertext{where we can use the definition of the change of variables to see}
    &= \int q_\epsilon(\er) \log q_\epsilon\left(\mathbf{T}_{xr}\left(\frac{\wx - \vmu_\text{prior}}{\vsigma_\text{prior}}\right)\right).
\end{align}
That is, we can compute the expectation by finding the hyperspherical coordinate-system density of the transformed normalized weights.
The transformation is standard and widely used, but computationally inconvenient, making it much more efficient to use a Gaussian prior where possible.
We can estimate it using a Monte Carlo approximation:
\begin{align}
\approx \frac{1}{N}\sum_{i=1}^{N} \log q_\epsilon\left(\mathbf{T}_{xr}\left(\frac{\wx - \vmu_\text{prior}}{\vsigma_\text{prior}}\right)\right).
\end{align}
Recalling \cref{eq:pdf} which states the probability density function of the Radial distribution in hyperspherical coordinates
\begin{equation}
    q(\boldsymbol{\epsilon}^{(r)}) = \frac{1}{\sqrt{2\pi}}e^{-\frac{\epsilon_0^2}{2}} \cdot \prod_{i=1}^{D-1}\sin(\eri_{i})^{D-i}, 
\end{equation}
we can then compute the cross-entropy term of the loss using an explicit coordinate transformation and this expression.
This is much more computationally cumbersome than the multivariate Gaussian case, but is tractable.

\section{Experimental Settings}\label{a:experimental_settings}
\subsection{Diabetic Retinopathy Settings}\label{a:dr_hypers}
\textit{Authors' Note: The version of the diabetic retinopathy dataset used in this paper (as described below) is slightly different from the one used in the latest release of the Diabetic Retinopathy benchmark (as of May 2021).
Radial BNNs have been included in a \href{https://github.com/google/uncertainty-baselines/tree/master/baselines/diabetic_retinopathy_detection}{more recent benchmarking effort} in which they do not perform quite as well as we found originally.}

The diabetic retinopathy data are publicly available at \url{https://www.kaggle.com/c/diabetic-retinopathy-detection/data}.
We augment and preprocess them similarly to \citet{leibigLeveraging2017}.
The images for our main experiments in \cref{s:retinopathy} are downsampled to 512x512 while the smaller robustness experiment in \cref{s:robustness} uses images downsampled to 256x256
We randomly flip horizontally and vertically.
Then randomly rotate 180 degrees in either direction.
Then we pad by between 0 and 5\% of the width and height and randomly crop back down to the intended size.
We then randomly crop to between 90\% and 110\% of the image size, padding with zeros if needed.
We finally resize again to the intended size and normalize the means and standard deviations of each channel separately based on the training set means and standard deviations.
The training set has 44,594 RGB images.
There are 7,026 validation and 10,000 test images. 

The smaller model used for robustness experiments is loosely inspired by VGG-16, with only 16 channels, except that it is a Bayesian neural network with mean and standard deviations for each weight, and that instead of fully connected networks at the end it uses a concatenated global mean and average pool.
The larger model used in the main experiments is VGG-16 but with the concatenated global mean and average pool instead of fully connected layers as above.
The only difference is that we use only 46 channels, rather than 64 channels as in VGG-16, because the BNN has twice as many parameters as a similarly sized deterministic network, and we wanted to compare models with the same number of parameters.
For the dropout model we use VGG-16 with the full 64 channels, and similarly for each of the models in the deep ensemble.
The prior for training MFVI and Radial BNNs was a unit multivariate Gaussian.
(We also tried using the scale mixture prior used in \citet{blundellWeight2015} and found it made no difference.)
Instead of optimizing $\sigma$ directly we in fact optimize $\rho$ such that $\sigma = \log(1 + e^{\rho})$ which guarantees that $\sigma$ is always positive.
In some cases, as described in the paper, the first epoch only trained the means and uses a NLL loss function.
This helps the optimization, but in principle can still allow the variances to train fully if early stopping is not employed (unlike reweighting the KL-divergence).
Thereafter, we trained using the full ELBO loss over all parameters.
Unlike some prior work using MFVI, we have not downweighted the KL-divergence during training.

For the larger models, we searched for hyperparameters using Bayesian optimization.
We searched between 0 and -10 as the initial value of $\rho$ (equivalent to $\sigma$ values of $\log(2)$ and $~2\cdot10^{-9}$).
For the learning rate we considered $10^{-3}$ to $10^{-5}$ using Adam with a batch size of 16.
Otherwise, hyperparameters we based on exploration from the smaller model.

We then computed the test scores using a Monte Carlo estimate from averaging 16 samples from the variational distribution.
We estimate the model's uncertainty about a datapoint using the mutual information between the posterior's parameters and the prediction on a datapoint.
This estimate is used to rank the datapoints in order of confidence and compute the model's accuracy under the assumption of referring increasingly many points to medical experts.

For the smaller models, we performed an extensive random hyperparameter search.
We tested each configuration with both MFVI and Radial BNNs.
We tested each configuration for both an SGD optimizer and Amsgrad.
When training with SGD we used Nesterov momentum 0.9 and uniformly sampled from 0.01, 0.001 and 0.0001 as learning rates, with a learning rate decay each epoch of either 1.0 (no decay), 0.98 or 0.96.
When training with Amsgrad we uniformly sampled from learning rates of 0.001, 0.0001, and 0.00001 and did not use decay.
We uniformly selected batch sizes from 16, 32, 64, 128, and 256.
We uniformly selected the number of variational distribution samples used to estimate the loss from 1, 2, and 4.
However, because we discarded all runs where there was insufficient graphics memory, we were only able to test up to 64x4 or 256x1 and batch sizes above 64 were proportionately less likely to appear in the final results.
We selected the initial variance from $\rho$ values of -6, -4, -2, or 0.
We also tried reducing the number of convolutional channels by a factor of $5/8$ or $3/8$ and found that this did not seem to improve performance.
We ran our hyperparameter search runs for 150 epochs.
We selected the best hyperparameter configurations based on the best validation accuracy at any point during the training.
We trained the models for 500 epochs but selected the models saved from 300 epochs as all models had started to overfit by the end of training.
For MFVI, this was using the SGD optimizer with learning rate 0.001, decay rate 0.98 every epoch, batch size 16, 4 variational samples for estimating the loss during training and $\rho$ of -6.
This outperformed the others by a significant margin.
Using our code on a V100 GPU with 8 vCPUs and an SSD this took slightly over 13 hours to train each model.
For the radial posterior, this was the Adam optimizer with learning rate 0.0001, batch size 64, 1 variational sample for estimating the loss during training and a $\rho$ of -6.
Using our code on the same GPU, this took slightly over 3h to run.
However, for the radial posterior there were very many other configurations with similar validation accuracies (one of the advantages of the posterior).

For the experiment shown in Figure \ref{fig:training_loss}, we have selected slightly different hyperparameters in order to train more quickly.
For both models, we use Adam with learning rate 0.0001 and train for 500 epochs.
The models have $5/8$ the number of channels of VGG-16.
The models are trained with batch size 64 and 4 variational samples to estimate the loss and its standard deviation.

\ifSubfilesClassLoaded{
\bibliographystyle{plainnat}
\bibliography{thesis_references}}{}
\end{document}

\chapter{Appendix to Evaluating Bayesian Deep Learning}
\label{chp:a:evaluation}

\newcommand{\nn}{f_\vtheta}
\newcommand{\ed}{\hat{p}(\rvx, \ry)}
\newcommand{\sed}{\tilde{p}(\rvx, \ry)}
\newcommand{\pr}{r}
\newcommand{\er}{\hat{R}}
\newcommand{\Rt}{\tilde{R}}
\newcommand{\Rs}{\tilde{R}_{\text{LURE}}}
\newcommand{\Rp}{\tilde{R}_{\text{PURE}}}
\newcommand{\Dpool}{\data_{\textup{pool}}}
\newcommand{\Dtrain}{\data_{\textup{train}}}
\newcommand{\ssu}{\sigma_{\textup{LURE}}}
\newcommand{\rhat}{\hat{r}}
\newcommand{\var}{\Var}
\newcommand{\muD}{\mu_{m|i,\data}}
\newcommand{\mugD}{\mu_{|\data}}
\newcommand{\mumgD}{\mu_{m|\data}}
\newcommand{\mukgD}{\mu_{k|\data}}
\newcommand*\circled[1]{\tikz[baseline=(char.base)]{
		\node[shape=circle,draw,inner sep=1pt] (char) {#1};}}

\section{Experimental Details}
\subsection{Linear Regression}\label{a:linear}
\begin{figure}
   \begin{subfigure}{0.48\textwidth}
      \centering
      \includegraphics[width=\textwidth]{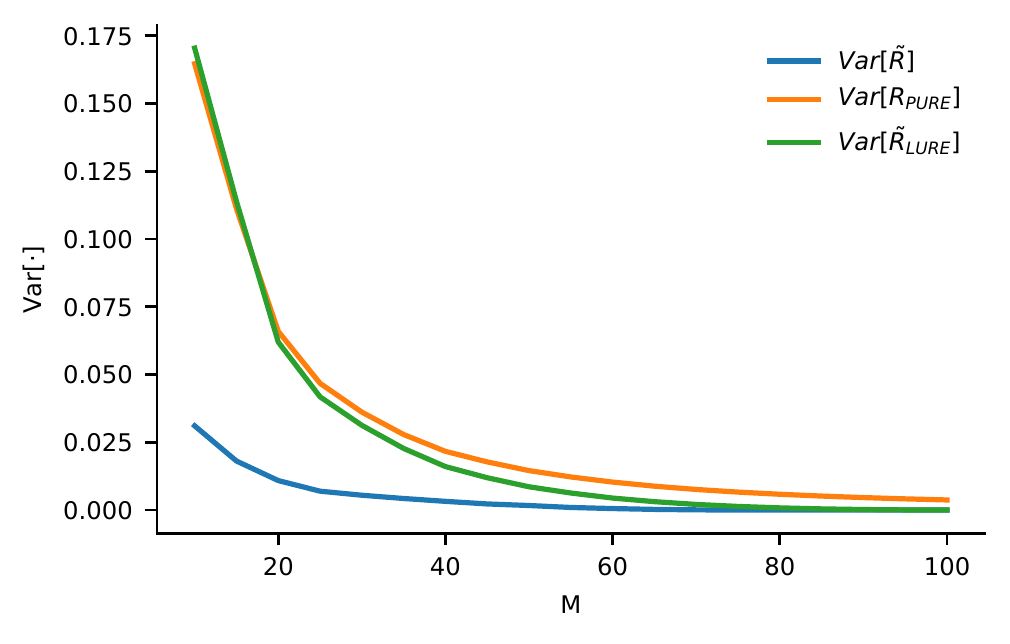}
      \vspace{-6mm}
      \caption{\textbf{Linear Regression}}
      \label{fig:var_linear_no_fit}
   \end{subfigure}
   \hfill
   \begin{subfigure}{0.48\textwidth}
      \centering
      \includegraphics[width=\textwidth]{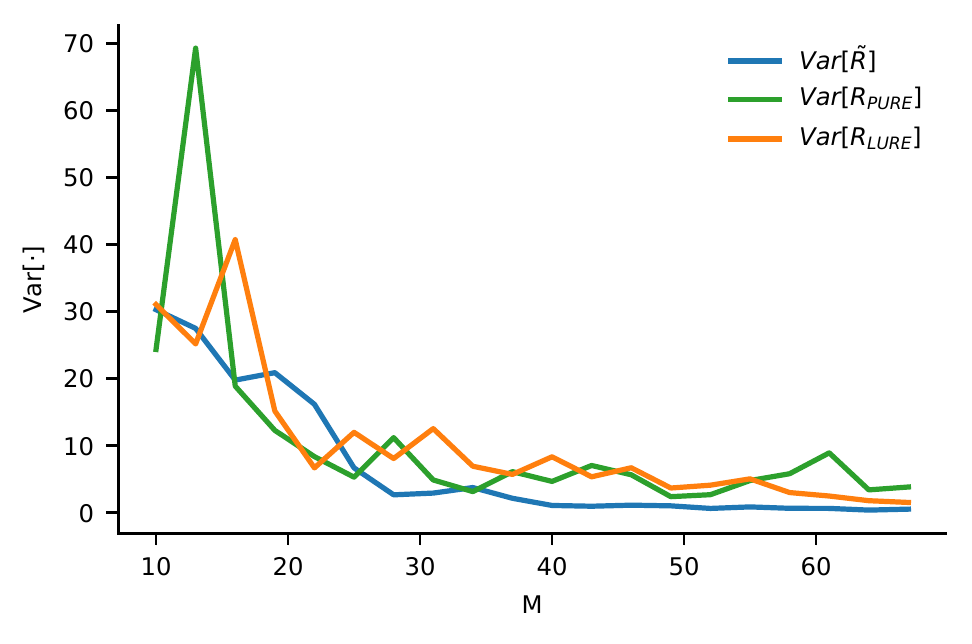}
      \vspace{-6mm}
      \caption{\textbf{Bayesian Neural Network}}
      \label{fig:var_mnist_no_fit}
   \end{subfigure}
   \caption[Variance of the estimators]{For linear regression (a) the biased estimator has the lowest variance, and $\Rs$ improves on $\Rp$. (b) But for the BNN the variances are more comparable, with $\Rs$ the lowest.}
\end{figure}
Our training dataset contains a small cluster of points near $x=-1$ and two larger clusters at $0\leq x \leq 0.5$ and $1 \leq x \leq 1.5$, sampled proportionately to the `true' data distribution.
The data distribution from which we select data in a Rao-Blackwellised manner has a probability density function over $\rx$ equal to:
\begin{equation}
   P(\rx= X) = \begin{cases} 0.12 &-1.2 \leq \rx \leq -0.8\\
      0.95 &0.0 \leq \rx \leq 0.5 \\
      0.95 &1.0 \leq \rx \leq 1.5
   \end{cases}
\end{equation}
while the distribution over $\ry$ is then induced by:
\begin{equation}
   \ry = \text{max}(0, x) \cdot \left(|x|^{\frac{3}{2}} +\frac{\sin(20x)}{4}\right).
\end{equation}

We set $N=101$, where there are 5 points in the small cluster and 96 points in each of the other two clusters, and consider $10 \leq M \leq 100$.
We actively sample points without replacement using a geometric heuristic that scores the quadratic distance to previously sampled points and then selects points based on a Boltzman distribution with $\beta=1$ using the normalized scores.

\begin{figure}
   \centering
   ~~~~~~~~\begin{subfigure}[b]{0.42\textwidth}
      \centering
      \includegraphics[width=\textwidth]{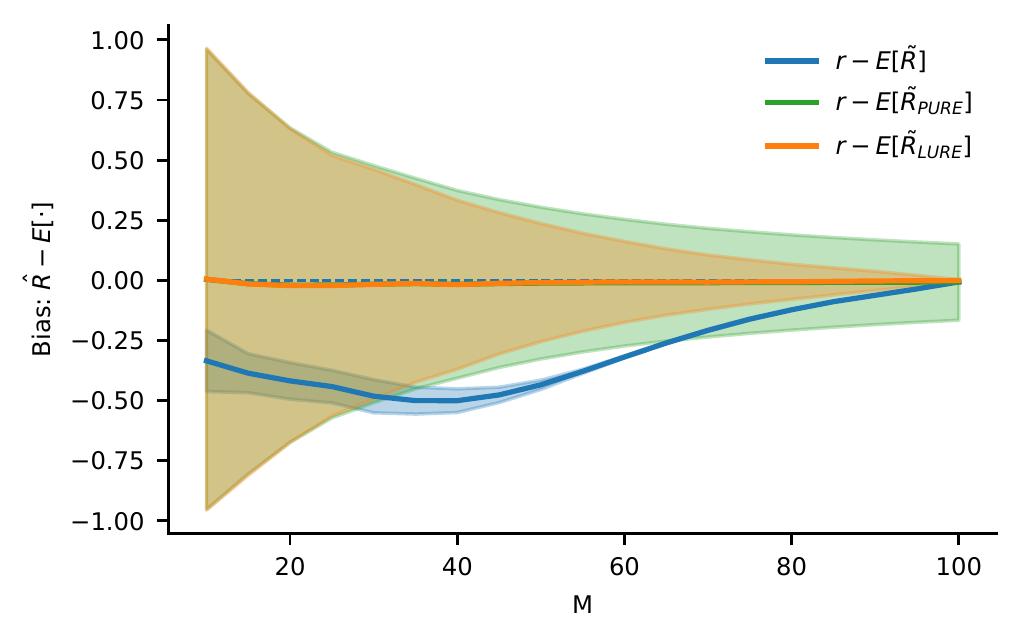}
      \vspace{-6mm}
      \caption{\textbf{Bias (like Fig.\ \ref{fig:bias_linear_no_fit}).}}
      \label{fig:eg_no_fit}
   \end{subfigure}
   \hfill
   \begin{subfigure}[b]{0.42\textwidth}
      \centering
      \includegraphics[width=\textwidth]{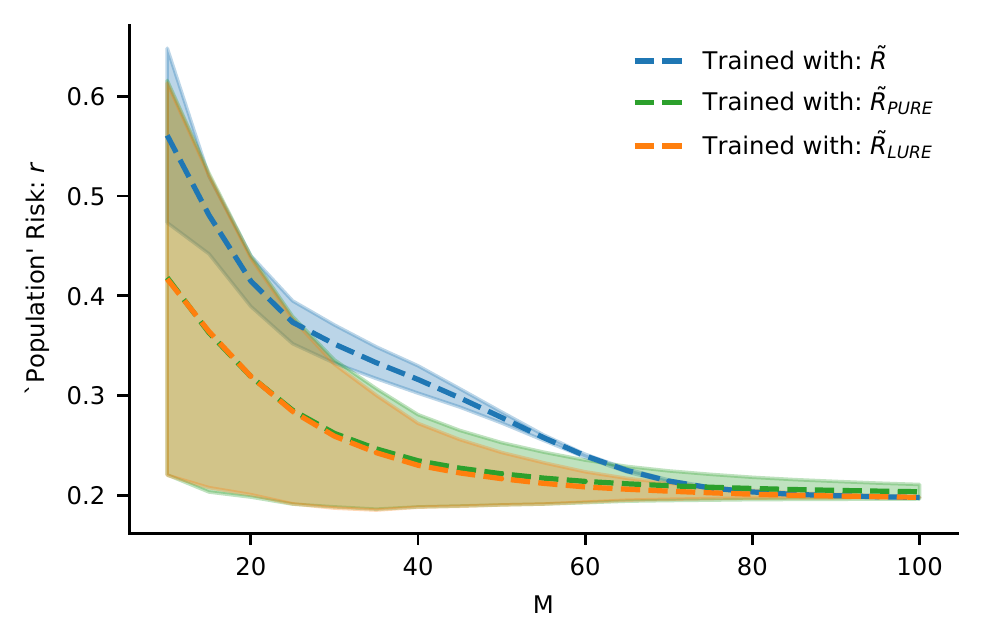}
      \vspace{-6mm}
      \caption{\textbf{TestMSE (like Fig.\ \ref{fig:linear_active_learning}).}}
      \label{fig:eg_fit}
   \end{subfigure}
   \caption[Epsilon-greedy proposal distributions]{Adopting an alternative proposal distribution---here an epsilon-greedy adaptation of a distance-based measure---does not change the overall picture for linear regression.}
   \label{fig:epsilon_greedy}
   \vspace{-4mm}
\end{figure}

Here, we also show in Figure \ref{fig:epsilon_greedy} results that are collected using an epsilon-greedy acquisition proposal.
The results are aligned with those from the other acquisition distribution we consider in the main body of the paper.
This proposal selects the point that is has the highest total distance to all previously selected points with probability 0.9 and uniformly at random with probability $\epsilon = 0.1$.
That is, the acquisition proposal is given by:
\begin{equation}
   P(i_m = j; i_{1:m-1}, \Dpool) = \begin{cases} 1 - \epsilon + \frac{\epsilon}{\abs{\Dpool}} &\argmax_{j \notin \Dtrain} \sum_{k \in \Dtrain} \abs{x_k - x_j}\\
      \frac{\epsilon}{\abs{\Dpool}} & \text{otherwise}
   \end{cases}
\end{equation}
where of course $\Dtrain$ are the $i_{1:m-1}$ elements of $\Dpool$.

For all graphs we use 1000 trajectories with different random seeds to calculate error bars.
Although, of course, each regression and scoring is deterministic, the acquisition distribution is stochastic.

Although the variance of the estimators can be inferred from Figure \ref{fig:bias_linear_no_fit}, we also provide Figure \ref{fig:var_linear_no_fit} which displays the variance of the estimator directly.
\begin{table}[]
   \resizebox{\textwidth}{!}{
   \begin{tabular}{@{}ll@{}}
   \toprule
   Hyperparameter & Setting description                                             \\ \midrule
   Architecture                                     & Convolutional Neural Network                                                             \\
   Conv 1                          & 1-16 channels, 5x5 kernel, 2x2 max pool \\
   Conv 2                                      & 16-32 channels, 5x5 kernel, 2x2 max pool\\
   Fully connected 1 & 128 hidden units \\
   Fully connected 2 & 10 hidden units \\
   Loss function & Negative log-likelihood \\
   Activation                                       & ReLU                          \\
   Approximate Inference Algorithm                  & Radial BNN Variational Inference \citep{farquharRadial2020}\\
   Optimization algorithm                           & Amsgrad \citep{reddiConvergence2018}       \\
   Learning rate                                    & $5\cdot10^{-4}$                                      \\
   Batch size                                       & 64                                                              \\
   Variational training samples                     & 8                                                              \\
   Variational test samples                         & 8                                                              \\
   Variational acquisition samples                         & 100                                                             \\
   Epochs per acquisition& up to 100 (early stopping patience=20), with 1000 copies of data\\
   Starting points & 10 \\
   Points per acquisition & 1\\
   Acquisition proposal distribution & $q(i_m;i_{1:m-1}, \Dpool) = \frac{e^{Ts_i}}{\sum e^{Ts_i}}$\\
   Temperature: $T$ & 10,000 \\
   Scoring scheme: $s$ & BALD (M.I. between $\rvtheta$ and output distribution)\\
   Variational Posterior Initial Mean               & \citet{heDeep2016}                         \\
   Variational Posterior Initial Standard Deviation & $\log[1 + e^{-4}]$          \\
   Prior                                       & $\mathcal{N}(0, 0.25^2)$                                                               \\
   Dataset                                          & Unbalanced MNIST\\
   Preprocessing                                    & Normalized mean and std of inputs.                             \\
   Validation Split                                 & 1000 train points for validation\\
   Runtime per result & 2-4h\\
   Computing Infrastructure & Nvidia RTX 2080 Ti\\
   \bottomrule
   \end{tabular}
   }
   \caption[Hyperparameters for active MNIST experiments]{Experimental Setting---Active MNIST.}
   \label{tbl:hypers-mnist}
   \end{table}
\subsection{Bayesian Neural Network}\label{a:bayesian_neural_network}
We train a Bayesian neural network using variational inference \citep{jordanIntroduction1999}.
In particular, we use the radial Bayesian neural network approximating distribution \citep{farquharRadial2020}.
The details of the hyperparameters used for training are provided in Table \ref{tbl:hypers-mnist}.

\begin{figure}[t]
   \centering
   \vspace{-4mm}
   \begin{subfigure}[b]{0.32\textwidth}
      \centering
      \includegraphics[width=\textwidth]{05_evaluation/active_learning/bias_fit/risk_5_linear.pdf}
      \caption{\textbf{Linear regression. Test MSE.}}
      \label{fig:linear_active_learning}
   \end{subfigure}
   \hfill
   \begin{subfigure}[b]{0.32\textwidth}
      \centering
      \includegraphics[width=\textwidth]{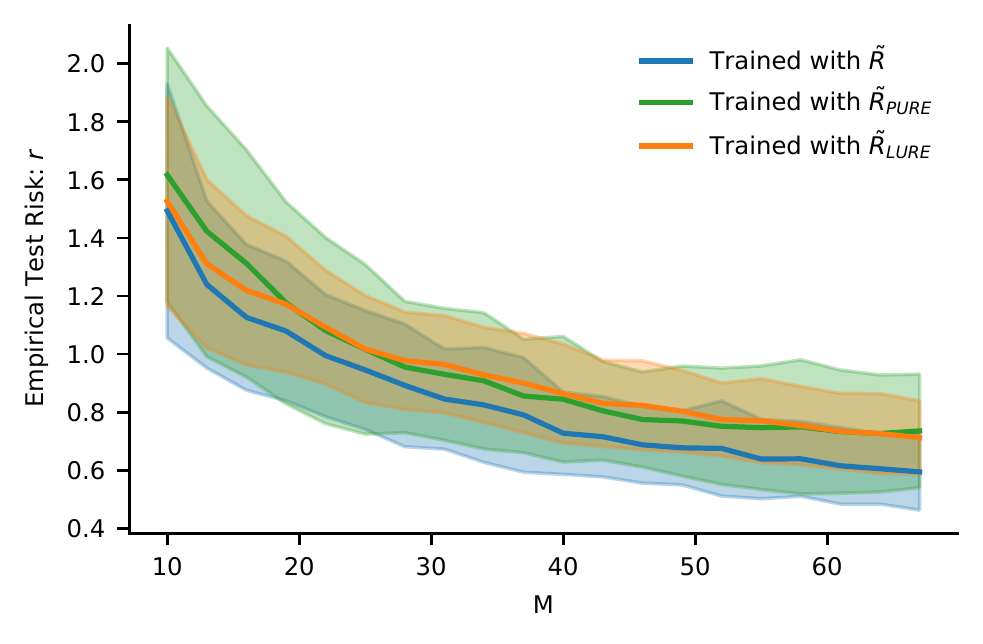}
      \caption{\textbf{MNIST: Test NLL.}}
      \label{fig:mnist_active_learning}
   \end{subfigure}
   \hfill
   \begin{subfigure}[b]{0.32\textwidth}
      \centering
      \includegraphics[width=\textwidth]{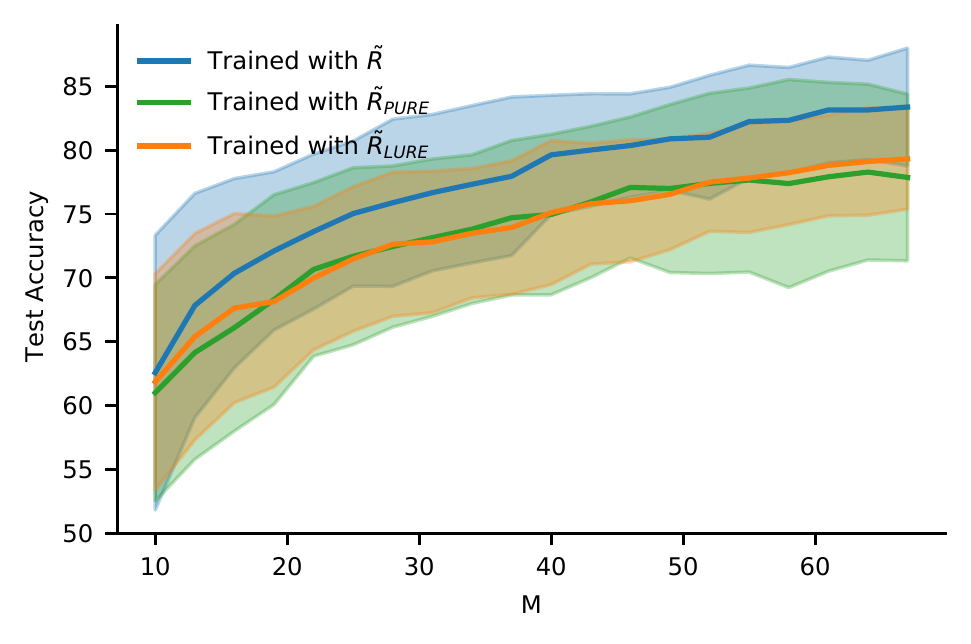}
      \caption{\textbf{MNIST: Test Acc.}}
      \label{fig:mnist_active_learning_accuracy}
   \end{subfigure}
   \medskip
   \begin{subfigure}[b]{0.32\textwidth}
      \centering
      \includegraphics[width=\textwidth]{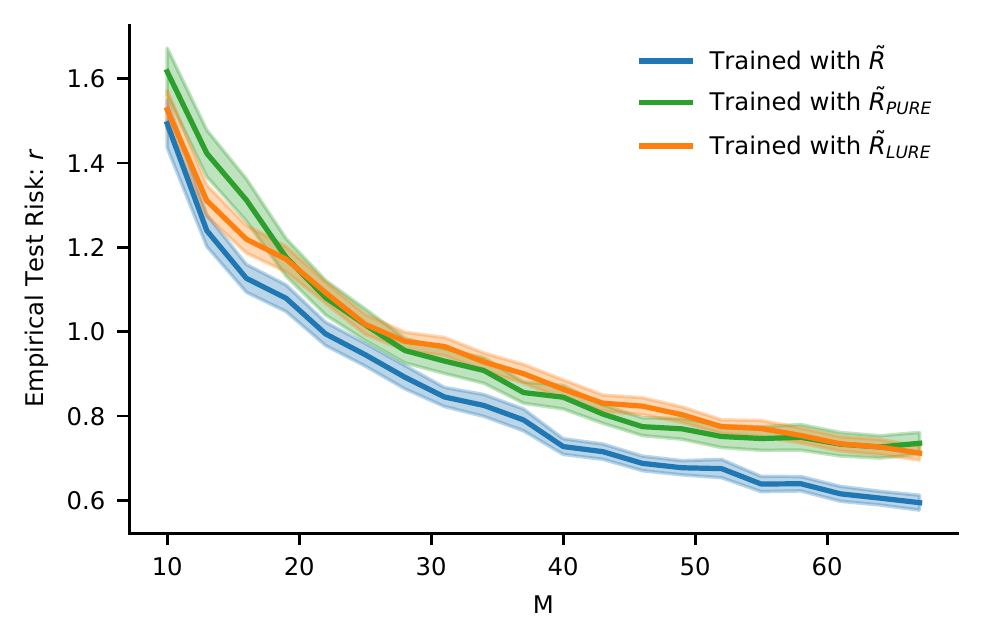}
      \caption{\textbf{MNIST: Test NLL. Standard Error}}
      \label{fig:mnist_active_learning_standard}
   \end{subfigure}
   \hfill
   \begin{subfigure}[b]{0.32\textwidth}
      \centering
      \includegraphics[width=\textwidth]{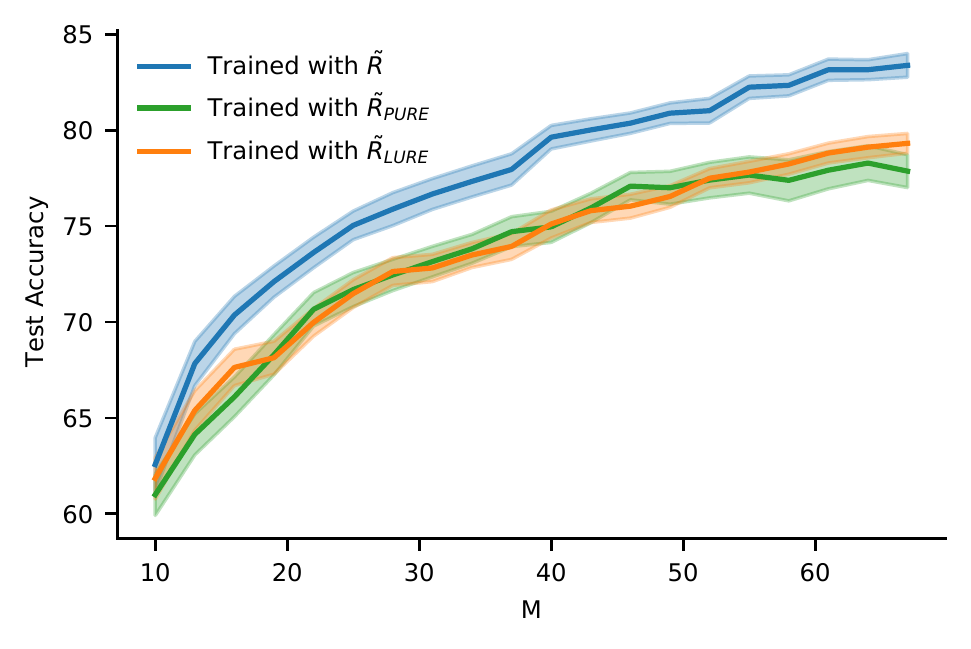}
      \caption{\textbf{MNIST: Test Acc. Standard Error}}
      \label{fig:mnist_active_learning_accuracy_standard_error}
   \end{subfigure}
   \hfill
   \begin{subfigure}[b]{0.32\textwidth}
      \centering
      \includegraphics[width=\textwidth]{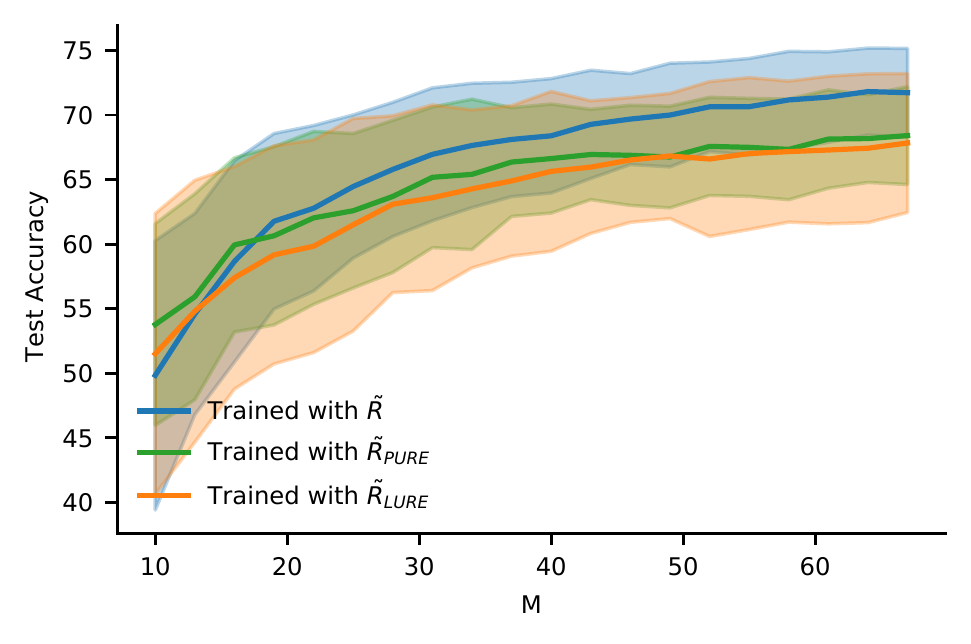}
      \caption{\textbf{FashionMNIST: Test Acc.}}
      \label{fig:fashion_mnist_active_learning_accuracy}
   \end{subfigure}
   \medskip
   \begin{subfigure}[b]{0.32\textwidth}
      \centering
      \includegraphics[width=\textwidth]{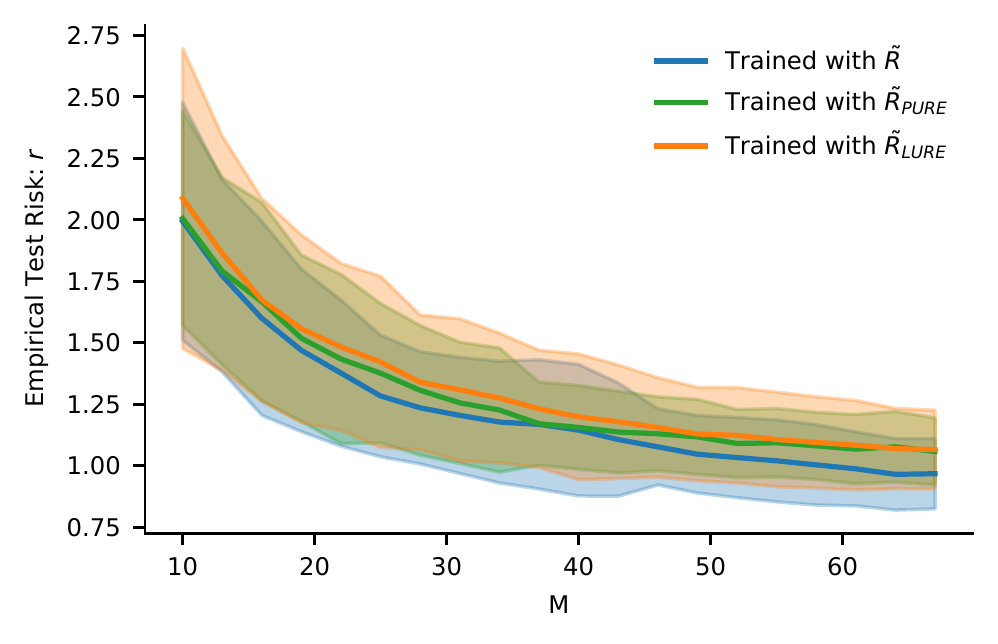}
      \caption{\textbf{FashionMNIST: Test NLL.}}
      \label{fig:fashion_mnist_active_learning}
   \end{subfigure}
   \hfill
   \begin{subfigure}[b]{0.32\textwidth}
      \centering
      \includegraphics[width=\textwidth]{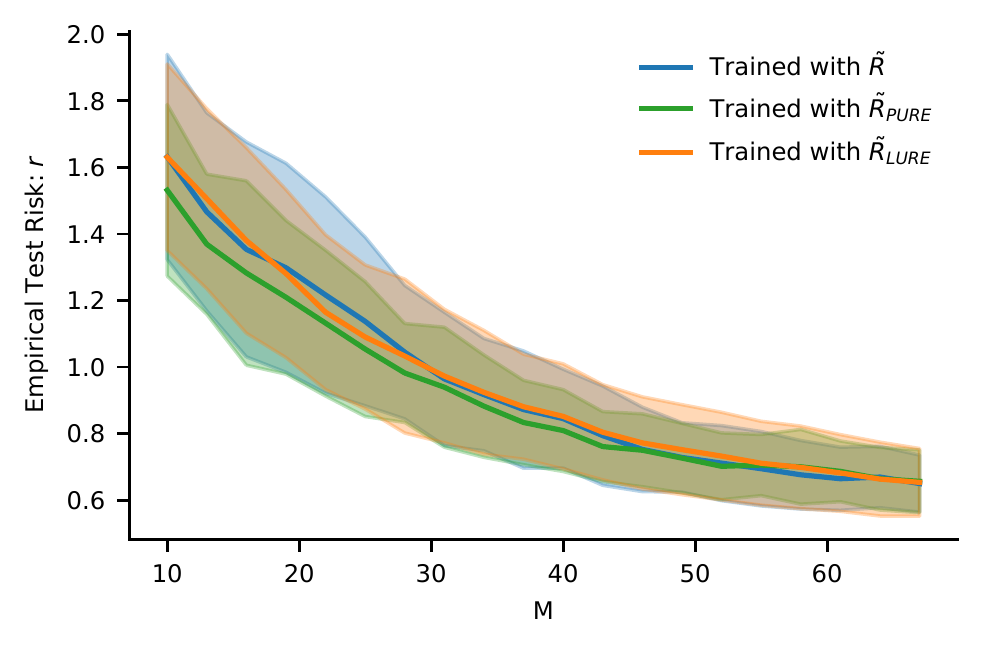}
      \caption{\textbf{MNIST (MCDO): Test NLL.}}
      \label{fig:mcdo_mnist_active_learning}
   \end{subfigure}
   \hfill
   \begin{subfigure}[b]{0.32\textwidth}
      \centering
      \includegraphics[width=\textwidth]{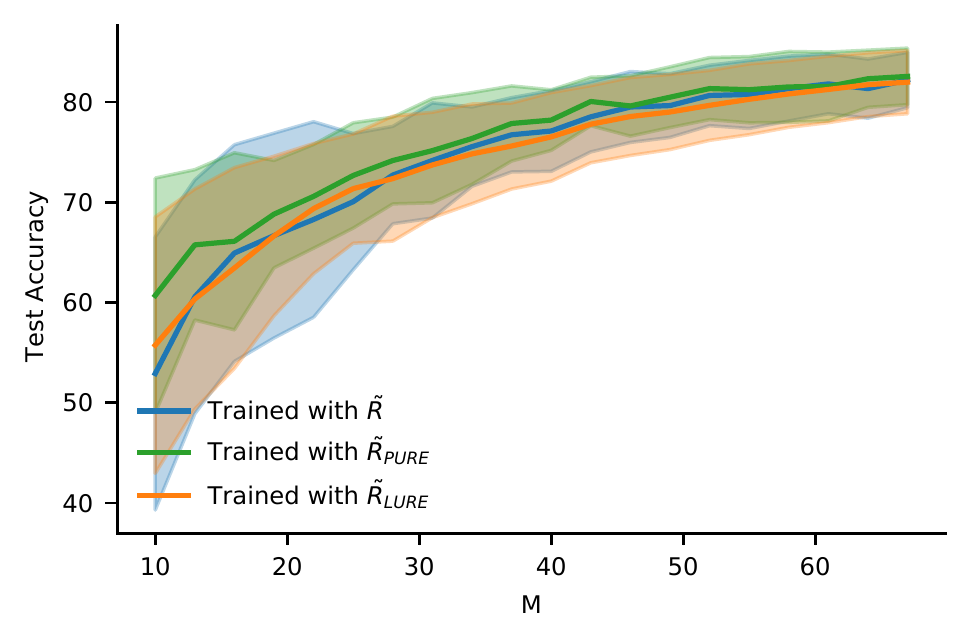}
      \caption{\textbf{MNIST (MCDO): Test Acc.}}
      \label{fig:mcdo_mnist_active_learning_accuracy}
   \end{subfigure}
   \medskip
   \begin{subfigure}[b]{0.32\textwidth}
      \centering
      \includegraphics[width=\textwidth]{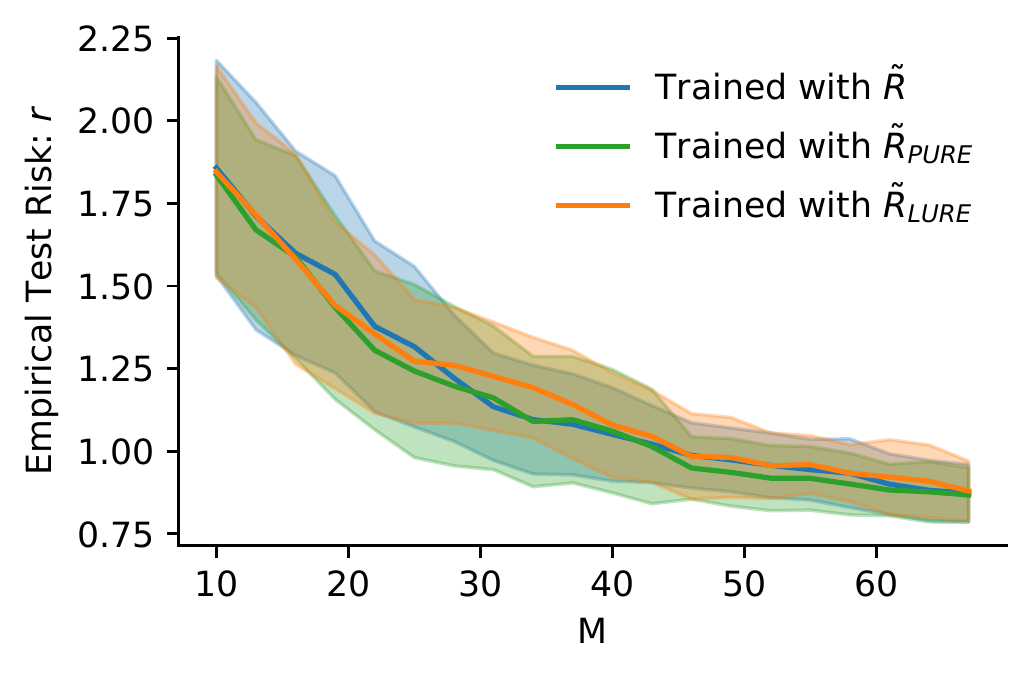}
      \caption{\textbf{FashionMNIST (MCDO): Test NLL.}}
      \label{fig:mcdo_fashion_mnist_active_learning}
   \end{subfigure}
   \hfill
   \begin{subfigure}[b]{0.32\textwidth}
      \centering
      \includegraphics[width=\textwidth]{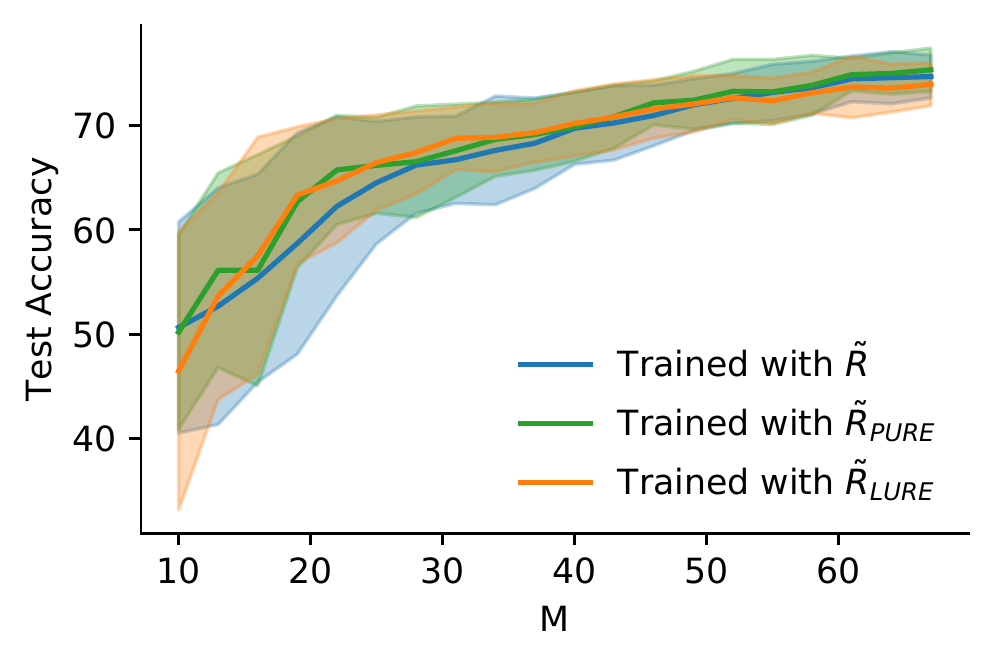}
      \caption{\textbf{FashionMNIST (MCDO): Test Acc.}}
      \label{fig:mcdo_fashion_mnist_active_learning_accuracy}
   \end{subfigure}
   \hfill
   \begin{subfigure}[b]{0.32\textwidth}
      \centering
      \includegraphics[width=\textwidth]{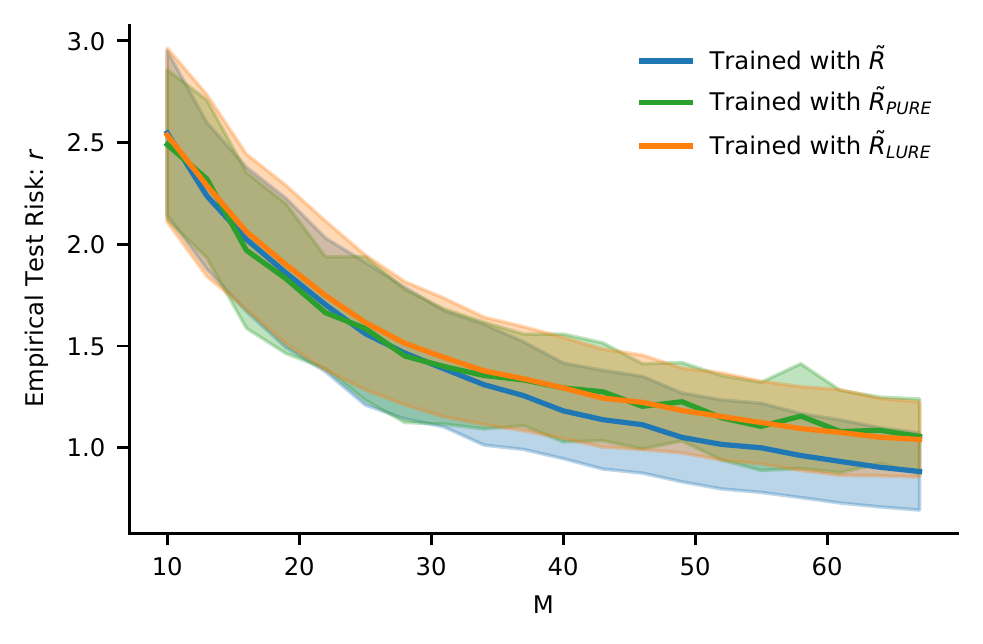}
      \caption{\textbf{MNIST (Balanced): Test NLL.}}
      \label{fig:balanced_mnist_active_learning}
   \end{subfigure}
   \caption[]{For linear regression, the models trained with $\Rp$ or $\Rs$ have lower `population' risk. In contrast, BNNs trained with $\Rs$ or $\Rp$ perform either similarly (h, i) or slightly worse (b-g), even though they remove bias and have lower variance. Shading is one standard deviation to show variation, except for (d-e) where we show standard error to demonstrate the significance of the differences. For (a) 1000 samples and `$r$' estimated on 10,100 points from distribution. Otherwise 45 samples and `$r$' estimated on the test dataset.}
   \label{fig:bias_with_fitting}
\end{figure}

The unbalanced dataset is constructed by first noising 10\% of the training labels, which are assigned random labels, and then selecting a subset of the training dataset such that the numbers of examples of each class is proportional to the ratio (1., 0.5, 0.5, 0.2, 0.2, 0.2, 0.1, 0.1, 0.01, 0.01)---that is, there are 100 times as many zeros as nines in the unbalanced dataset.
(Figure \ref{fig:balanced_mnist_active_learning} shows a version of this experiment which uses a balanced dataset instead, in order to make sure that any effects are not entirely caused by this design choice.)
In fact, we took only a quarter of this dataset in order to speed up acquisition (since each model must be evaluated many times on each of the candidate datapoints to estimate the mutual information).
1000 validation points were then removed from this pool to allow early stopping.
The remaining points were placed in $\Dpool$.
We then uniformly selected 10 points from $\Dpool$ to place in $\Dtrain$.
Adding noise to the labels and using an unbalanced dataset is designed to mimic the difficult situations that active learning systems are deployed on in practice, despite the relatively simple dataset.
However, we used a simple dataset for a number of reasons.
Active learning is very costly because it requires constant retraining, and accurately measuring the properties of estimators generally requires taking large numbers of samples.
The combination makes using more complicated datasets expensive.
In addition, because our work establishes a lower bound on architecture complexity for which correcting the active learning bias is no longer valuable, establishing that lower bound with MNIST is in fact a stronger result than showing a similar result with a more complex model.

\begin{figure}
   \begin{subfigure}{0.48\textwidth}
      \centering
      \includegraphics[width=\textwidth]{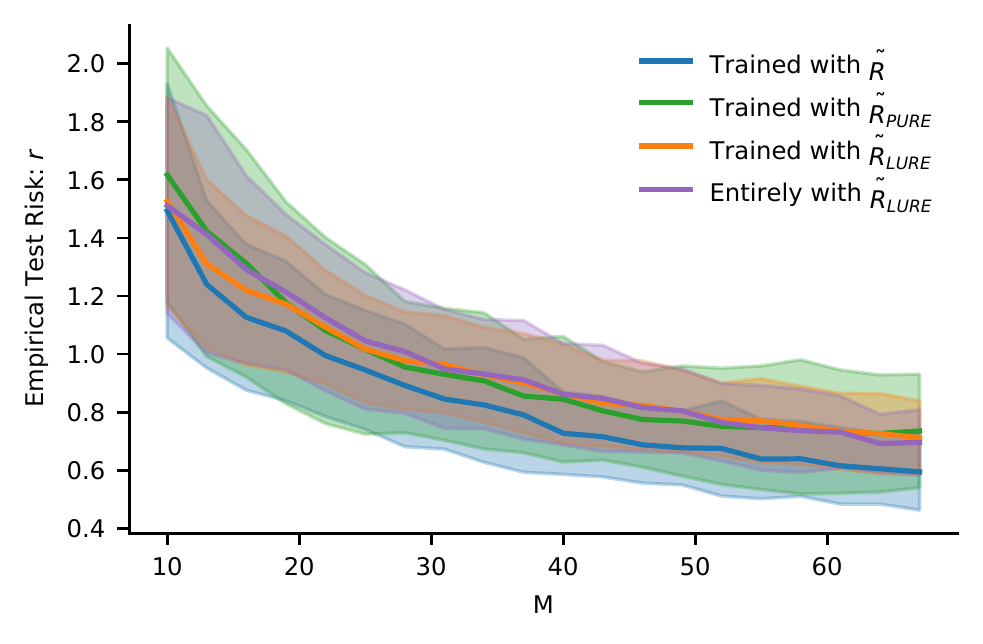}
      \vspace{-6mm}
      \caption{\textbf{Test loss}}
      \label{fig:rsure_mnist_loss}
   \end{subfigure}
   \hfill
   \begin{subfigure}{0.48\textwidth}
      \centering
      \includegraphics[width=\textwidth]{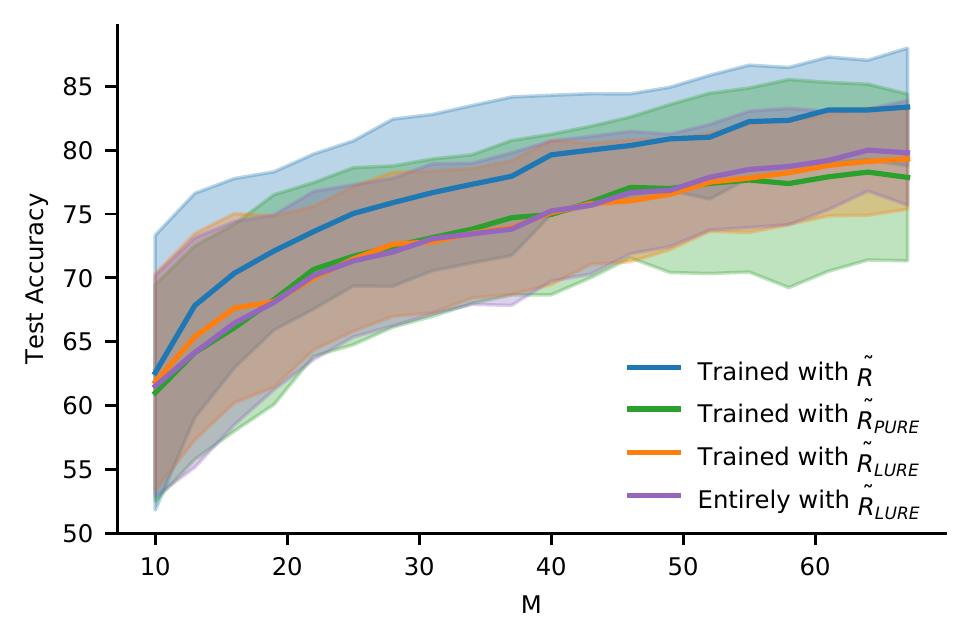}
      \vspace{-6mm}
      \caption{\textbf{Test accuracy}}
      \label{fig:rsure_mnist_acc}
   \end{subfigure}
   \caption[Using unbiased estimators throughout acquisition and training]{We contrast the effect of using $\Rs$ throughout the entire acquisition procedure and training (rather than using the same acquisition procedure based on $\tilde{R}$ for all estimators). The purple test performance and orange are nearly identical, suggesting the result is not sensitive to this choice.}
\end{figure}

The active learning loop then proceeds by:
\begin{enumerate}
   \item training the neural network on $\Dtrain$ using $\tilde{R}$;
   \item scoring $\Dpool$;
   \item sampling a point to be added to $\Dtrain$;
   \item Every 3 points, we separately trained models on $\Dtrain$ using $\tilde{R}$, $\Rp$, and $\Rs$ and evaluate them.
\end{enumerate}   
This ensures that all of the estimators are on data collected under the same sampling distribution for fair comparison.
As a sense-check, in Figures \ref{fig:rsure_mnist_loss} and \ref{fig:rsure_mnist_acc} we show an alternate version in which the first step trains with $\Rs$ instead of $\tilde{R}$, and find that this does not have a significant effect on the results.


\begin{figure}
   \begin{subfigure}{0.48\textwidth}
      \centering
      \includegraphics[width=\textwidth]{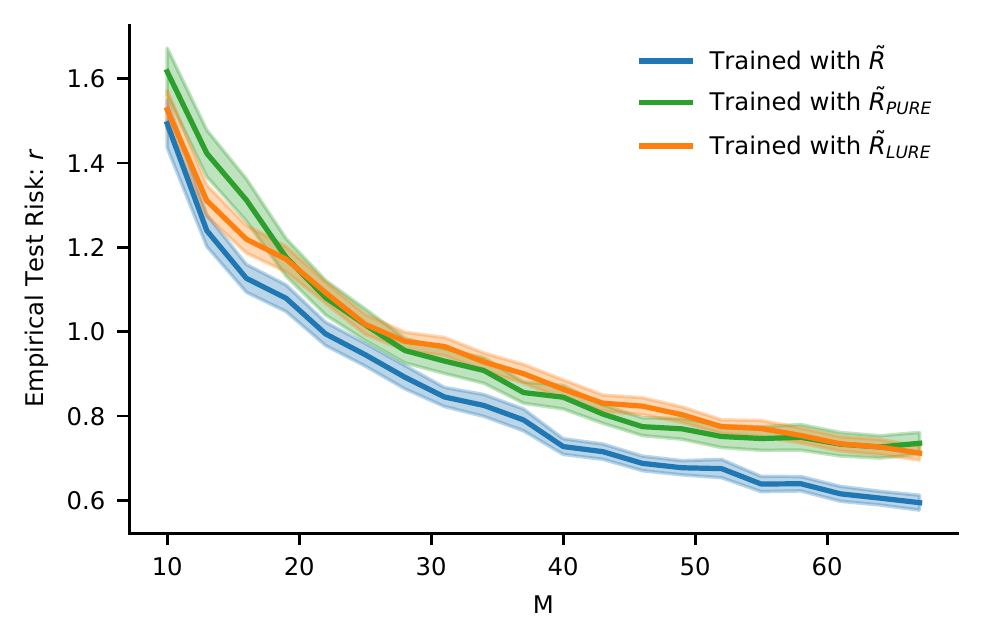}
      \vspace{-6mm}
      \caption{\textbf{Test loss}}
      \label{fig:se_mnist_loss}
   \end{subfigure}
   \hfill
   \begin{subfigure}{0.48\textwidth}
      \centering
      \includegraphics[width=\textwidth]{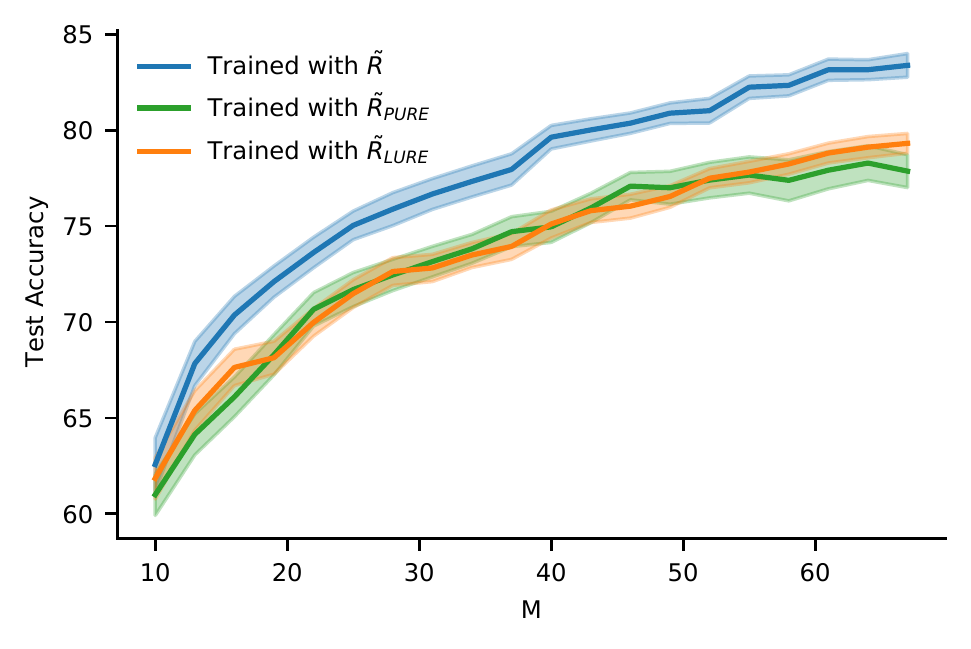}
      \vspace{-6mm}
      \caption{\textbf{Test accuracy}}
      \label{fig:se_mnist_acc}
   \end{subfigure}
   \caption[Active learning plots using standard error to demonstrate differences]{Versions of Figures \ref{fig:mnist_active_learning} and \ref{fig:mnist_active_learning_accuracy} shown with standard errors (45 points) instead of standard deviations. This makes it clearer that the biased $\tilde{R}$ has better performance, even if only marginally so.}
\end{figure}

When we compute the bias of a fixed neural network in Figure \ref{fig:bias_mnist_no_fit}, we train a single neural network on 1000 points.
We then sample evaluation points using the acquisition proposal distribution from the test dataset and evaluate the bias using those points.

In Figures \ref{fig:se_mnist_loss} and \ref{fig:se_mnist_acc} we review the graphs shown in Figures \ref{fig:mnist_active_learning} and \ref{fig:mnist_active_learning_accuracy}, this time showing standard errors in order to make clear that the biased $\tilde{R}$ estimator has better performance, while the earlier figures show that the performance is quite variable.

\begin{figure}[t]
   \centering
   \vspace{-4mm}
   \begin{subfigure}[b]{0.32\textwidth}
      \centering
      \includegraphics[width=\textwidth]{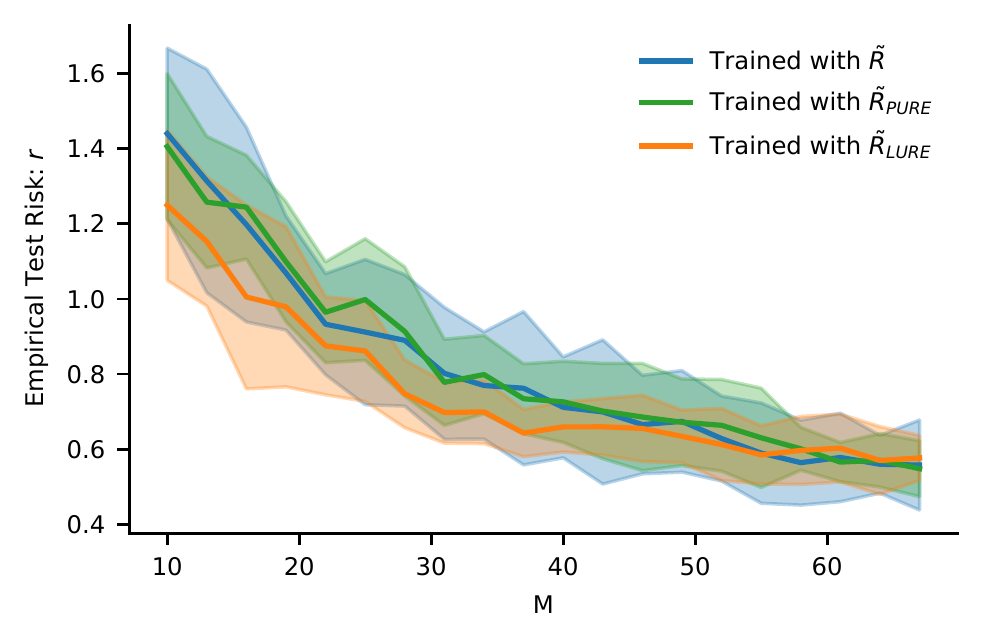}
      \vspace{-6mm}
      \caption{\textbf{T=5000 NLL.}}
      \label{fig:5000nll}
   \end{subfigure}
   \hfill
   \begin{subfigure}[b]{0.32\textwidth}
      \centering
      \includegraphics[width=\textwidth]{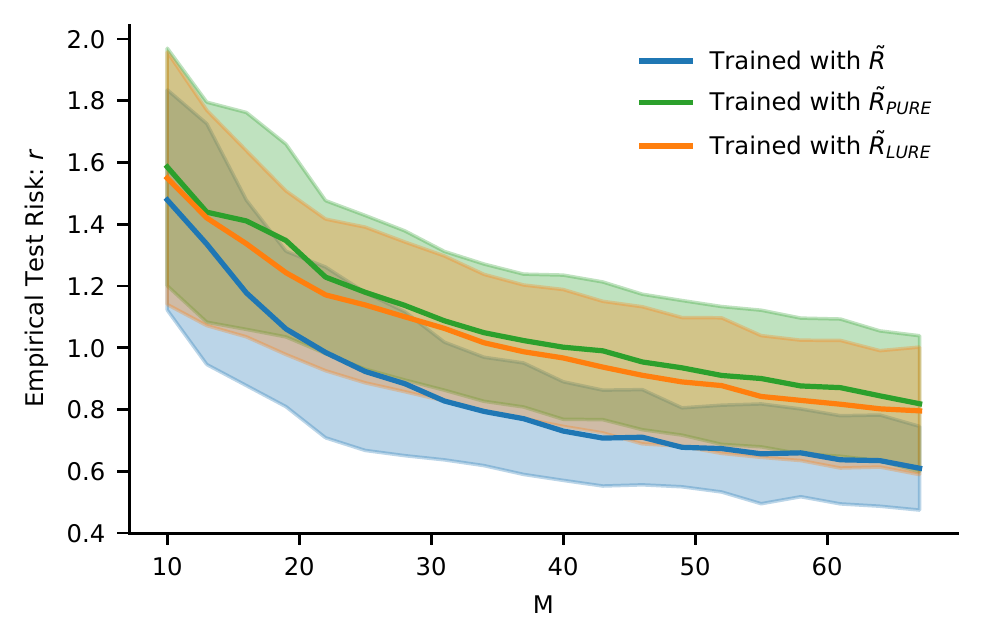}
      \vspace{-6mm}
      \caption{\textbf{T=15000 NLL.}}
      \label{fig:15000nll}
   \end{subfigure}
   \hfill
   \begin{subfigure}[b]{0.32\textwidth}
      \centering
      \includegraphics[width=\textwidth]{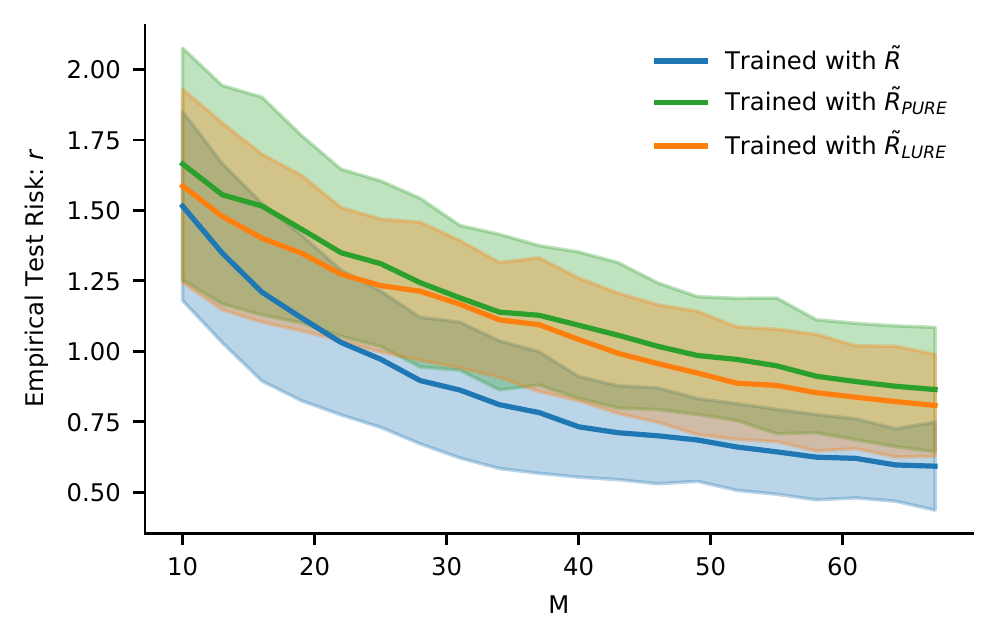}
      \vspace{-6mm}
      \caption{\textbf{T=20000 NLL.}}
      \label{fig:20000nll}
   \end{subfigure}
   \medskip
   \begin{subfigure}[b]{0.32\textwidth}
      \centering
      \includegraphics[width=\textwidth]{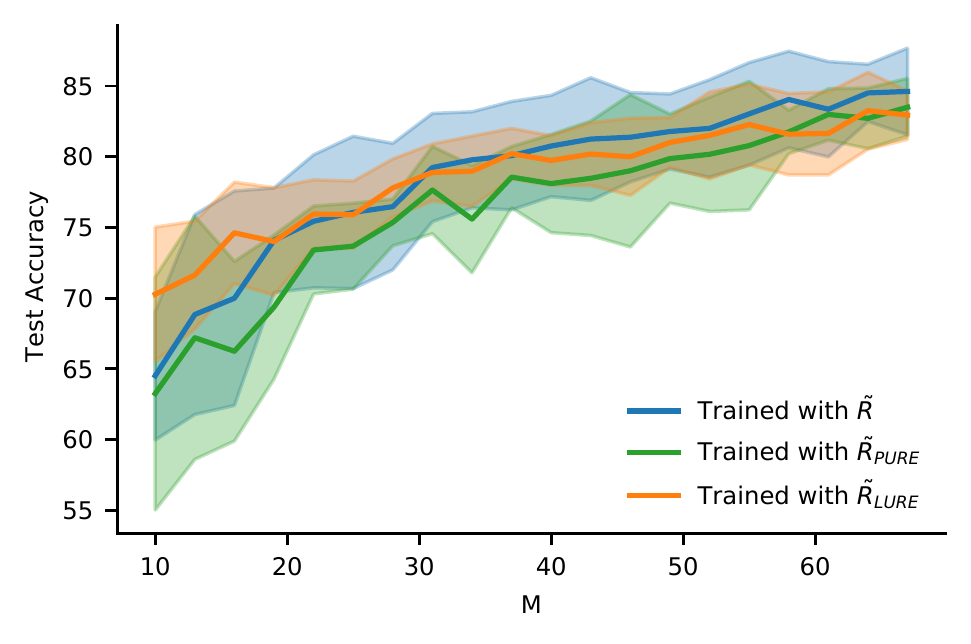}
      \vspace{-6mm}
      \caption{\textbf{T=5000 Acc.}}
      \label{fig:5000acc}
   \end{subfigure}
   \hfill
   \begin{subfigure}[b]{0.32\textwidth}
      \centering
      \includegraphics[width=\textwidth]{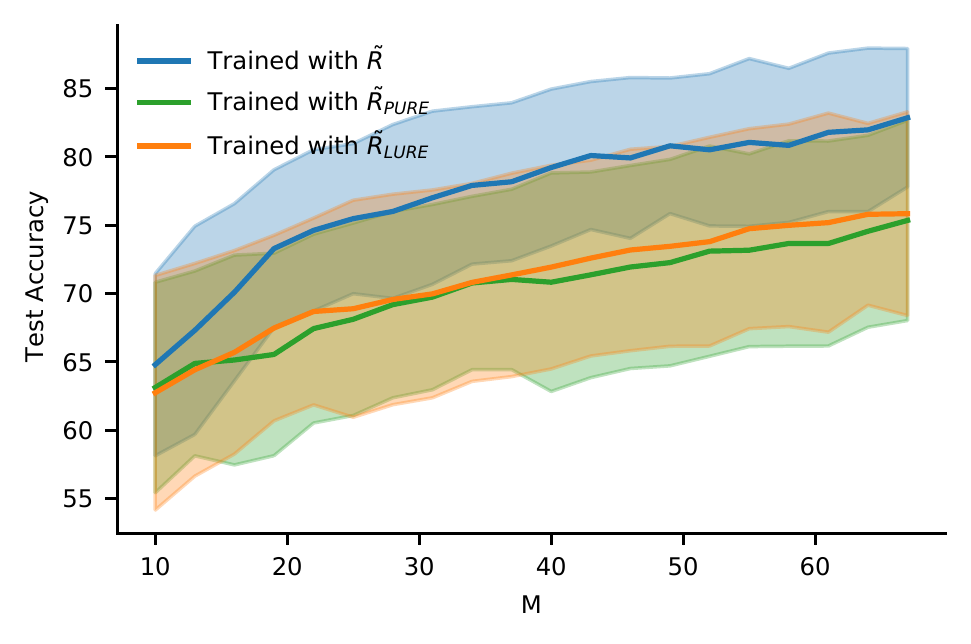}
      \vspace{-6mm}
      \caption{\textbf{T=15000 Acc.}}
      \label{fig:15000acc}
   \end{subfigure}
   \hfill
   \begin{subfigure}[b]{0.32\textwidth}
      \centering
      \includegraphics[width=\textwidth]{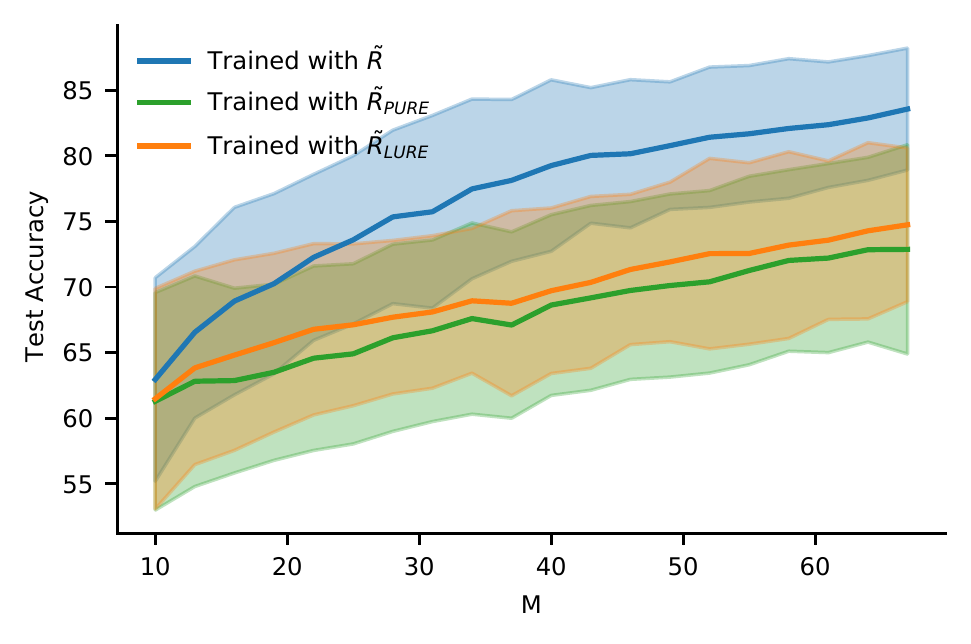}
      \vspace{-6mm}
      \caption{\textbf{T=20000 Acc.}}
      \label{fig:20000acc}
   \end{subfigure}
   \vspace{-2mm}
   \caption[Temperature ablation for unbiased active learning]{Higher temperatures approach a deterministic acquisition function. These also tend to increase the variance of the risk estimator because the weight associated with unlikely points increases, when it happens to be selected. The overall pattern seems fairly consistent, however.}
   \label{fig:temperature_ablation}
   \vspace{-4mm}
\end{figure}

We considered a range of alternative proposal distributions.
In addition to the Boltzman distribution which we used, we considered a temperature range between 1,000 and 20,000 finding it had relatively little effect.
Higher temperatures correspond to more certainly picking the highest mutual information point, which approaches a deterministic proposal.
We found that because the mutual information had to be estimated, and was itself a random variable, different trajectories still picked very different sets of points.
However, for very high temperatures the estimators became higher variance, and for lower temperatures, the acquisition distribution became nearly uniform.
In Figure \ref{fig:temperature_ablation} we show the results of networks trained with a variety of temperatures other than the 10,000 ultimately used.
We also considered a proposal which was simply proportional to the scores, but found this was also too close to sampling uniformly for any of the costs or benefits of active learning to be visible.

\begin{figure}[t]
   \centering
   \vspace{-4mm}
   \begin{subfigure}[b]{0.32\textwidth}
      \centering
      \includegraphics[width=\textwidth]{05_evaluation/active_learning/bias_fit/fashion/None-weighted_test_accuracy.pdf}
      \vspace{-6mm}
      \caption{\textbf{FashionMNIST: Accuracy.}}
      \label{fig:fashionacc}
   \end{subfigure}
   \hfill
   \begin{subfigure}[b]{0.32\textwidth}
      \centering
      \includegraphics[width=\textwidth]{05_evaluation/active_learning/bias_fit/mcdo/mcdo_bnn-weighted_test_accuracy.pdf}
      \vspace{-6mm}
      \caption{\textbf{MNIST (MCDO): Acc.}}
      \label{fig:mcdoacc}
   \end{subfigure}
   \hfill
   \begin{subfigure}[b]{0.32\textwidth}
      \centering
      \includegraphics[width=\textwidth]{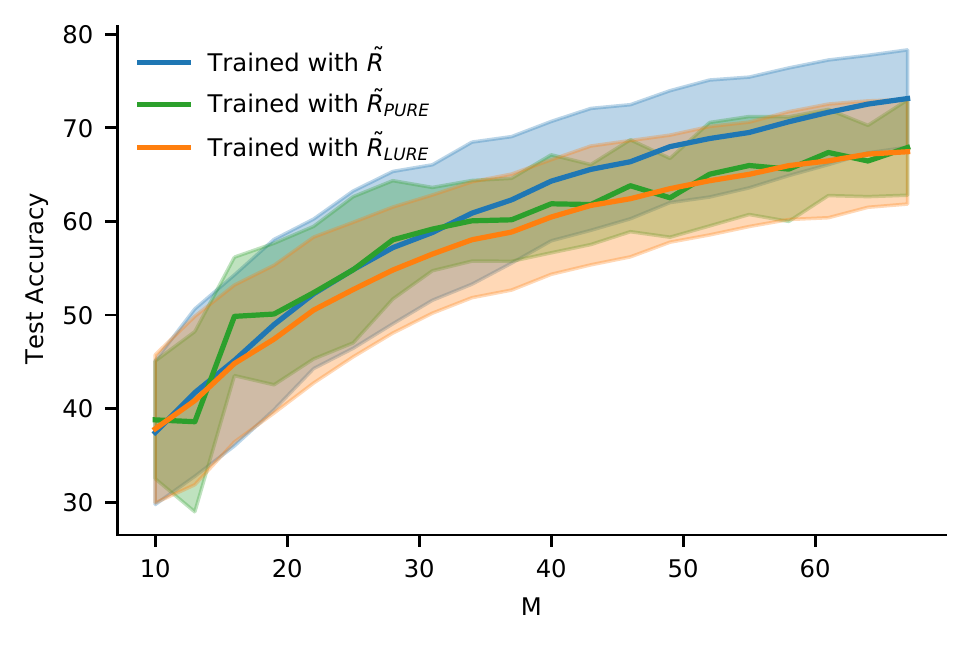}
      \vspace{-6mm}
      \caption{\textbf{MNIST (Balanced): Acc.}}
      \label{fig:balancedacc}
   \end{subfigure}
   \medskip
   \begin{subfigure}[b]{0.32\textwidth}
      \centering
      \includegraphics[width=\textwidth]{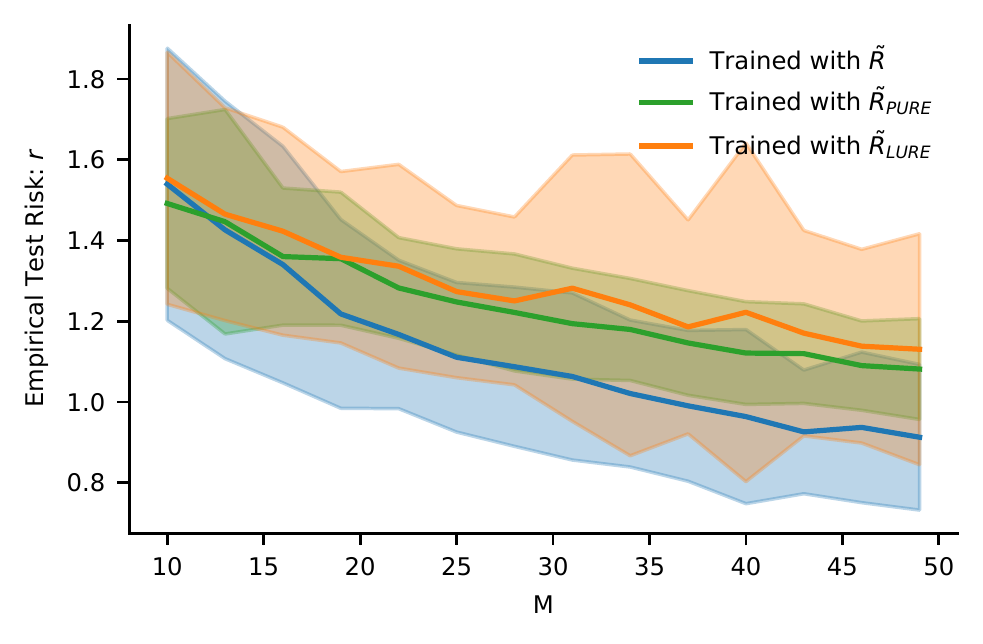}
      \vspace{-6mm}
      \caption{\textbf{MNIST (MLP): NLL}}
      \label{fig:mlpnll}
   \end{subfigure}
   \qquad
   \begin{subfigure}[b]{0.32\textwidth}
      \centering
      \includegraphics[width=\textwidth]{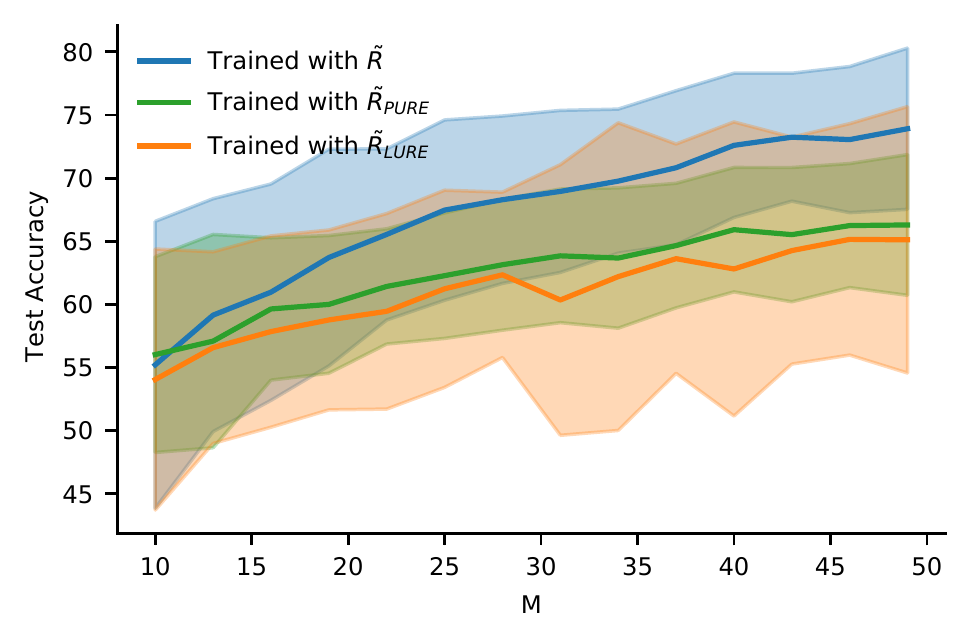}
      \vspace{-6mm}
      \caption{\textbf{MNIST (MLP): Accuracy}}
      \label{fig:mlpacc}
   \end{subfigure}
   \vspace{-2mm}
   \caption[Further unbiased active learning downstream performance ablations]{Further downstream performance experiments. (a)-(c) are partners to Figures \ref{fig:fashion_mnist_active_learning}, \ref{fig:mcdo_mnist_active_learning}, and \ref{fig:balanced_mnist_active_learning}. (d) and (e) show similar results for a smaller multi-layer perceptron (with one hidden layer of 50 units). In all cases the results broadly mirror the results in the main paper.}
   \label{fig:ablations}
\end{figure}

We considered Monte Carlo dropout as an alternative approximating distribution \citep{galDropout2015} (see Figures \ref{fig:mcdo_mnist_active_learning} and \ref{fig:mcdoacc}).
We found that the mutual information estimates were compressed in a fairly narrow range, consistent with the observation by \citet{osbandRandomized2018} that Monte Carlo dropout uncertainties do not necessarily converge unless the dropout probabilities are also optimized \citep{galConcrete2017}.
While this might be good enough when only the \emph{relative} score is needed in order to calculate the argmax, for our proposal distribution we would ideally prefer to have good \emph{absolute} scores as well.
For this reason, we chose the richer approximate posterior distribution instead.

Last, we considered a different architecture, using a full-connected neural network with a single hidden layer with 50 units, also trained as a Radial BNN.
This showed higher variance in downstream performance, but was broadly similar to the convolutional architecture (see Figures \ref{fig:mlpnll} and \ref{fig:mlpacc}).

\section{Deep Active Learning In Practice}\label{a:active_learning_practice}
\begin{table}[t]
    \resizebox{\textwidth}{!}{
    \begin{tabular}{lcccl}
    \toprule
    Reference & Application & Corrects Bias & Acknowledges Bias & Notes \\ \midrule
\citep{senerActive2018}&             &               &                   &       \\
\citep{shenDeep2018} &  \Checkmark &               &                   &       \\
\citep{beluchPower2018} &             &               &                   &       \\
\citep{hautActive2018}&\Checkmark&               &                   &       \\
\citep{sinhaVariational2019}&             &               &                   &       \\
\citep{siddhantDeep2018}&\Checkmark&                   &\Checkmark\\
\citep{ghosalWeakly2019}&\Checkmark&               &                   &       \\
\citep{yangLeveraging2018}&\Checkmark&               &                   &       \\
\citep{yooLearning2019}&             &               &                   &       \\
\citep{kirschBatchBALD2019}&             &               &                   &       \\
\citep{huangCostEffective2018}&             &               &\Checkmark&       \\
\citep{wenComparison2018}&\Checkmark&               &                   &       \\
\citep{chenDeep2019}&\Checkmark&               &                   &Discusses bias in $\Dpool$.\\
\citep{zhangBayesian2019}&\Checkmark&               &                   &Discusses bias in $\Dpool$.\\
\citep{kellenbergerHalf2019}&\Checkmark&               &                   &       \\
\bottomrule

    \end{tabular}}
    \caption[Overview of active learning bias in prior work]{Existing applications of deep active learning rarely acknowledge the bias introduced by actively sampling points and do not, to the best of our knowledge, try to correct it.}
    \label{tbl:existing_active_learning}
    \end{table}
In Table \ref{tbl:existing_active_learning}, we show an informal survey of highly cited papers citing \citet{galDeep2017}, which introduced active learning to computer vision using deep convolutional neural networks.
Across a range of papers including theory papers as well as applications ranging from agriculture to molecular science only two papers acknowledged the bias introduced by actively sampling and none of the papers took steps to address it.
It is worth noting, though, that at least two papers motivated their use of active learning by observing that they expected their training data to already be unrepresentative of the population data and saw active learning as a way to address \emph{that} bias.
This does not quite work, unless you explicitly assume that the actively chosen distribution is more like the population distribution, but is an interesting phenomenon to observe in practical applications of active learning.
\ifSubfilesClassLoaded{
\bibliographystyle{plainnat}
\bibliography{thesis_references}}{}
\end{document}

\end{document}